\documentclass[10pt,journal,compsoc]{IEEEtran}

\ifCLASSOPTIONcompsoc
\usepackage[nocompress]{cite}
\else
\usepackage{cite}
\fi

\ifCLASSINFOpdf

\else

\fi

\usepackage{graphicx}
\usepackage{amsmath,amssymb} % define this before the line numbering.
\usepackage{color}
\usepackage[pagebackref=true,breaklinks=true,citecolor=blue,linkcolor=blue,letterpaper=true,colorlinks,bookmarks=false]{hyperref}

\usepackage{xspace}
\usepackage{color}

\definecolor{gray}{rgb}{0.35,0.35,0.35}
\definecolor{MyBlue}{rgb}{0,0.2,0.8}
\definecolor{MyRed}{rgb}{0.8,0.2,0}
\definecolor{MyGreen}{rgb}{0.0,0.5,0.1}
\definecolor{MyGray}{rgb}{0.4,0.4,0.4}

\usepackage{graphicx}
\graphicspath{{figures/}{figures/teaser}{figures/result}{figures/illustration}}

\long\def\ignorethis#1{}

\newlength\figcca%
\newlength\figccb%
\newlength\figccc%
\newlength\figccd

\newcommand{\vf}{{\bf f}}

\newcommand{\vx}{{\bf x}}

\newcommand{\vN}{{\bf N}}

\newcommand{\vP}{{\bf P}}

\newcommand{\vW}{{\bf W}}

\newcommand{\vS}{{\bf S}}
\newcommand{\vT}{{\bf T}}

\def\onedot{. }

 \def\vs{\emph{vs}\onedot}

\usepackage{times}
\usepackage{mathptmx}
\usepackage{multirow}
\usepackage{paralist}
\usepackage{array}
\usepackage{balance}
\usepackage{algorithm}
\usepackage{algorithmicx}
\usepackage{algpseudocode}
\usepackage{url}
\usepackage{enumitem}
\usepackage{booktabs}

\newlength\figwidth

\begin{document}

\title{Tracking Persons-of-Interest via\\ Unsupervised Representation Adaptation}

\author{Shun~Zhang,
Jia-Bin~Huang, %~\IEEEmembership{Member,~IEEE},
Jongwoo~Lim, %~\IEEEmembership{Member,~IEEE},
Yihong~Gong, %~\IEEEmembership{Senior Member,~IEEE},
Jinjun~Wang, \\ %~\IEEEmembership{Member,~IEEE},
Narendra Ahuja, %~\IEEEmembership{Fellow,~IEEE},
and~Ming-Hsuan~Yang %,~\IEEEmembership{Senior~Member,~IEEE
% }

\IEEEcompsocitemizethanks{
\IEEEcompsocthanksitem S. Zhang is with School of Electronics and Information, Northwestern Polytechnical University, Xi'an, P.R. China, 710072. E-mail: szhang@nwpu.edu.cn.
\IEEEcompsocthanksitem
J.-B. Huang is with Department of Electrical and Computer Engineering, Virginia Tech, Blacksburg, VA, 24060. E-mail: jbhuang@vt.edu.
\IEEEcompsocthanksitem
J. Lim is with the Division of Computer Science and Engineering, Hanyang University, Seoul, Korea, 133-791. E-mail: jlim@hanyang.ac.kr.
\IEEEcompsocthanksitem
Y. Gong and J. Wang are with the Institute of Artificial Intelligence and Robotics, Xi'an Jiaotong University, Xi'an, P.R. China, 710049. E-mail: \{ygong,jinjun\}@mail.xjtu.edu.cn.
\IEEEcompsocthanksitem
N. Ahuja is with the Department of Electrical and Computer Engineering, University
of Illinois at Urbana-Champaign, Urbana, IL. E-mail: n-ahuja@illinois.edu.
\IEEEcompsocthanksitem
M.-H. Yang is with the School of Engineering, University of California at Merced, Merced, CA, 95344. E-mail: mhyang@ucmerced.edu.
%\protect \\
% note need leading \protect in front of \\ to get a newline within \thanks as
% \\ is fragile and will error, could use \hfil\break instead.
}% <-this % stops an unwanted space
%\thanks{Manuscript received April 19, 2005; revised August 26, 2015.}
}

\IEEEtitleabstractindextext{
\begin{abstract}
Multi-face tracking in unconstrained videos is a challenging problem as faces of one person often appear drastically different in multiple shots due to significant variations in scale, pose, expression, illumination, and make-up.
Existing multi-target tracking methods often use low-level features which are not sufficiently discriminative for identifying faces with such large appearance variations.
In this paper, we tackle this problem by learning discriminative, video-specific face representations using convolutional neural networks (CNNs).
Unlike existing CNN-based approaches which are only trained on large-scale face image datasets offline, we use the contextual constraints to generate a large number of training samples for a given video, and further adapt the pre-trained face CNN to specific videos using discovered training samples.
Using these training samples, we optimize the embedding space so that the Euclidean distances correspond to a measure of semantic face similarity via minimizing a triplet loss function.
With the learned discriminative features, we apply the hierarchical clustering algorithm to link tracklets across multiple shots to generate trajectories.
We extensively evaluate the proposed algorithm on two sets of TV sitcoms and YouTube music videos, analyze the contribution of each component, and demonstrate significant performance improvement over existing techniques.
\end{abstract}

}

\maketitle

\IEEEdisplaynontitleabstractindextext

% For peer review papers, you can put extra information on the cover
% page as needed:
% \ifCLASSOPTIONpeerreview
% \begin{center} \bfseries EDICS Category: 3-BBND \end{center}
% \fi
%
% For peerreview papers, this IEEEtran command inserts a page break and
% creates the second title. It will be ignored for other modes.
\IEEEpeerreviewmaketitle
\IEEEraisesectionheading{\section{Introduction}\label{sec:introduction}}

%\paragraph{What's the problem?}
\IEEEPARstart{M}{ulti-target} tracking (MTT) aims at locating all targets of interest (e.g., faces, pedestrians, players, and cars) and inferring their trajectories in a video over time while maintaining their identities.
This problem is at the core of numerous computer vision applications such as video surveillance, robotics, and sports analysis.
Multi-face tracking is one important domain of MTT that can be applied to high-level video understanding tasks such as face recognition, content-based retrieval, and interaction analysis.

\begin{figure}[t]
\setlength{\abovecaptionskip}{0mm}
\setlength{\belowcaptionskip}{-3mm}
\centering
\includegraphics[width=0.5\textwidth]{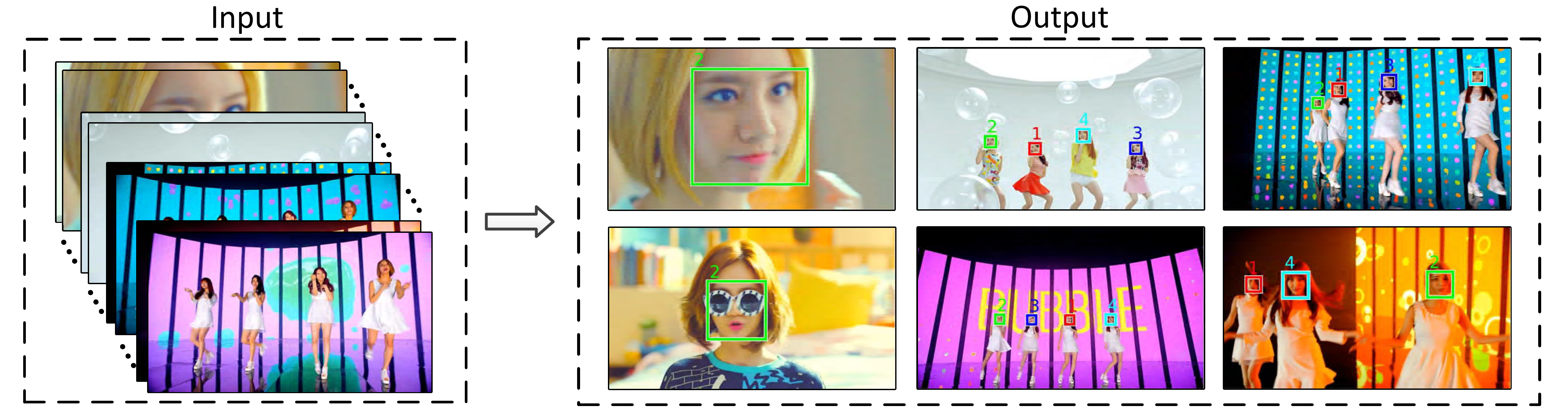}
\vspace{-3mm}
\caption{
\textbf{Multi-face tracking}.
We tackle the problem of tracking multiple faces of people while maintaining their identities in \textit{unconstrained} videos.
Such videos consist of many shots from different cameras.
The main challenge is to address large appearance variations of faces from different shots due to changes in pose, view angle, scale, makeup, illumination, camera motion and heavy occlusions.
}
\label{fig:Faces}
\vspace{-5mm}
\end{figure}

%\paragraph{Challenges in multi-face tracking}
The problem of multi-face tracking is particularly challenging in unconstrained scenarios where the videos are generated from multiple moving cameras with different views or scenes as shown in Figure~\ref{fig:Faces}.
Examples include automatic character tracking in movies, TV sitcoms, or music videos.
It has attracted increasing attention in recent years due to the fast-growing popularity of such videos on the Internet.
Unlike tracking in the \emph{constrained} counterparts (e.g., a video from a single stationary or moving camera) where the main challenge is to deal with occlusions and intersections, multi-face tracking in \emph{unconstrained} videos needs to address the following issues:
(1) A video often consists of many shots and the contents of two neighboring shots may be dramatically different;
(2) It entails dealing with re-identifying faces with large appearance variations due to changes in scale, pose, expression, illumination, and makeup in different shots or scenes; and
(3) The results of face detection may be unreliable due to low resolution, occlusion, nonrigid deformation, motion blurring and complex backgrounds.

%\paragraph{How people addressed the problem? Issues with existing approaches}
Multi-target tracking has been extensively studied in the literature with the primal focus on humans.
Recent approaches often address multi-face tracking by tracking-by-detection techniques.
These methods first apply an object detector to locate faces in every frame, and
apply data association approaches~\cite{brendel2011multiobject,collins2012multitarget,yang2012multi,zhang2008global,zhao2012tracking}
that use visual cues (e.g., appearance, position, motion, and size) in an affinity model to link detections or tracklets (track fragments) into trajectories.
Such methods are effective when the targets are continuously detected and when the camera is either stationary or slowly moving.
However, for unconstrained videos with many shot changes and intermittent appearance of targets, the data association problem becomes
more difficult because the assumptions such as appearance and size consistency, and continuous motion no longer hold in neighboring shots.
Therefore, the design of discriminative features plays a critical role in identifying faces \textit{across} shots in unconstrained scenarios.

Existing MTT methods~\cite{yang2012multi,zhang2008global,zhao2012tracking} use combinations of low-level features such as color histograms, Haar-like features, or HOG~\cite{dalal2005histograms} to construct an appearance model for each target.
However, these hand-crafted features often are not sufficiently discriminative to identify faces with large appearance changes.
For example, low-level features extracted from faces of two different persons under the same pose (e.g., frontal poses) are likely more similar than those extracted from faces of the same person under different poses (e.g., frontal and profile poses).

%\paragraph{Discriminative Face Features}
Deep convolutional neural networks (CNNs) have demonstrated significant performance improvements on recognition tasks, e.g., image classification~\cite{krizhevsky2012imagenet}.
The features extracted from the activation of a pre-trained CNN have been shown to be effective for generic visual recognition tasks~\cite{donahue2013decaf}.
In particular, CNN-based features have shown impressive performance on face recognition and verification tasks~\cite{sun2014deep,sun2014deep1,schroff2015facenet,hu2014discriminative}.
These models are often trained using large-scale face recognition datasets in a fully supervised manner and then serve as feature extractors for unseen face images.
However, these models may not achieve good performance in unconstrained videos as the visual domains of the training and test sets may be significantly different.

In this paper, we address this domain shift by adapting a pre-trained CNN to the \textit{specific} videos.
Due to the lack of manual annotations of target identities, we collect a large number of training samples of faces by exploiting contextual constraints of tracklets in the video.
With these automatically discovered training samples, we adapt the pre-trained CNN so that the Euclidean distance between the embedded features reflects the semantic distance between face images.
We incorporate these discriminative features into a hierarchical agglomerative clustering algorithm to link tracklets across multiple shots into trajectories.
We analyze the contribution of each component in the proposed algorithm and demonstrate the effectiveness of the learned features to identify characters in 10 long TV sitcom episodes and singers in 8 challenging music videos.
We further apply our adaptive feature learning approach to other objects (e.g., pedestrians) and show competitive performance on pedestrian tracking across cameras.
The preliminary results have been presented in \cite{shun2016tracking}.

%\paragraph{Our contributions, The significance of results}
We make the following contributions in this work:
\begin{compactitem}
\item Unlike existing work that uses linear metric learning on hand-crafted features, we account for large appearance variations of faces in videos by learning video-specific features with the deep contrastive and triplet-based metric learning on automatically discovered samples through contextual constraints.
\item We propose an improved triplet loss function which helps the learned model simultaneously pulls positive pairs closer and pushes away negative samples from the positive pairs.
\item In contrast to prior work that often uses face tracks with false positives manually removed, we take raw video as the input and perform detection, tracking, clustering, and feature adaptation in a fully automatic way.
\item We develop a new dataset with 8 music videos from YouTube containing annotations of 3,845 face tracklets and 117,598 face detections.
This benchmark dataset is challenging (with frequent shot changes, large appearance variations, and rapid camera motion) and crucial for
evaluating multi-face tracking algorithms in unconstrained environments.
\item We demonstrate the proposed adaptive feature learning approach can be extended to tracking other objects, and present empirical results on pedestrian tracking across cameras using the DukeMTMC dataset~\cite{ristani2016performance}.
\end{compactitem}

% ================================================
% Related Work
% ================================================
\vspace{-3mm}
\section{Related Work and Problem Context}
\vspace{-1mm}
\label{sec:related}

\vspace{1mm}
\noindent \textbf{Multi-target tracking.}
In recent years, numerous multi-target tracking methods have been developed
by applying a pre-learned object detector to locate instances in every frame, and determine the trajectories by solving
a \emph{data association} problem~\cite{brendel2011multiobject, collins2012multitarget,yang2012multi,zhang2008global,zhao2012tracking}.
A plethora of global optimization methods have been developed for association based on
the Viterbi decoding scheme~\cite{andriluka2008people}, Hungarian algorithm~\cite{stauffer2003estimating, perera2006multi, kaucic2005unified}, quadratic Boolean programming~\cite{leibe2007coupled}, maximum weight-independent sets~\cite{brendel2011multiobject}, linear programming~\cite{jiang2007linear, berclaz2011multiple}, energy minimization~\cite{Andri:EnergyMini:2011, andriyenko:discrete:2012}, and min-cost network flow~\cite{zhang2008global}.
Some methods tackle the data association problem with a hierarchical association framework.
For example, Xing et al. \cite{xing2009multi} propose a two-stage association method to combine local and global tracklets association to track multiple targets.
Huang et al. \cite{HuangC:HierAsso:2008} propose a three-level association approach by
first linking the detections from consecutive frames into short tracklets at the bottom level and then applying iterative Hungarian algorithm and an EM algorithm at higher levels.
Yang et al. \cite{yang2012online} extend the three-level work in~\cite{HuangC:HierAsso:2008} through learning an online discriminative appearance model.

\begin{figure*}[t]
\setlength{\abovecaptionskip}{0.cm}
\setlength{\belowcaptionskip}{-3mm}
\centering
\includegraphics[width=.8\textwidth]{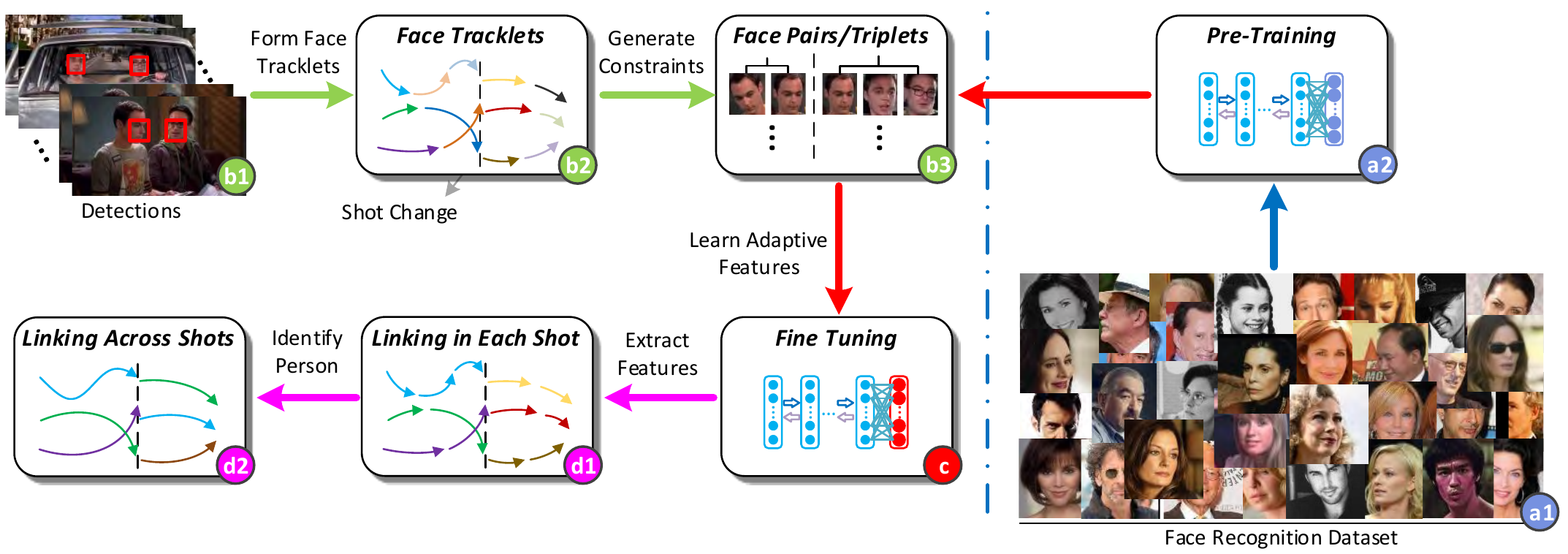}
\caption{
\textbf{Algorithm pipeline}. Our multi-face tracking algorithm has four main steps:
(a) Pre-training a CNN on a large-scale face recognition dataset to learn identity-preserving features,
(b) Generating face pairs or face triplets from the tracklets in a specific video with the proposed spatio-temporal constraints and contextual constraints,
(c) Adapting the pre-trained CNN to learn video-specific features from the automatically generated training samples, and
(d) Linking tracklets within each shot and then across shots to form the face trajectories.
}
\label{fig:Outline}
\vspace{-3mm}
\end{figure*}

Data association can also be formulated as a linear assignment problem.
Existing algorithms typically integrate appearance and motion cues into an affinity model to infer and link detections (or tracklets) into trajectories~\cite{brendel2011multiobject,zhang2008global,andriyenko:discrete:2012,huang2006robust,liyuan2007cascadePF}.
However, these MTT methods do not perform well in unconstrained videos where abrupt changes across different shots occur, and the assumptions of smooth appearance change no longer hold.

%\paragraph{Features in conventional multi-target tracking}
To identify targets across shots, discriminative appearance features are required to discern targets in various circumstances.
Most existing multi-target tracking methods \cite{zhang2008global,ben2011tracking,HuangC:HierAsso:2008,li2009learning,wu2013simultaneous} use color histograms as features and Bhattacharyya distance or correlation coefficient as affinity measures.
Several mehtods~\cite{andriyenko:discrete:2012,Andri:EnergyMini:2011,roth2012robust,wang2014tracklet,kuo2011does} use hand-crafted features, e.g.,
Haar-like~\cite{viola2001rapid}, SIFT~\cite{fulkerson2008localizing,lowe2004distinctive}, HOG~\cite{dalal2005histograms} features,
or combination~\cite{roth2012robust,wang2014tracklet,kuo2011does}.
For robustness, some approaches~\cite{zhang2015multi,kuo2010multi,yang2012online,yang2012multi} adaptively select the most discriminative features for a specific video~\cite{breitenstein2009robust,collins2005online,grabner2006line}.
However, all these hand-crafted feature representations are not tailored for faces, and thus are less effective at handling the large appearance variations in unconstrained scenarios.

%\subsection{Visual Constraints in Multi-target Tracking}
\vspace{1mm}
\noindent \textbf{Visual constraints in multi-target tracking.}
Several approaches \cite{cinbis2011unsupervised,tapaswi2014total,wang2014tracklet,wu2013simultaneous,wu2013constrained} exploit visual constraints from videos for improving tracking performance.
These visual constraints are often derived from the spatio-temporal relationships among the extracted tracklets~\cite{cinbis2011unsupervised,wu2013constrained}.
Two types of constraints are commonly used:
1) all samples in the same tracklet represent the same object and
2) a pair of tracklets in the same frame indicates that two different objects are present.
Prior work either uses these constraints implicitly for learning a cast-specific metric \cite{cinbis2011unsupervised,wang2014tracklet}; or explicitly for linking cluster or tracklet ~\cite{wu2013constrained,wu2013simultaneous}.

Numerous cues from contextual constraints have also been used for tracking, e.g., clothing~\cite{el2010face}, script~\cite{bauml2013semi,sivic2009you,Everingham06a}, speech~\cite{paul2014conditional}, gender~\cite{zhou2015multi}, video editing style~\cite{tapaswi2014total}, clustering prior~\cite{tang2015face}, and dynamic clustering constraints~\cite{zhang2016joint}.
For examples, the methods in \cite{ramanan2007leveraging,tapaswi2012knock,anguelov2007contextual} incorporate clothing appearance for improving person identification accuracy.
Lin et al.~\cite{lin2010joint} present a probabilistic context model to jointly tag people across multiple domains of people, events, and locations.
The work~\cite{Everingham06a,anguelov2007contextual} exploits speaker analysis to improve face labeling.

In this work, we exploit visual constraints generated in a way similar to~\cite{wu2013constrained,tapaswi2014total,xiao2014weighted,zhang2016joint}.
The proposed algorithm different from previous methods in three aspects.
First, existing approaches often rely on hand-crafted features and learn a linear transformation over the extracted features, which may not be effective in modeling large appearance variation of faces.
We learn discriminative a face feature representation specific to a video by adapting all layers in a deep neural network.
Second, previous work often uses face tracks with false positives removed manually~\cite{wu2013simultaneous,cinbis2011unsupervised,wu2013constrained,tapaswi2014total,xiao2014weighted}.
In contrast, our approach takes a raw video as the input and perform detection, tracking, clustering, and feature adaptation without these pre-processing steps.
Third, unlike the work in~\cite{zhang2016joint} that discovers negative pairs by thresholding based on feature distances, we discover negative pairs by transitively propagating the relationship among co-occurred tracklets using the proposed contextual constraints.

\vspace{1mm}
\noindent \textbf{CNN-based representation learning.}
Recent face recognition and verification methods focus on learning identity-preserving feature representations from deep neural networks.
While the models may differ, these CNN-based face representations (e.g., DeepID~\cite{sun2014deep}, DeepFace~\cite{taigman2014deepface}, FaceNet~\cite{schroff2015facenet}, VGG-Face~\cite{Parkhi2015VGGFace}) are learned by training CNNs using large-scale datasets in a fully supervised manner.
These CNNs then operate as feature extractors for face recognition, identification, and face clustering.
In this work, we also use a CNN to learn identity-preserving features from a face recognition dataset.
The main difference lies in that we further adapt the pre-trained representation to a specific video, thereby further improve the specificity of the model and enhance discriminative strength.
In addition, we introduce a symmetric triplet-based loss function and demonstrate its effectiveness over the commonly used contrastive loss and triplet loss.

\vspace{1mm}
\noindent \textbf{Long-term object tracking.}
The goal of long-term object tracking~\cite{kalal2012tracking,pernici2012facehugger} is to
locate a specific target over time even when the target leaves and re-enters the scene.
These trackers perform well on various types of targets such as cars and faces.
However, online trackers are designed to handle scenes recorded by a stationary or slow-moving camera and thus not effective in tracking faces in unconstrained videos for two reasons.
First, these trackers are prone to drift due to online model update with noisy examples.
Second, hand-crafted features are not sufficiently discriminative to re-identify faces across shots.
We tackle the first issue by processing the video offline, i.e., apply a face detector in every frame and associate all tracklets in the video.
For the second issue, we learn adaptive discriminative representation to account for large appearance variations of faces across shots or scenes.

\vspace{-3mm}
\section{Algorithmic Overview}
\vspace{-1mm}
\label{sec_overview}

Our goal is to track multiple faces across shots in an unconstrained video while maintaining identities of the persons of interest.
To achieve this, we learn discriminative features that are adapted to the appearance variations in the specific videos.
We then use a hierarchical clustering algorithm to link tracklets across shots into long trajectories.
The main steps of the proposed algorithm are summarized below and in Figure~\ref{fig:Outline}.
\vspace{1mm}
\begin{compactenum}[(a)]
\item \textbf{Pre-training}: We pre-train a CNN model based on the AlexNet~\cite{krizhevsky2012imagenet}
using an external face dataset to learn identity-preserving features (Section~\ref{sec_pretrain}).
\item \textbf{Discovering training samples}:
We detect shot changes and divide a video into non-overlapping shots.
Within each shot, we apply a face detector and link adjacent detections into short tracklets.
We discover a large collection of training samples (in pairs or triplets) from tracklets based on the spatio-temporal and contextual constraints (Section~\ref{sec_constraints}).
\item \textbf{Learning video-specific features}:
We adapt the pre-trained CNN model using the automatically discovered training samples to account for large appearance changes of faces pertaining to a specific video (Section~\ref{sec_metricLearning}).
We present an improved triplet loss to enhance the discriminative ability of the learned features.
\item \textbf{Linking tracklets}:
Within each shot, we use a conventional multi-face tracking method to link tracklets into short trajectories.
We use a hierarchical clustering algorithm to link trajectories across shots.
Finally, we assign the tracklets in each cluster with the same identity (Section~\ref{sec_MTT}).
\end{compactenum}

%==========================================================================

\vspace{-3mm}
\section{Learning Discriminative Features}
\vspace{-1mm}
\label{sec_adaptive}

In this section, we present the algorithmic details for learning video-specific features.
After describing how the generic face features are obtained from the pre-training step, we introduce the process of discovering training examples and learning discriminative features using the proposed symmetric triplet loss function.

\vspace{-3mm}
\subsection{Supervised Pre-training}\label{sec_pretrain}
\vspace{-1mm}

%\noindent\textbf{Supervised Pre-training.}
We learn identity-preserving features by pre-training a deep neural network on a large-scale face recognition dataset.
Based on the AlexNet architecture~\cite{krizhevsky2012imagenet}, we
replace the output layer with $K$ nodes where each node corresponds to a specific person.
We train the network on an external CASIA-WebFace dataset~\cite{yi2014learning} (494,414 images of 10,575 subjects) for face recognition in a fully supervised manner.
We select $9,427$ persons, $80\%$ of the images (431,300 images) for training and the remaining $20\%$ (47,140 images) as the validation set.
Each face image is normalized to $227 \times 227 \times 3$ pixels.
We use stochastic gradient descent with an initial learning rate of 0.01 that decreases by a factor of 10 for every 20,000 iterations using the Caffe~\cite{jia2014caffe} toolbox.

\vspace{-3mm}
\subsection{Discovering Training Samples}
\vspace{-1mm}
\label{sec_constraints}

\begin{figure}[t]
\setlength{\abovecaptionskip}{0.cm}
\setlength{\belowcaptionskip}{-0.4cm}
\centering
\includegraphics[width=.5\textwidth]{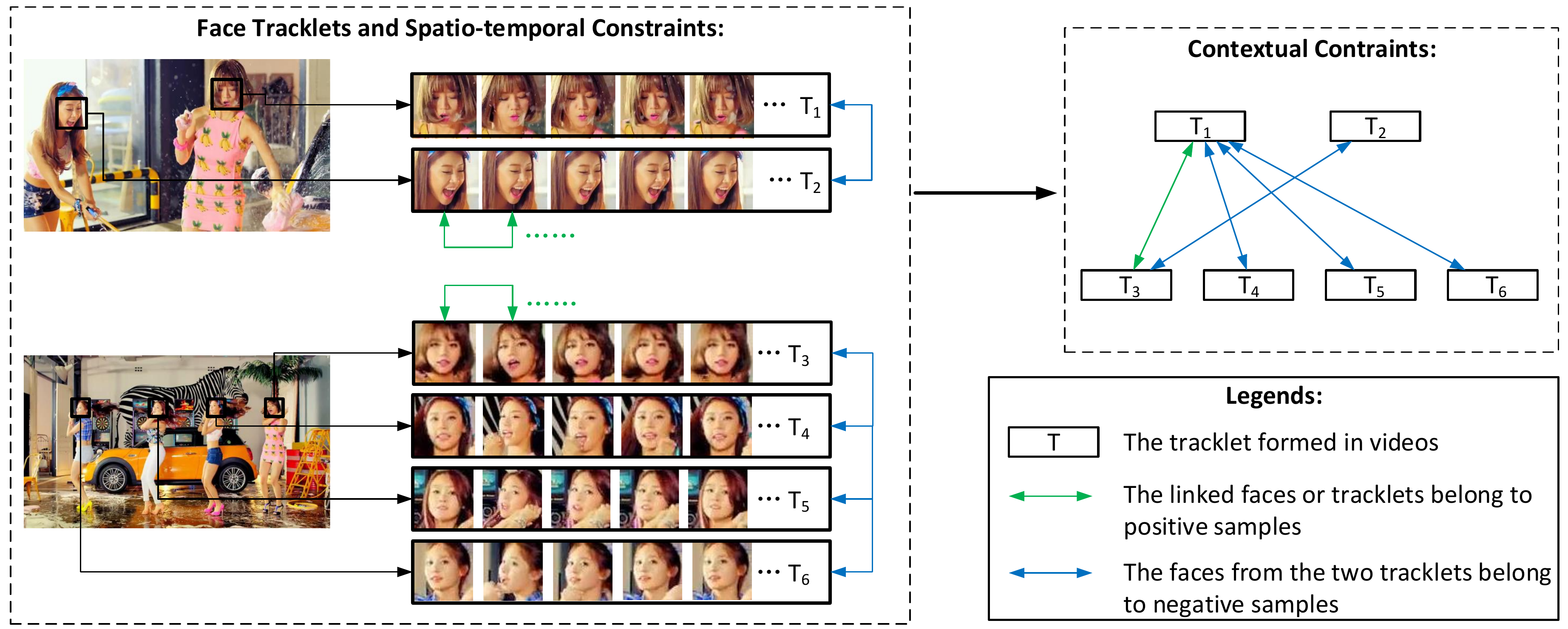}
\vspace{-1mm}
\caption{
% JB: The texts in the figure are way too small.
\textbf{Contextual constraints generation}. Here, we label the faces in $\vT_1$ and $\vT_3$ as the same identity given the sufficiently high similarity between the contextual features of $\vT_1$ and $\vT_3$. With this additional constraint, we can propagate the constraints transitively and derive that the faces from $\vT_1$ and $\vT_4$ (or $\vT_5$, $\vT_6$) are in fact belong to different identities, and the faces from $\vT_3$ and $\vT_2$ are from different people.
}
\label{fig:Contextual}
\vspace{-3mm}
\end{figure}

\vspace{1mm}
\noindent\textbf{Shot detection and tracklets linking.}
We first use a shot change detection method to divide each input video into non-overlapping shots.\footnote{http://sourceforge.net/projects/shot-change/}
Next, we use a face detector~\cite{mathias2014face} to locate faces in each frame.
Given the face detections for each frame, we use a two-threshold strategy~\cite{HuangC:HierAsso:2008} to generate tracklets within each shot by linking the detected faces in adjacent frames based on similarities in appearances, positions, and scales.
Note that the two-threshold strategy for linking detections could be replaced by more sophisticated methods, e.g., tracking using particle filters \cite{huang2006robust,breitenstein2009robust}.
All tracklets shorter than five frames are discarded.
The extracted face tracklets are formed in a conservative manner with limited temporal spans up to the length of each shot.

\vspace{1mm}
\noindent\textbf{Spatio-temporal constraints.}
Existing methods typically exploit spatio-temporal constraints from tracklets to generate training samples from the video.
Given a set of tracklets, we can discover a large collection of positive and negative training sample pairs belonging to the same/different persons: (1) all pairs of faces in one tracklet are from one person and (2) two face tracklets that appear in the same frame contain faces of different persons.

Let $\vT^i = \{\vx^{i}_1, \ldots,\vx^{i}_{n_i}\}$ denote the $i$-{th} face tracklet of length $n_i$.
We generate a set of positive pairs $\vP^+$ by collecting all within-tracklet face pairs:
\begin{align}
\vP^+=\{(\vx^{i}_k,\vx^{i}_l)\}, ~\mathrm{s.t.} ~\forall k,l=1 , \ldots,n_i, ~k \neq l.
\end{align}
Similarly, if tracklets $\vT^i$ and $\vT^j$ overlap in some frames, we can generate a set of negative pairs $\vN^-$ by collecting all between-tracklet face pairs:
\begin{align}
\vN^-=\{(\vx^{i}_k,\vx^{j}_l)\}, ~\mathrm{s.t.} ~\forall k=1, \ldots,n_i, ~\forall l=1, \ldots,n_j.
\end{align}

\vspace{1mm}
\noindent\textbf{Contextual constraints.}
With the spatio-temporal constraints, we can obtain a large number of face pairs without manual labeling.
These training pairs, however, may have some biases.
First, the positive (within-tracklet) pairs occur close in time (e.g., only several frames apart in one shot), which means that the positive face pairs often have small appearance variations.
Second, the negative pairs are all generated from tracklets that co-occur in the \emph{same} shot.
Consequently, we are not able to train the model so that it can distinguish or link faces \emph{across} shots (as we do not have training samples for these cases).

\begin{figure}[t]
\setlength{\abovecaptionskip}{1mm}
\setlength{\belowcaptionskip}{-0.2cm}
\centering
\begin{tabular}{{c}{c}}
\hspace{0mm}\includegraphics[width=3.5cm]{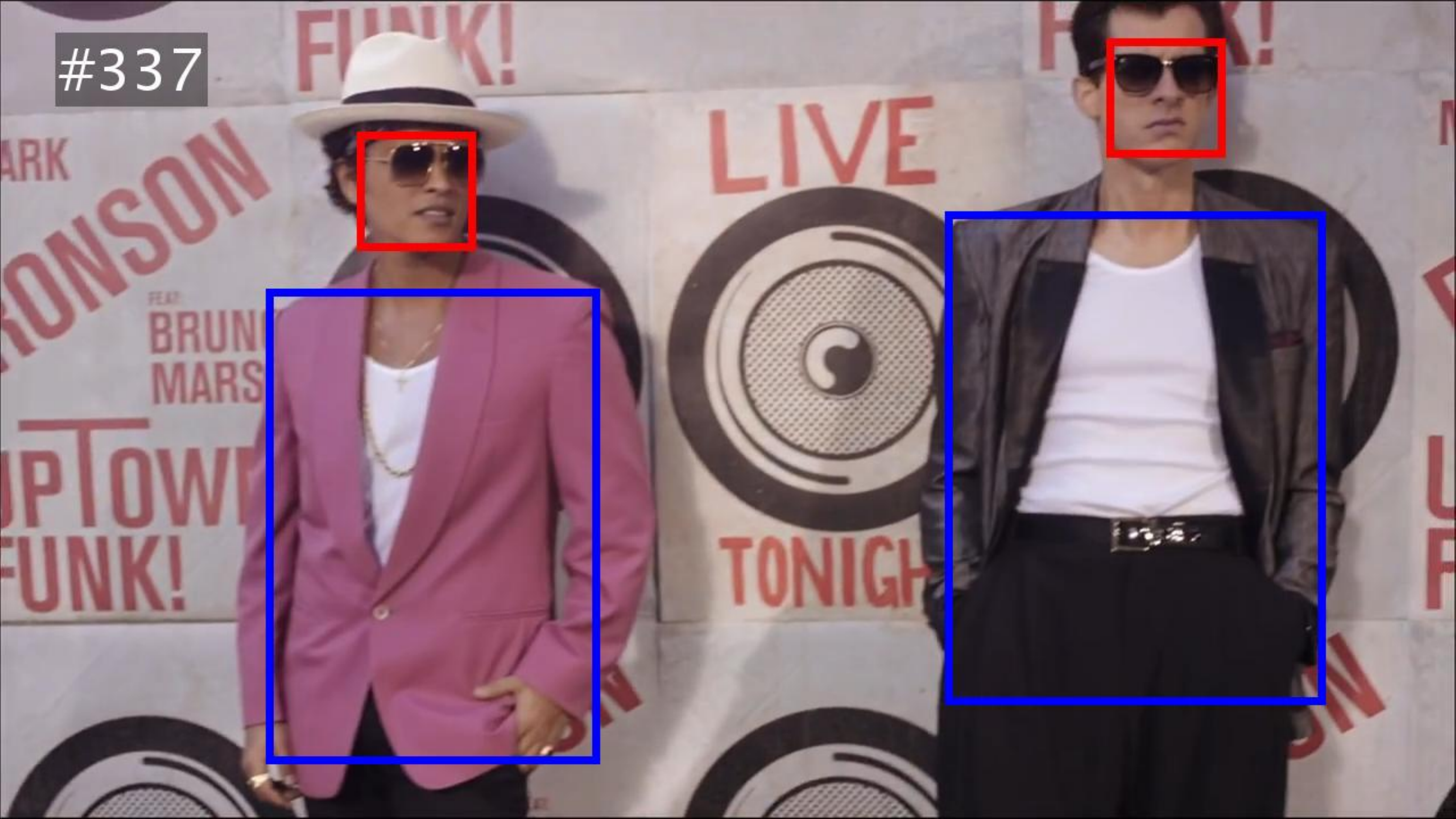} &
\hspace{1.5mm}\includegraphics[width=3.5cm]{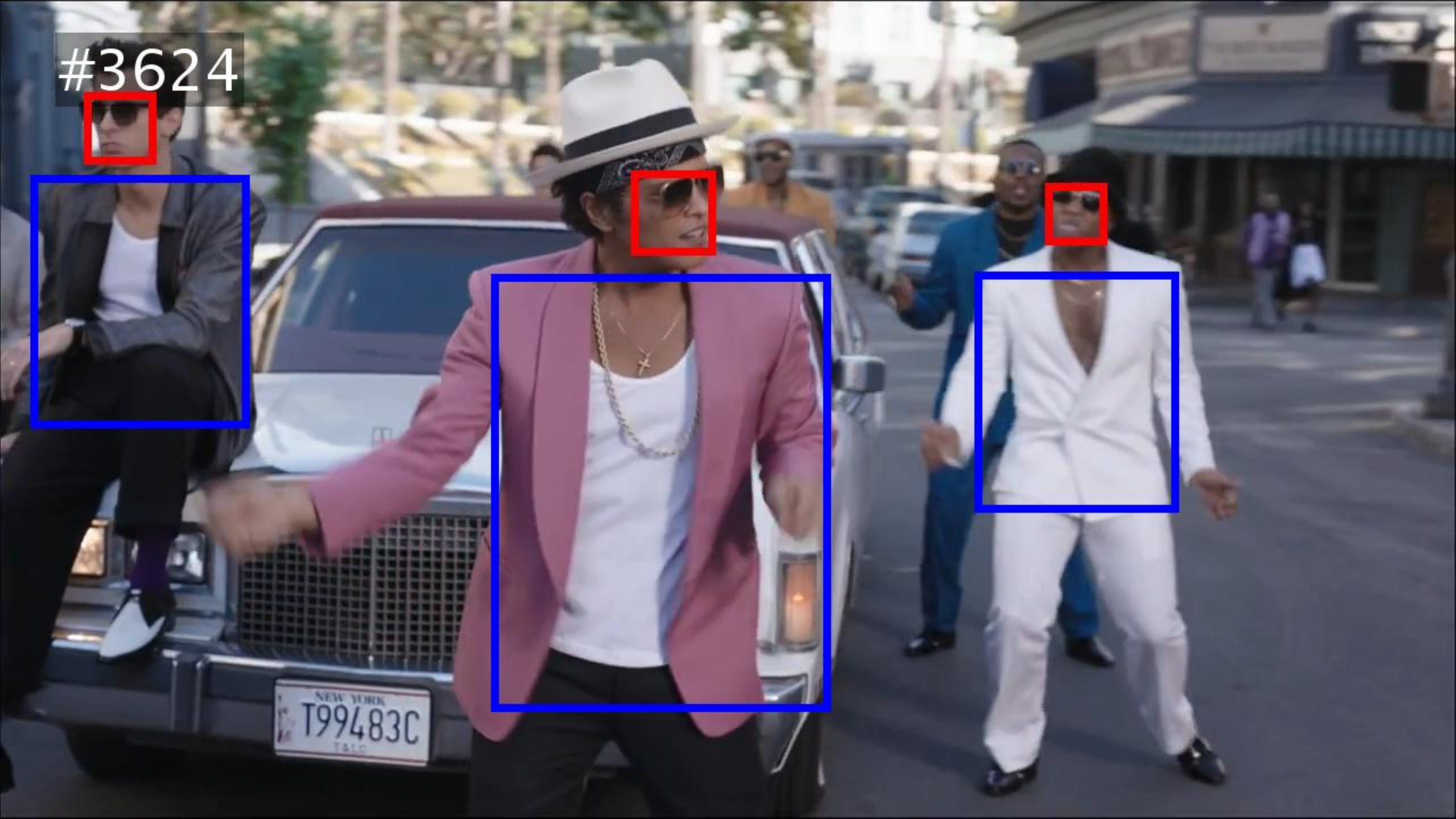} \\
\end{tabular}
\caption{
\textbf{Face detections and clothing regions}.
% The red boxes are the face detections. The blue rectangles denote the clothing regions.
% where we search for the clothing regions.
}
\label{fig:Clothing}
\vspace{-3mm}
\end{figure}

To address these problems, we mine additional positive and negative face pairs for learning our video-specific features.
The idea is to exploit contextual information beyond facial regions for identifying the person across shots.
Specifically, we identify the clothing region following~\cite{du2016face} and extract features using the AlexNet model.
Given the $i$-th face detection $\vx_i$ in one frame, we locate the torso region $s_i$ by using a probabilistic mask $I(p\in \vs_i|\vx_i)$ (See~\ref{fig:Clothing}), where $p$ is a pixel in the current frame.
The mask is learned from the statistics of body part's spatial relationship on the Human in 3D dataset~\cite{bourdev2009poselets}.
We then concatenate the face features and the clothing feature together.
We then apply the HAC algorithm (see Section~\ref{sec:linking}) to group tracklets and label the grouped tracklets with high confidence as positive pairs.
These grouped tracklets generally contain faces with the similar clothing in the different shots or scenes and thus provide additional positive tracklet pairs.
In addition, by leveraging the additional positive pairs, we can discover more negative pairs by \emph{transitively} propagating the relationship among tracklets.
For example, suppose we know tracklets A and B represent different persons, an additional positive constraint on tracklets A and C automatically implies that tracklet B and C are different persons.

Figure~\ref{fig:Contextual} illustrates the generation process of contextual constraints .
Here, the tracklets $\vT_1$ and $\vT_2$ co-occur in one shot and the tracklets $\vT_3$, $\vT_4$, $\vT_5$ and $\vT_6$ co-occur in another shot.
Using only spatio-temporal constraints, we are not able to obtain training samples from different shots.
As a result, the tracklets $\vT_1$ and $\vT_4$ may be incorrectly identified as the same person.
However, from the contextual cues, we may be able to identify that the tracklets $\vT_1$ and $\vT_3$ are the same person.
Using this additional positive constraints, we can automatically generate additional negative constraints, e.g., $\vT_1$ is a different person from $\vT_4$, $\vT_5$ and $\vT_6$.

\vspace{-3mm}
\subsection{Learning Adaptive Discriminative Features}
\vspace{-1mm}
\label{sec_metricLearning}

With the discovered training pairs from applying both contextual and spatio-temporal constraints, we optimize the embedding function $\vf(\cdot)$ such that the distance $D(\vf(\vx_1), \vf(\vx_2))$ in the embedding space reflects the semantic similarity of two face images $\vx_1$ and $\vx_2$:
\begin{align}\label{eq_Euclid}
D(\vf(\vx_1), \vf(\vx_2)) = \|\vf(\vx_{1})-\vf(\vx_{2})\|_2^2.
\end{align}
We set the feature dimension of $\vf(\cdot)$ as 64 in all of our experiments.
We first describe two commonly used loss functions for optimizing the embedding space: (1) contrastive loss and (2) triplet loss, and then present
a symmetric triplet loss function for feature learning.

\begin{figure}[t!]
\setlength{\abovecaptionskip}{0.cm}
\setlength{\belowcaptionskip}{-1mm}
\centering
%\vspace{-1mm}
\includegraphics[width=0.4\textwidth]{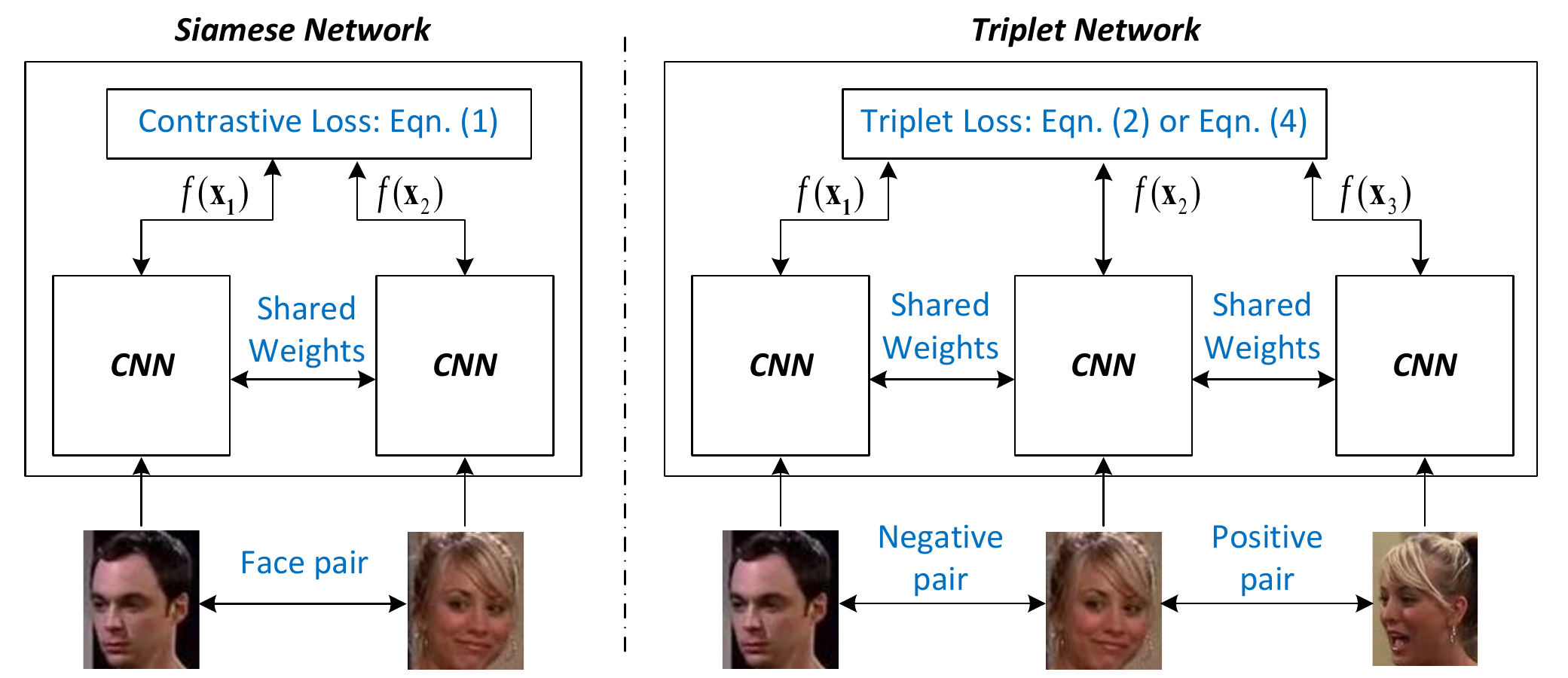}
\caption{
\textbf{Siamese vs. Triplet network}. Illustration of the Siamese network (left) with pairs as inputs and the triplet network (right) with triplets as inputs for learning discriminative features adaptively. The Siamese network consists of two CNNs and uses a contrastive loss. The Triplet network consists of three CNNs and uses a triplet loss. The CNNs in each network share the same architectures and parameters and are initialized with parameters of the CNN pre-trained on the large-scale face recognition dataset.
}
\label{fig:Siamese}
\vspace{-3mm}
\end{figure}

\vspace{1mm}
\noindent \textbf{Contrastive loss.}
The Siamese network~\cite{chopra2005learning,hadsell2006dimensionality} consists of two identical CNNs with the shared architecture and parameters as shown in Figure~\ref{fig:Siamese}.
Minimizing the contrastive loss function encourages small distance of two images of the same person and large distance otherwise.
Denote $(\vx_{1}, \vx_{2})\in\{\vP^+,\vN^-\}$ as a pair of training images generated with the spatio-temporal constraints.
Similar to \cite{chopra2005learning,hadsell2006dimensionality}, the contrastive loss function is:
\setlength{\abovedisplayskip}{3pt}
\begin{align} L_{p} =
\begin{cases}
\frac{1}{2} D(\vf(\vx_1),\vf(\vx_2)) & \quad \text{if } (\vx_{1},\vx_{2})\in\vP^+\\
\frac{1}{2} \max(0,\tau-D(\vf(\vx_1),\vf(\vx_2)) & \quad \text{if } (\vx_{1},\vx_{2})\in\vN^-\\
\end{cases},
\setlength{\belowdisplayskip}{3pt}
\end{align}
where $\tau$ ($\tau=1$ in all our experiments) is the margin.
Intuitively, if $\vx_{1}$ and $\vx_{2}$ are from the same person,
the loss is $\frac{1}{2}D(\vf(\vx_1),\vf(\vx_2))$ and
we aim to decrease $D(\vf(\vx_1),\vf(\vx_2))$.
Otherwise, we increase $D(\vf(\vx_1),\vf(\vx_2))$ until it is larger than the margin $\tau$.

\vspace{1mm}
\noindent \textbf{Triplet loss.}
The triplet-based network~\cite{schroff2015facenet} consists of three identical
CNNs with the shared architecture and parameters as shown in Figure~\ref{fig:Siamese}.
One triplet consists of two face images of the same person and one face image from another person.
We generate a set of triplets $\vS$ from two tracklets $\vT^i$ and $\vT^j$ belonging to different persons:
$\vS=\{(\vx^{i}_k,\vx^{i}_l,\vx^{j}_m)\}, ~\mathrm{s.t.}~\forall k,l=1, \ldots, n_i, ~k \neq l, ~\forall m=1, \ldots,n_j$.
Here we aim to ensure that the embedded distance of the positive pair $(\vx^{i}_k,\vx^{i}_l)$ is closer than that of the negative pair $(\vx^{i}_k,\vx^{j}_m)$ by a distance margin $\alpha$ ($\alpha=1$).
For one triplet, the triplet loss is of the form:
\setlength{\abovedisplayskip}{3pt}
\begin{align}\label{eq_TripLoss}
L_{t} = \frac{1}{2}\max\left(0,D(\vf(\vx^{i}_k),\vf(\vx^{i}_l))-D(\vf(\vx^{i}_k),\vf(\vx^{j}_m))+\alpha\right).
\end{align}

\begin{figure}[t]
\setlength{\abovecaptionskip}{1mm}
\setlength{\belowcaptionskip}{-0.2cm}
\centering
\begin{tabular}{{c}{c}}
\hspace{-5.5mm}\includegraphics[width=3.5cm]{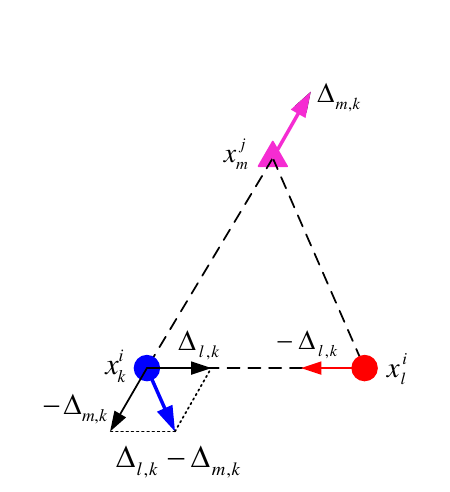} &
\hspace{-6.5mm}\includegraphics[width=3.5cm]{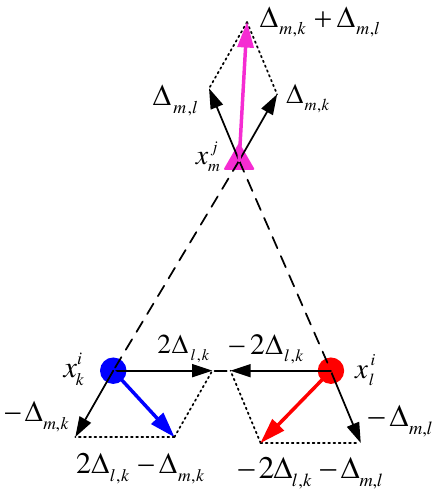} \\
\hspace{-5.5mm} (a) \scriptsize{Conventional triplet loss} &
\hspace{-6.5mm} (b) \scriptsize{SymTriplet loss} \\
\end{tabular}
\caption{
\textbf{Triplet loss vs. SymTriplet loss}. Illustration of the negative partial gradient direction to the triplet sample.
(a) the conventional triplet loss;
(b) the SymTriplet loss.
The triplet samples $\vx^i_k$, $\vx^i_l$ and $\vx^j_m$ are highlighted with blue, red and magenta colors, relatively.
The circles denote faces from the same person whereas the triangle denotes a different person.
The gradient directions are color-coded.
}
\label{fig:SymLoss}
\vspace{-3mm}
\end{figure}

\vspace{1mm}
\noindent \textbf{Symmetric triplet loss.}
The conventional triplet loss in (\ref{eq_TripLoss}), takes only two of the three distances into consideration: $D(\vf(\vx^{i}_k),\vf(\vx^{i}_l))$ and $D(\vf(\vx^{i}_k),\vf(\vx^{j}_m))$ although there are three distances between each pair.
We illustrate the problem of the conventional triplet loss by analyzing the gradients of the loss function.
We denote the difference vector between the triplet ($\vx^i_k$, $\vx^i_l$ and $\vx^j_m$):
\begin{equation}
\Delta_{l,k} \rm{=} \vf(\vx^i_l)\rm{-}\vf(\vx^i_k), \
\Delta_{m,k}\rm{=} \vf(\vx^j_m)\rm{-}\vf(\vx^i_k), \
\Delta_{m,l}\rm{=} \vf(\vx^j_m)\rm{-}\vf(\vx^i_l).
\end{equation}
For non-zero triplet loss in (\ref{eq_TripLoss}), we can compute the gradients as
\begin{footnotesize}
\setlength{\abovedisplayskip}{3pt}
\begin{equation}
\frac{\partial L_t}{\partial \vf(\vx^i_k)} \rm{=} -(\Delta_{l,k}\rm{-}\Delta_{m,k}), \
\frac{\partial L_t}{\partial \vf(\vx^i_l)} \rm{=} \Delta_{l,k}, \
\frac{\partial L_t}{\partial \vf(\vx^j_m)} \rm{=} -\Delta_{m,k}.
\end{equation}
\end{footnotesize}
Figure~\ref{fig:SymLoss}(a) shows the negative gradient directions for each sample.
There are two issues with the triplet loss in (\ref{eq_TripLoss}).
First, the loss function pushes the negative data point $\vx^j_m$ away from only one of the positive pair $\vx^i_k$ rather than both $\vx^i_k$ and $\vx^i_l$.
Second, the gradients on the positive pair $(\vx^i_k,\vx^i_l)$ are not symmetric with respect to the negative data $\vx^i_k$.

To address these issues, we propose a symmetric triplet loss function (SymTriplet) by considering all three distances as:
\begin{footnotesize}
\begin{align}
\label{eq_SymLoss}
L_s \rm{=} \max\left[0,~D(\vf(\vx^{i}_k),\vf(\vx^{i}_l))\rm{-}\frac{1}{2}(D(\vf(\vx^{i}_k),\vf(\vx^{j}_m))
\rm{+}D(\vf(\vx^{i}_l),\vf(\vx^{j}_m)))\rm{+}\alpha \right],
\end{align}
\end{footnotesize}
\hspace{-1mm} where $\alpha$ is the distance margin.
The gradients induced from the proposed SymTriplet loss are
\begin{equation}
\frac{\partial L_s}{\partial \vf(\vx^i_k)} \rm{=} \rm{-}(2\Delta_{l,k}\rm{-} \Delta_{m,k}),
\frac{\partial L_s}{\partial \vf(\vx^i_l)} \rm{=} 2\Delta_{l,k} \rm{+} \Delta_{m,l},
\frac{\partial L_s}{\partial \vf(\vx^j_m)} \rm{=} \rm{-}(\Delta_{m,k} \rm{+} \Delta_{m,l}).
\end{equation}
Figure~\ref{fig:SymLoss}(b) shows the negative gradient directions.
The proposed SymTriplet loss directly optimizes the embedding space such that the positive pair are pulled closer to each other and the negative sample ($\vx^j_m$) is pulled away from the two positive samples ($\vx^i_k,\vx^i_l$).
This property allows us to improve the discriminative strength of the learned features.

\vspace{-3mm}
\subsection{Training Algorithm}
\vspace{-1mm}
We train the triplet network model with the SymTriplet loss function
and stochastic gradient decent method with momentum.
We compute the derivatives of (\ref{eq_SymLoss}) as follows:
\begin{align}\label{eq_derivW}
\frac{\partial L_s}{\partial \vW}=
\begin{cases}
\frac{\partial\widetilde{L_s}}{\partial \vW} & L_s>0, \\
0 & L_s=0,
\end{cases}
\end{align}
where
\begin{footnotesize}
\begin{align} \label{eq_derivW2}
\frac{\partial\widetilde{L_s}}{\partial \vW}=
&2(\vf(\vx^i_k)-\vf(\vx^i_l))\frac{\partial \vf(\vx^i_k) - \partial \vf(\vx^i_l)}{\partial \vW}
-(\vf(\vx^i_k)-\vf(\vx^j_m))\frac{\partial \vf(\vx^i_k) - \partial \vf(\vx^j_m)}{\partial \vW} \nonumber \\
&-(\vf(\vx^i_l)-\vf(\vx^j_m))\frac{\partial \vf(\vx^i_l) - \partial \vf(\vx^j_m)}{\partial \vW}.
\end{align}
\end{footnotesize}
\hspace{-2mm}We can compute the gradients from each input triplet examples given the values of $\vf(\vx^i_k)$, $\vf(\vx^i_l)$, $\vf(\vx^j_m)$ and $\frac{\partial \vf(\vx^i_k)}{\partial \vW}$, $\frac{\partial \vf(\vx^i_l)}{\partial \vW}$, $\frac{\partial \vf(\vx^j_m)}{\partial \vW}$, which can be obtained using the standard forward and backward propagations separately for each image in the triplet examples.
We summarize the main training steps in Algorithm~\ref{alg_Training}.

\begin{algorithm}[t!]
\caption{Stochastic gradient descent with SymTriplet loss}\label{alg_Training}
\begin{algorithmic}[1]
\State{\textbf{Input:}} Training samples $\{(\vx^i_k, \vx^i_l, \vx^j_m)\}$.

\State{\textbf{Output:}} Network parameters $\vW$,

\For {$t = 1 \to $ Max number of iterations}

$\frac{\partial L_s}{\partial \vW} = 0$

\For { all training triplet samples $(\vx^i_k, \vx^i_l, \vx^j_m)$}
\State Compute $\vf(\vx^i_k)$, $\vf(\vx^i_l)$ and $\vf(\vx^j_m)$ by forward propagation;
\State Compute $\frac{\partial \vf(\vx^i_k)}{\partial \vW}$, $\frac{\partial \vf(\vx^i_l)}{\partial \vW}$ and $\frac{\partial \vf(\vx^j_m)}{\partial \vW}$ by back propagation;
\State Compute $\frac{\partial L_s}{\partial \vW}$ according to (\ref{eq_derivW}) and (\ref{eq_derivW2}).
\EndFor
\State Update the parameters $\vW^t \leftarrow \vW^{t-1}-\lambda_t\frac{\partial L_s}{\partial \vW}$
\EndFor
\end{algorithmic}
\end{algorithm}

%\section{Multi-face Tracking via Hierarchical Tracklet Linking}

%====================================================================
\vspace{-3mm}
\section{Multi-face Tracking via Tracklet Linking}
\vspace{-1mm}
\label{sec_MTT}

We use a two-step procedure to link face tracklets generated in Section~\ref{sec_constraints}:
(1) linking the face tracklets within each shot into shot-level tracklets, and
(2) merging shot-level tracklets across multiple shots into trajectories.

\vspace{-3mm}
\subsection{Linking Tracklets Within Each Shot}
\vspace{-1mm}
We use a typical multi-object tracking framework for linking tracklets within each shot.
First, we extract features from each detected face using the learned deep network.
We measure the linking probabilities between two tracklets using temporal, kinematic and appearance information.
Then, we use the Hungarian algorithm to determine a globally optimal label assignment~\cite{HuangC:HierAsso:2008,xing2009multi} and link tracklets with the same label are linked into shot-level tracklets.
% JB: Please make sure you do so.
%The full algorithmic details are described in the supplementary material.

\vspace{-3mm}
\subsection{Linking Tracklets Across Shots}
\label{sec:linking}
\vspace{-1mm}

To link tracklets across multiple shots, we apply a bottom-up hierarchical agglomerative clustering (HAC) algorithm with a stopping threshold and learned appearance features as follows:
\begin{compactenum}[(a)]
\item Given $N$ tracklets in all shots $\Gamma=\{\vT^1,\vT^2, \ldots,\vT^N\}$, we start with treating each tracklet as a singleton cluster.
\item We evaluate all pairwise distances between two tracklets and then use the mean distance metric as the similarity measure.
Given $\vT^i = \{\vx^{i}_1, \ldots,\vx^{i}_{n_i}\}$ and $\vT^j = \{\vx^{j}_1, \ldots,\vx^{j}_{n_j}\}$, the distance $D^{ij}$ is defined as:
\begin{equation}
D^{ij} = \frac{1}{n_i}\frac{1}{n_j}\sum_{k=1}^{n_i}\sum_{l=1}^{n_j}\|\vf(\vx^{i}_k)-\vf(\vx^{j}_l)\|_2^2\,,
\end{equation}
where $\vx^{i}_k$ denotes the $k$-th face detection in the $i$-th tracklet, and $\vf(\vx^{i}_k)$ denotes the feature extracted from the embedding layer in the Triplet network.
\item For trackles which have overlapped frames, we set the corresponding distance as infinity.
\item We first determine the pair of clusters that has the shortest distance and merge into a new cluster.
We update all distances from the new cluster to all other clusters.
For those clusters which have overlapped frames with the new cluster, the corresponding distances to the new cluster are set to infinity.
\item Repeat (d) until the shortest distance is larger than a threshold $\theta$.
\end{compactenum}

We remove clusters containing less than 4 tracklets and less than 50 frames.
The tracklets in each cluster are labeled with the same identity to form trajectories.

%============================================================

\vspace{-3mm}
\section{Experimental Results}\label{sec_exp}
\vspace{-1mm}

We first describe the implementation details, datasets, and evaluation metrics.
Next, we present the evaluation results of the proposed algorithm against the state-of-the-art methods.
More experimental results and videos are available in the supplementary material 
at  \url{http://vllab1.ucmerced.edu/~szhang/FaceTracking/}.
The source code and annotate datasets will be made available to the public.

\vspace{-3mm}
\subsection{Implementation Details}
\vspace{-1mm}

\vspace{1mm}
\noindent \textbf{CNN fine-tuning.} We adapt the pre-trained CNN with the proposed SymTriplet loss.
For feature embedding, we replace the classification layer in the pre-trained network with 64 output nodes.
We use stochastic gradient descent with the momentum term set to 0.9.
For the network training, we set a fixed learning rate to 0.00001 for finetuning and a weight decay of 0.0001.
We use a mini-batch size 128 and train the network for 2,000 epochs.

\vspace{1mm}
\noindent \textbf{Linking tracklets.} For determining the threshold $\theta$ in Section~\ref{sec:linking}, we perform parameter sweeping (with an interval of 0.1) using cross-validation.
For features trained from the Siamese network, we empirically set the threshold of the HAC algorithm as $\theta=0.4$.
For features trained from the triplet network, we use $\theta = 5$ for both the triplet and SymTriplet loss functions.

\vspace{-3mm}
\subsection{Datasets}
\vspace{-1mm}
We evaluate the proposed algorithm on three types of videos containing multiple persons:
\begin{enumerate}
\item Videos in a laboratory setting: Frontal~\cite{wu2013simultaneous}
\item TV sitcoms: The Big Bang Theory (BBT)~\cite{wu2013simultaneous,wu2013constrained} and the Buffy the Vampire Slayer (BUFFY)~\cite{sivic2009you,Everingham06a,du2016face} datasets
\item Ｍusic videos from YouTube
\end{enumerate}

\vspace{1mm}
\noindent\textbf{Frontal video.} \textsc{Frontal} is a short video in a constrained scene acquired indoors with a fixed camera.
Four persons facing the camera move around and occlude each other.

\vspace{1mm}
\noindent\textbf{BBT dataset.}
We select the first 7 episodes from Season 1 of the Big Bang Theory TV Sitcom (referred to as \textsc{BBT01-07}).
Each video is about 23 minutes long with the main cast of 5-13 people and is recorded mostly indoors.
The main difficulty lies in identifying faces of the same person from frequent changes of camera views and scenes, where
there are large appearance variations in viewing angle, pose, scale, and illumination.

\vspace{1mm}
\noindent\textbf{BUFFY dataset.}
The BUFFY dataset has been widely evaluated in the context of automatic face labeling~\cite{sivic2009you,Everingham06a,du2016face}.
The dataset contains three episodes (episode 2, 5 and 6) from Season 5 of the TV series Buffy the Vampire Slayer (referred to as \textsc{BUFFY02}, \textsc{BUFFY05}, and \textsc{BUFFY06}).
Each video is about 40 minutes long with the main cast of 13-19 people.
The illumination condition in this video dataset is more challenging than that in the BBT dataset as it contains many scenes with dim light.

\vspace{1mm}
\noindent\textbf{Music video dataset.}
We introduce a new dataset of 8 music videos from YouTube.
It is challenging to track multiple faces in these videos due to large variations caused by frequent shot/scene changes, large appearance variations, and rapid camera motion.
Three sequences (\textsc{T-ara}, \textsc{Westlife} and \textsc{Pussycat Dolls}) are recorded from live music performance with multiple cameras in different views.
The other sequences (\textsc{Bruno Mars}, \textsc{Apink}, \textsc{Hello Bubble}, \textsc{Darling} and \textsc{Girls Aloud}) are MTV videos.
Faces in these videos often undergo large appearance variations due to changes in pose, scale, makeup, illumination, camera motion, and occlusions.

\vspace{-3mm}
\subsection{Evaluation Metrics}\label{metric}
\vspace{-1mm}

We evaluate the proposed method in two main aspects.
First, to evaluate the effectiveness of the learned video-specific features, we use a bottom-up HAC algorithm to merge pairs of tracklets until all tracklets have been merged into the pre-defined number of clusters (i.e., the actual number of people in the video).
We measure the quality of clustering using the weighted purity:
\begin{align}
W = \frac{1}{M}\sum_{c}{m_c\cdot p_c},
\end{align}
where each cluster $c$ contains $m_c$ elements and its purity $p_c$ is measured as the fraction of the largest number of faces from the same person to $m_c$, and $M$ denotes the total number of faces in the video.

Second, we evaluate the method with the metrics commonly used in multi-target tracking~\cite{zhang2015multi},
including Recall, Precision, F1, FAF, IDS, Frag, MOTA, and MOTP.
% JB: Move the definition to Supplementary material
We list the definitions of these metrics in the supplementary material available at
\url{http://vllab1.ucmerced.edu/~szhang/FaceTracking/}.
The up and down arrows indicate whether higher or lower scores are better for each metric.

\vspace{-3mm}
\subsection{Evaluation on Features}
\vspace{-1mm}
We evaluate the proposed adaptive features against several alternatives summarized in Table~\ref{tab:feas_list}.

\begin{table*}[t]
\scriptsize
\centering
\caption{Summary of the evaluated features.}
\vspace{-1mm}
\begin{tabular}{l>{\centering}m{1.5cm}>{\centering}m{2.0cm}>{\centering}m{2.2cm}p{7.0cm}}
\toprule
Method Name & Feature Dimension & Architecture & Training Loss & Description \\ \midrule
HOG~\cite{dalal2005histograms} & 4,356 & - & - & A conventional hand-crafted feature \\
AlexNet~\cite{krizhevsky2012imagenet} & 4,096 & AlexNet & Softmax loss & A generic feature representation \\
Pre-trained & 4,096 & AlexNet & Softmax loss & Face representation trained on the WebFace dataset \\
VGG-Face~\cite{Parkhi2015VGGFace} & 4,096 & VGG-16 & Softmax loss & A publicly available face descriptor \\
VGG-Face-ULDML & 4,096 & 16-layer VGG & ULDML~\cite{cinbis2011unsupervised} & A Mahalanobis mapping from the VGG-Face features \\
Ours-Triplet & 64 & AlexNet & Triplet loss & Trained with traditional spatio-temporal constraints \\
Ours-Siamese & 64 & AlexNet & Contrastive loss & Trained with traditional spatio-temporal constraints \\
Ours-SymTriplet & 64 & AlexNet & SymTriplet loss & Trained with traditional spatio-temporal constraints \\
Ours-SymTriplet-Contx & 64 & AlexNet & SymTriplet loss & Trained with the contextual constraints \\
Ours-SymTriplet-BBT02 & 64 & AlexNet & SymTriplet loss & Trained on the BBT02 video with spatio-temporal constraints. \\
\bottomrule
\end{tabular}
\label{tab:feas_list}
\vspace{-1mm}
\end{table*}

\begin{table*}[t]
\scriptsize
\centering
\caption{Clustering results on 7 BBT videos and 3 BUFFY videos. The weighted purity of each video is measured on the ideal number of clusters. \textbf{\color{red}Red} text indicates the best and \underline{\color{blue}blue} text indicates the second-best performance.
}
\vspace{-1mm}
\begin{tabular}{@{}lccccccccccc@{}}
\toprule
&\multicolumn{7}{c}{BBT dataset} & & \multicolumn{3}{c}{BUFFY dataset}\\
\cmidrule{2-8} \cmidrule{10-12}
Methods & BBT01 & BBT02 & BBT03 & BBT04 & BBT05 & BBT06 & BBT07 && BUFFY02&BUFFY05&BUFFY06\\ \midrule
HOG~\cite{dalal2005histograms}&0.37& 0.31&0.37&0.36&0.29&0.26&0.30&&0.21&0.38&0.25\\
AlexNet~\cite{krizhevsky2012imagenet}&0.47&0.31&0.45&0.36&0.29&0.26&0.39&&0.33&0.37& 0.26\\
Pre-trained & 0.86 &0.71&0.73&0.59&0.51&0.50&0.73&&0.26&0.42&0.32\\
VGG-Face~\cite{Parkhi2015VGGFace}&0.91 &0.85 &0.85 &0.54 &0.65 &0.46 &0.79&&0.22&0.51&0.41\\
VGG-Face-ULDML & 0.92 & \underline{\color{blue}0.87} & 0.86 & 0.60 & 0.68 & 0.23 & 0.85 &&0.34&0.54&0.43\\ \midrule
Ours-Siamese &\underline{\color{blue}0.94}&\textbf{\color{red}0.95}&0.87&0.74&0.70&0.70&0.89&&0.44&0.67&0.61\\
Ours-Triplet &\underline{\color{blue}0.94}&\textbf{\color{red}0.95}&\underline{\color{blue}0.92}&0.74&0.68&0.70&0.89&&0.45&0.66&0.70\\
Ours-SymTriplet & \underline{\color{blue}0.94} & \textbf{\color{red}0.95}& \underline{\color{blue}0.92} & \underline{\color{blue}0.78}& \underline{\color{blue}0.85} & \underline{\color{blue}0.75}& \underline{\color{blue}0.91}&&\underline{\color{blue}0.46} &\underline{\color{blue}0.68} &\underline{\color{blue}0.73}\\
Ours-SymTriplet-Contx& \textbf{\color{red}0.95} & \textbf{\color{red}0.95}& \textbf{\color{red}0.93} & \textbf{\color{red}0.84}& \textbf{\color{red}0.86} & \textbf{\color{red}0.83}& \textbf{\color{red}0.92}&&\textbf{\color{red}0.58}&\textbf{\color{red}0.70}&\textbf{\color{red}0.75}\\ \midrule
Ours-SymTriplet-BBT02& 0.90 &\textbf{\color{red}0.95} &0.87 &0.74 &0.79 &0.67 &0.88 &&-&-&-\\
\bottomrule
\end{tabular}
\label{tab:feats_videos}
\vspace{-1mm}
\end{table*}

\begin{table*}[t]
\scriptsize
\centering
\caption{Clustering results on 8 music videos. The weighted purity of each video is measured on the ideal number of clusters.
}
\vspace{-1mm}
\begin{tabular}{@{}lcccccccc@{}}
\toprule
&\multicolumn{8}{c}{Music dataset}\\
\cmidrule{2-9}
Methods & T-ara & Pussycat Dolls & Bruno Mars & Hello Bubble & Darling & Apink & Westlife & Girls Aloud \\ \midrule
HOG~\cite{dalal2005histograms}&0.22& 0.28&0.36&0.33&0.20&0.20&0.27&0.29\\
AlexNet~\cite{krizhevsky2012imagenet}&0.24&0.32&0.35&0.31&0.19&0.21&0.37&0.30\\
Pre-trained & 0.31 &0.31&0.49&0.34&0.25&0.28&0.32&0.33\\
VGG-Face~\cite{Parkhi2015VGGFace}&0.23 &0.46 &0.44 &0.29 &0.21 &0.24 &0.27 &0.31\\
VGG-Face-ULDML & 0.26 & 0.44 & 0.47 & 0.34 & 0.28 & 0.26 & 0.41 & 0.32 \\ \midrule
Ours-Siamese&\underline{\color{blue}0.69}&0.77&0.88&0.54&0.46&0.48&0.54&0.67\\
Ours-Triplet &0.68&0.77&0.83&0.60&0.49&0.60&0.52&0.67\\
Ours-SymTriplet & \underline{\color{blue}0.69} & \underline{\color{blue}0.78}
& \underline{\color{blue}0.90} & \underline{\color{blue}0.64}
& \underline{\color{blue}0.70} & \underline{\color{blue}0.72}
& \underline{\color{blue}0.56} & \underline{\color{blue}0.69}\\
Ours-SymTriplet-Contx & \textbf{\color{red}0.84} & \textbf{\color{red}0.83}
& \textbf{\color{red}0.91} & \textbf{\color{red}0.69}
& \textbf{\color{red}0.72} & \textbf{\color{red}0.74}
& \textbf{\color{red}0.66} & \textbf{\color{red}0.75}\\
\bottomrule
\end{tabular}
\label{tab:feats_videos1}
\vspace{-1mm}
\end{table*}

\vspace{1mm}
\noindent \textbf{Adaptive features vs. off-the-shelf features.}
We evaluate the proposed features (Ours-Siamese, Ours-Triplet, Ours-SymTriplet and Ours-SymTriplet-Contx) adapted to a specific video against the off-the-shelf features (HOG, AlexNet, pre-trained, and VGG-Face) in Table~\ref{tab:feats_videos} and~\ref{tab:feats_videos1}.
We show that identity-preserving features (pre-trained and VGG-Face) trained on face datasets offline achieve better performance over generic feature representation (e.g., AlexNet and HOG).
Our video-specific features trained with Siamese and triplet networks achieve favorable performance than other alternatives, highlighting the importance of learning video-specific features.
For example, in the \textsc{Daring} sequence, the proposed method with the Ours-SymTriplet-Contx features achieves the weighted purity of 0.76, significantly outperforming the off-the-shelf features, i.e., VGG-Face: 0.20, AlexNet: 0.18 and HOG: 0.19.
Overall, the results with the proposed features are more than twice as accurate as that using off-the-shelf features in music videos.
For the BBT dataset, the proposed feature adaptation consistently outperforms that with off-the-shelf features.

\vspace{1mm}
\noindent \textbf{Measuring the effectiveness of features via clustering}
Here, we validate the effectiveness the proposed features compared to the baselines.
Figure~\ref{fig:feats_eval} shows the results in terms of clustering purity versus the number of clusters on 7 BBT sequences and 5 music videos.
The ideal line (purple dash line) means that all faces are correctly grouped with weighted purity $W_C = 1$.
For more effective features, the weighted purity measures approach to 1 at a faster rate.
For each feature type, we show the weighted purity at the ideal number cluster (i.e., number of people in a video) in the legend.
%

% JB: Remove (a,b,c,...) They are meaningless. Change the typesetting to minipage as shown in Figure{fig:samples1} for maximum figure.
\setlength{\figwidth}{0.15\textwidth}
\begin{figure*}[t]
\scriptsize
\centering
\begin{tabular}{{c}{c}{c}{c}{c}{c}}
\hspace{-2.5mm}\includegraphics[width=\figwidth]{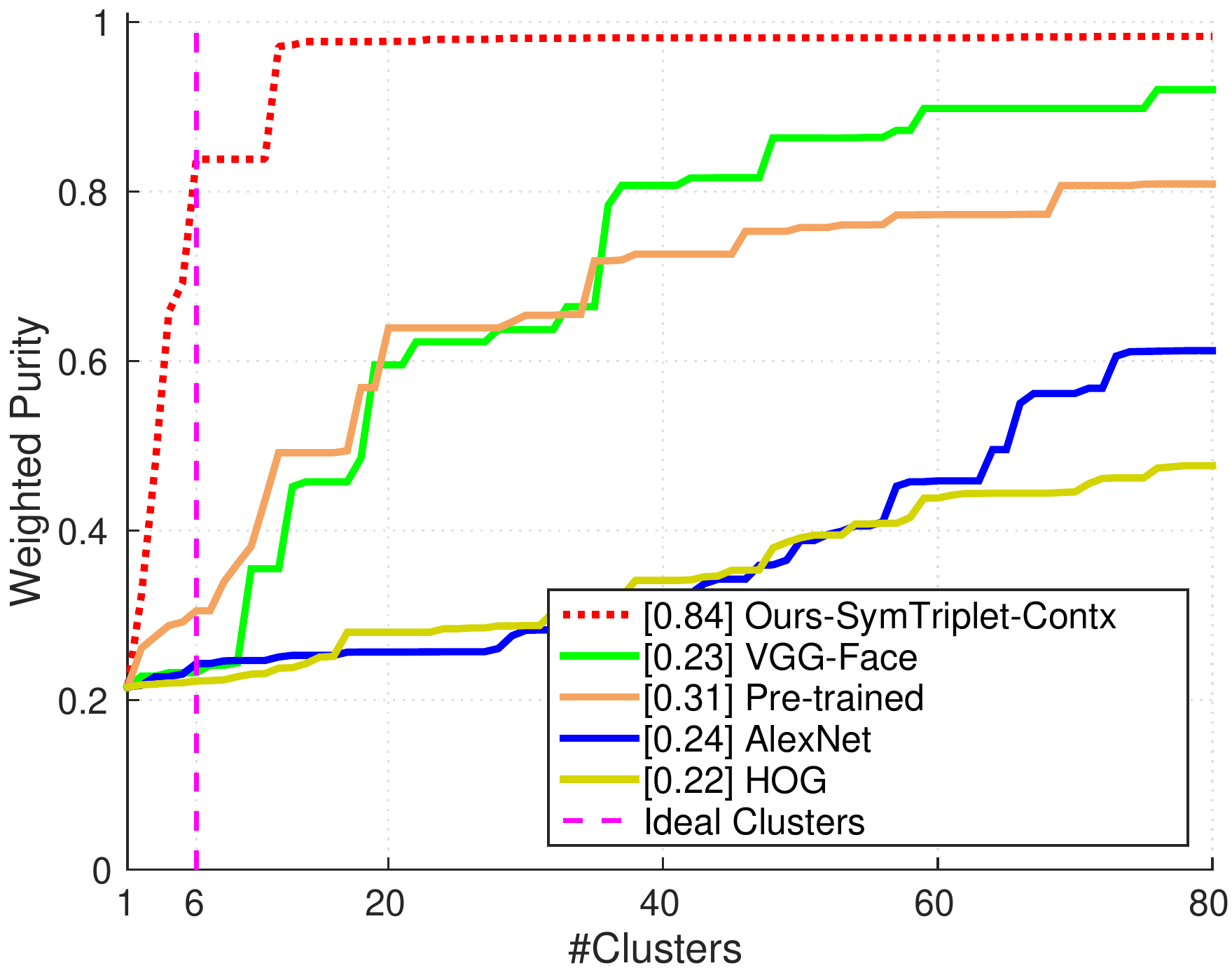} &
\hspace{-2.5mm}\includegraphics[width=\figwidth]{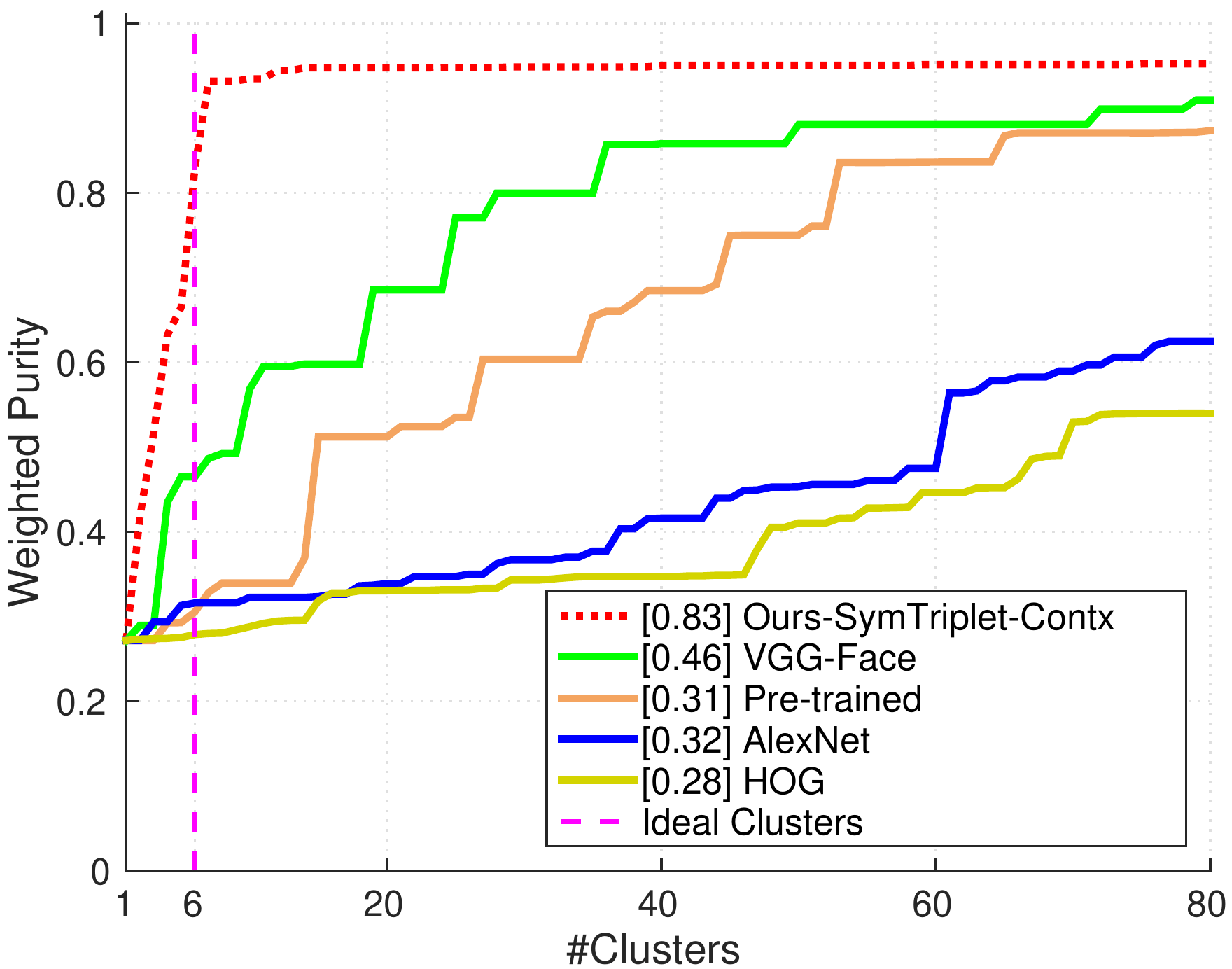} &
\hspace{-2.5mm}\includegraphics[width=\figwidth]{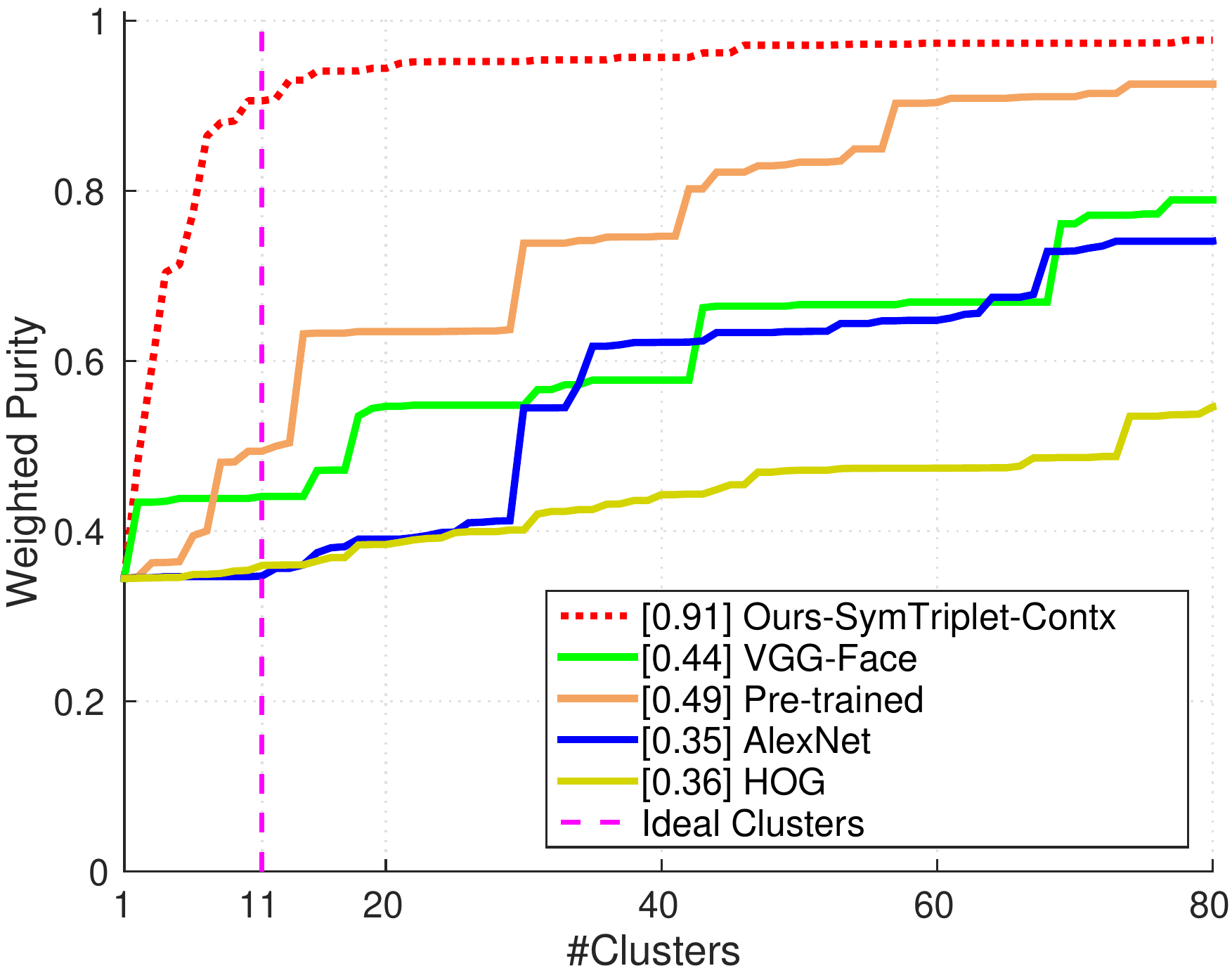} &
\hspace{-2.5mm}\includegraphics[width=\figwidth]{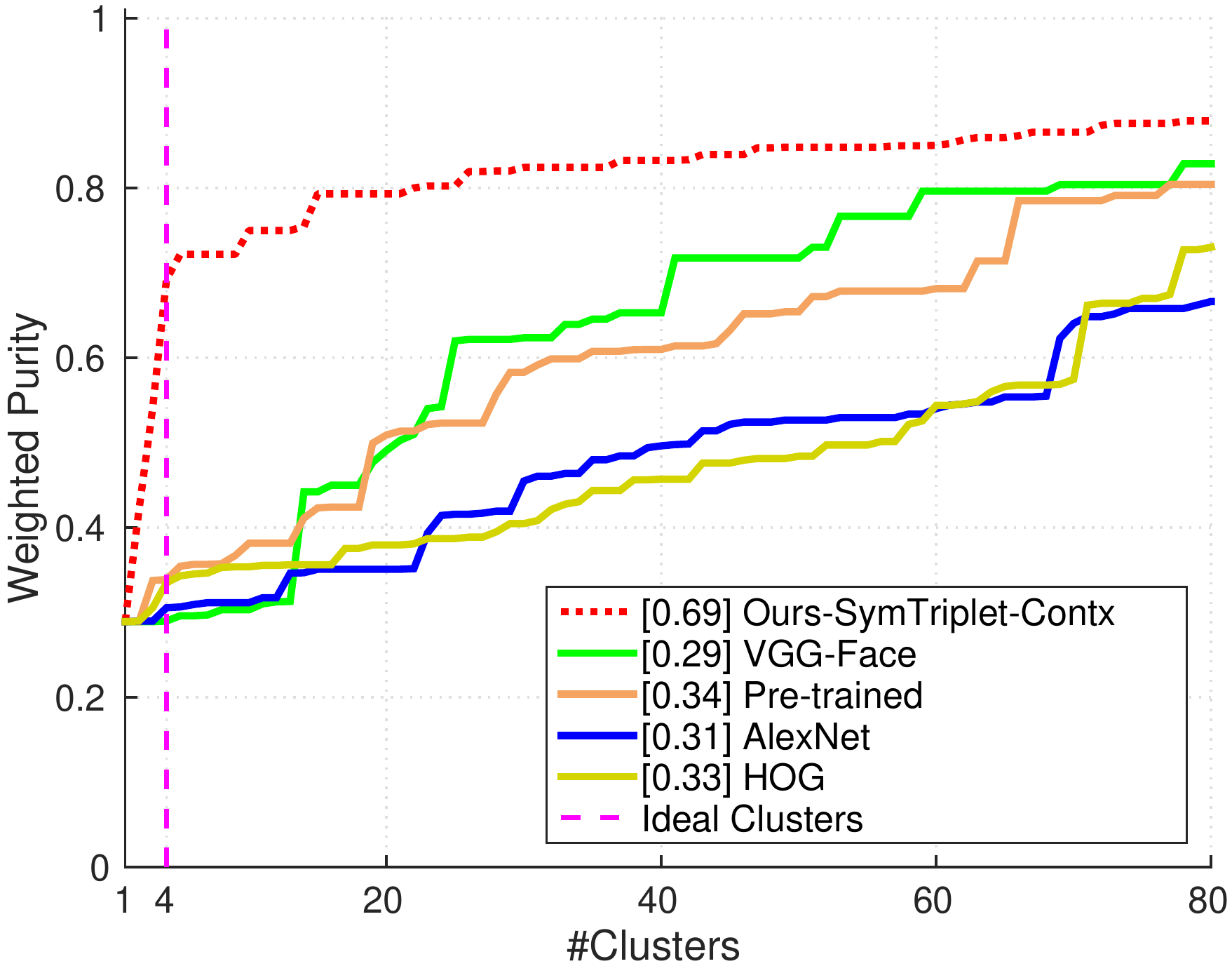} &
\hspace{-2.5mm}\includegraphics[width=\figwidth]{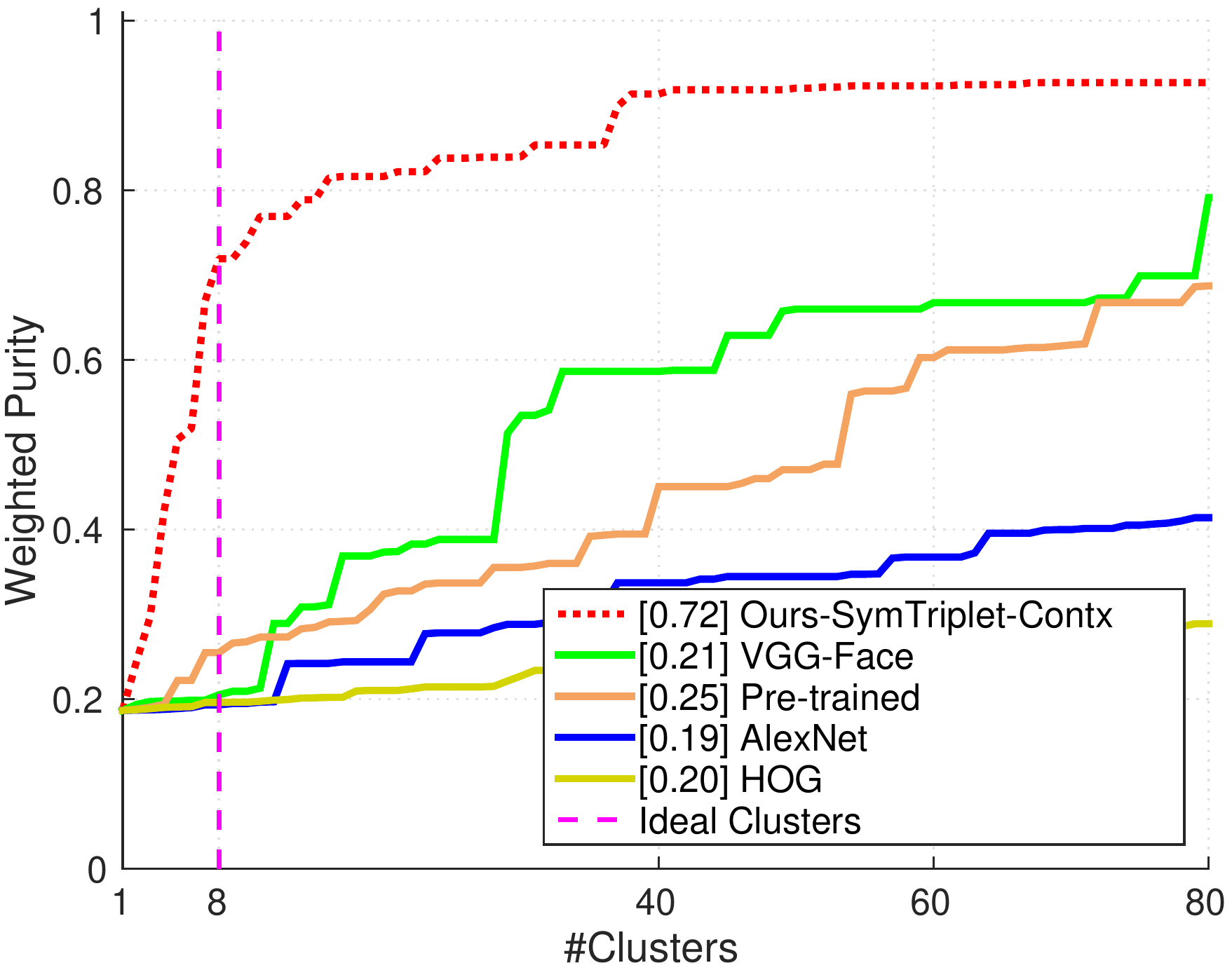} &
\hspace{-2.5mm}\includegraphics[width=\figwidth]{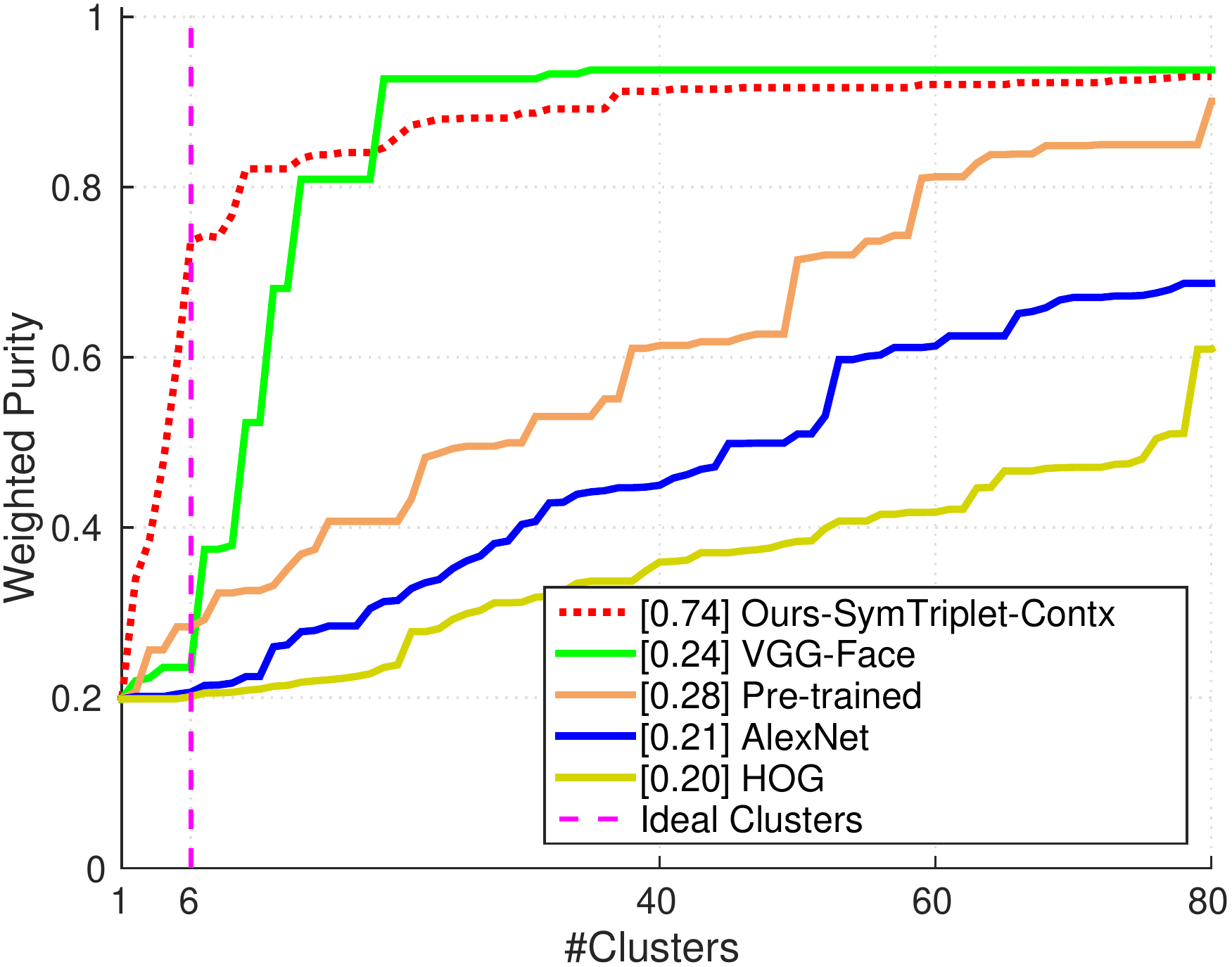} \\
\hspace{-2.5mm}(a) T-ara &
\hspace{-2.5mm}(b) Pussycat Dolls &
\hspace{-2.5mm}(c) Bruno Mars &
\hspace{-2.5mm}(d) HelloBubble &
\hspace{-2.5mm}(e) Darling &
\hspace{-2.5mm}(f) Apink \\
\hspace{-2.5mm}\includegraphics[width=\figwidth]{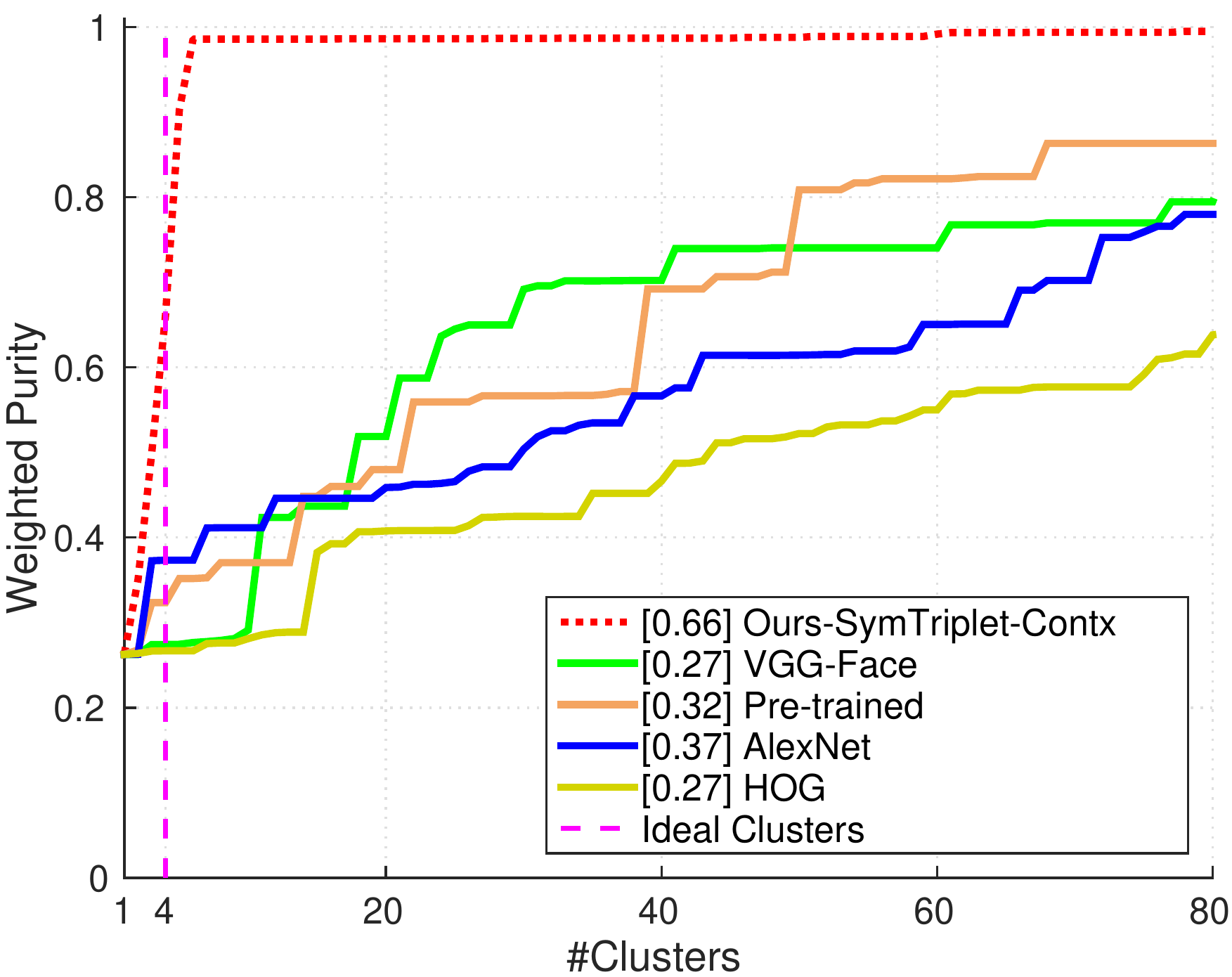} &
\hspace{-2.5mm}\includegraphics[width=\figwidth]{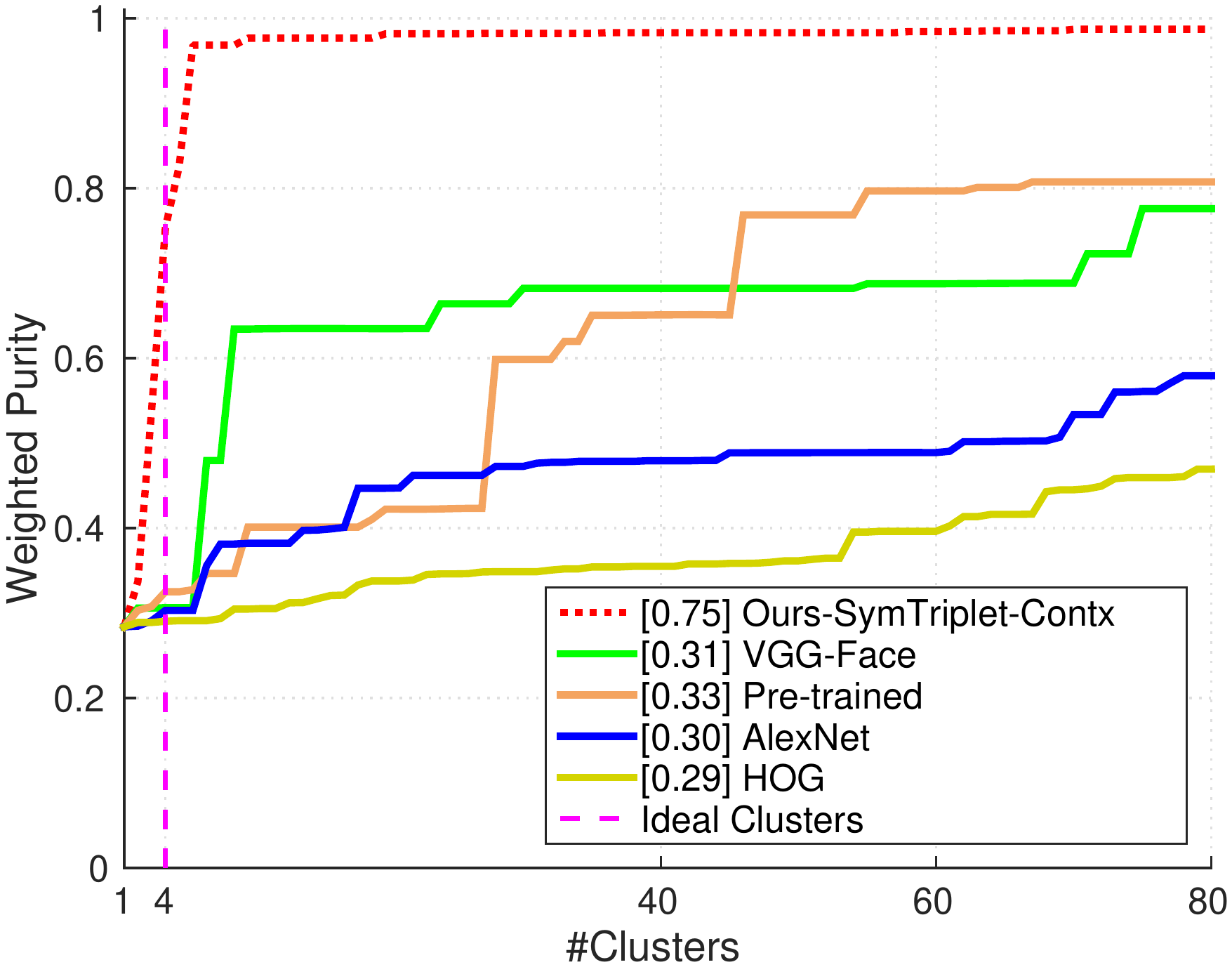} &
\hspace{-2.5mm}\includegraphics[width=\figwidth]{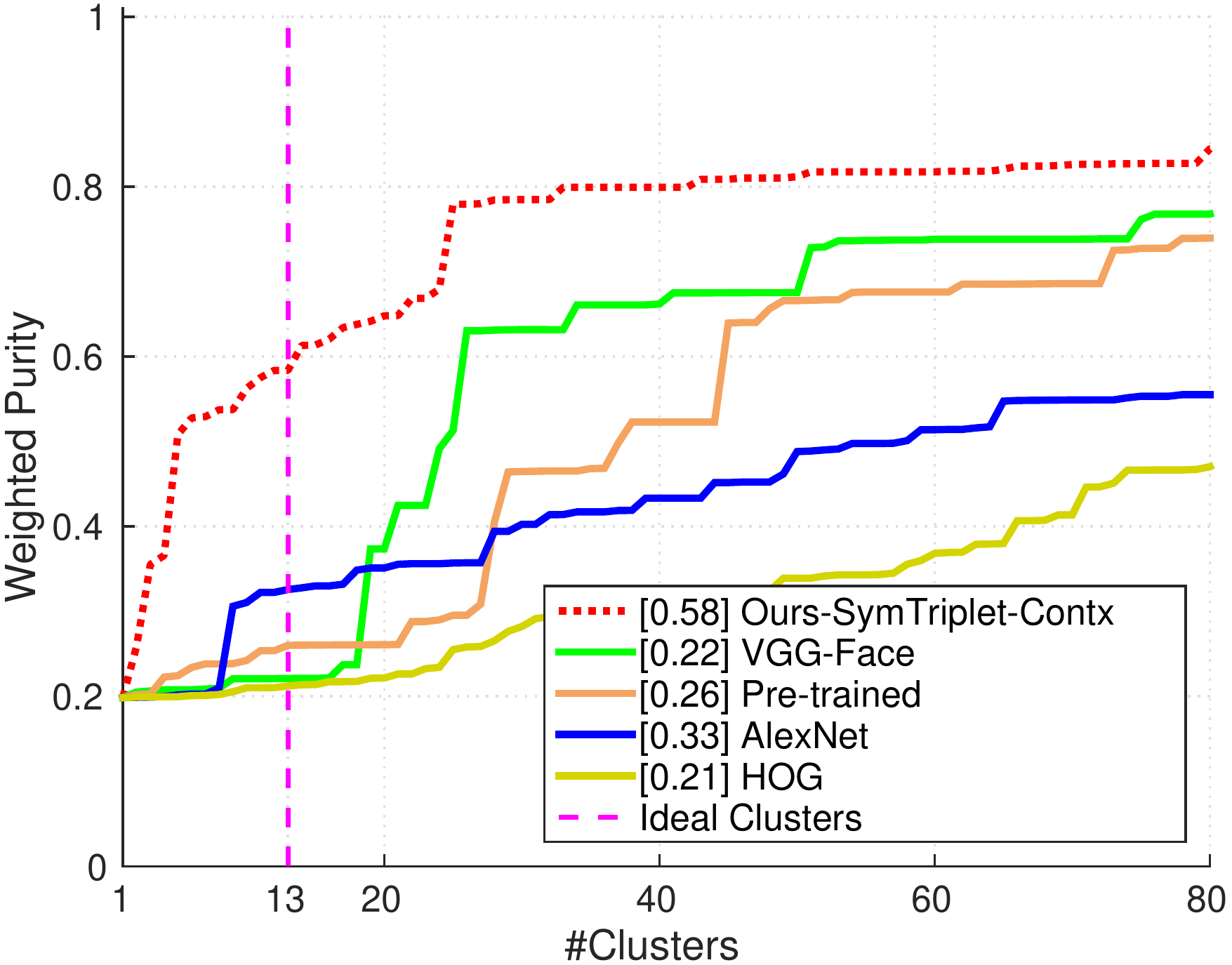} &
\hspace{-2.5mm}\includegraphics[width=\figwidth]{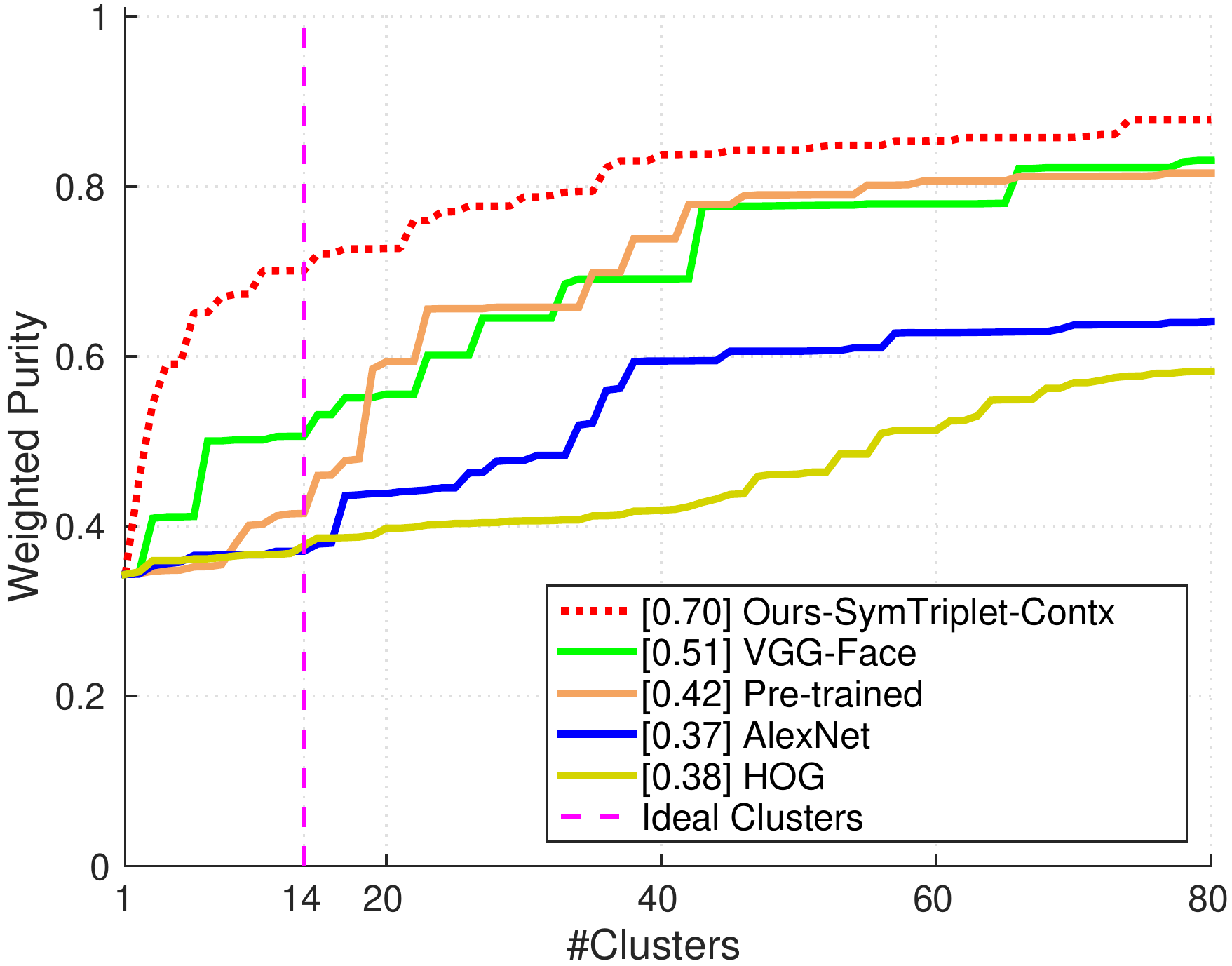} &
\hspace{-2.5mm}\includegraphics[width=\figwidth]{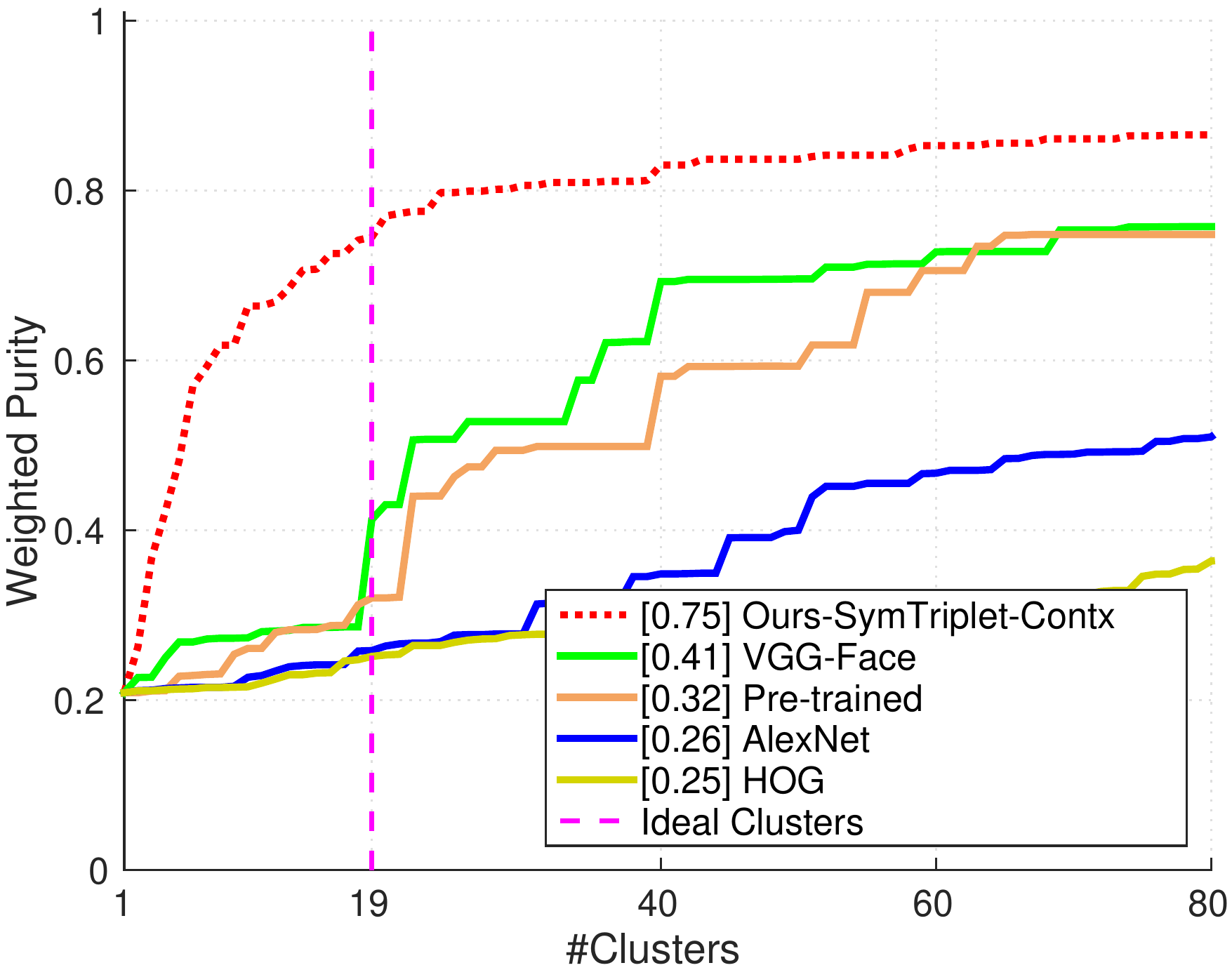} &
\hspace{-2.5mm}\includegraphics[width=\figwidth]{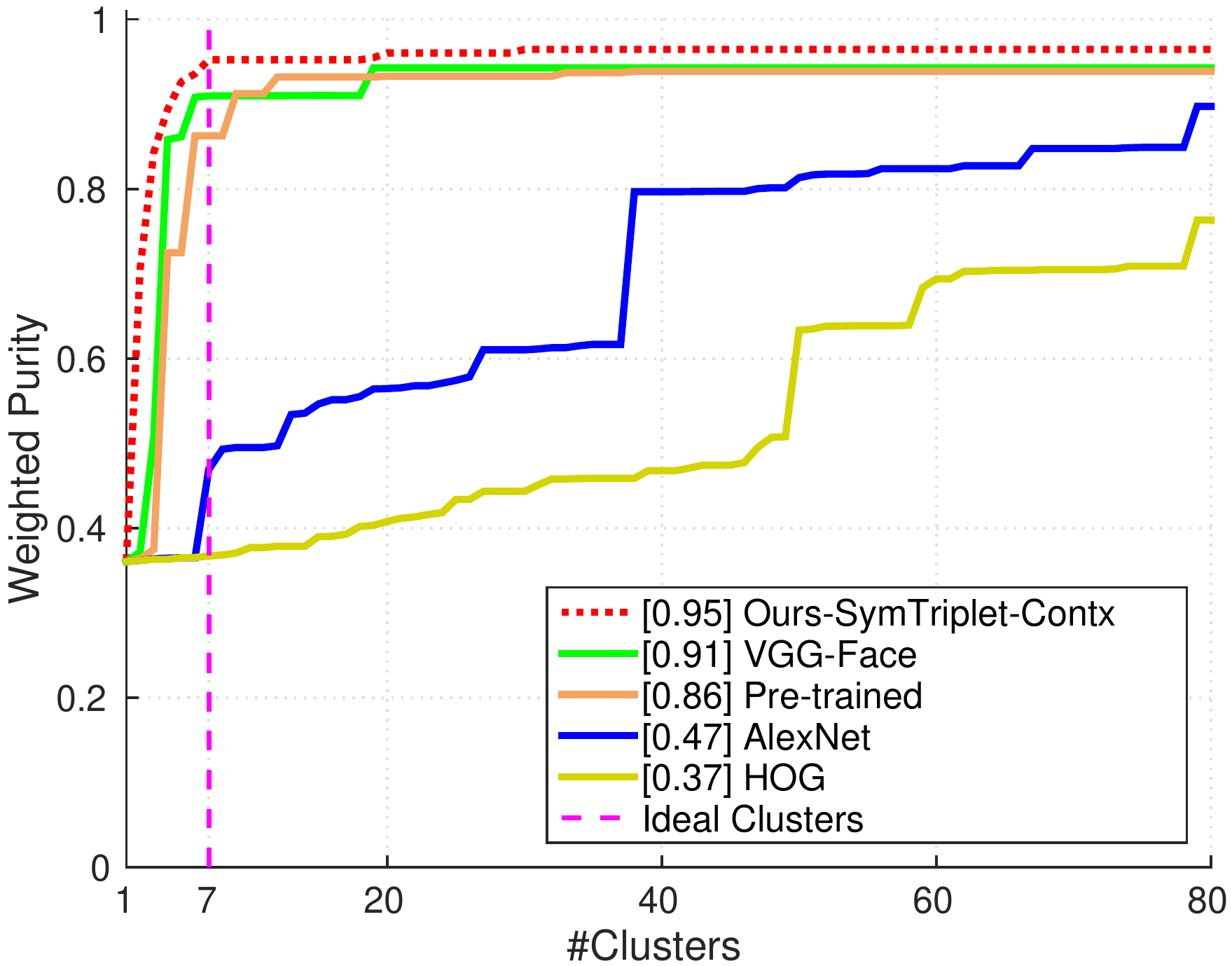} \\
\hspace{-2.5mm}(g) Westlife &
\hspace{-2.5mm}(h) Girls Aloud &
\hspace{-2.5mm}(i) BUFFY02 &
\hspace{-2.5mm}(j) BUFFY05 &
\hspace{-2.5mm}(k) BUFFY06 &
\hspace{-2.5mm}(l) BBT01 \\
\hspace{-2.5mm}\includegraphics[width=\figwidth]{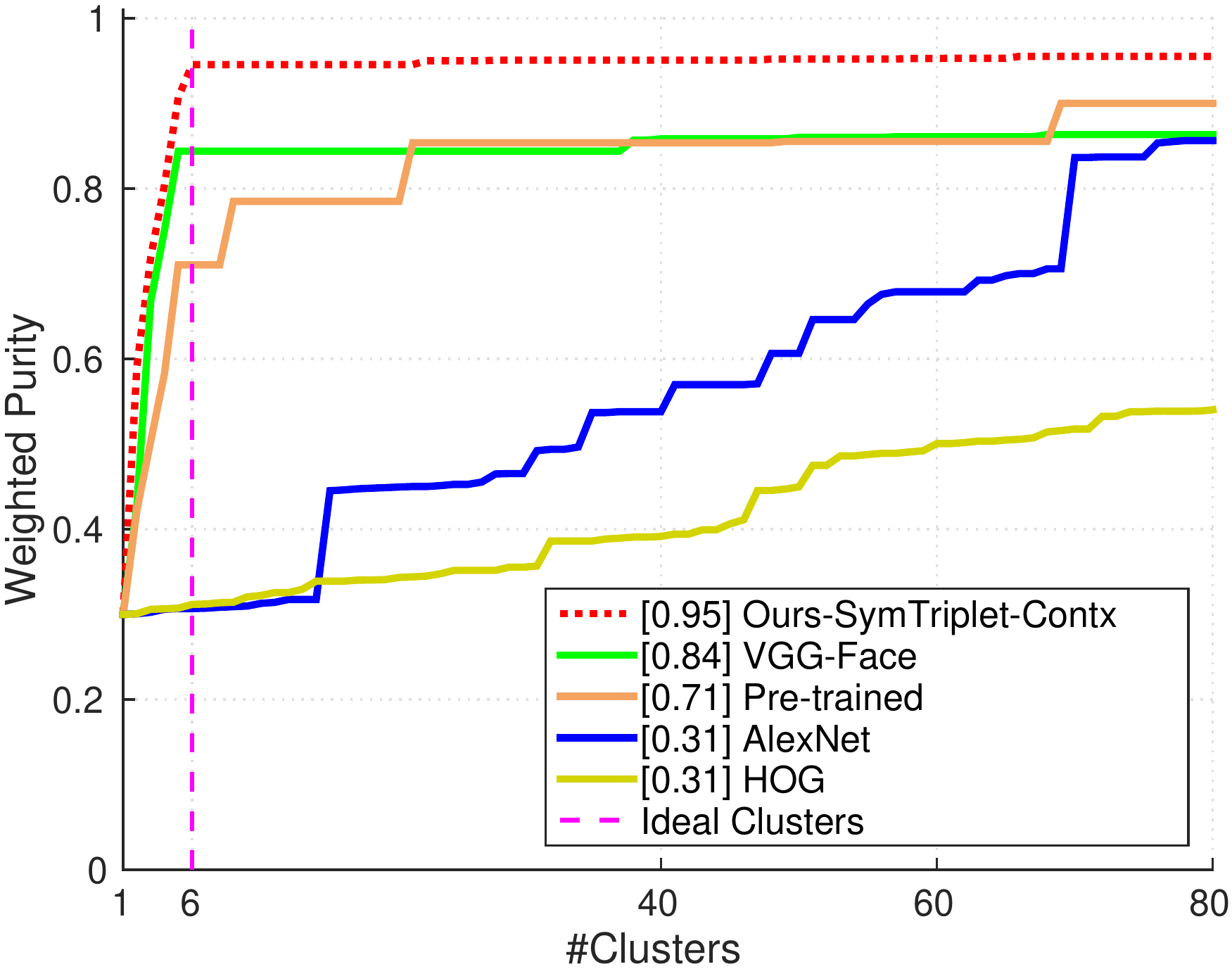} &
\hspace{-2.5mm}\includegraphics[width=\figwidth]{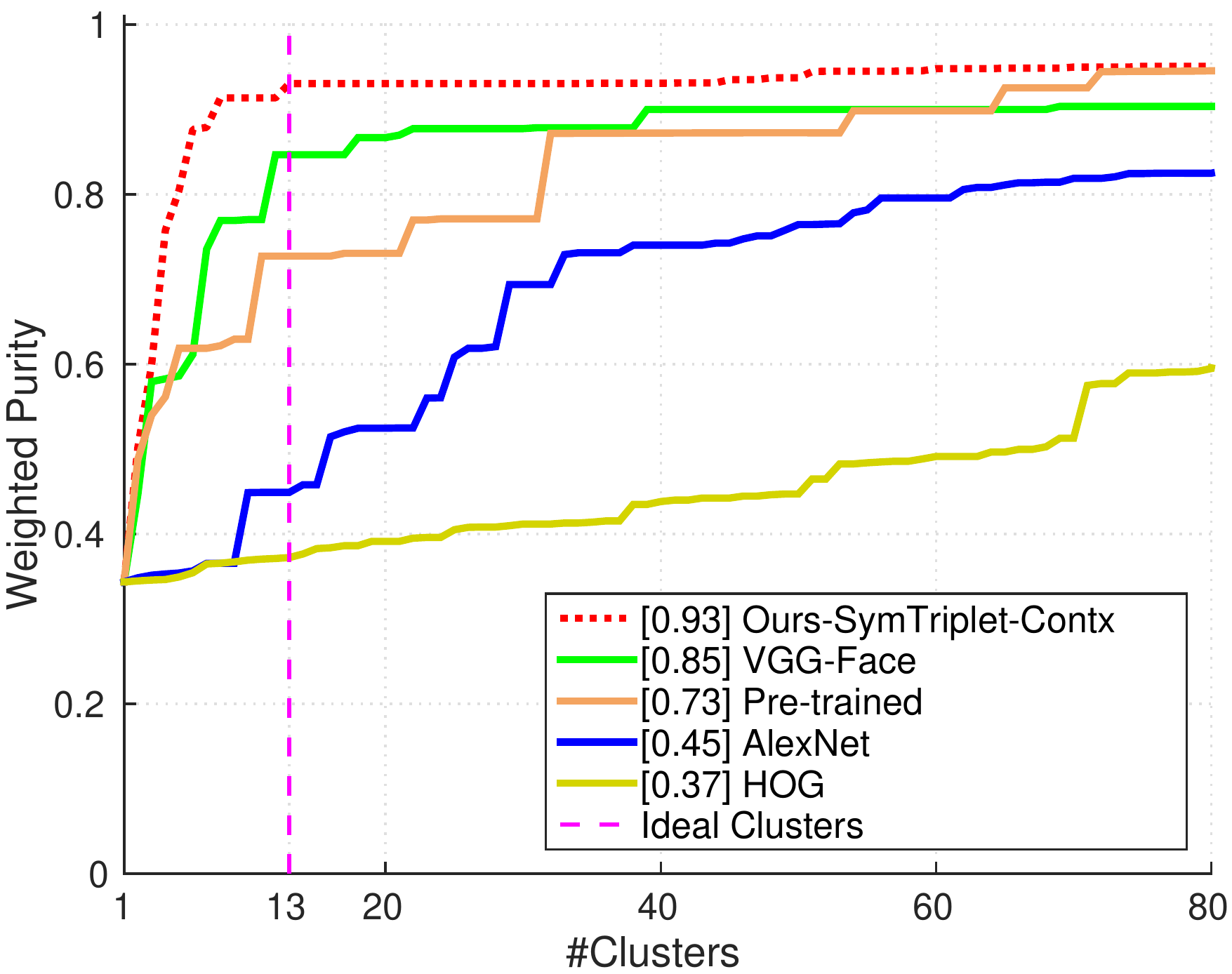} &
\hspace{-2.5mm}\includegraphics[width=\figwidth]{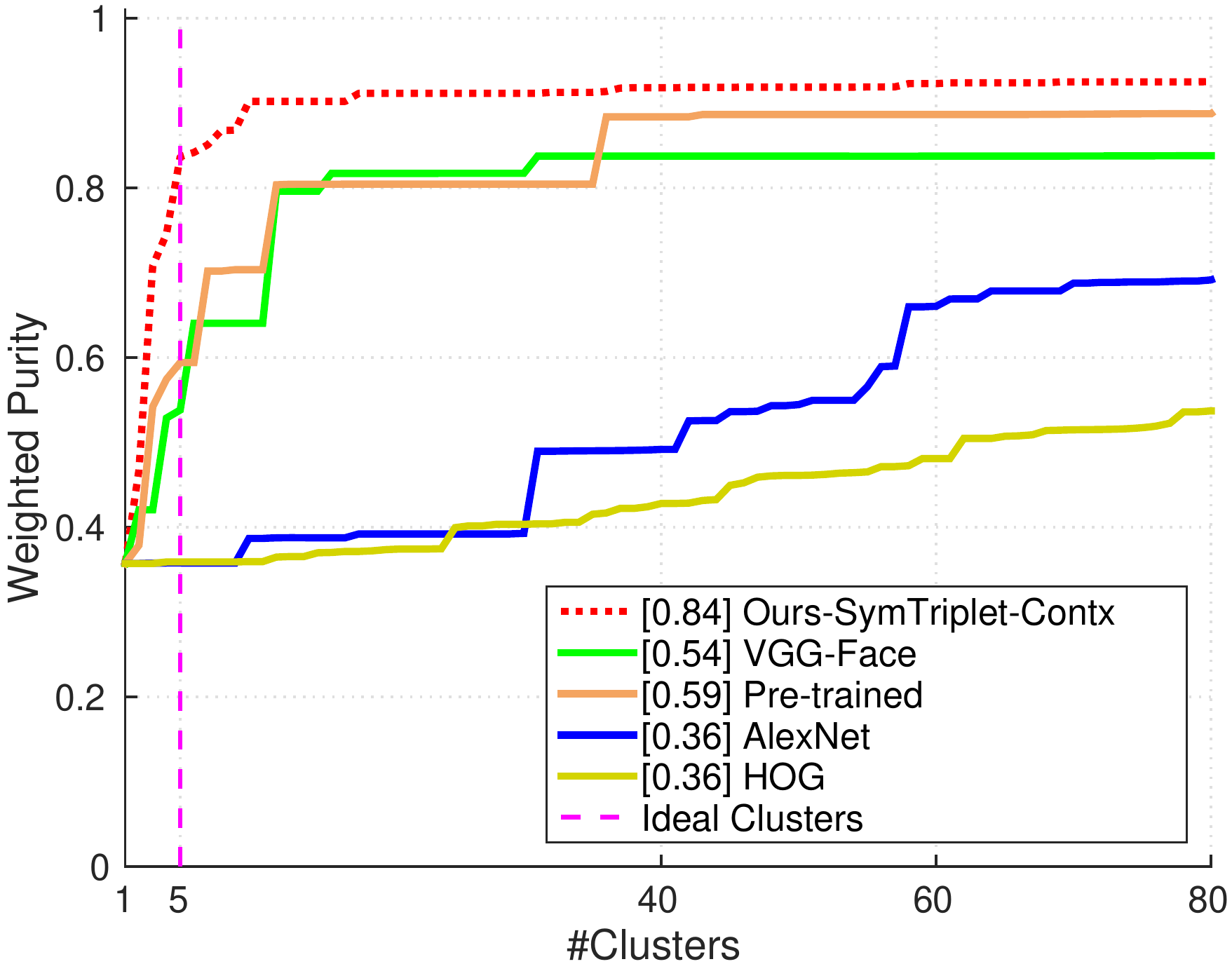} &
\hspace{-2.5mm}\includegraphics[width=\figwidth]{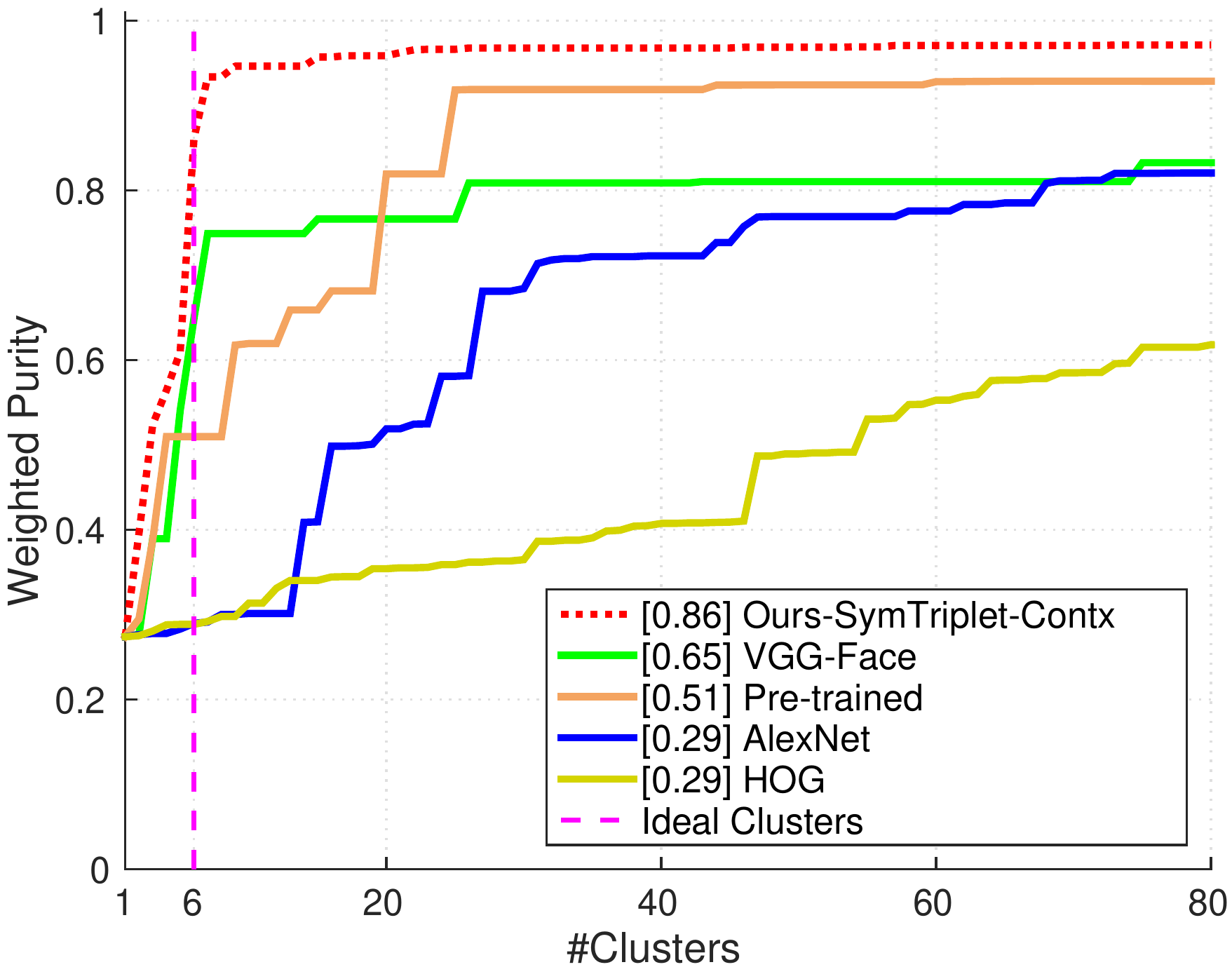} &
\hspace{-2.5mm}\includegraphics[width=\figwidth]{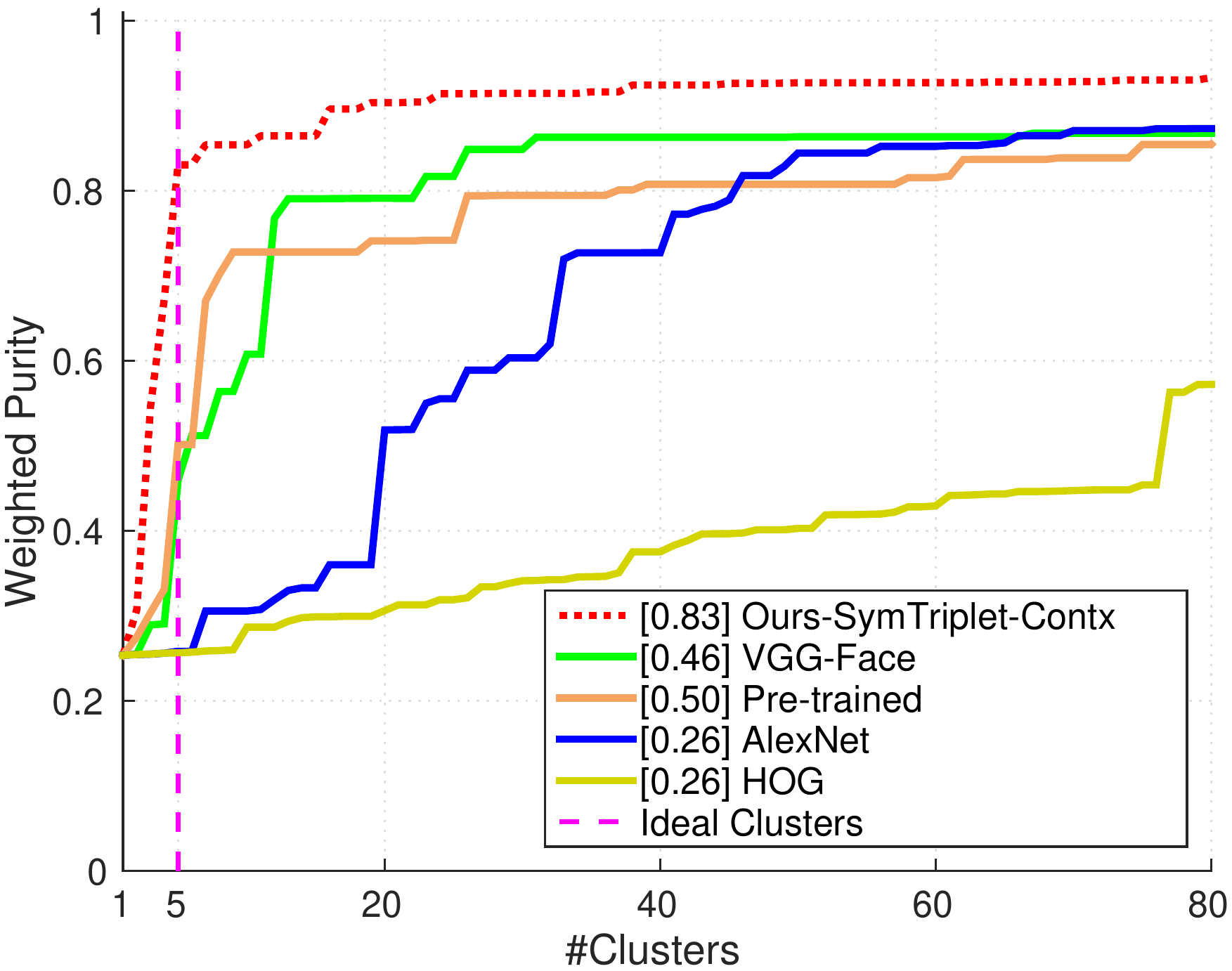} &
\hspace{-2.5mm}\includegraphics[width=\figwidth]{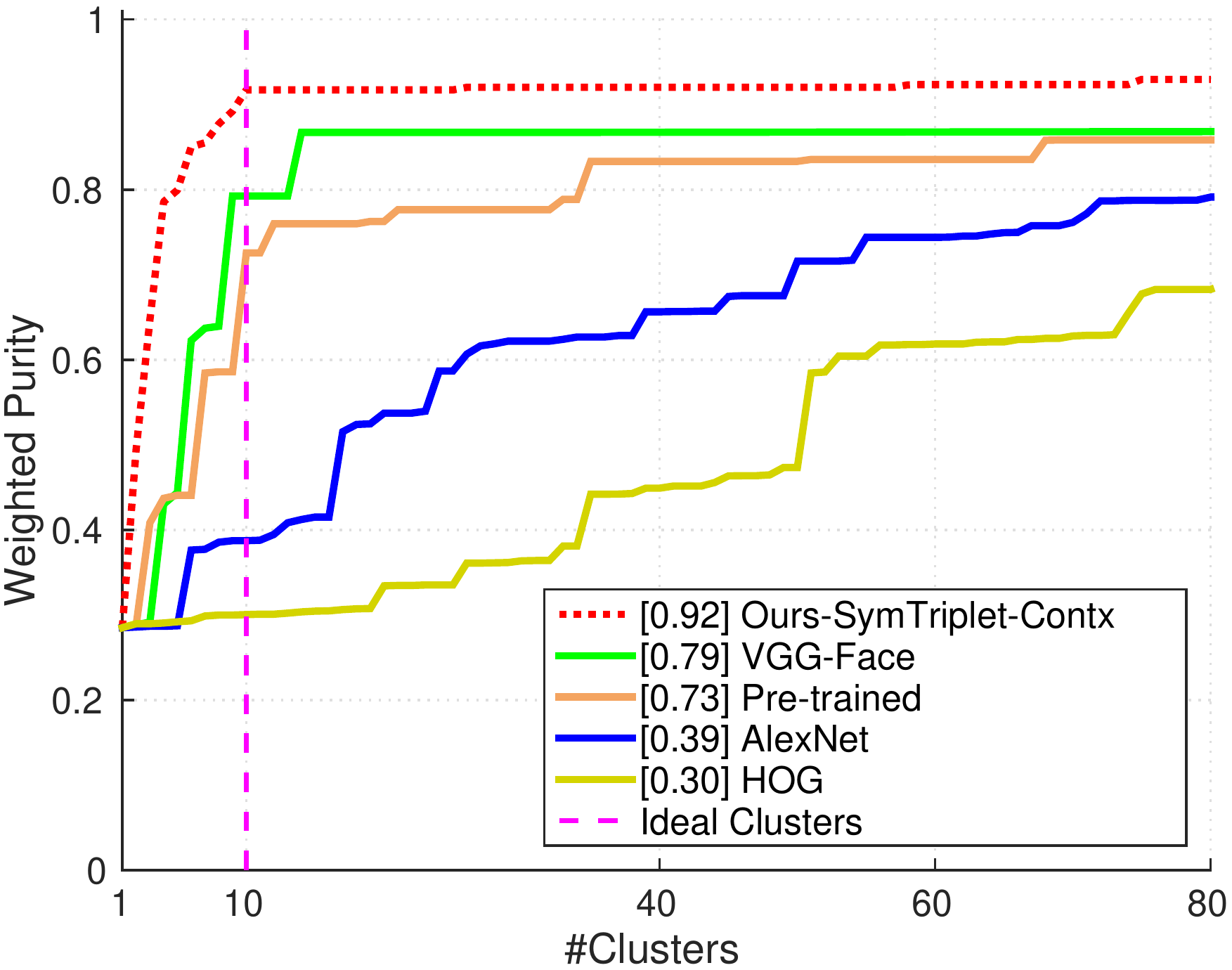} \\
\hspace{-2.5mm}(m) BBT02 &
\hspace{-2.5mm}(n) BBT03 &
\hspace{-2.5mm}(o) BBT04 &
\hspace{-2.5mm}(p) BBT05 &
\hspace{-2.5mm}(q) BBT06 &
\hspace{-2.5mm}(r) BBT07
\end{tabular}
\caption{
% JB: The legend is redundant and too small to see in the figures.
% Use \crule[green]{0.3cm}{0.3cm} Method 1, \crule[blue]{0.3cm}{0.3cm} Method 2 in the caption instead.
\textbf{Clustering performance}. The clustering purity versus the number of clusters in comparison with different features on YouTube music video, Big Bang Theory and BUFFY datasets.
The ideal line indicates that all faces are correctly grouped into ideal clusters, and its corresponding weighted purity is equal to 1.
For the more effective feature, its purity approximates to 1 faster with the increase in the number of clusters.
The legend contains the purities at the ideal number of clusters for each feature.
}
\label{fig:feats_eval}
\vspace{-3mm}
\end{figure*}

\begin{figure}[t]
\setlength{\abovecaptionskip}{1mm}
\setlength{\belowcaptionskip}{-1mm}
\footnotesize
\centering
\begin{tabular}{{c}{c}}
\includegraphics[width=0.4\linewidth]{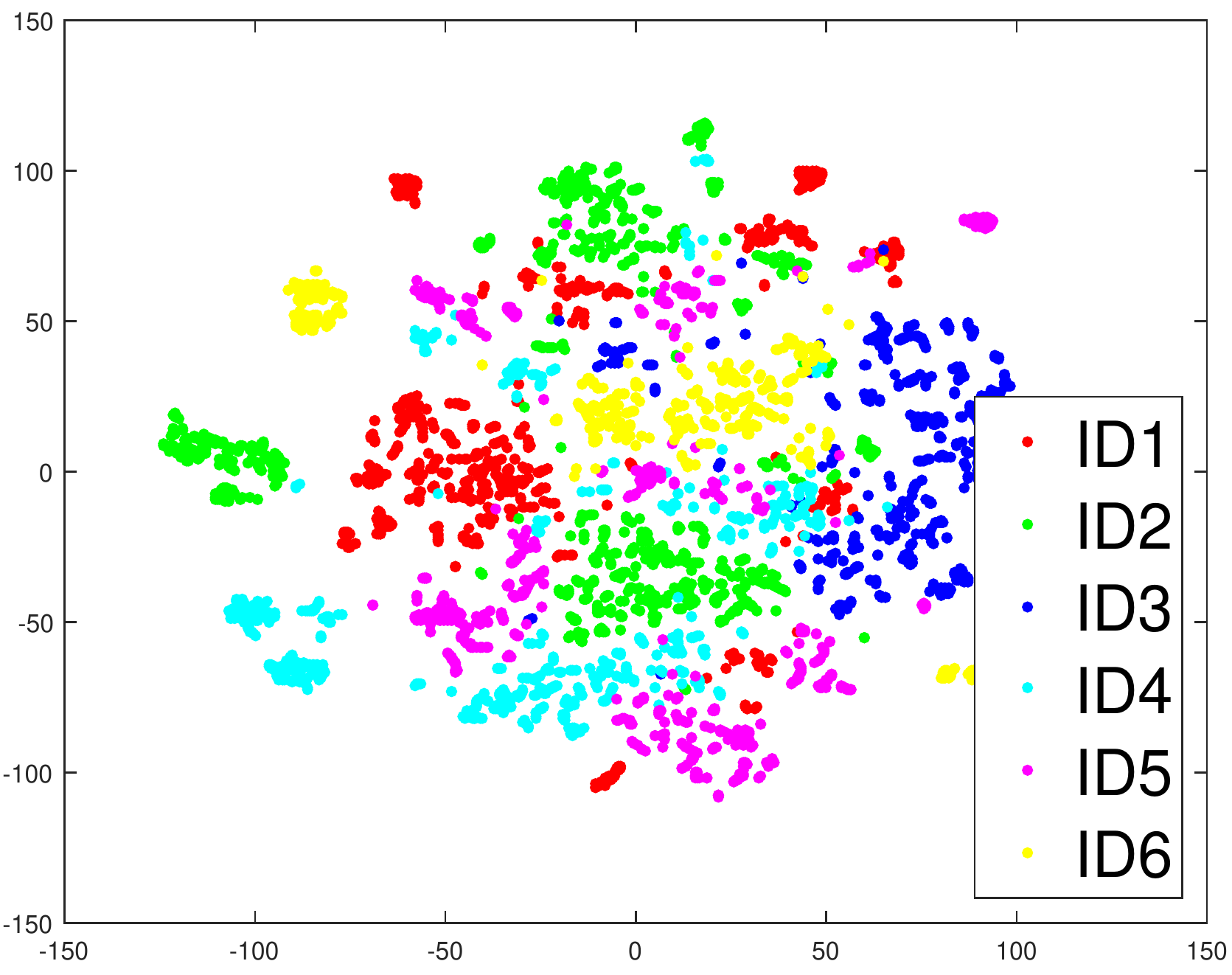} &
\hspace{0.5mm}\includegraphics[width=0.4\linewidth]{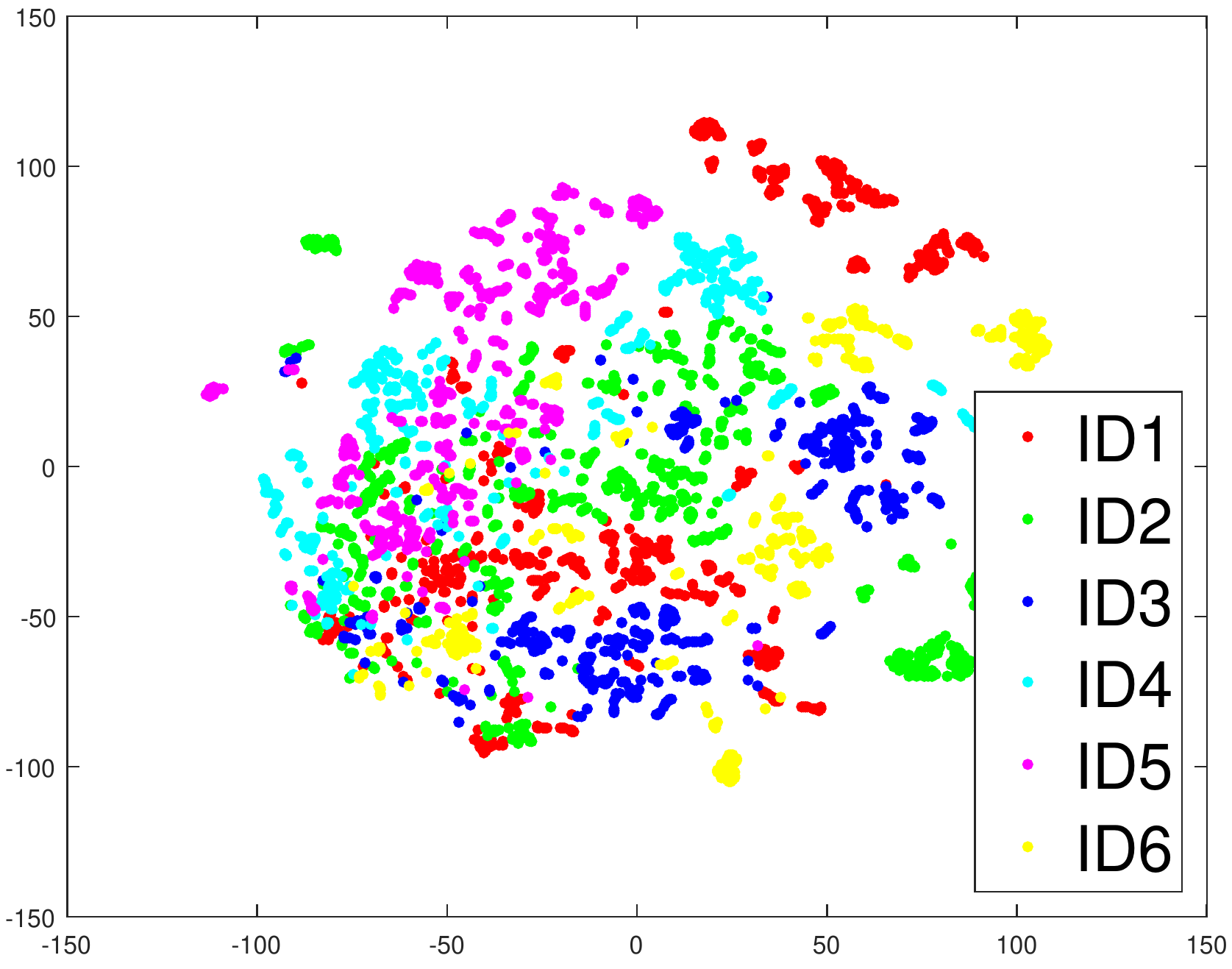} \\
HOG (4356-D) &
\hspace{0.5mm}AlexNet (4096-D) \\
\hspace{0.5mm}\includegraphics[width=0.4\linewidth]{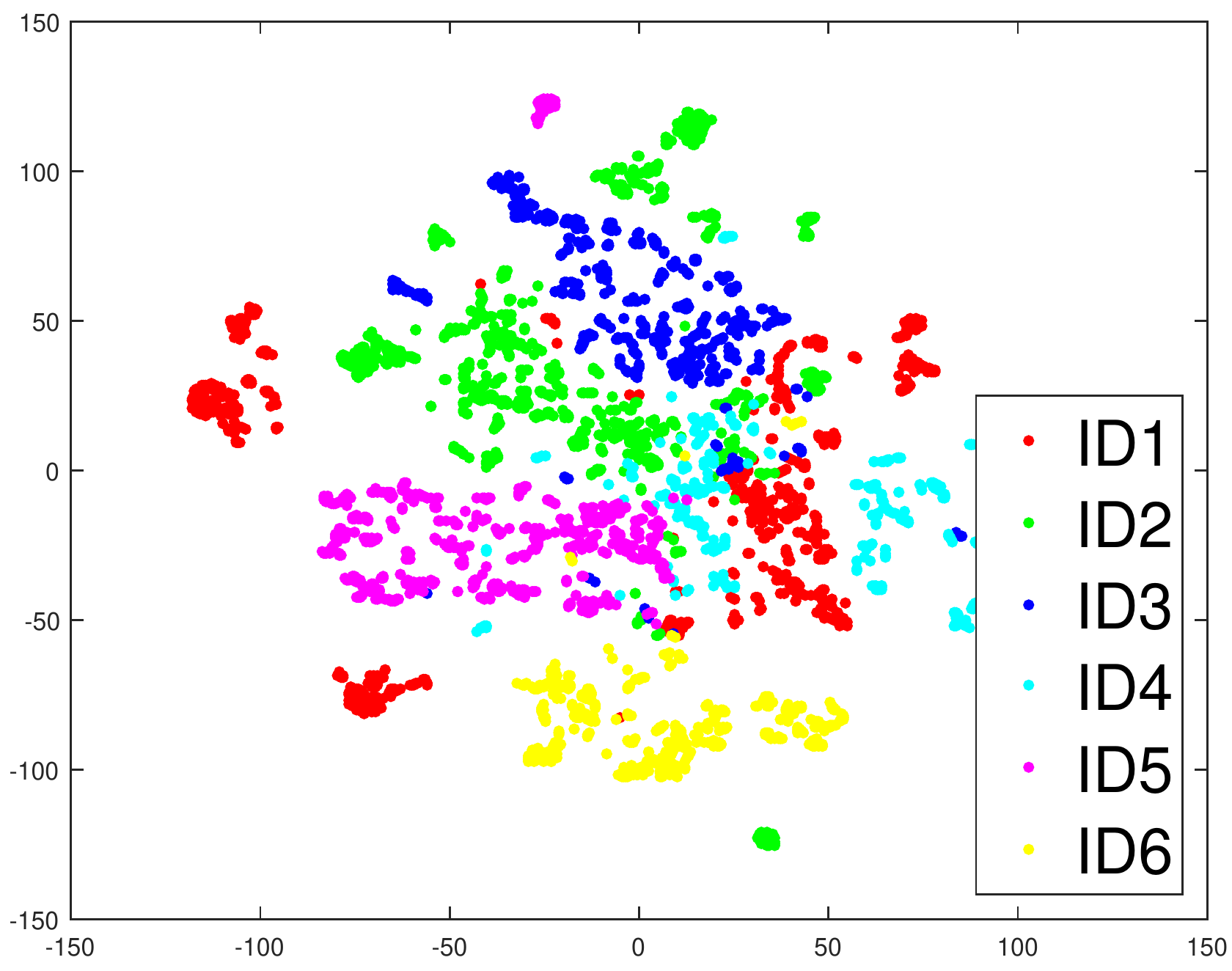} & % \hspace{5mm}
\hspace{0.5mm}\includegraphics[width=0.4\linewidth]{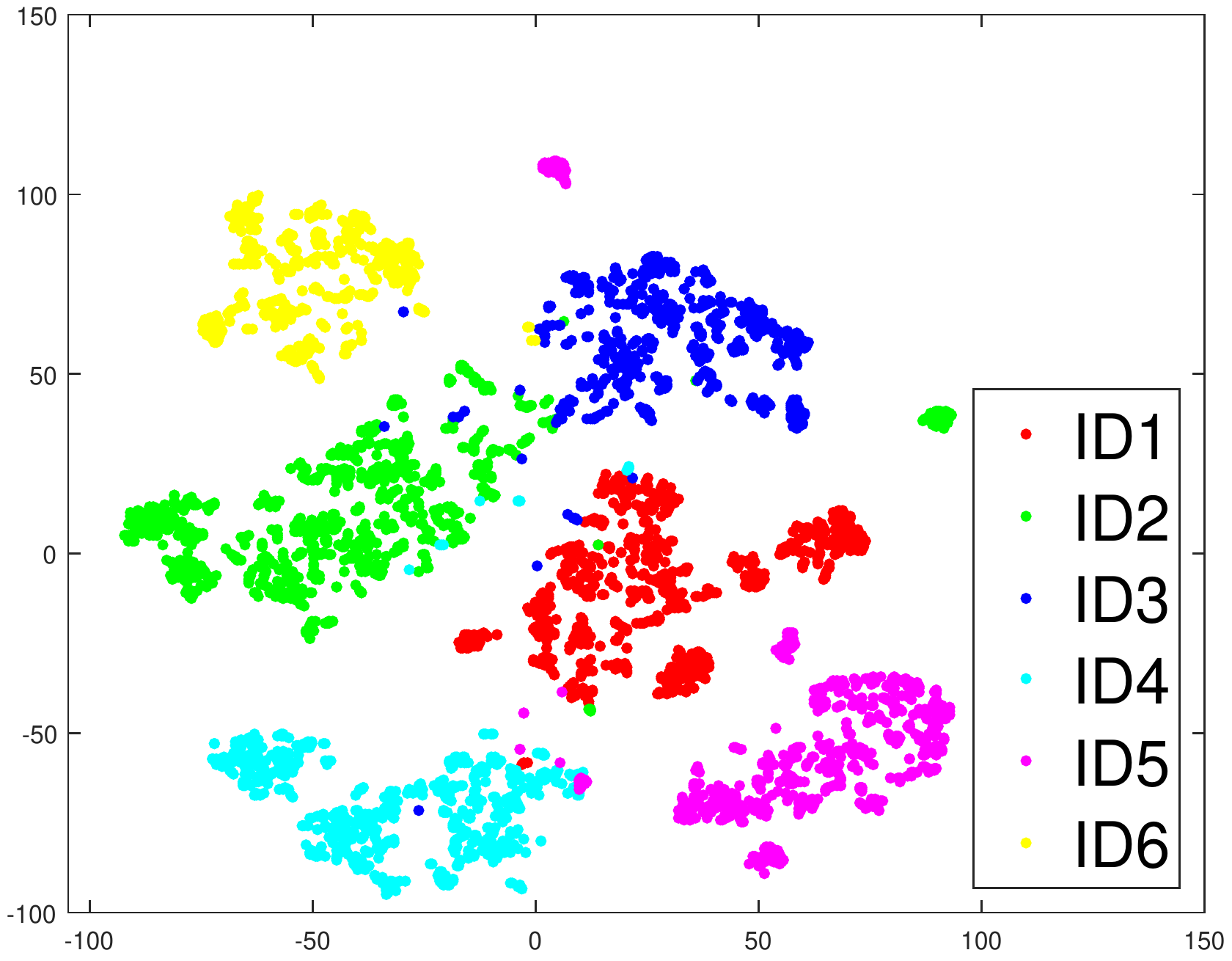} \\
\hspace{0.5mm}Pre-trained (4096-D)&
\hspace{0.5mm}Ours-SymTriplet (64-D)\\
\end{tabular}
\caption{\textbf{2D tSNE visualization}. 2D tSNE visualization of all face features from the proposed fine-tuned CNN for adapting video-specific variations, compared with HOG, AlexNet, and pre-trained features. \textsc{T-ara} has 6 main casts. The faces of different people are color coded.
}
\label{fig:tSNE}
\vspace{-3mm}
\end{figure}

Figure~\ref{fig:tSNE} shows 2D visualization of extracted features from the \textsc{T-ara} using the t-SNE algorithm~\cite{van2008visualizing}.
The visualization illustrates the difficulty in handling large appearance variations in unconstrained videos.
For HOG features, there exist no clear cluster structures, and faces of the same person are scattered around.
Although the AlexNet and pre-trained features increase inter-person distances, the clusters of the same person do not appear in close proximity.
In contrast, the proposed adaptive features form tighter clusters for the same person and greater separation between different persons.

\vspace{1mm}
\noindent \textbf{Nonlinear and linear metric learning.}
% JB: We should comment on why we cannot use Our model + ULMDL. The results you show have TWO factors (architecture and metric learning)
Unlike several existing approaches~\cite{cinbis2011unsupervised,wu2013simultaneous,wu2013constrained,tapaswi2014total,xiao2014weighted} that rely on hand-crafted features and linear metric learning, we use a deep nonlinear metric learning by finetuning all layers to learn discriminative face representations.
To demonstrate the contribution of the nonlinear metric learning, we compare our adaptive features with VGG-Face-ULDML which learns Mahalanobis distance on the VGG-Face features in Table~\ref{tab:feats_videos}.
We show that the proposed method with adaptive features and the nonlinear metric achieve higher clustering purity than VGG-Face-ULDML on all videos.
For example, on the \textsc{T-ara} sequence, the clustering purity by Ours-SymTriplet-Contx and VGG-Face-ULDML is 0.84, and 0.26, respectively.

\vspace{1mm}
\noindent \textbf{SymTriplet and conventional Siamese/Triplet loss.}
We demonstrate the effectiveness of the proposed SymTriplet loss (Ours-SymTriplet) with comparisons to the contrastive loss (Ours-Siamese) and the triplet loss (Ours-Triplet) on all videos in Table~\ref{tab:feats_videos}.
The proposed method with the SymTriplet loss performs well against the other methods since positive sample pairs are pulled closer and negative samples are pushed away from the positive pairs.
For example, on the BBT05 sequence, Ours-SymTriplet (0.85) achieves higher clustering purity than Ours-Triplet (0.68) and Ours-Siamese (0.70);

\vspace{1mm}
\noindent \textbf{Contextual and spatio-temporal constraints.}
We evaluate the effectiveness of contextual constraints.
Using the SymTriplet loss, we compare the features learned from using only spatio-temporal constraints (Ours-SymTriplet), and \emph{both} contextual and spatio-temporal constraints (Ours-SymTriplet-Contx).
Table~\ref{tab:feats_videos} shows that Ours-SymTriplet-Contx achieves better performance when compared with Ours-SymTriplet on all videos.
We attribute the performance improvement to the additional positive and negative face pairs discovered through contextual cues and the transitive constraint propagation.

\vspace{1mm}
\noindent \textbf{Comparisons with other face clustering algorithms.}
We compare our method with five recent state-of-the-art face clustering algorithms
\cite{wu2013simultaneous,cinbis2011unsupervised,wu2013constrained,xiao2014weighted,zhang2016joint} on the \textsc{Frontal}, \textsc{BBT01}, \textsc{BUFFY02}, and \textsc{Notting Hill} videos.
%All the methods exploit visual constraints from tracklets for face clustering.
%
Table~\ref{tab:cluster_eval} shows the clustering accuracy over faces and tracklets
(using the same datasets and metrics as \cite{wu2013simultaneous,wu2013constrained}).\footnote{The code and data of some methods, e.g., \cite{tapaswi2014total} are not available.}
In contrast to the methods in~\cite{wu2013simultaneous,cinbis2011unsupervised,wu2013constrained,xiao2014weighted} which learn linear transformations over the extracted features, our work learns nonlinear metrics by adapting all layers of the CNN and performs favorably on the \textsc{Frontal}, \textsc{BBT01}, and \textsc{BUFFY02} sequences.
Both \cite{zhang2016joint} and our method discover more informative face pairs to adapt the pre-trained models to learn discriminative face representations and achieve similar clustering performance on the \textsc{BUFFY02} and \textsc{Notting Hill} videos.

\begin{table*}[t]
\scriptsize
\centering
\caption{
Clustering accuracy on the Frontal, BBT01, BUFFY02 and Notting Hill videos. We compare our results with three baseline features and five other state-of-the-art face clustering methods~\cite{wu2013simultaneous,cinbis2011unsupervised,wu2013constrained,xiao2014weighted,zhang2016joint}
based on the same face tracks input and metrics as in~\cite{wu2013simultaneous,wu2013constrained}.}
\vspace{-1mm}
%\begin{tabular} {lccccc}
\begin{tabular}{@{}lccccccccc@{}}
\toprule
\multirow{2}{*}{\textbf{Method}} &\multicolumn{2}{c}{Frontal}
& & \multicolumn{2}{c}{BBT01} & & \multicolumn{1}{c}{BUFFY02} & &\multicolumn{1}{c}{Notting Hill}\\
\cmidrule{2-3} \cmidrule{5-6} \cmidrule{8-8} \cmidrule{10-10}
& faces& tracklets & & faces & tracklets & & faces & & faces\\ \midrule
HOG~\cite{dalal2005histograms}&0.411&0.402 &
& 0.495 & 0.472 & & 0.304 & &0.451\\
AlexNet~\cite{krizhevsky2012imagenet}&0.591&0.435 &
& 0.716 & 0.698 & & 0.426 & &0.634\\
Pre-trained &0.777
&0.381 & & \underline{\color{blue}0.747} & \underline{\color{blue}0.775} & & 0.516 & &0.791\\ \midrule
Cinbis-ICCV-11~\cite{cinbis2011unsupervised}&0.844&0.861 & &0.581 & 0.565 && 0.416 & &0.732\\
Wu-CVPR-13~\cite{wu2013constrained}&0.950&0.907 & &0.626 & 0.596 && 0.503 & &0.844\\
Wu-ICCV-13~\cite{wu2013simultaneous} &0.950&0.907 & &0.665 & 0.668 && - & &-\\
Xiao-ECCV-14~\cite{xiao2014weighted}&\underline{\color{blue}0.962}&\underline{\color{blue}0.938} & &0.694 & 0.721 && 0.628 & &0.963\\
Zhang-ECCV-16~\cite{zhang2016joint} &-& &-&-&-&&\underline{\color{blue}0.921} & &\textbf{\color{red}0.990}\\ \midrule
Ours-SymTriplet-Contx & \textbf{\color{red}0.998} & \textbf{\color{red}0.998} &
& \textbf{\color{red}0.946} & \textbf{\color{red}0.982} && \textbf{\color{red}0.926} & &\underline{\color{blue}0.980}\\
\bottomrule
\end{tabular}
\label{tab:cluster_eval}
\vspace{-3mm}
\end{table*}

\begin{figure}[t]
\setlength{\abovecaptionskip}{0.cm}
\setlength{\belowcaptionskip}{-3mm}
\centering
\includegraphics[width=.5\linewidth]{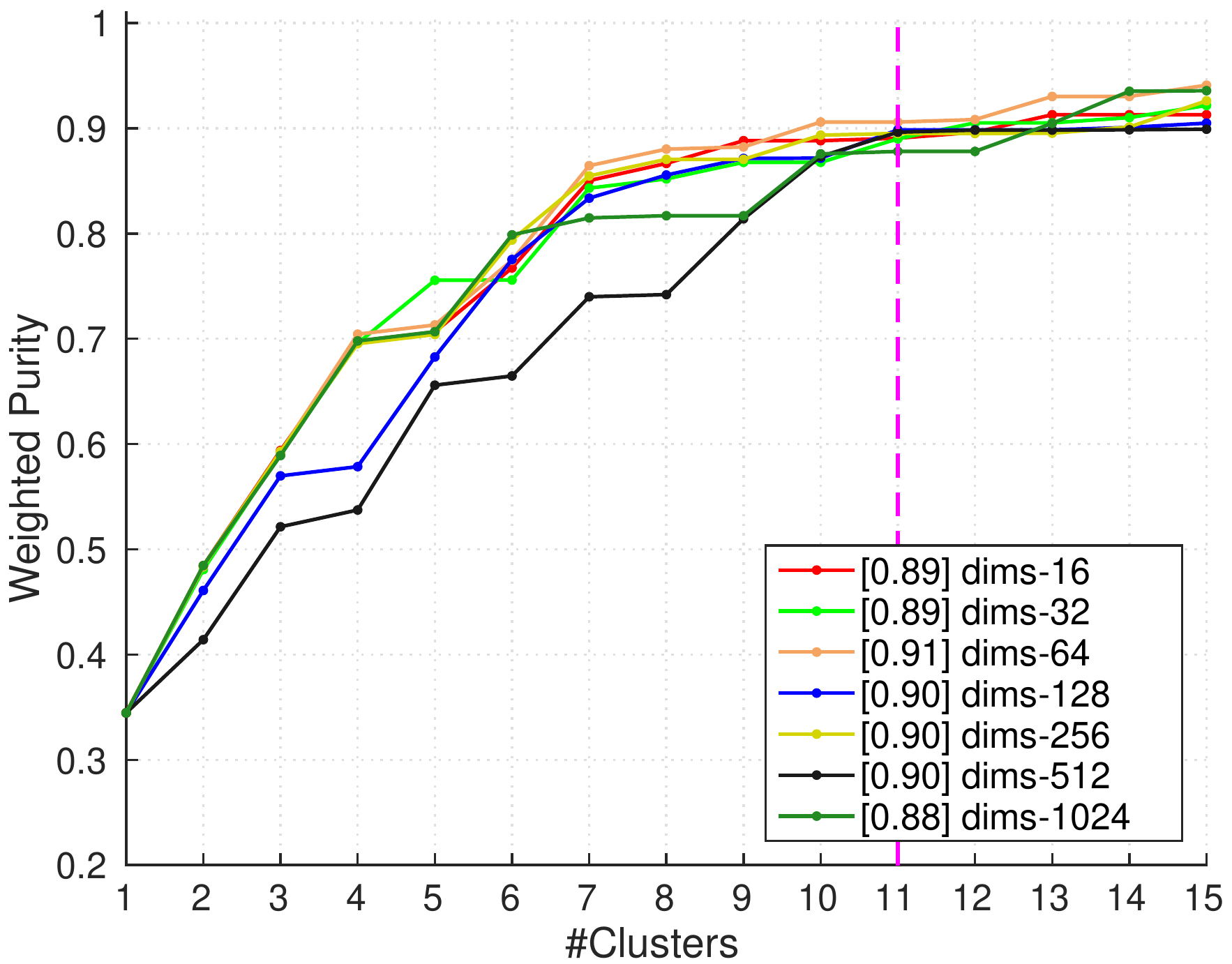}
\caption{
\textbf{Effect of feature dimensionality}. The legend shows the weighted purity over the number of clusters on \textsc{Bruno Mars} sequence.
}
\label{fig:FeaDims}
\vspace{-3mm}
\end{figure}

\vspace{1mm}
\noindent \textbf{Comparisons with different number of feature dimensions.}
We investigate the effect of the dimensionality of the embedded features.
Figure~\ref{fig:FeaDims} shows the clustering purity versus the number of clusters in comparison with a different number of feature dimensions on the sequence \textsc{Bruno Mars}.
In general, the clustering accuracy is not very sensitive to the selection of feature dimension.
However, we do observe that using large feature dimension (e.g., 512 and 1024) does not perform well compared to smaller ones.
We attribute this to the insufficient training samples.
The evaluation on feature dimension also validates the selection of using 64-dimensional features for accuracy and efficiency.

\vspace{-3mm}
\subsection{Multi-face Tracking}
\vspace{-1mm}

\vspace{1mm}
\noindent \textbf{Comparisons with the state-of-the-art multi-target trackers.}
We compare the proposed algorithm with several state-of-the-art MTT trackers including modified versions of TLD~\cite{kalal2012tracking}, ADMM~\cite{ayazoglu2012fast}, IHTLS~\cite{dicle2013way}, and methods by Wu et al.~\cite{wu2013simultaneous,wu2013constrained}.
The TLD~\cite{kalal2012tracking} scheme is a long-term single-target tracker which can re-detect targets of interest when targets leave and re-enter a scene.
We implement two extensions of TLD for multi-face tracking.
The first one is the mTLD scheme where in each sequence, we run multiple TLD trackers for all targets and each TLD tracker is initialized with the ground truth bounding box in the first frame.
For the second extension of TLD, we integrate the mTLD into our framework (referred to as Ours-mTLD).
We use the mTLD to generate shot-level trajectories within each shot instead of using
the two-threshold and Hungarian algorithms.
At the beginning of each shot, we initialize TLD trackers with untracked detections and link the detections in the following frames according to the overlap scores with TLD outputs.

Table~\ref{tab:MTT} shows quantitative results of the proposed algorithm, the mTLD~\cite{kalal2012tracking}, ADMM~\cite{ayazoglu2012fast}, and IHTLS~\cite{dicle2013way} on the BBT, BUFFY and music video datasets.
We also show the tracking results with the pre-trained features without adapting to a specific video.
Note that the results shown in Table~\ref{tab:MTT} are based on the overall evaluation.
% JB: Make sure that you do so.
We leave the results from each individual sequence in the supplementary material
(see \url{http://vllab1.ucmerced.edu/~szhang/FaceTracking/}).

The mTLD method does not perform well on both datasets in terms of recall, precision, F1, and MOTA metrics.
The ADMM~\cite{ayazoglu2012fast} and IHTLS~\cite{dicle2013way} schemes often generate numerous identity switches and fragments as both methods do not re-identify persons well when abrupt camera motions or shot changes occur.
The tracker with the pre-trained features is not effective to re-identify faces in different shots and achieve low MOTA.
The Ours-mTLD scheme has more IDS and Frag than the Ours-SymTriplet method.
The shot-level trajectories determined by the mTLD method are short and noisy since TLD trackers sometimes drift or do not perform well when large appearance changes occur.
In contrast, both Ours-SymTriplet and Ours-SymTriplet-Contx perform well
in terms of precision, F1, and MOTA metrics, with significantly fewer identity switches and fragments.

\begin{table*}[t]
\centering
\scriptsize
\caption{
Quantitative comparison with other state-of-the-art multi-target tracking methods on the BBT and music video datasets. }
\vspace{-1mm}
\begin{tabular}{@{}lcccccccc@{}}
\toprule
\multicolumn{9}{c}{BBT dataset} \\
\midrule
\textbf{Method} &\textbf{Recall}$\uparrow$ &\textbf{Precision}$\uparrow$ &\textbf{F1}$\uparrow$ &\textbf{FAF}$\downarrow$ &\textbf{IDS}$\downarrow$ &\textbf{Frag}$\downarrow$ &\textbf{MOTA}$\uparrow$ &\textbf{MOTP}$\uparrow$ \\ \midrule
mTLD~\cite{kalal2012tracking}
&1.1\%&8.1\%&1.9\%&\textbf{\color{red}0.18}&\textbf{\color{red}8}&\textbf{\color{red}83}
&-11.2\%&73.2\%\\
ADMM~\cite{ayazoglu2012fast}
&\textbf{\color{red}78.3\%}&56.8\%&65.8\%&0.49&2709&4623&39.5\%&72.7\%\\
IHTLS~\cite{dicle2013way}
&\underline{\color{blue}77.7\%}&63.4\%&69.8\%&0.49
&2648&4496&39.2\%&72.7\%\\
Pre-trained
&45.0\%&76.8\%&56.8\%&\underline{\color{blue}0.19}
&908&\underline{\color{blue}2435}&30.0\%&\textbf{\color{red}77.9\%}\\ \midrule
Ours-mTLD
&63.7\%&78.8\%&70.5\%&{0.24}
&1224&{3487}&44.6\%&\underline{\color{blue}77.6\%}\\
Ours-Siamese
& 74.5\%& \underline{\color{blue}81.4\%}& 77.8\% & {0.24} &884 &4051 &{56.1\%}& {77.4\%} \\
Ours-Triplet
& 76.2\%& 80.2\%& {78.1\%}& 0.27 &944 &4223 &55.8\%&77.3\% \\
Ours-SymTriplet
&76.6\%& {81.0\%}& \underline{\color{blue}78.7\%}& 0.26 &{846} &4261 &\underline{\color{blue}57.2\%}& 77.2\% \\
Ours-SymTriplet-Contx
&76.8\%& \textbf{\color{red}81.7\%}& \textbf{\color{red}79.2\%}& 0.23 &\underline{\color{blue}817} &4073 &\textbf{\color{red}59.6\%}& \underline{\color{blue}77.6\%}
\\
\bottomrule
\toprule
\multicolumn{9}{c}{BUFFY dataset} \\ \midrule
\textbf{Method} &\textbf{Recall}$\uparrow$ &\textbf{Precision}$\uparrow$ &\textbf{F1}$\uparrow$ &\textbf{FAF}$\downarrow$ &\textbf{IDS}$\downarrow$ &\textbf{Frag}$\downarrow$ &\textbf{MOTA}$\uparrow$ &\textbf{MOTP}$\uparrow$\\ \midrule
mTLD~\cite{kalal2012tracking}
&4.6\%&21.5\% &7.6\% &0.32
&\textbf{\color{red}192}&\textbf{\color{red}453}&-8.2\%&{69.1\%} \\
ADMM~\cite{ayazoglu2012fast}
&\textbf{\color{red}78.3\%}&64.9\% &70.9\%&0.37
&1420&2445&31.6\%&\underline{\color{blue}70.1\%} \\
IHTLS~\cite{dicle2013way}
&\underline{\color{blue}78.0\%}&68.9\%&\underline{\color{blue}73.2\%}&0.30
&1558&\underline{\color{blue}2424}&38.1\%&\textbf{\color{red}70.2\%} \\
Pre-trained
&52.3\%&72.1\%&60.6\%&\textbf{\color{red}0.12}
&405&{2672}&38.3\%&{68.8\%} \\ \midrule
Ours-mTLD
&65.4\%&73.5\%&69.2\%&0.27
&413&2503&45.3\%&\textbf{\color{red}70.2\%}\\
Ours-Siamese
&66.1\%& {74.3\%}&{70.0\%}& {0.19}& 389 &{2470} &{45.6\%}& \textbf{\color{red}70.2\%} \\
Ours-Triplet
&{67.3\%}& 74.6\%& 70.8\%& {0.20}& 388 &2462 &47.4\%& \textbf{\color{red}70.2\%} \\
Ours-SymTriplet
&{68.1\%}& \underline{\color{blue}74.7\%}& {71.2\%}& {0.19}& {363} &2460 &\underline{\color{blue}{47.6\%}}& \textbf{\color{red}70.2\%} \\
Ours-SymTriplet-Contx
&{70.9\%}& \textbf{\color{red}77.5\%}& \textbf{\color{red}74.0\%}& \underline{\color{blue}0.18}& \underline{\color{blue}293} &{2446} &\textbf{\color{red}49.4\%}& \textbf{\color{red}70.2\%} \\
\bottomrule
\toprule
\multicolumn{9}{c}{Music video dataset} \\ \midrule
\textbf{Method} &\textbf{Recall}$\uparrow$ &\textbf{Precision}$\uparrow$ &\textbf{F1}$\uparrow$ &\textbf{FAF}$\downarrow$ &\textbf{IDS}$\downarrow$ &\textbf{Frag}$\downarrow$ &\textbf{MOTA}$\uparrow$ &\textbf{MOTP}$\uparrow$\\\midrule
mTLD~\cite{kalal2012tracking}
&9.7\%&36.1\% &15.3\% &0.39
&\textbf{\color{red}280}&\textbf{\color{red}621}&-7.7\%&{68.4\%} \\
ADMM~\cite{ayazoglu2012fast}
&\textbf{\color{red}75.5\%}&61.8\% &68.0\%&0.50
&2382&2959&51.7\%&63.7\% \\
IHTLS~\cite{dicle2013way}
&\textbf{\color{red}75.5\%}&68.0\%&71.6\%&0.41
&2013&2880&56.2\%&63.7\% \\
Pre-trained
&60.1\%&88.8\%&71.7\%&\textbf{\color{red}0.17}
&931&\underline{\color{blue}2140}&51.5\%&\underline{\color{blue}79.5\%} \\ \midrule
Ours-mTLD
&69.1\%&88.1\%&77.4\%&0.21
&1914&2786&57.7\%&\textbf{\color{red}80.1\%}\\
Ours-Siamese
&71.5\%& {89.4\%}&{79.5\%}& \underline{\color{blue}0.19}& 986 &{2512} &{62.3\%}& 64.0\% \\
Ours-Triplet
&{71.8\%}& 88.8\%& 79.4\%& {0.20}& 902 &2546 &61.8\%& 64.2\% \\
Ours-SymTriplet
&{71.8\%}& \underline{\color{blue}89.7\%}& \underline{\color{blue}79.8\%}& \underline{\color{blue}0.19}& {699} &2563 &\underline{\color{blue}62.8\%}& {64.3\%} \\
Ours-SymTriplet-Contx
&\underline{\color{blue}73.2\%}& \textbf{\color{red}90.5\%}& \textbf{\color{red}80.9\%}& \underline{\color{blue}0.19}& \underline{\color{blue}625} &2417 &\textbf{\color{red}64.1\%}& {64.2\%} \\
\bottomrule
\end{tabular}
\label{tab:MTT}
\vspace{-3mm}
\end{table*}

%\vspace{1mm}
\noindent \textbf{Contribution of linking tracklets.}
We evaluate the design choices of the two-step linking process:
1) linking tracklets within the shot and 2) linking shot-level tracklets across shots.
To this end, we evaluate two other alternatives:
\begin{itemize}
\item Ours-SymTriplet-noT: without linking tracklets \emph{within} each shot
\item Ours-SymTriplet-noC: without clustering the shot-level tracklets \emph{across} different shots
\end{itemize}

Table~\ref{tab:MTT2} shows that both Ours-SymTriplet-noT and Ours-SymTriplet-noC decrease the performance in terms of Recall, F1, IDS, Frag and MOTA metrics.
Without linking tracklets within each shot, Ours-SymTriplet-noT cannot recover several missed faces, and thus yields lower performance on Recall.
Although some separated tracklets in each shot can be grouped together by the HAC clustering algorithm, many tracklets may be grouped incorrectly due to the lack of consideration of spatio-temporal coherence within each shot.
This explains the increase of IDS and Frag.
Without clustering tracklets across different shots, Ours-SymTriplet-noC assigns each short tracklet in each shot with different identities, which results in significantly more identity switches and lower MOTA.

\begin{table*}[t]
\scriptsize
\centering
\caption{
Quantitative comparison with Ours-SymTriplet-noT and Ours-SymTriplet-noC on the BBT datasets. }
\vspace{-1mm}
\begin{tabular}{@{}lcccccccc@{}}
\toprule
\multicolumn{9}{c}{BBT dataset} \\
\midrule
\textbf{Method} &\textbf{Recall}$\uparrow$ &\textbf{Precision}$\uparrow$ &\textbf{F1}$\uparrow$ &\textbf{FAF}$\downarrow$ &\textbf{IDS}$\downarrow$ &\textbf{Frag}$\downarrow$ &\textbf{MOTA}$\uparrow$ &\textbf{MOTP}$\uparrow$ \\ \midrule
Ours-SymTriplet-Contx
&\textbf{\color{red}76.8\%}& \textbf{\color{red}81.7\%}& \textbf{\color{red}79.2\%}& \textbf{\color{red}0.23} &\textbf{\color{red}817} &\textbf{\color{red}4073} &\textbf{\color{red}59.6\%}& \textbf{\color{red}77.6\%} \\
Ours-SymTriplet-noT
&65.3\%& \underline{\color{blue}81.3\%}& {72.4\%}& \underline{\color{blue}0.24} &\underline{\color{blue}952} &5109 &{43.1\%}& \underline{\color{blue}77.4\%} \\
Ours-SymTriplet-noC
&\underline{\color{blue}68.7\%}& {81.2\%}& \underline{\color{blue}74.4\%}& 0.25 &{1496} &\underline{\color{blue}4082} &\underline{\color{blue}46.2\%}& 77.2\% \\
\bottomrule
\end{tabular}
\label{tab:MTT2}
\vspace{-3mm}
\end{table*}

% JB: Change the format to Figure~\ref{fig:samples1}. No need to use hard-coded width and hspace.
\begin{figure*}[t]
\scriptsize
\centering
\begin{tabular}{{c}{c}{c}{c}{c}}
\hspace{-1.0mm}\includegraphics[width=3.4cm]{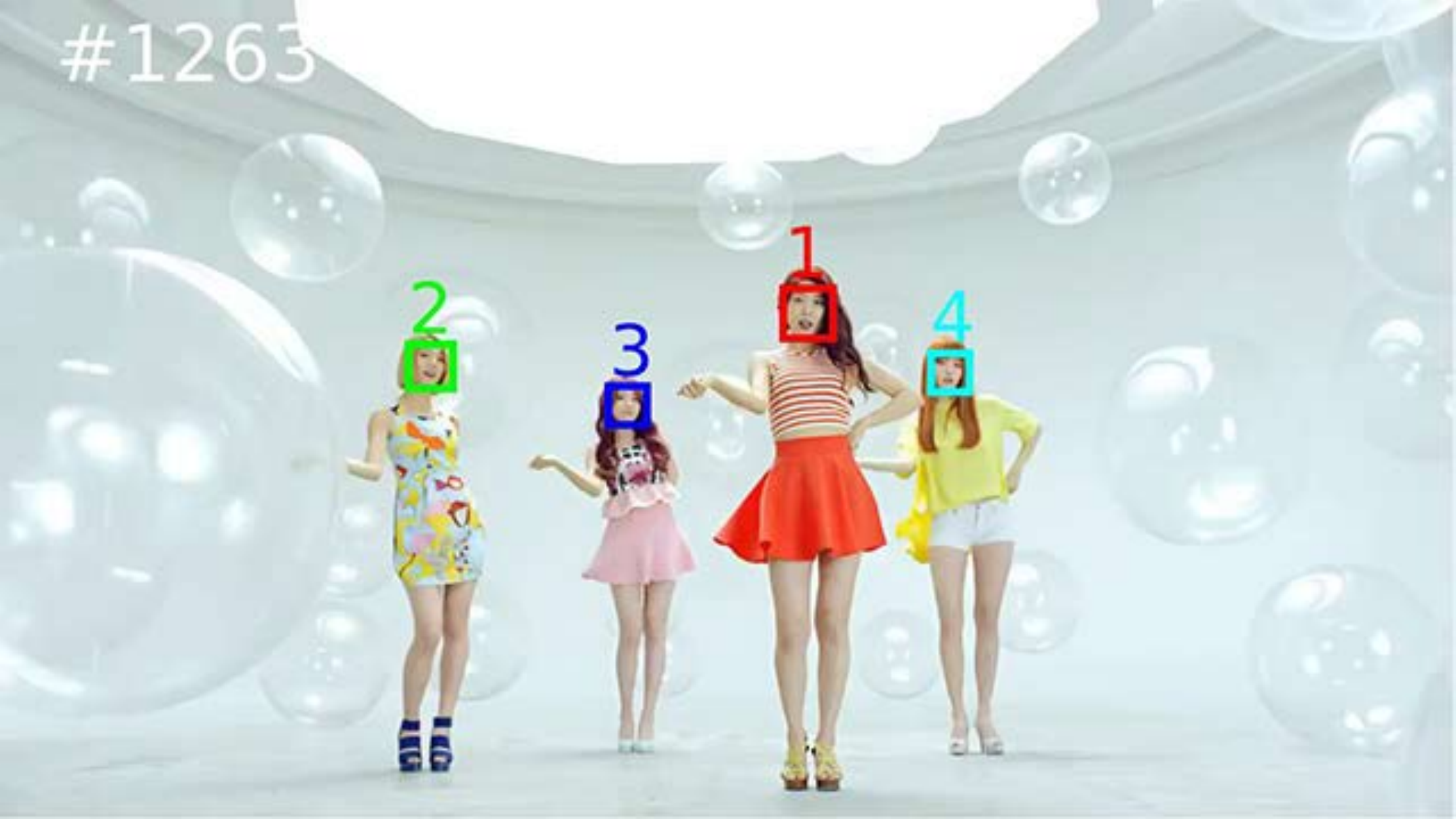} &
\hspace{-2.5mm}\includegraphics[width=3.4cm]{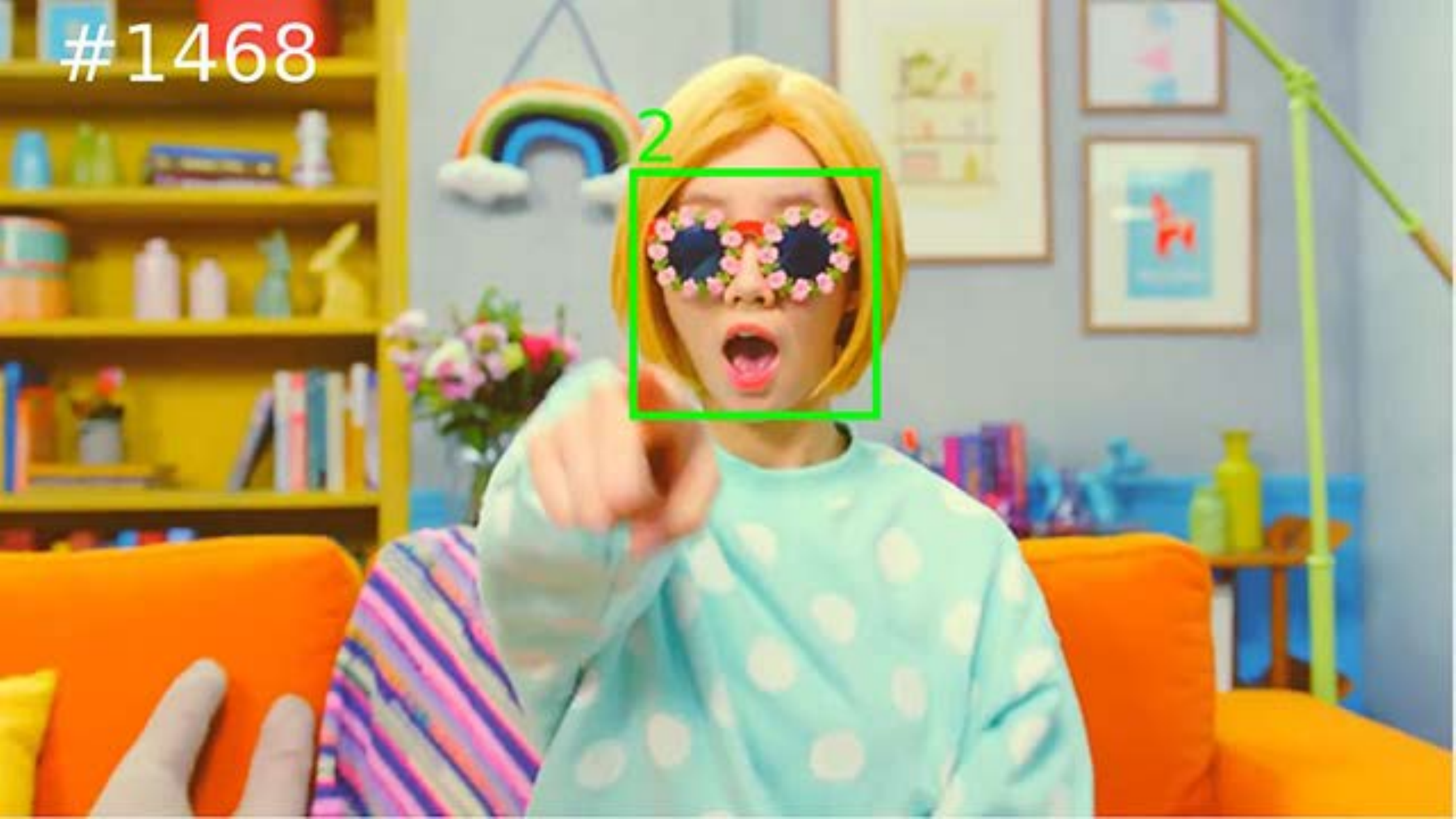} &
\hspace{-2.5mm}\includegraphics[width=3.4cm]{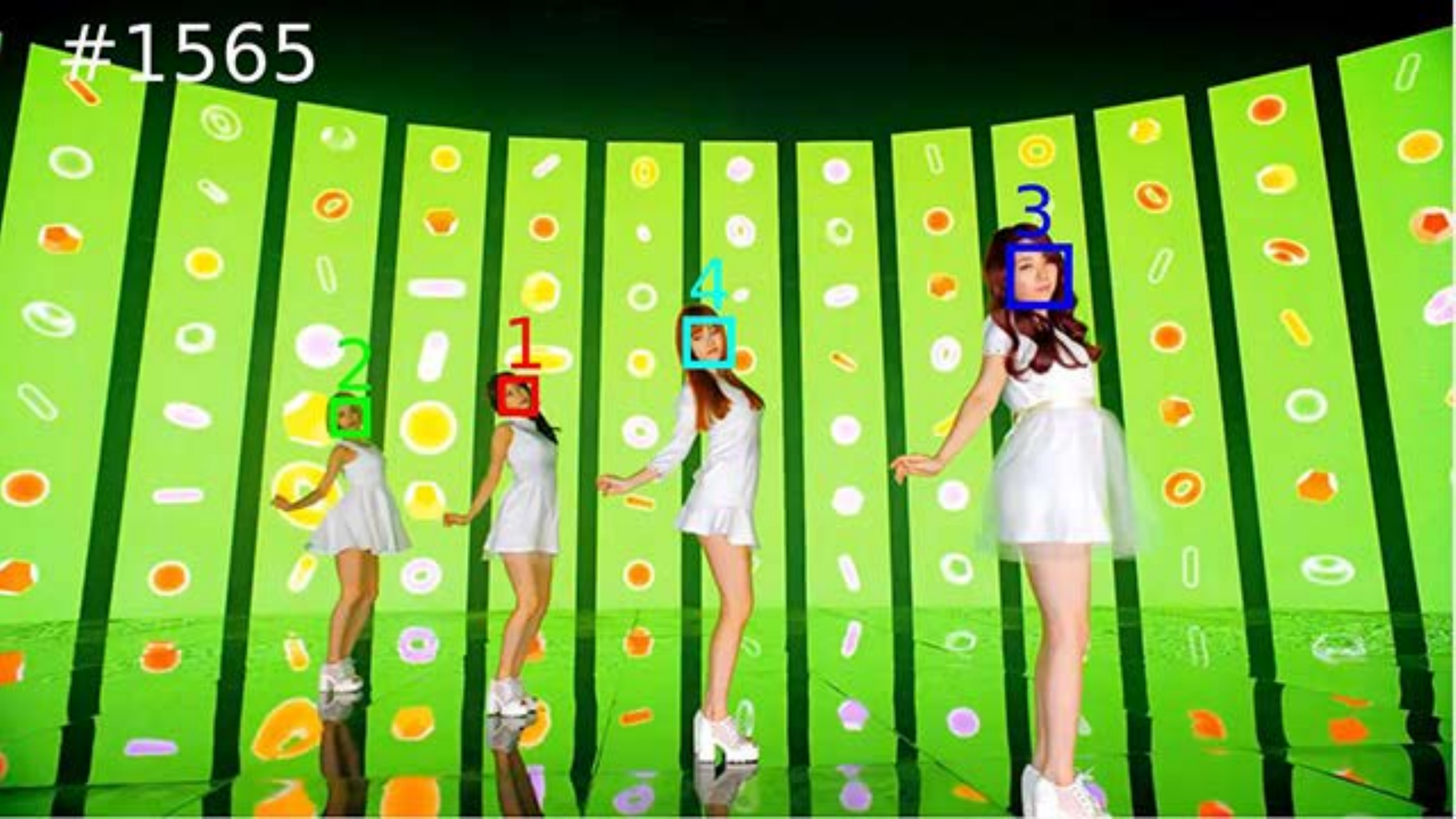} &
\hspace{-2.5mm}\includegraphics[width=3.4cm]{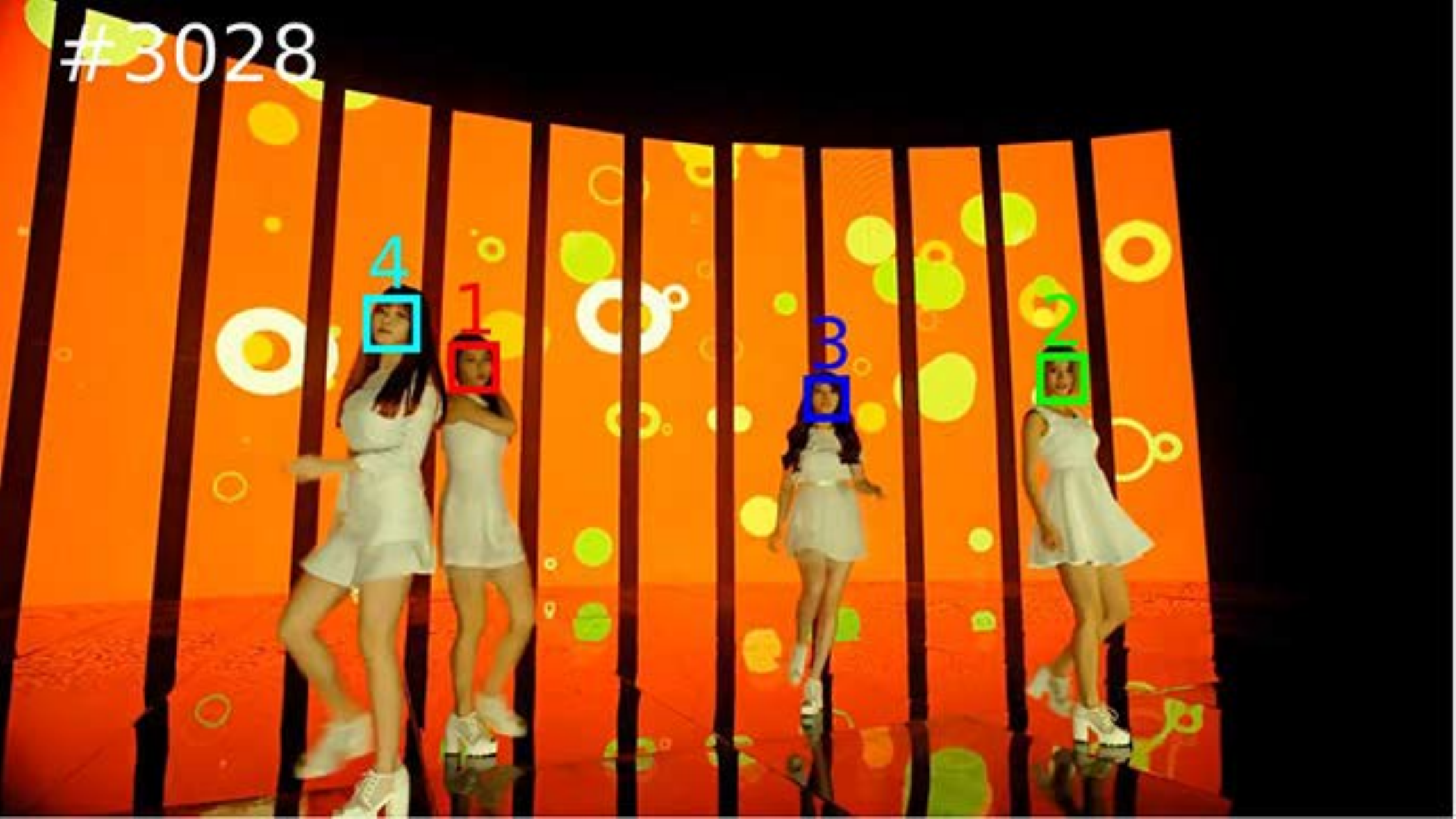} &
\hspace{-2.5mm}\includegraphics[width=3.4cm]{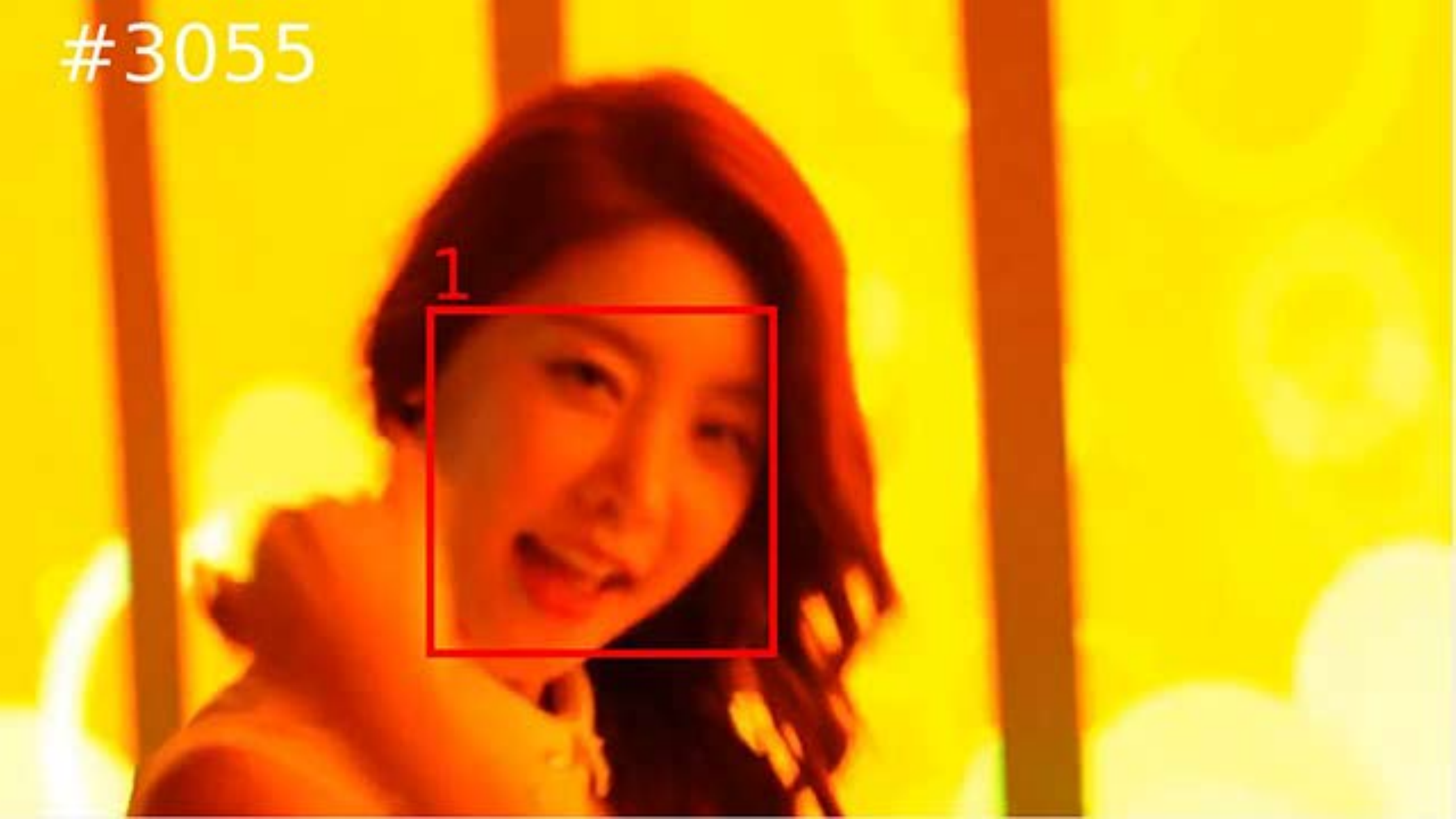} \\

\hspace{-1.0mm}\includegraphics[width=3.4cm]{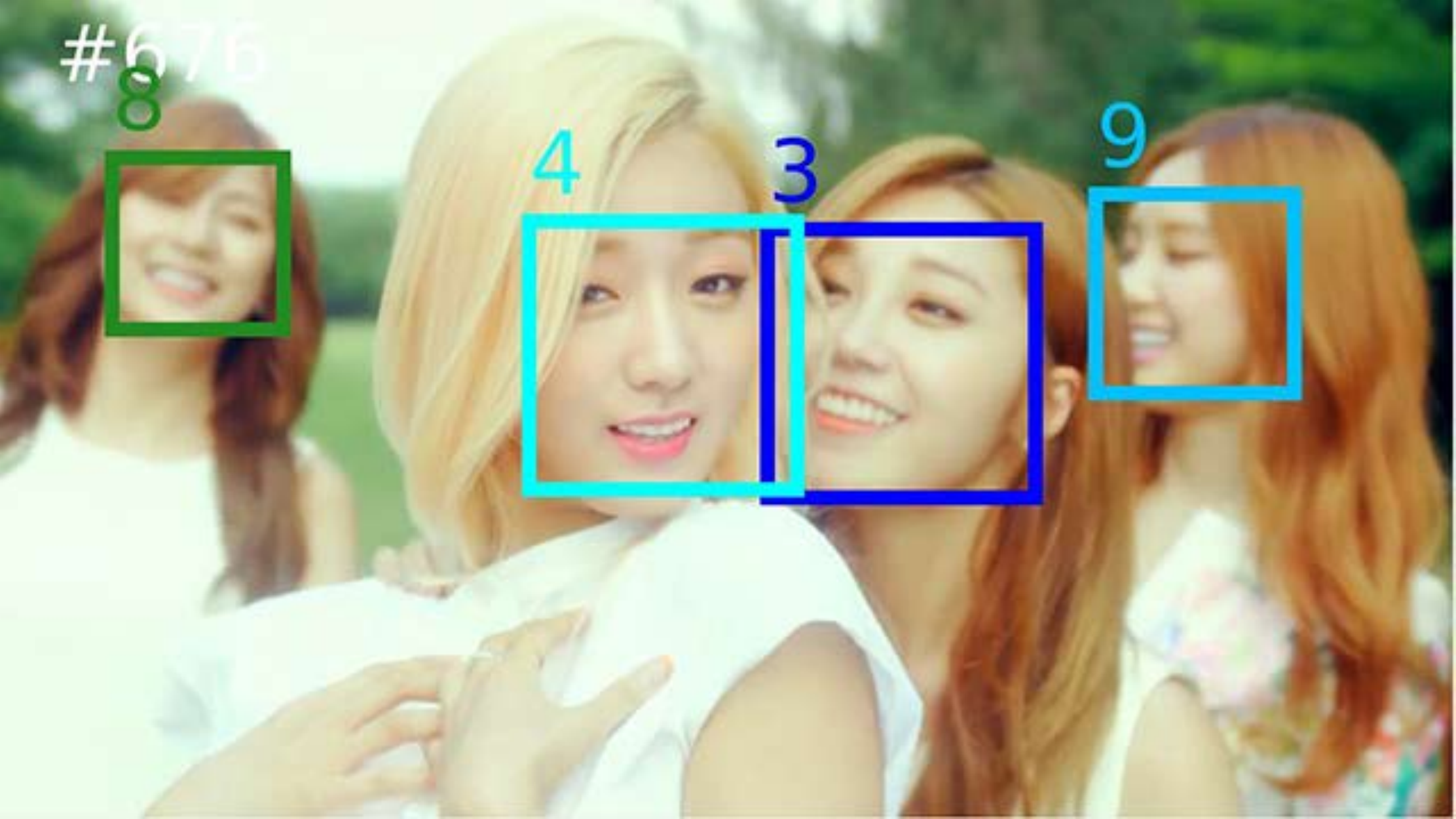} &
\hspace{-2.5mm}\includegraphics[width=3.4cm]{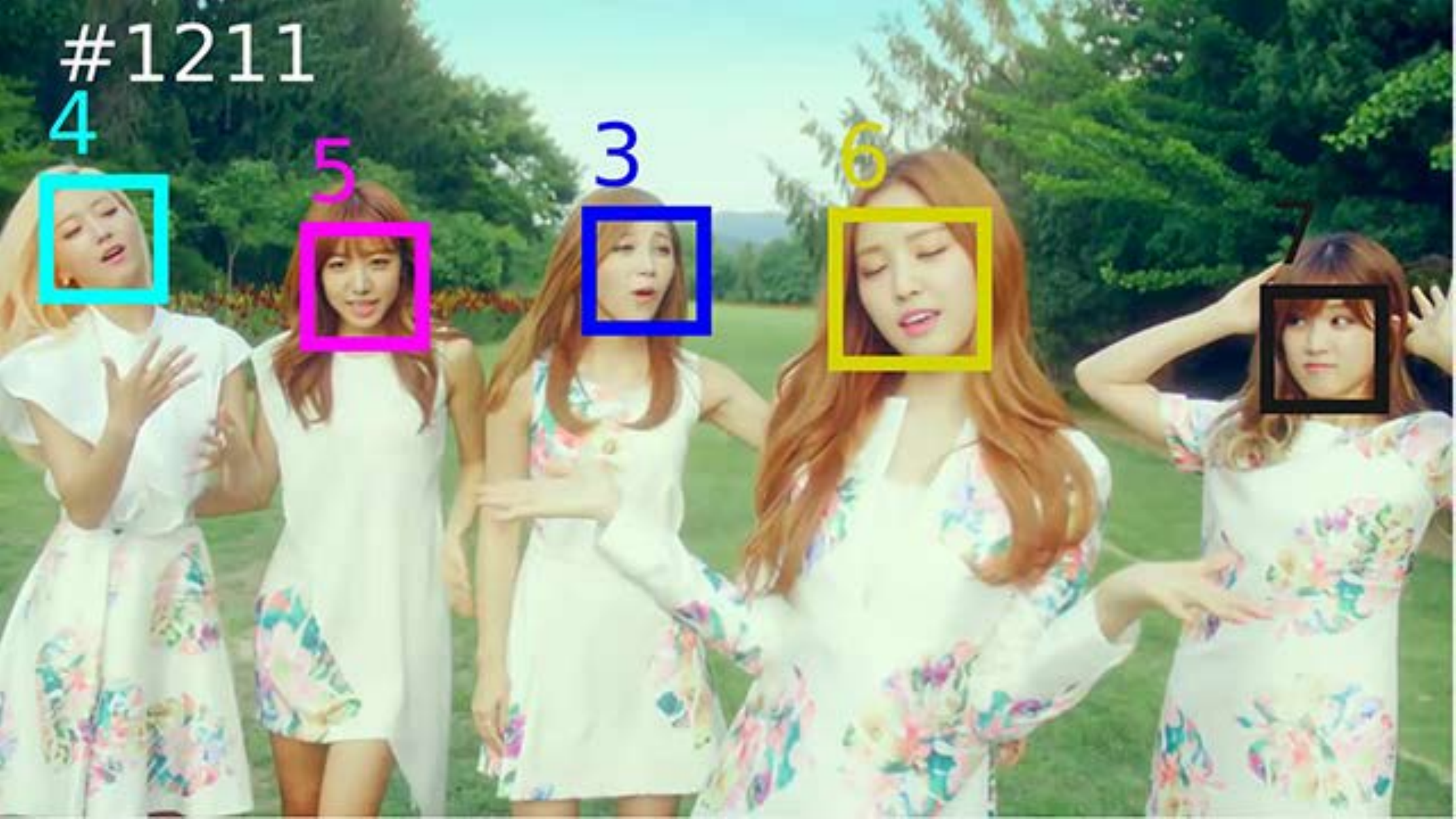} &
\hspace{-2.5mm}\includegraphics[width=3.4cm]{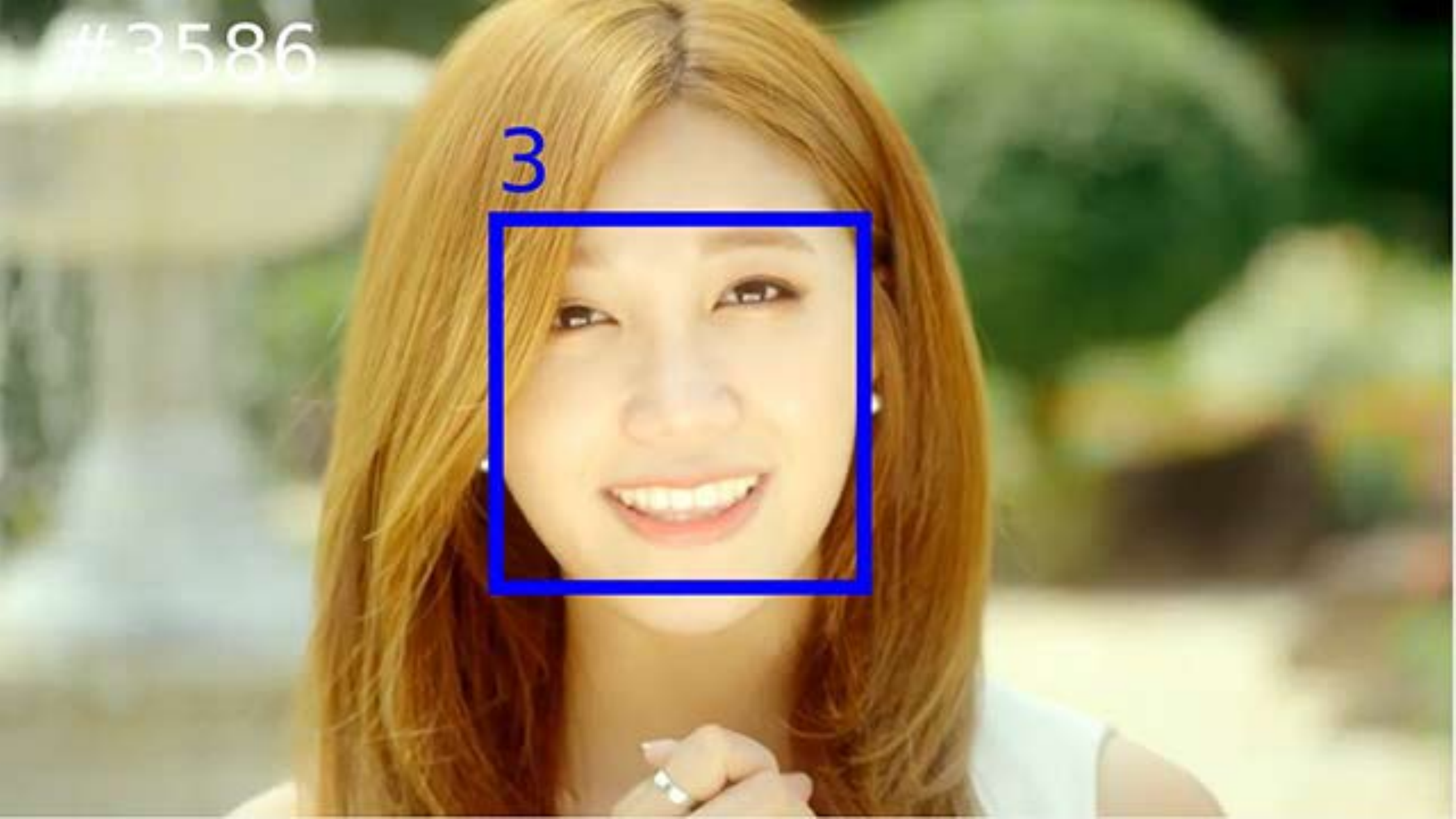} &
\hspace{-2.5mm}\includegraphics[width=3.4cm]{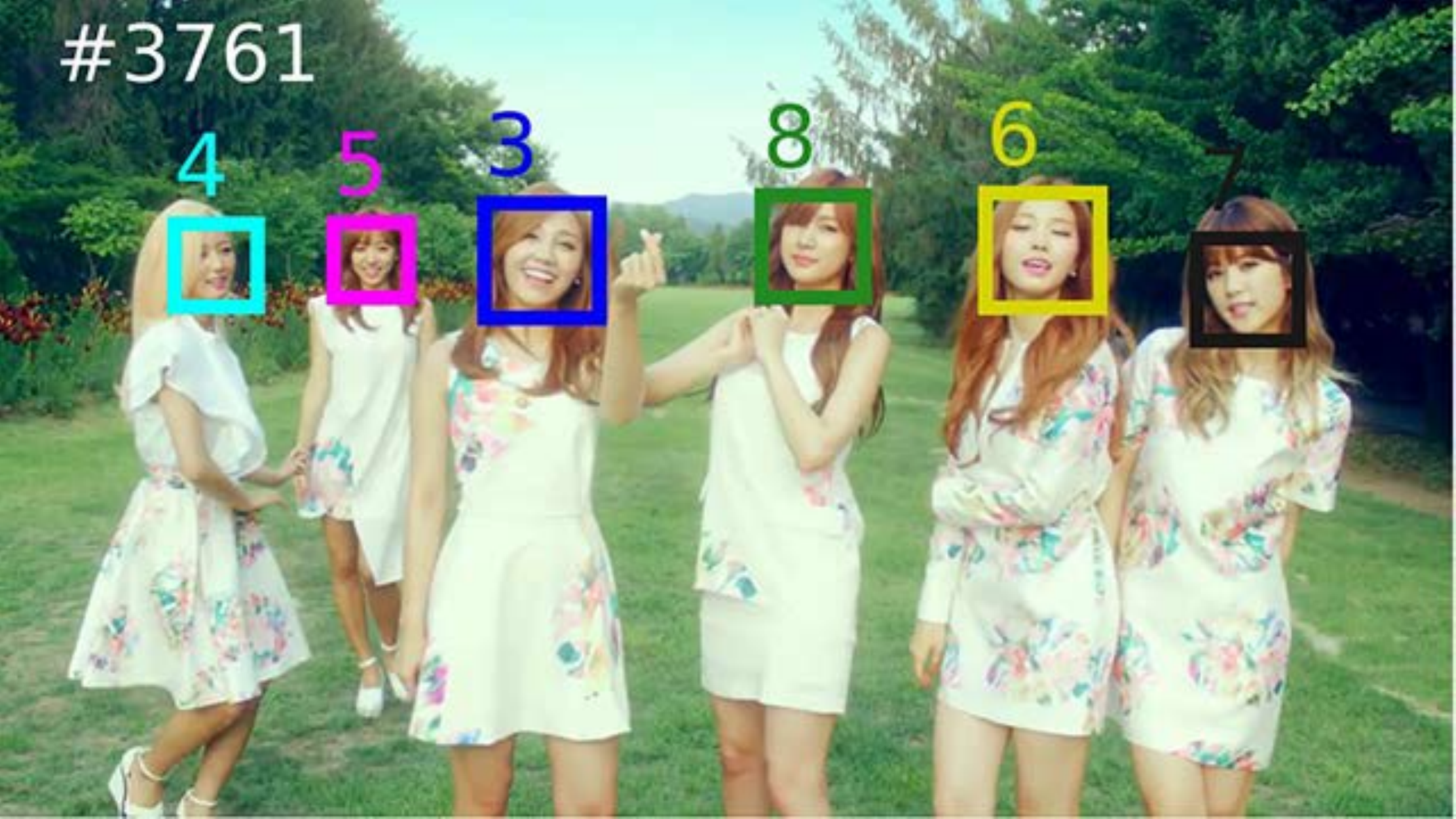} &
\hspace{-2.5mm}\includegraphics[width=3.4cm]{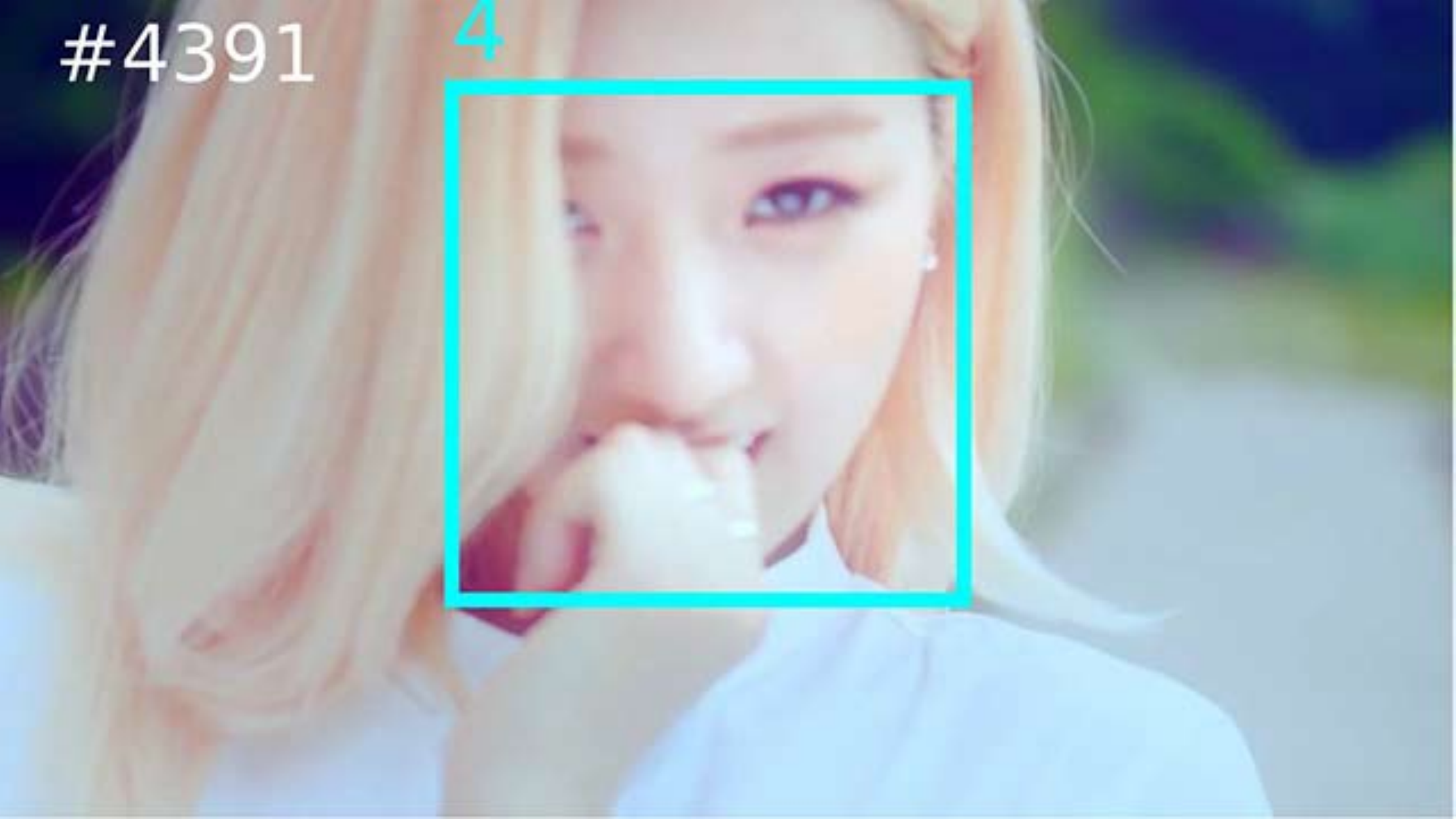}\\

\hspace{-1.0mm}\includegraphics[width=3.4cm]{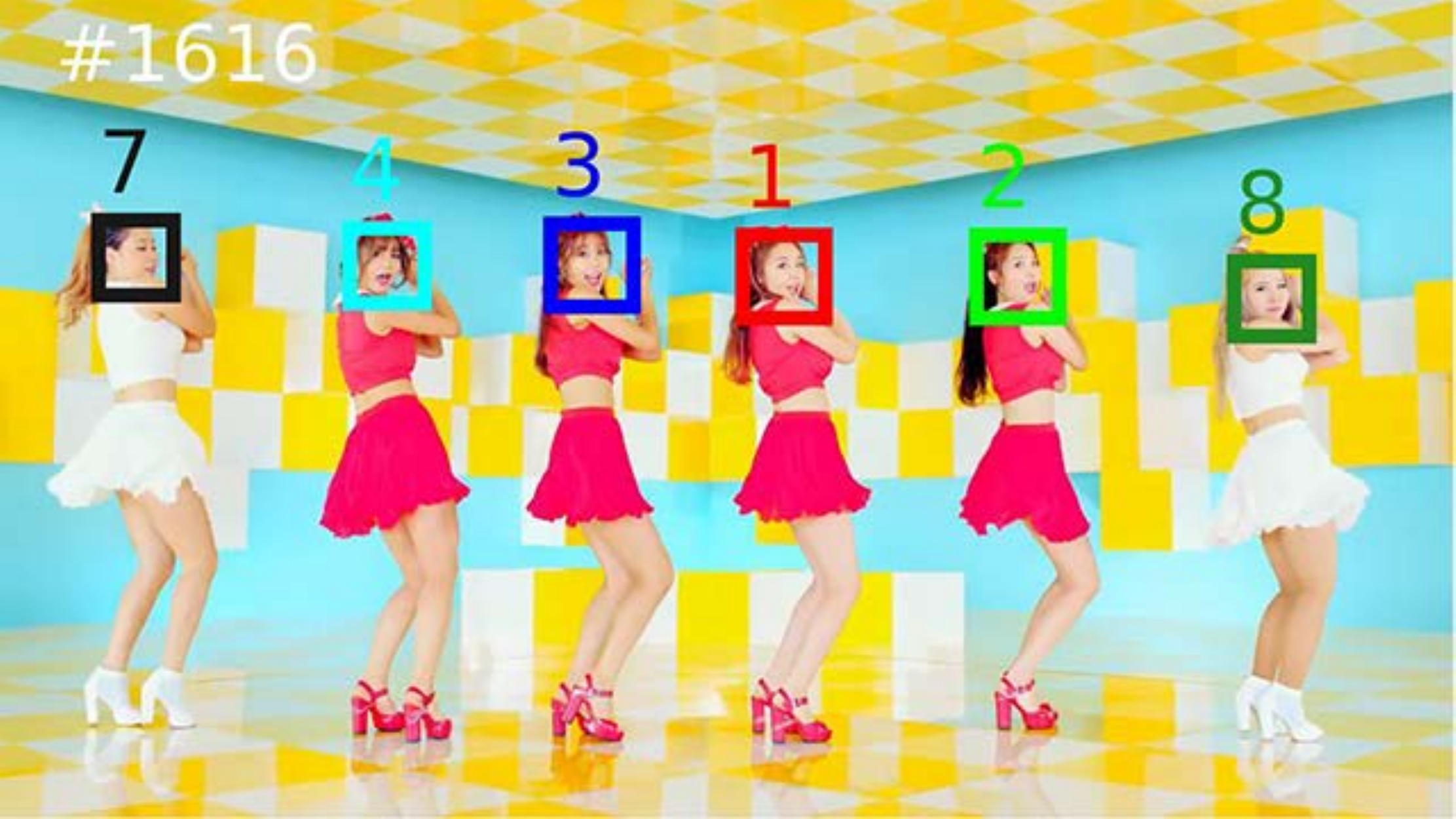} &
\hspace{-2.5mm}\includegraphics[width=3.4cm]{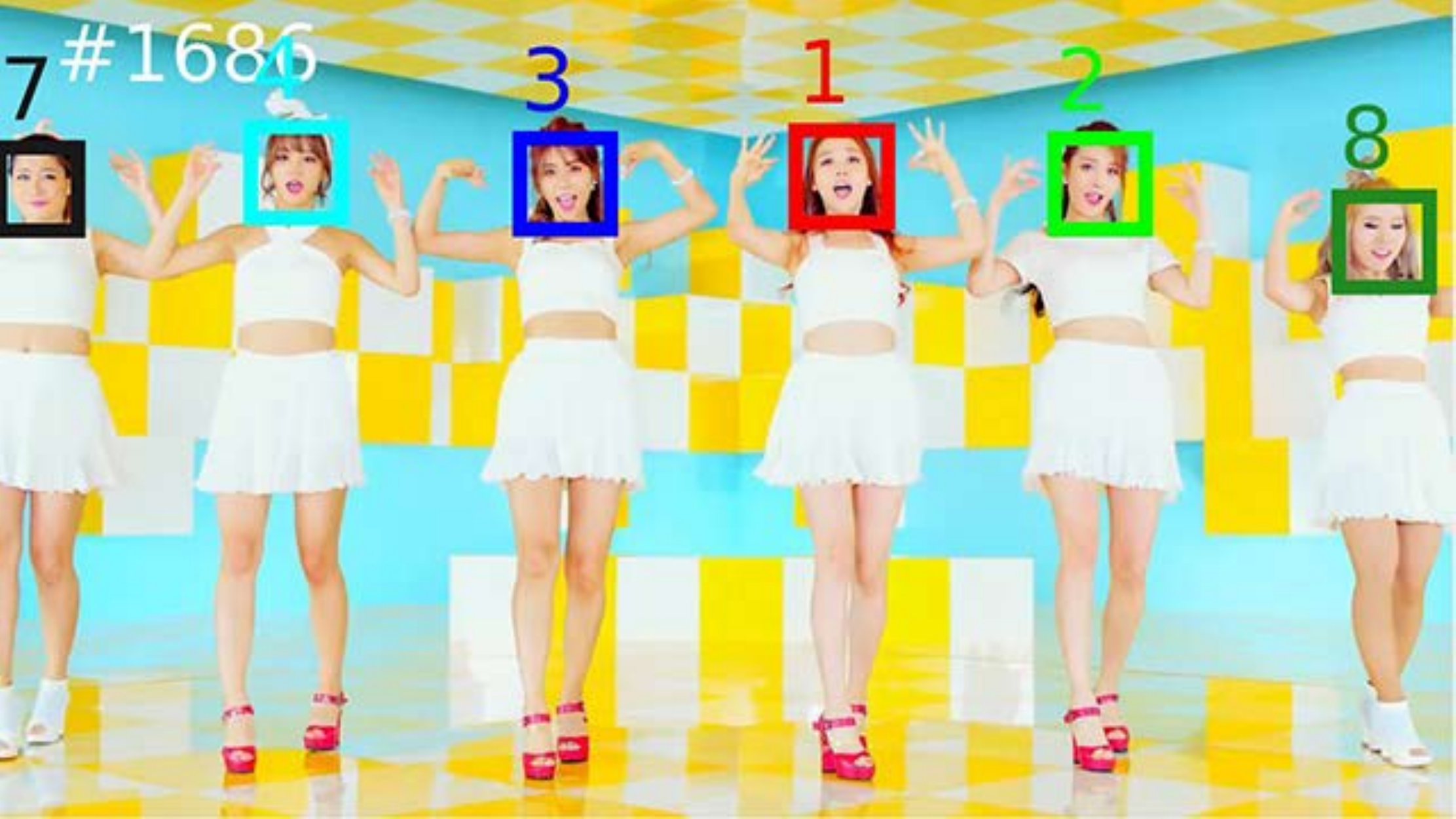} &
\hspace{-2.5mm}\includegraphics[width=3.4cm]{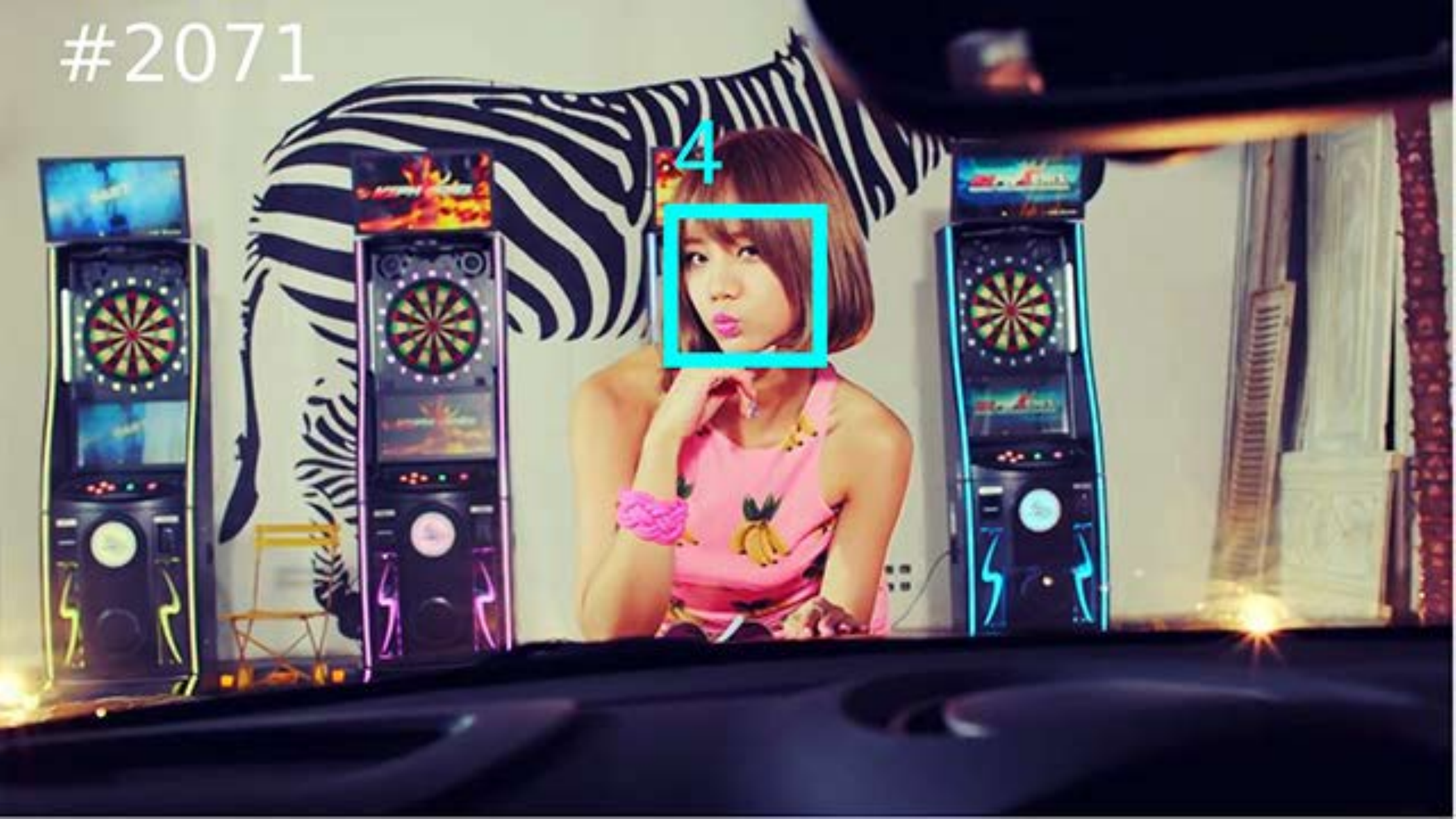} &
\hspace{-2.5mm}\includegraphics[width=3.4cm]{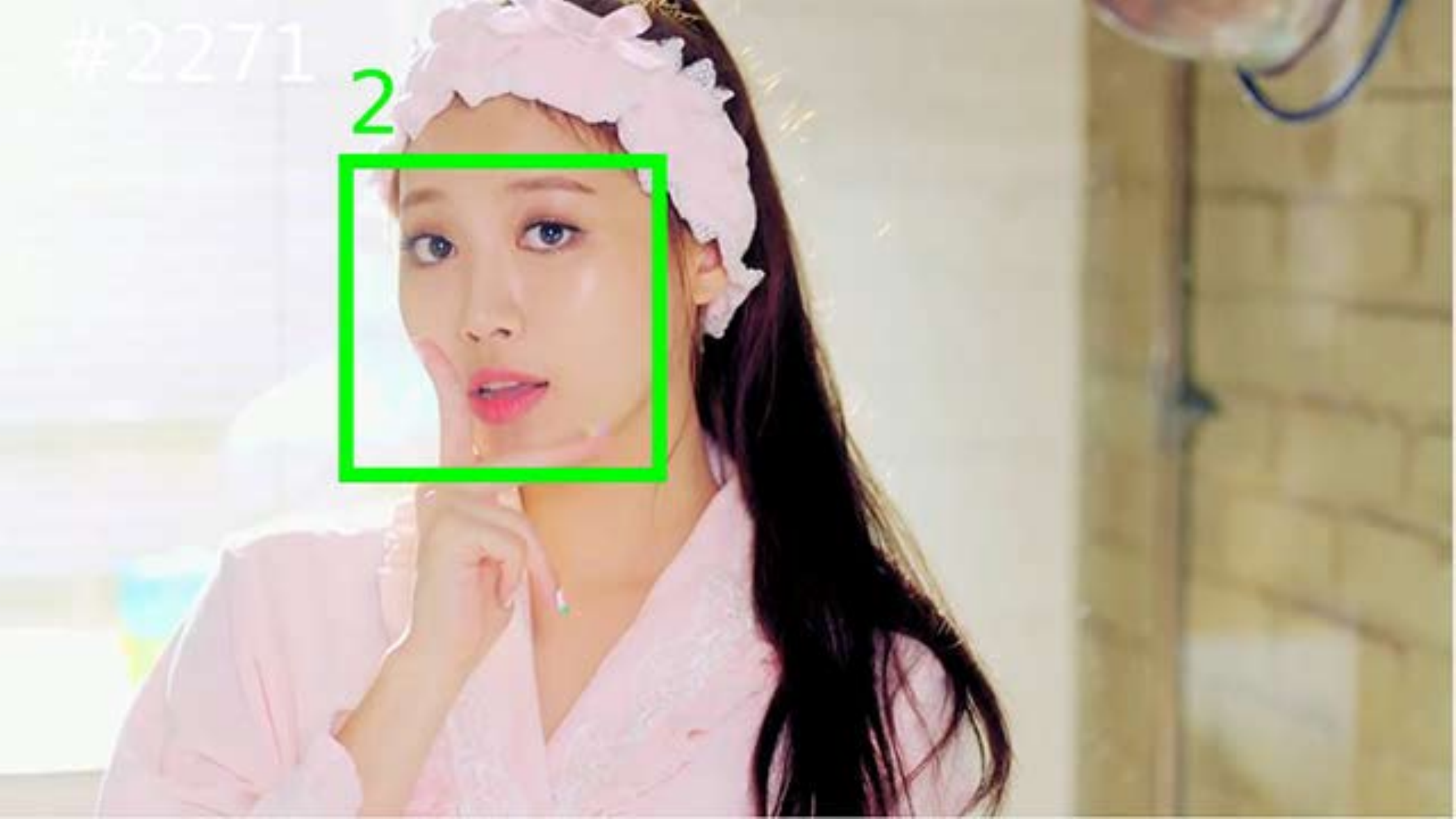} &
\hspace{-2.5mm}\includegraphics[width=3.4cm]{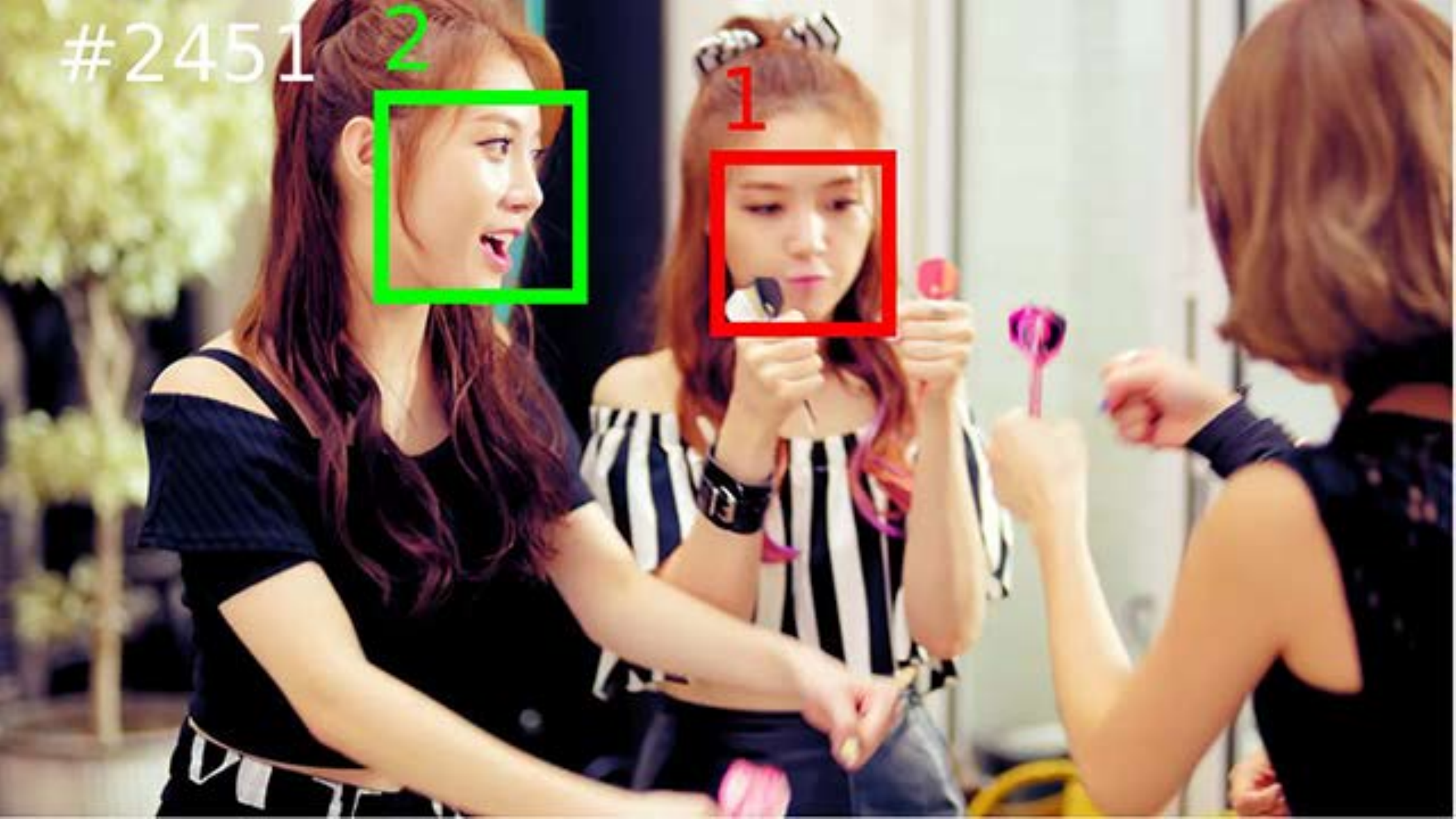}\\

\hspace{-1.0mm}\includegraphics[width=3.4cm]{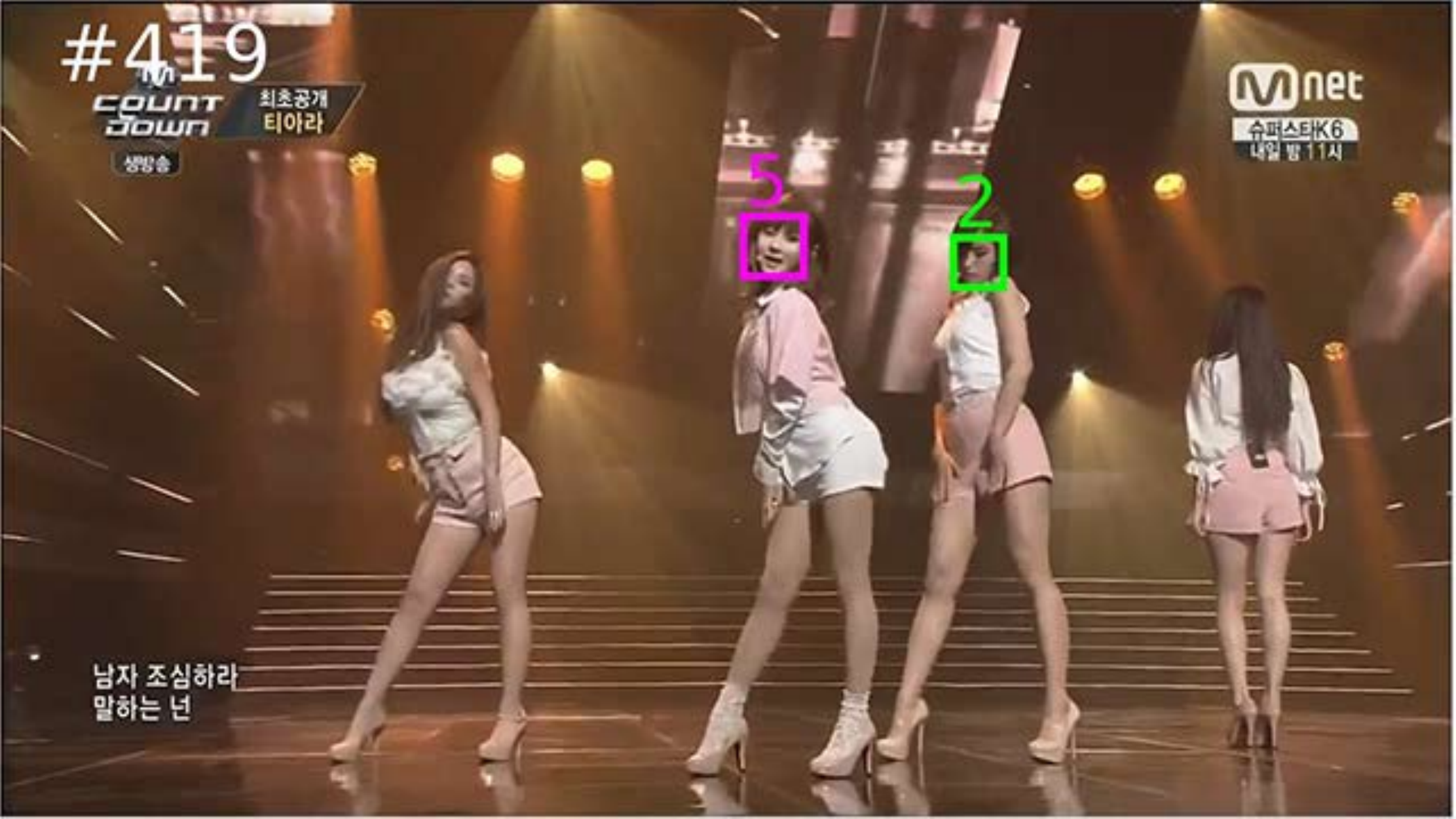} &
\hspace{-2.5mm}\includegraphics[width=3.4cm]{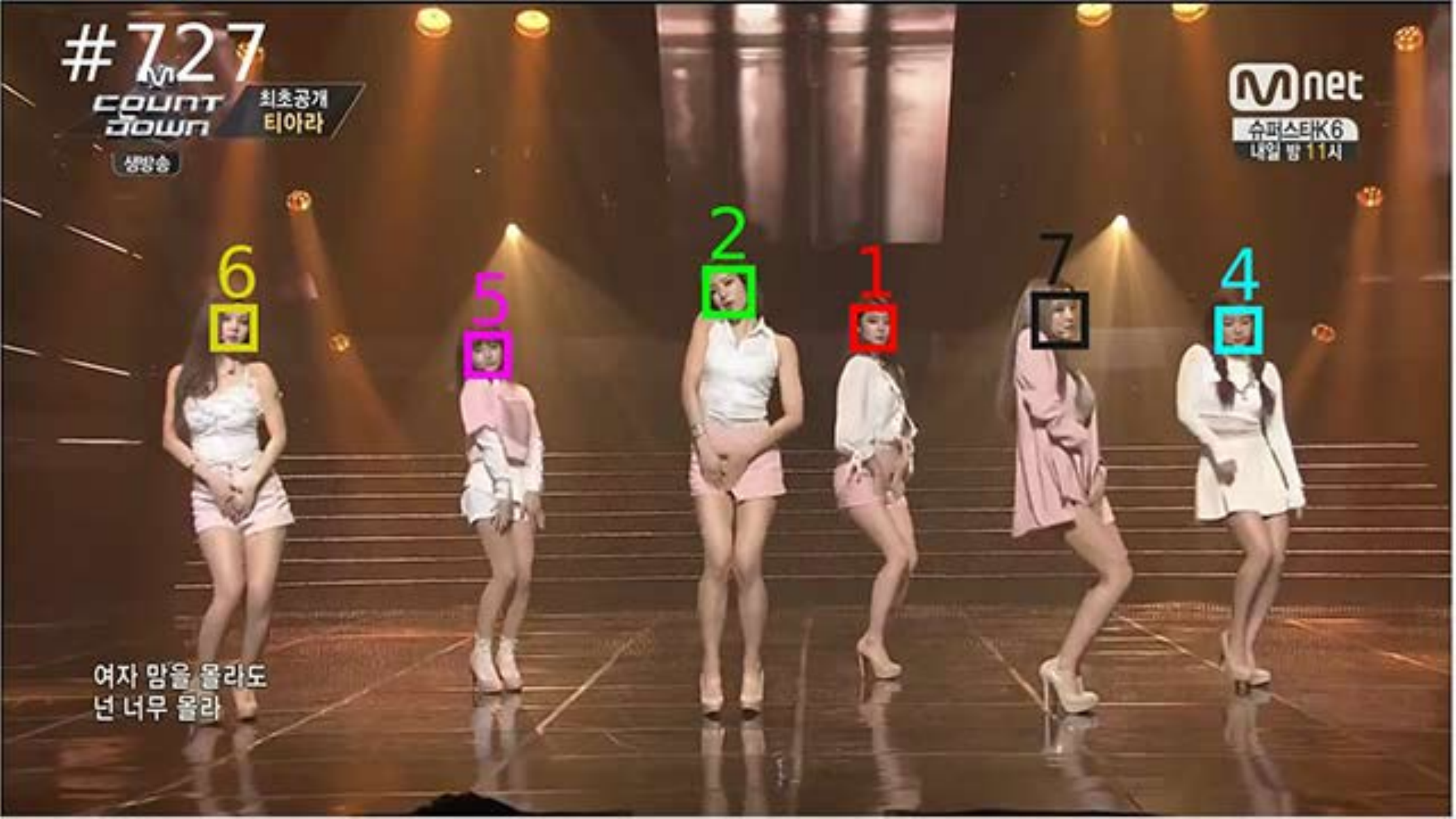} &
\hspace{-2.5mm}\includegraphics[width=3.4cm]{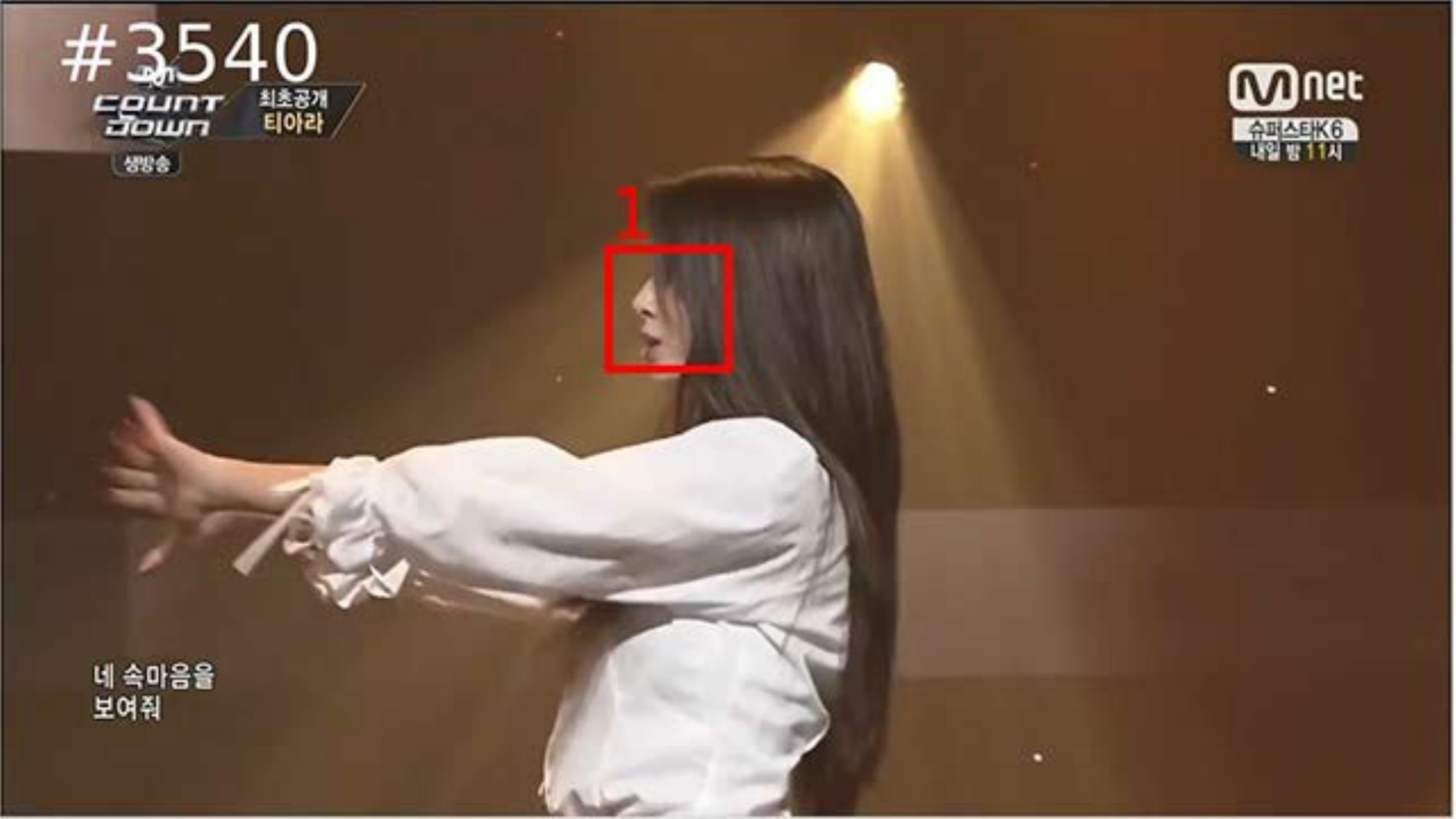} &
\hspace{-2.5mm}\includegraphics[width=3.4cm]{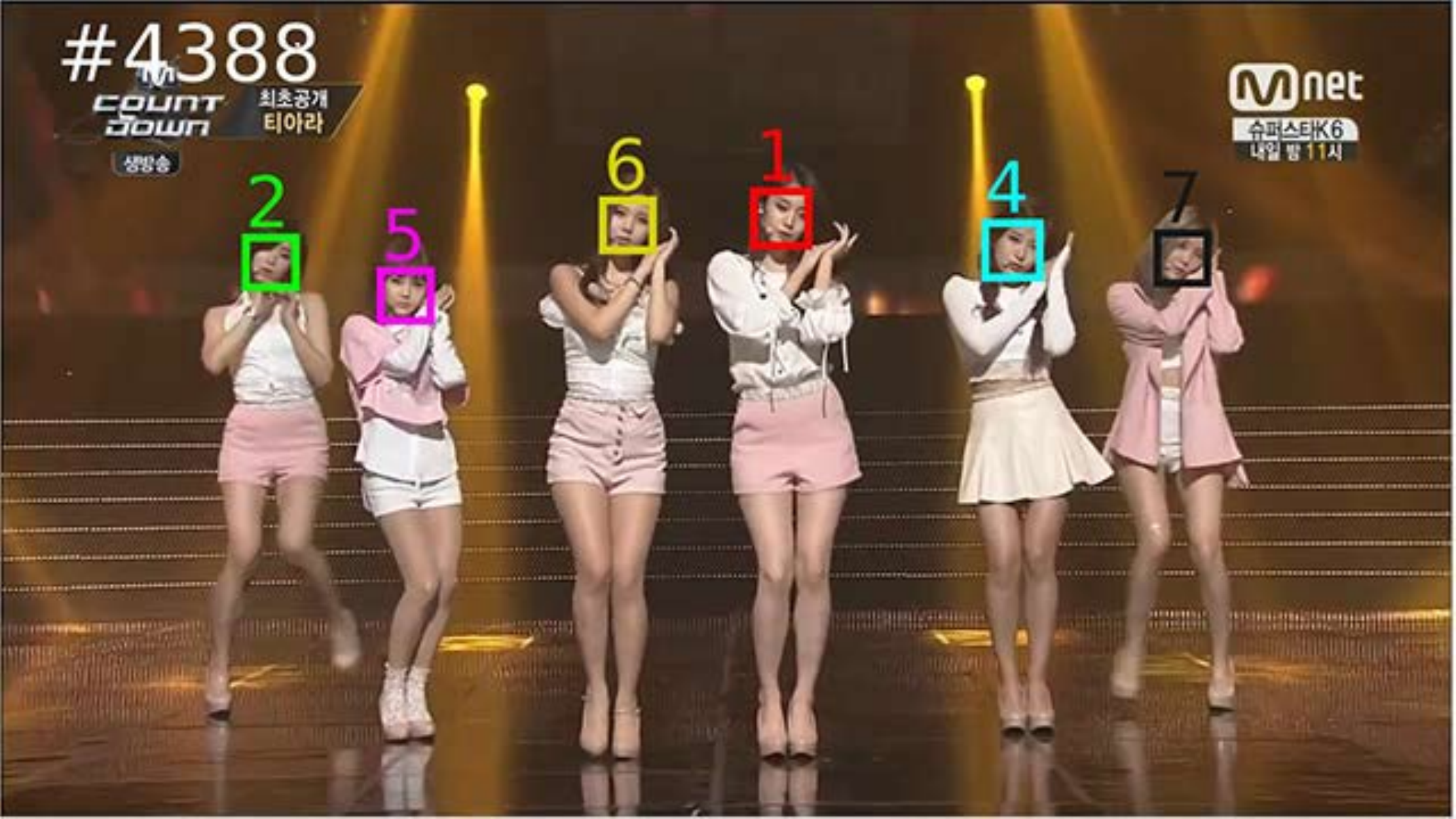} &
\hspace{-2.5mm}\includegraphics[width=3.4cm]{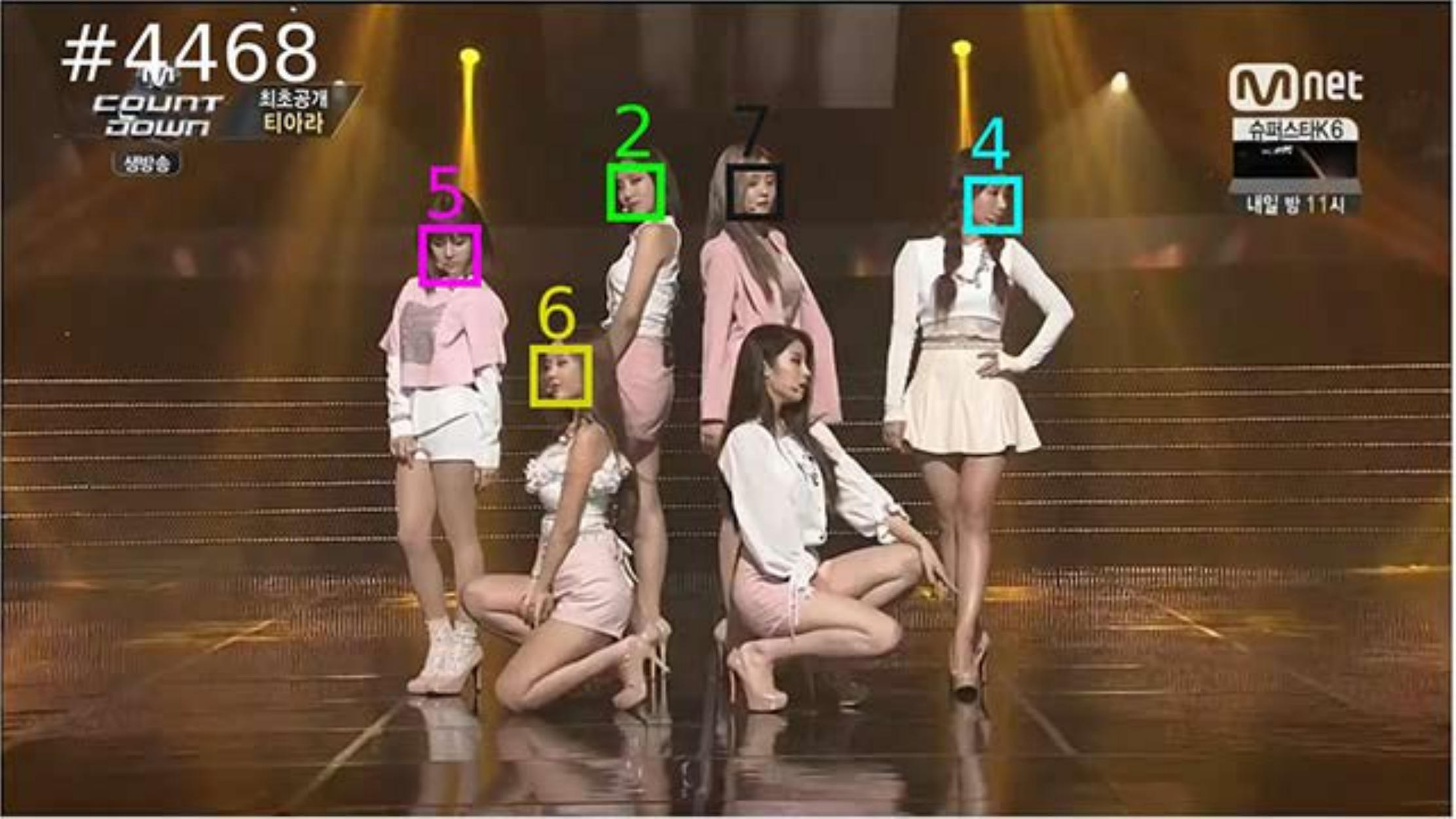} \\

\hspace{-1.0mm}\includegraphics[width=3.4cm]{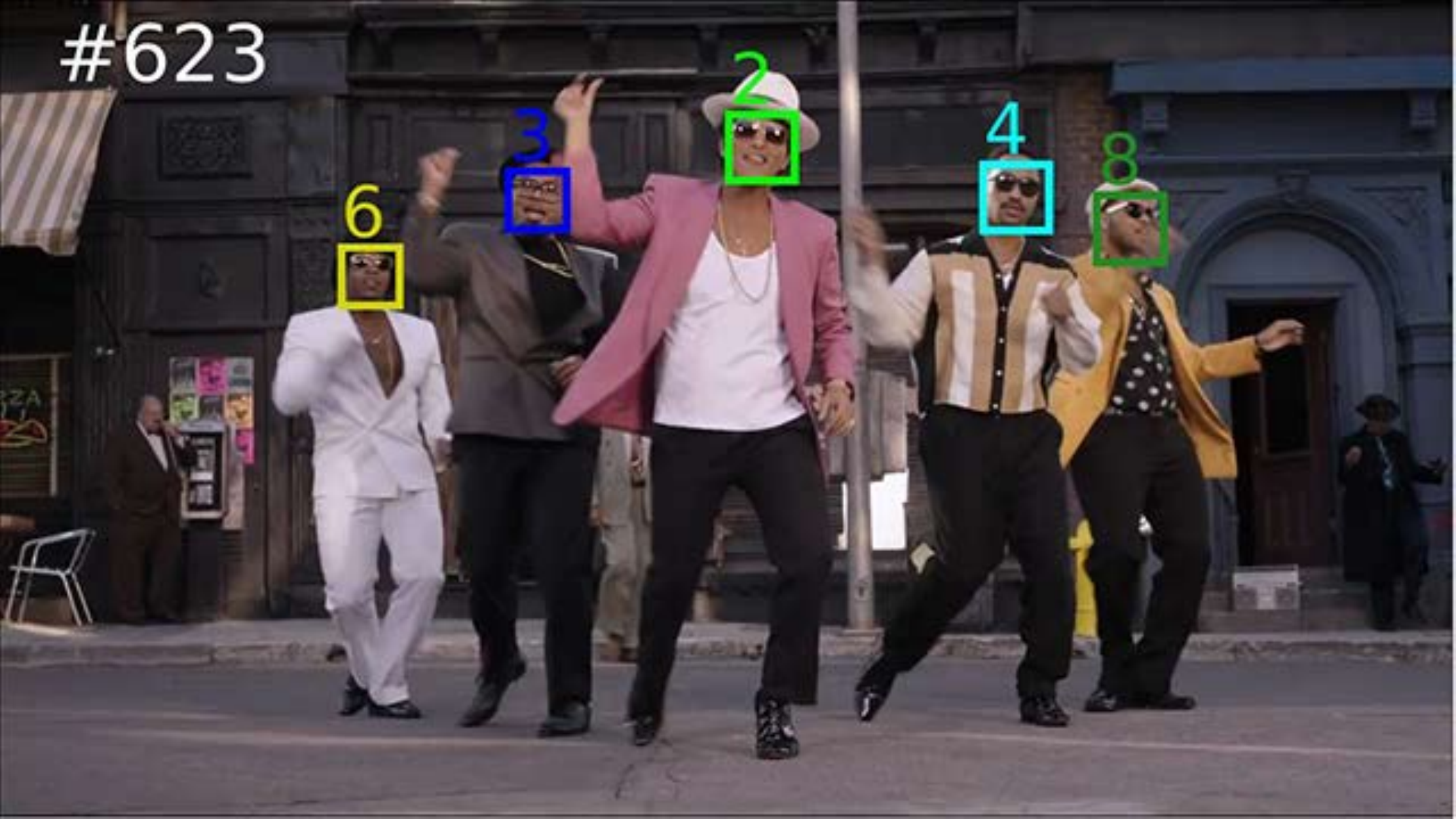} &
\hspace{-2.5mm}\includegraphics[width=3.4cm]{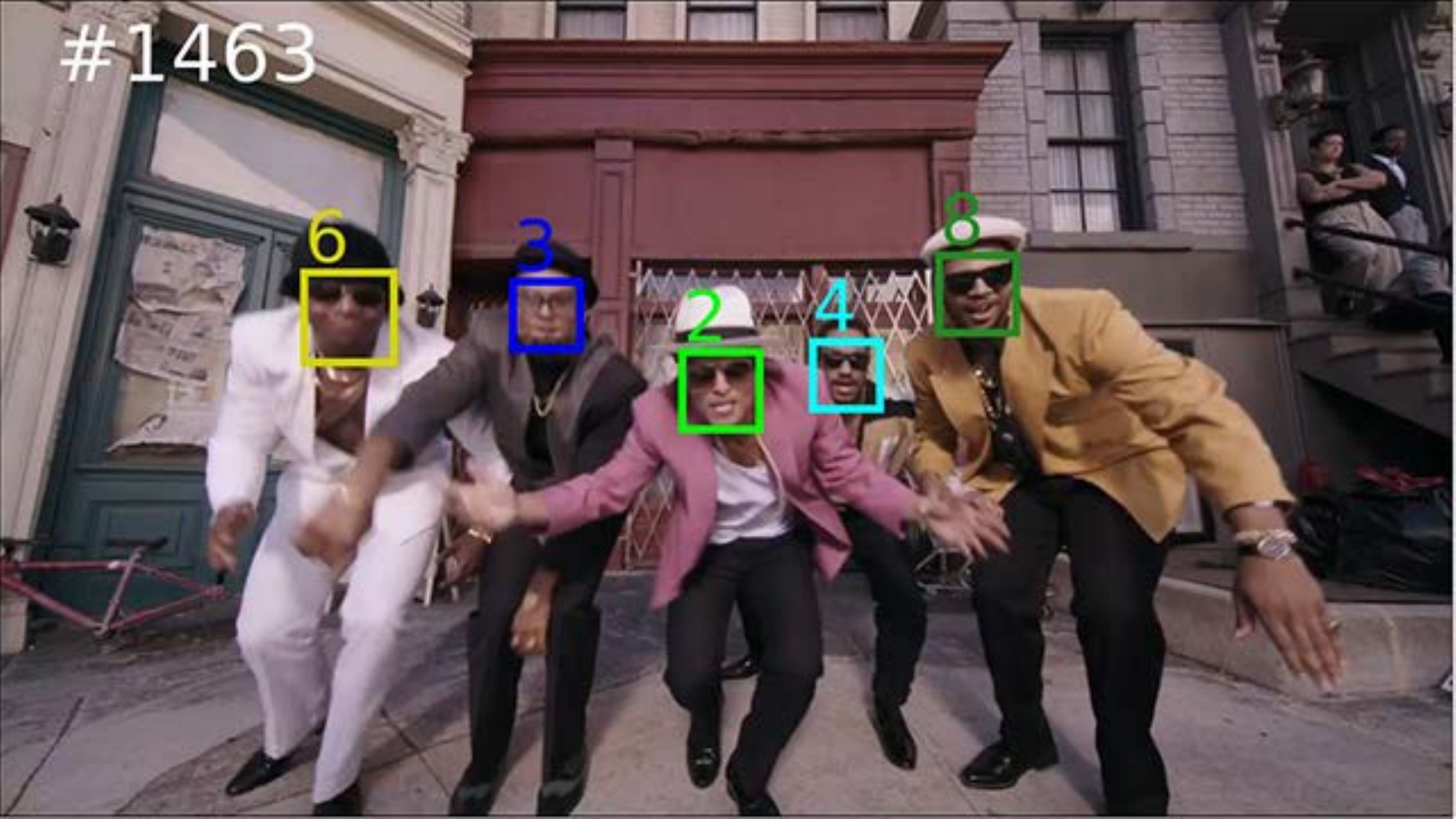} &
\hspace{-2.5mm}\includegraphics[width=3.4cm]{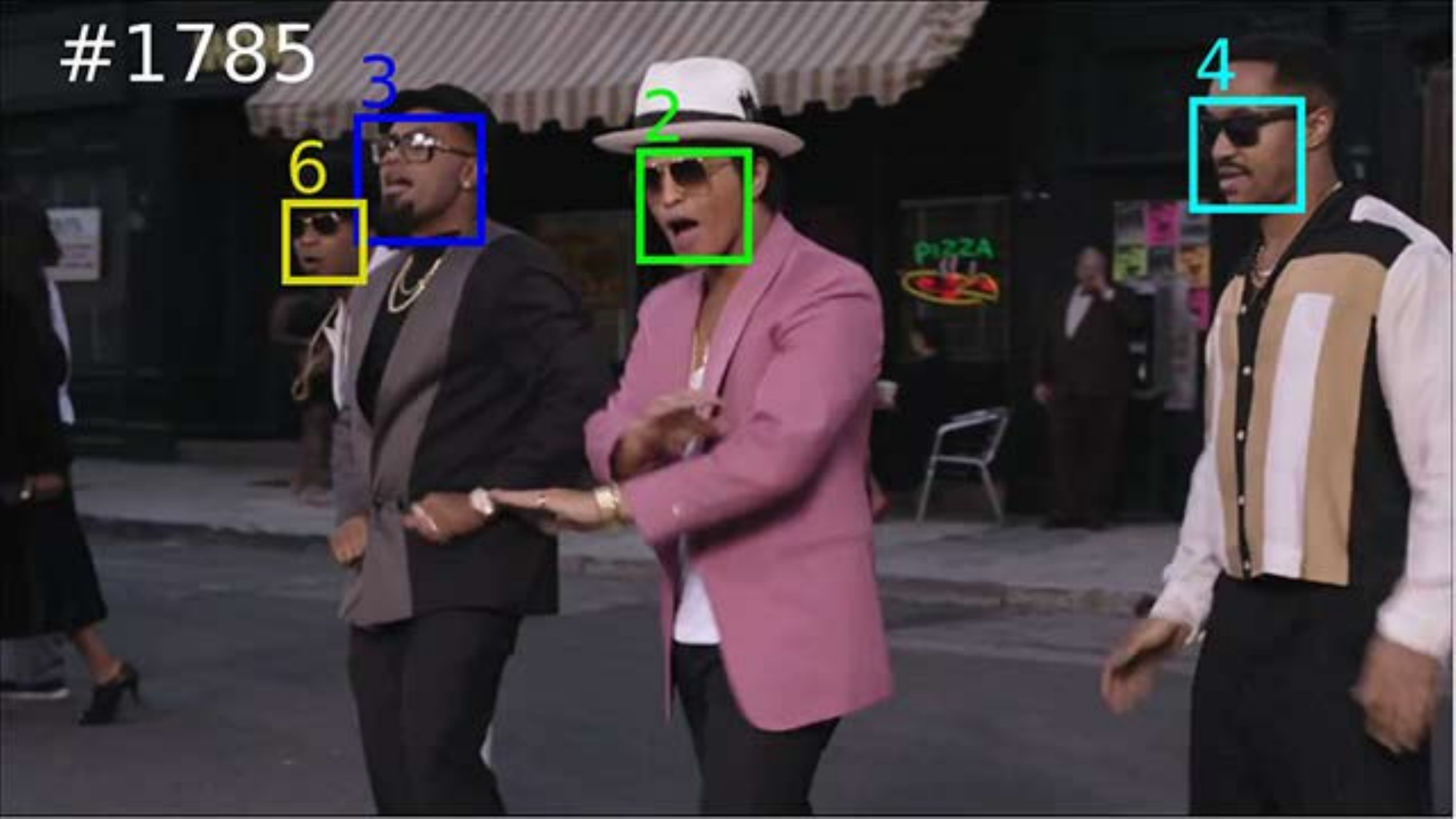} &
\hspace{-2.5mm}\includegraphics[width=3.4cm]{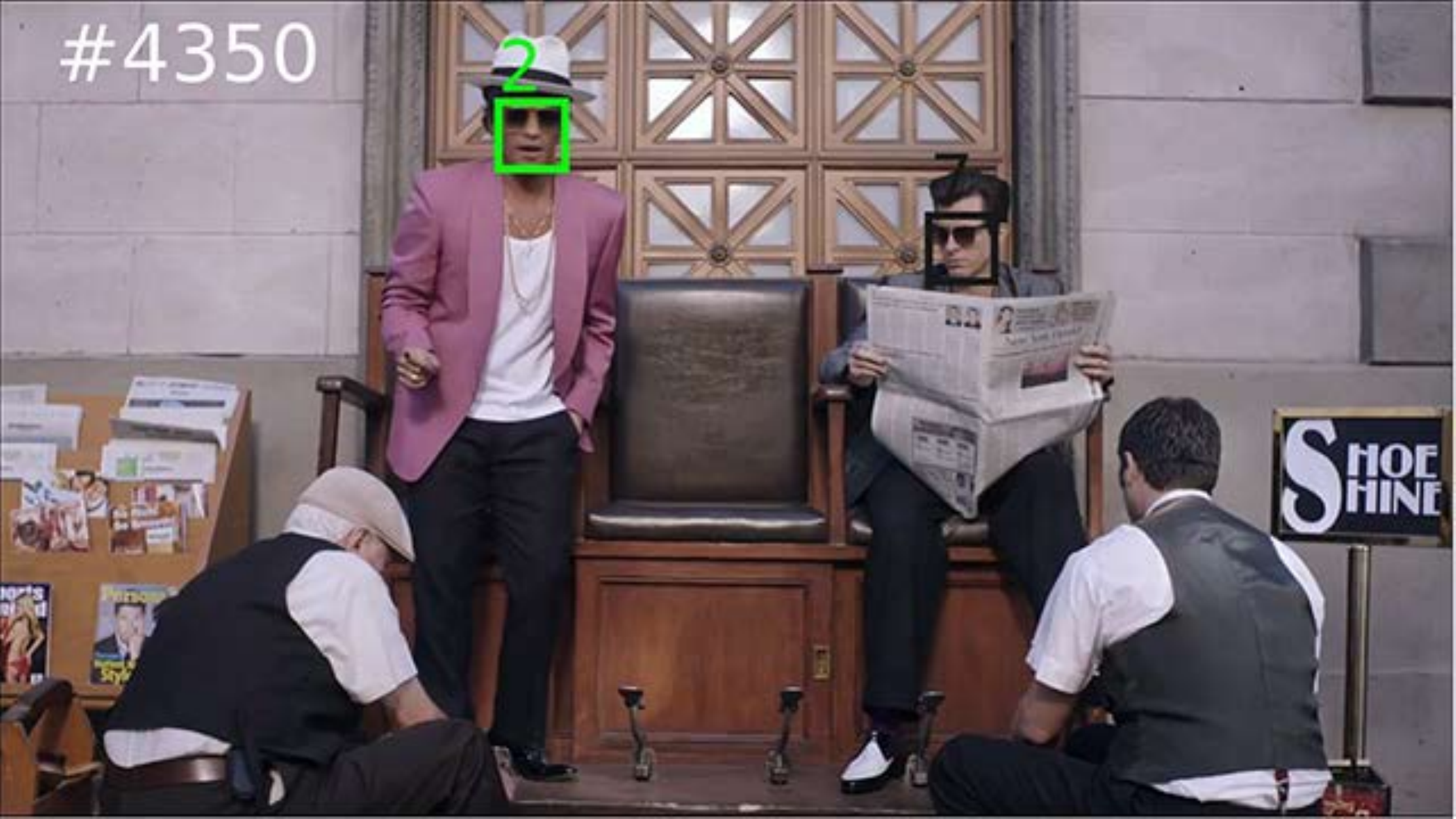} &
\hspace{-2.5mm}\includegraphics[width=3.4cm]{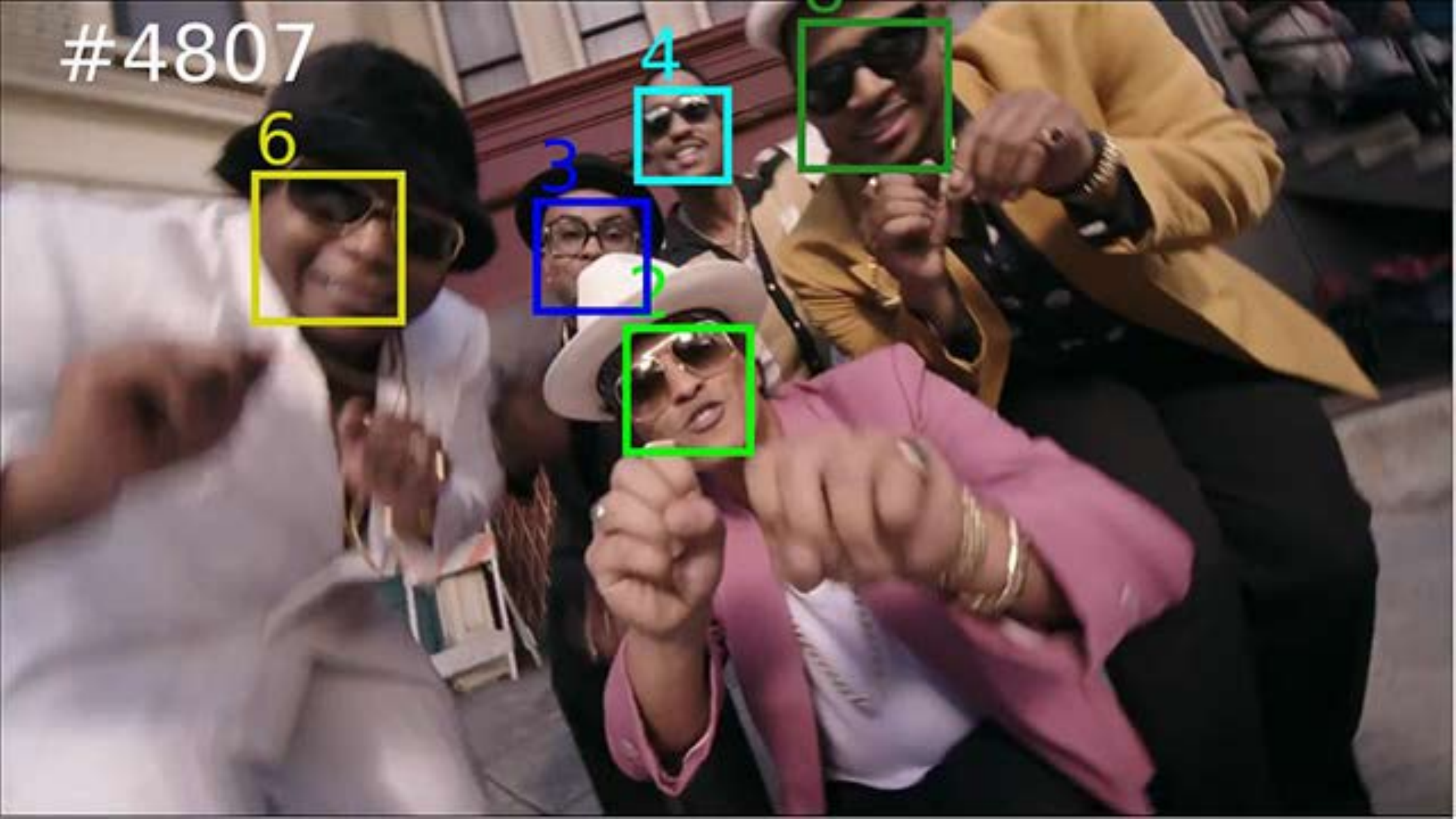} \\

\hspace{-1.0mm}\includegraphics[width=3.4cm]{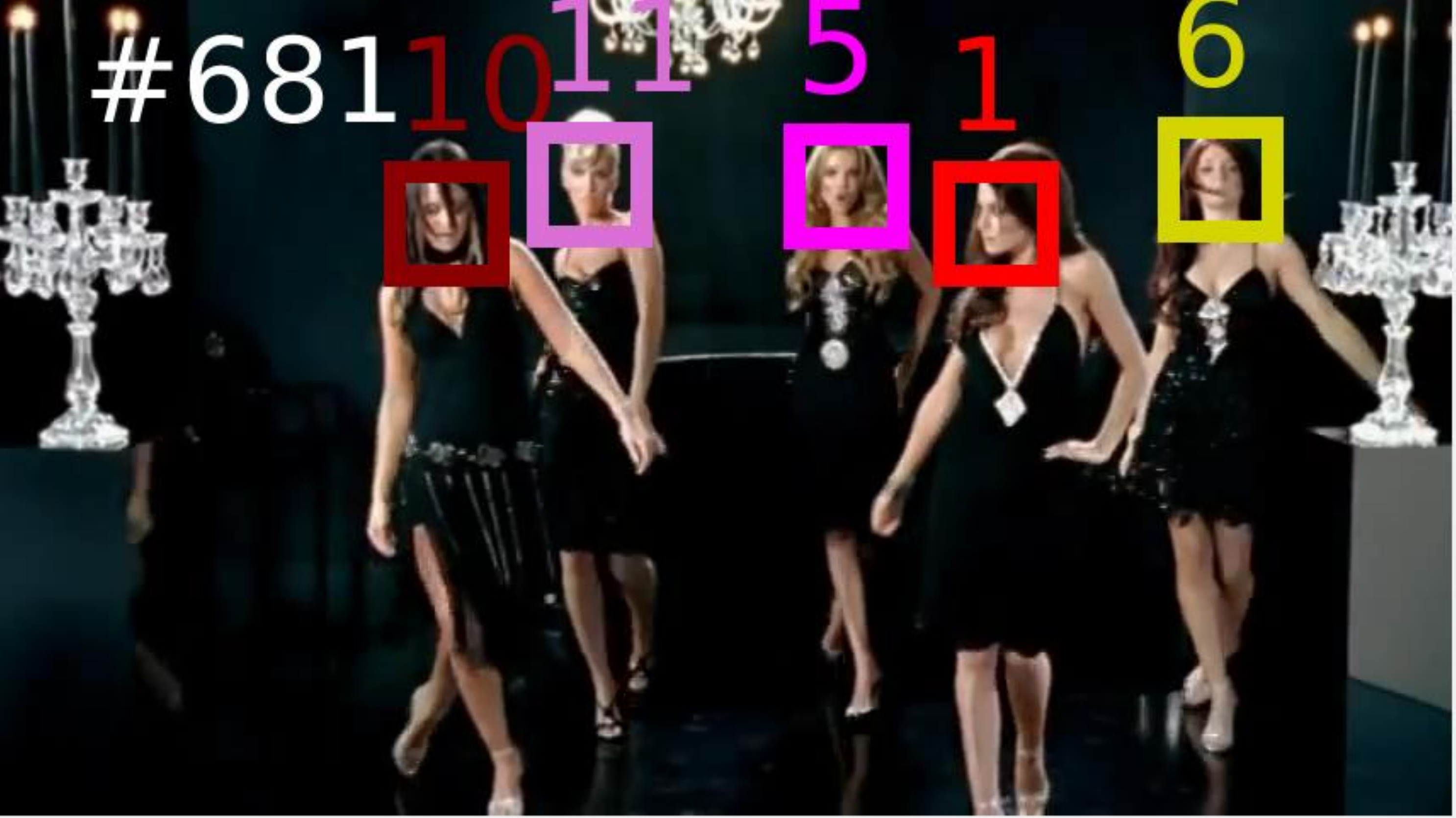} &
\hspace{-2.5mm}\includegraphics[width=3.4cm]{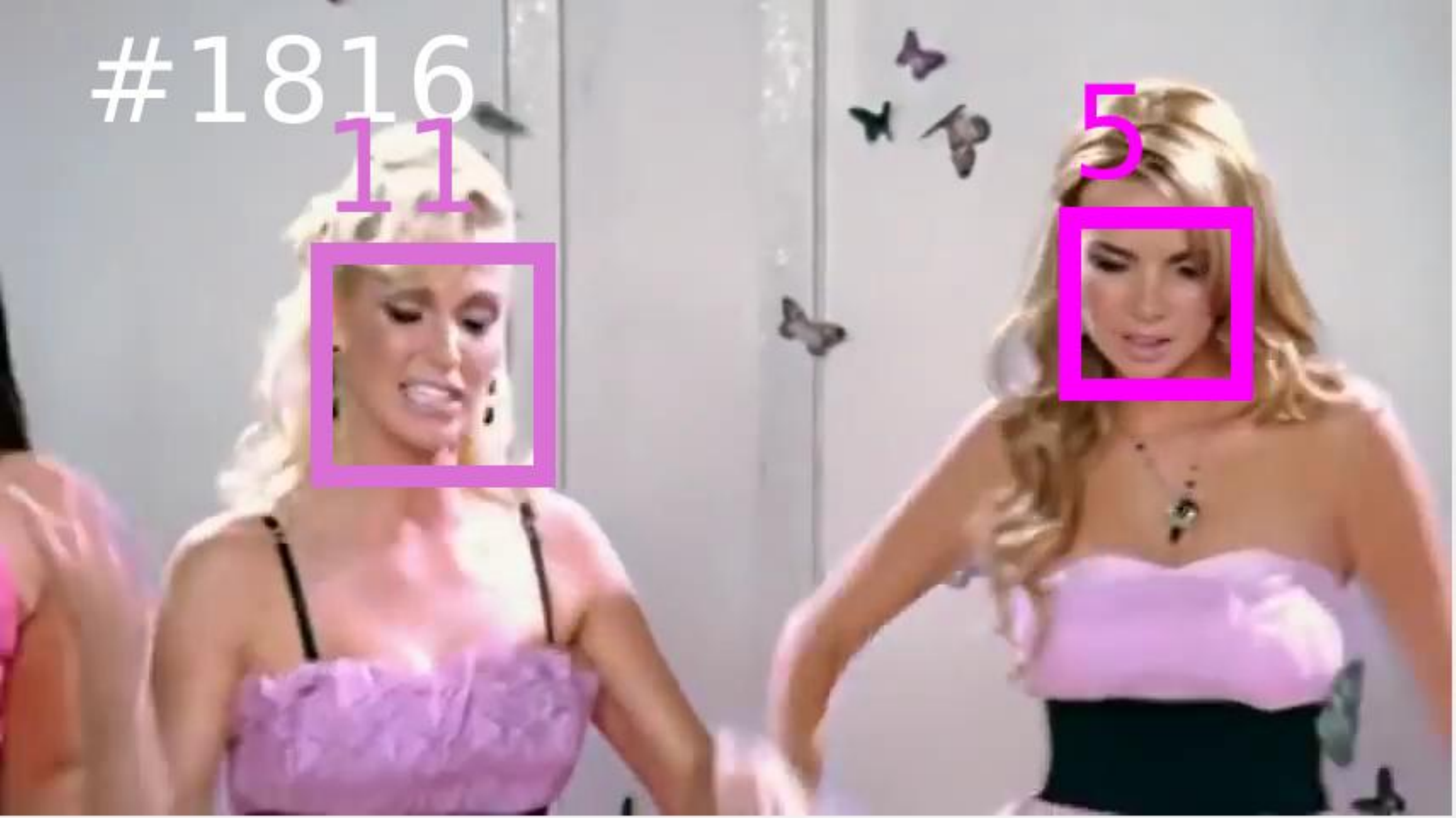} &
\hspace{-2.5mm}\includegraphics[width=3.4cm]{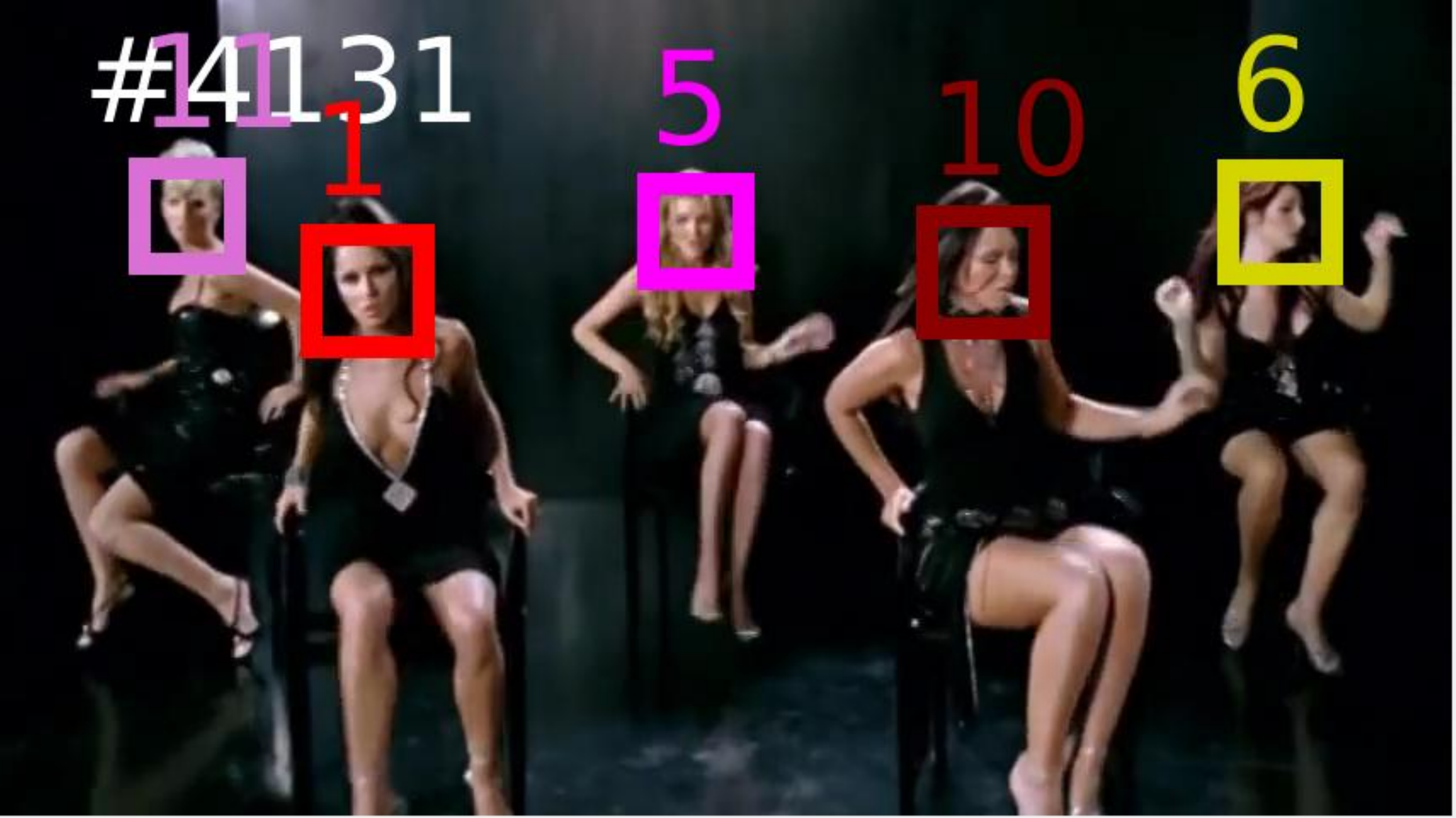} &
\hspace{-2.5mm}\includegraphics[width=3.4cm]{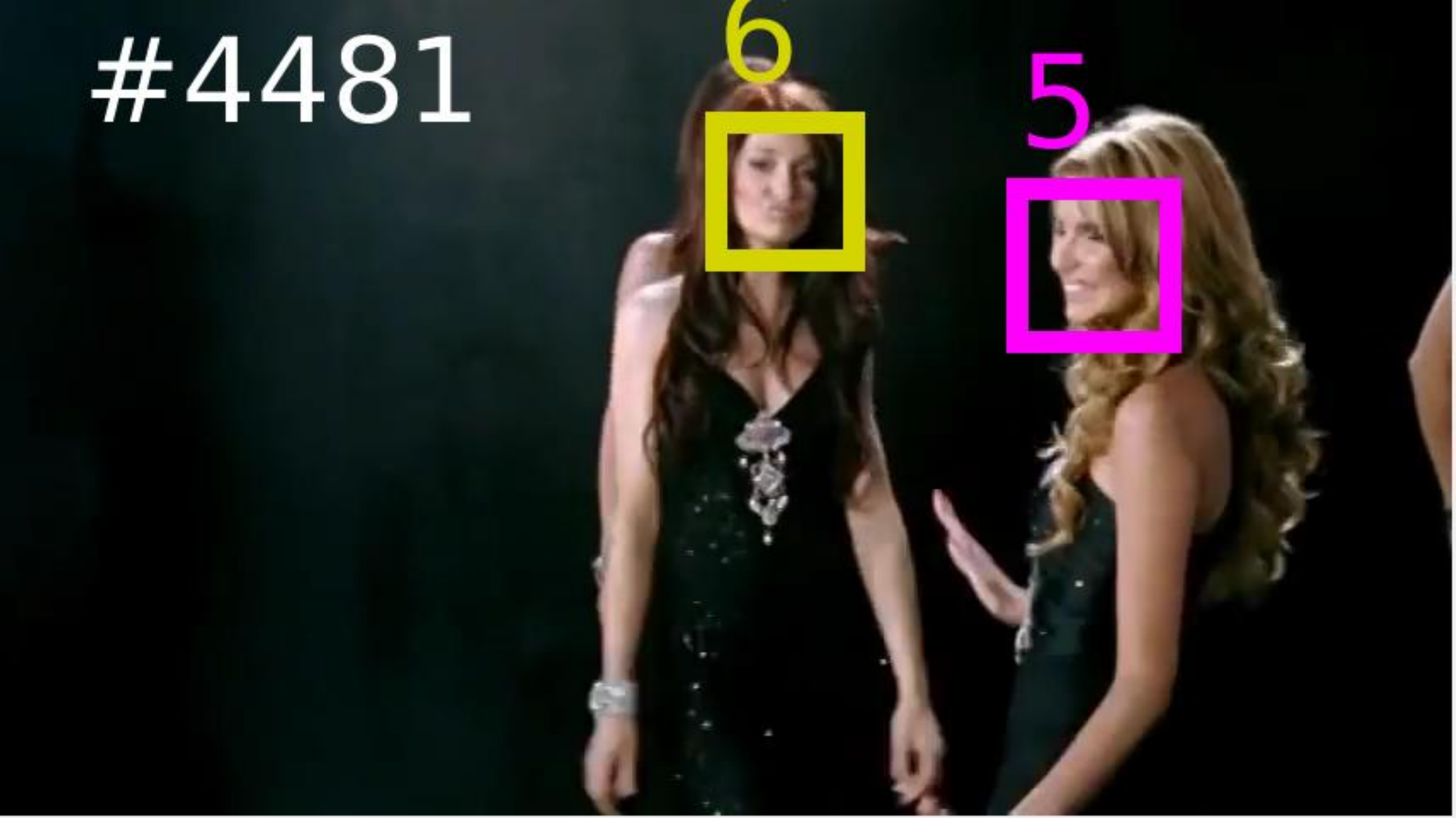} &
\hspace{-2.5mm}\includegraphics[width=3.4cm]{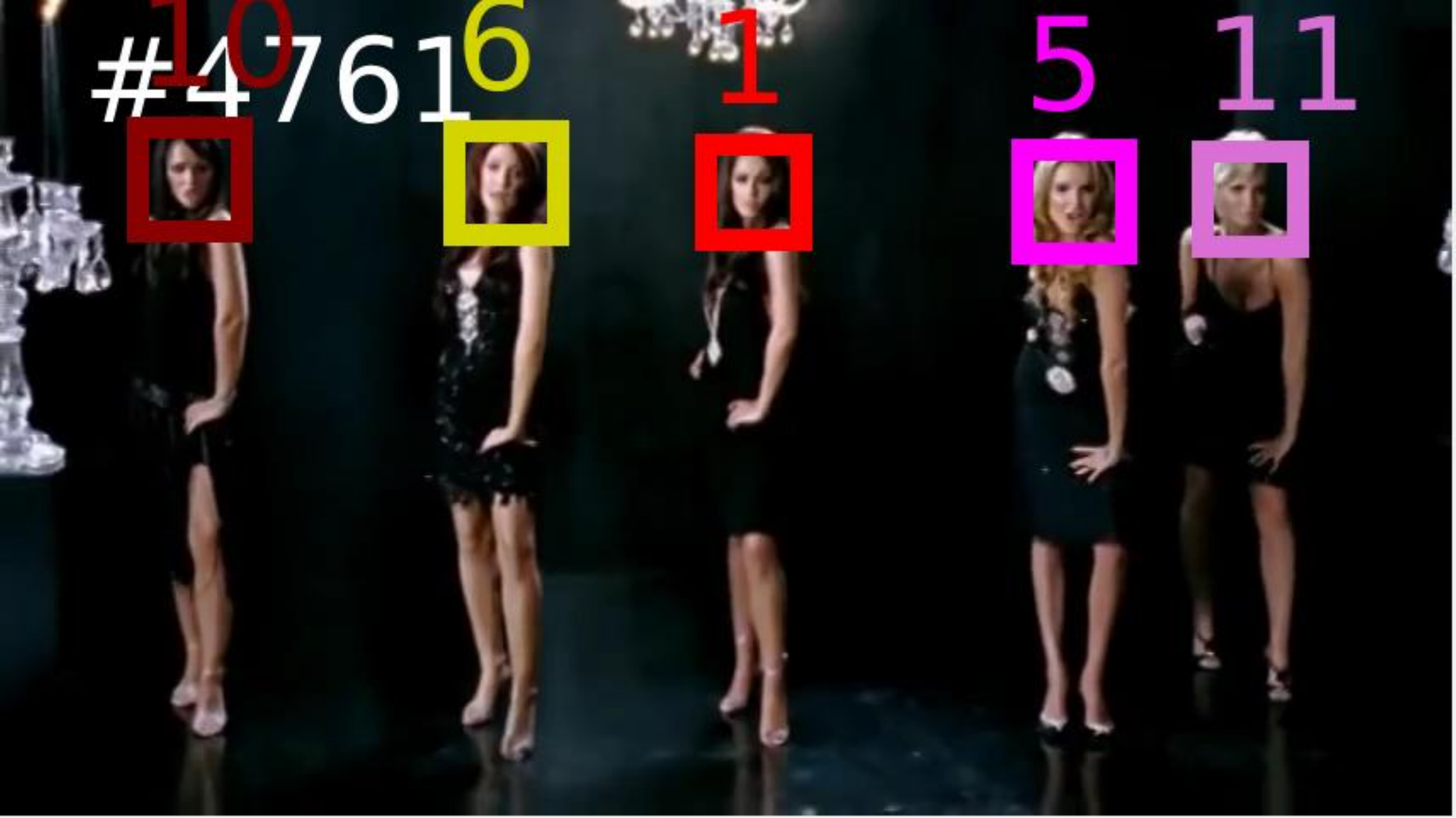} \\

\hspace{-1.0mm}\includegraphics[width=3.4cm]{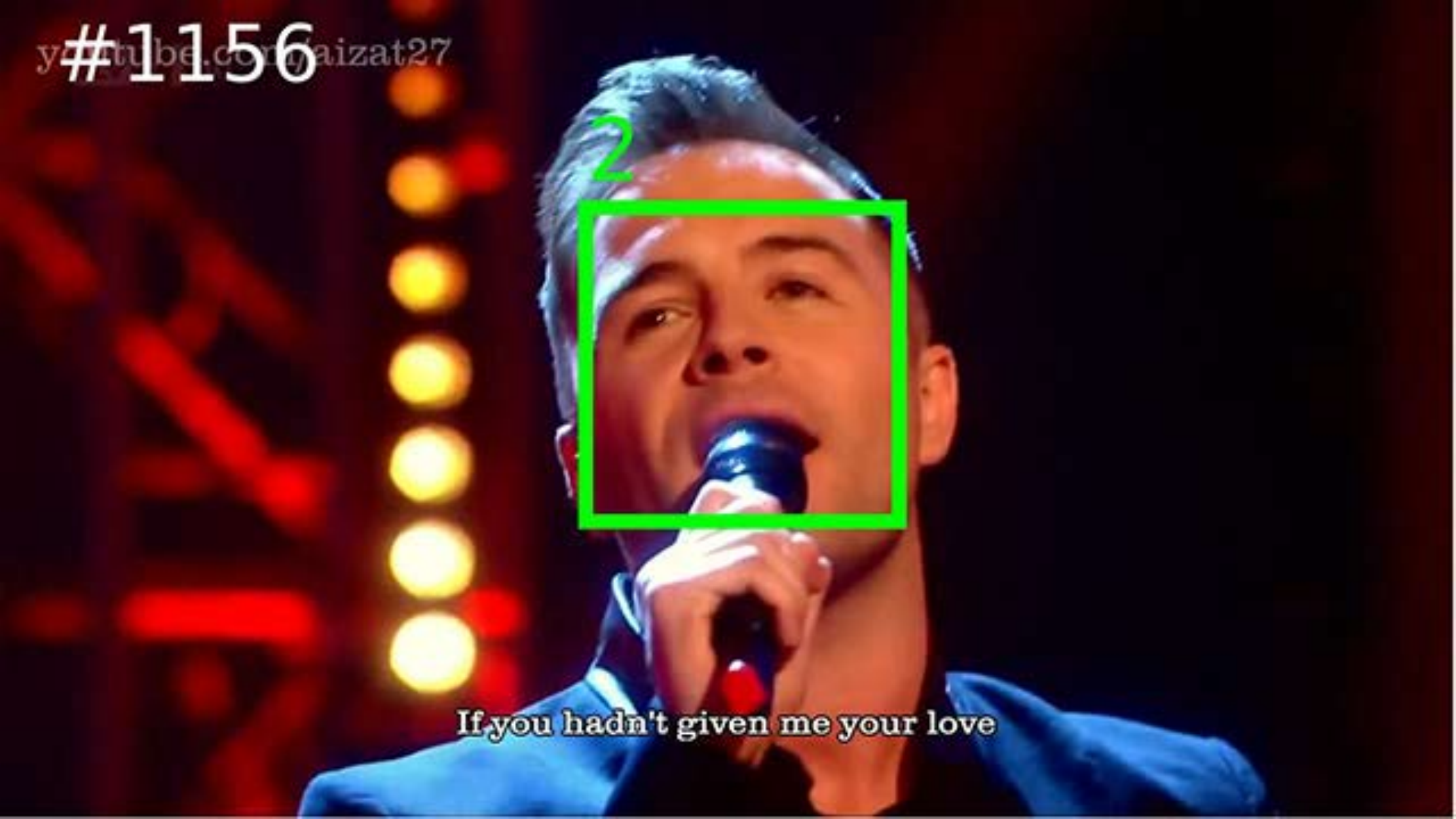} &
\hspace{-2.5mm}\includegraphics[width=3.4cm]{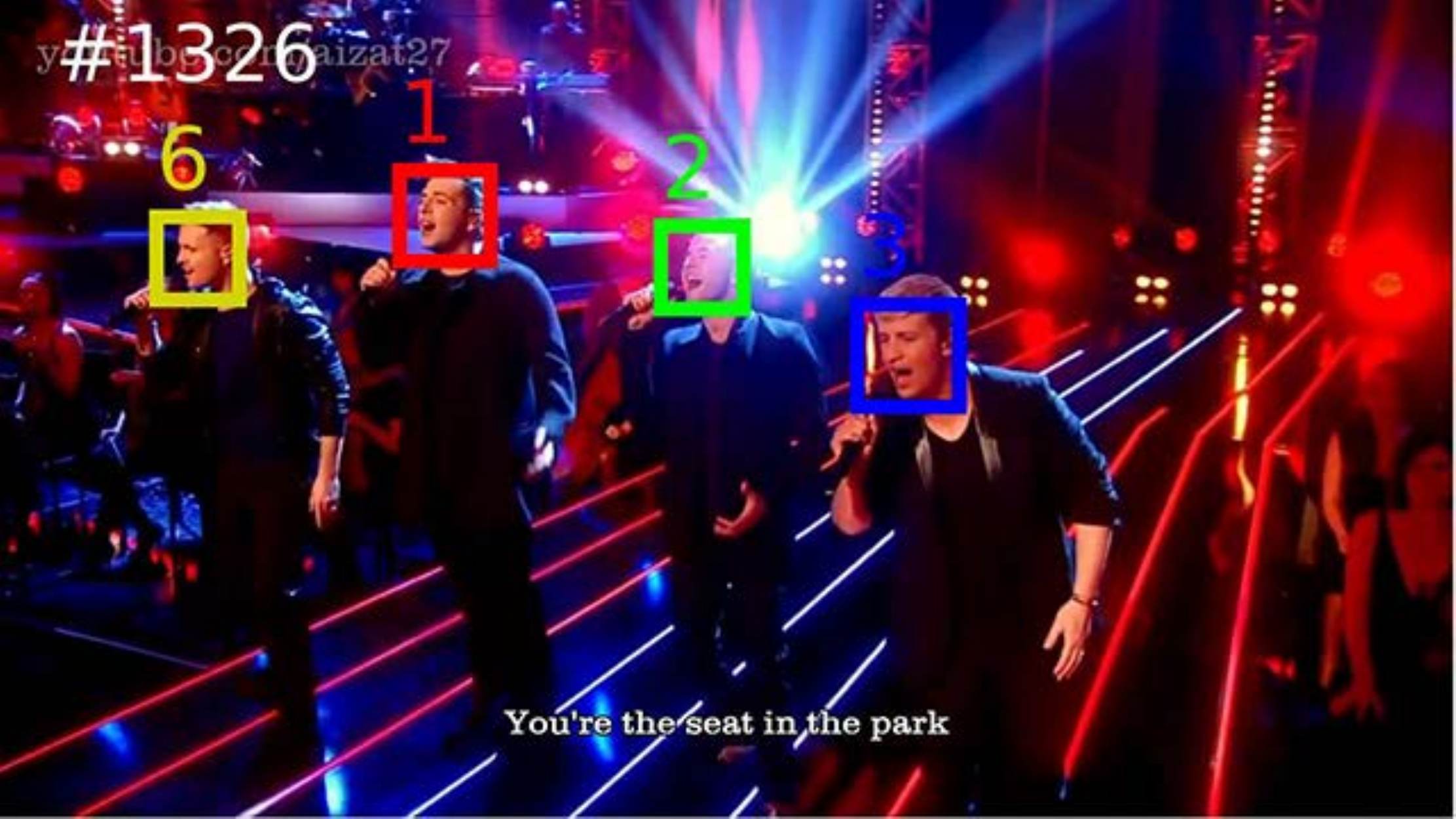} &
\hspace{-2.5mm}\includegraphics[width=3.4cm]{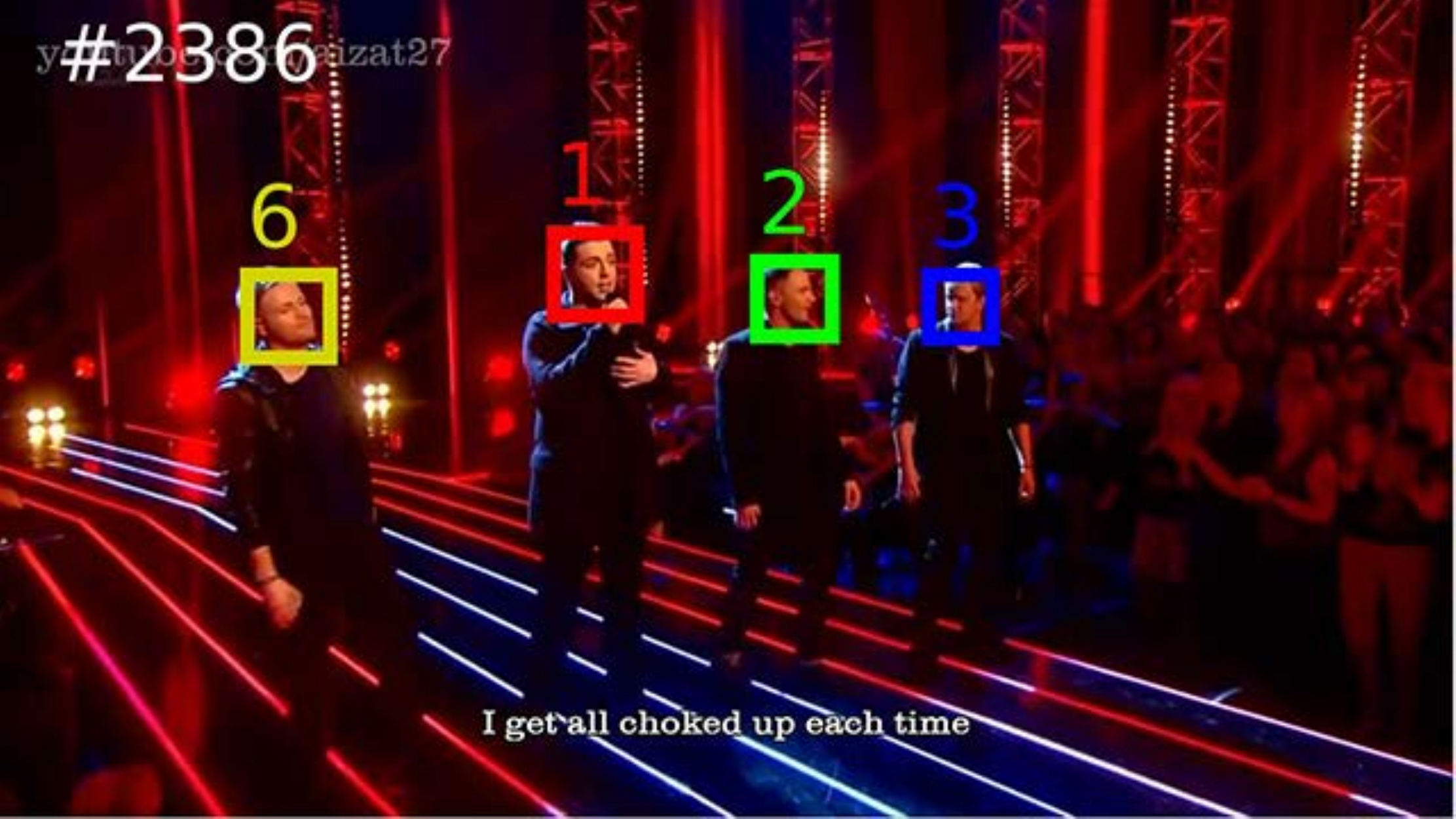} &
\hspace{-2.5mm}\includegraphics[width=3.4cm]{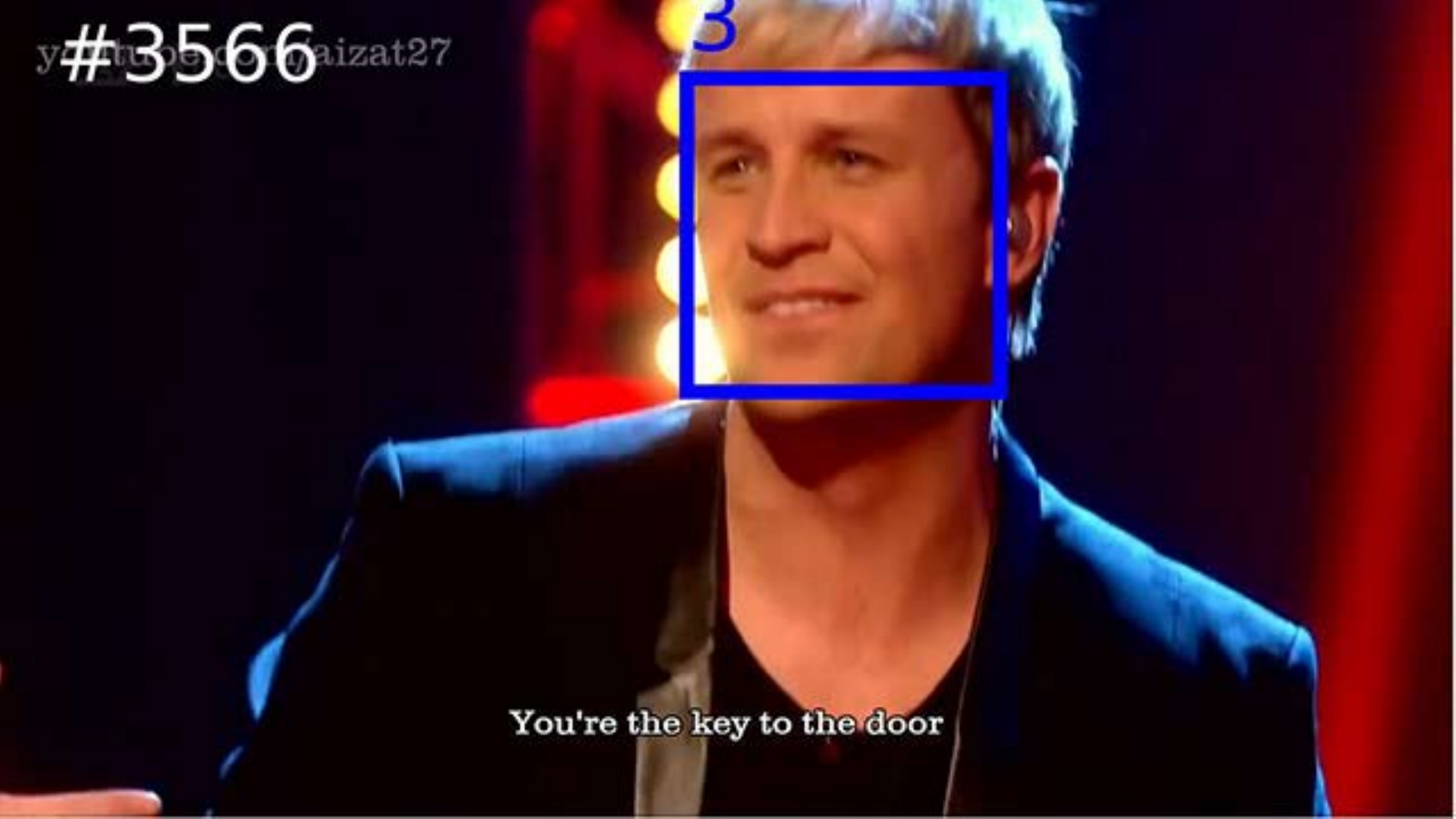} &
\hspace{-2.5mm}\includegraphics[width=3.4cm]{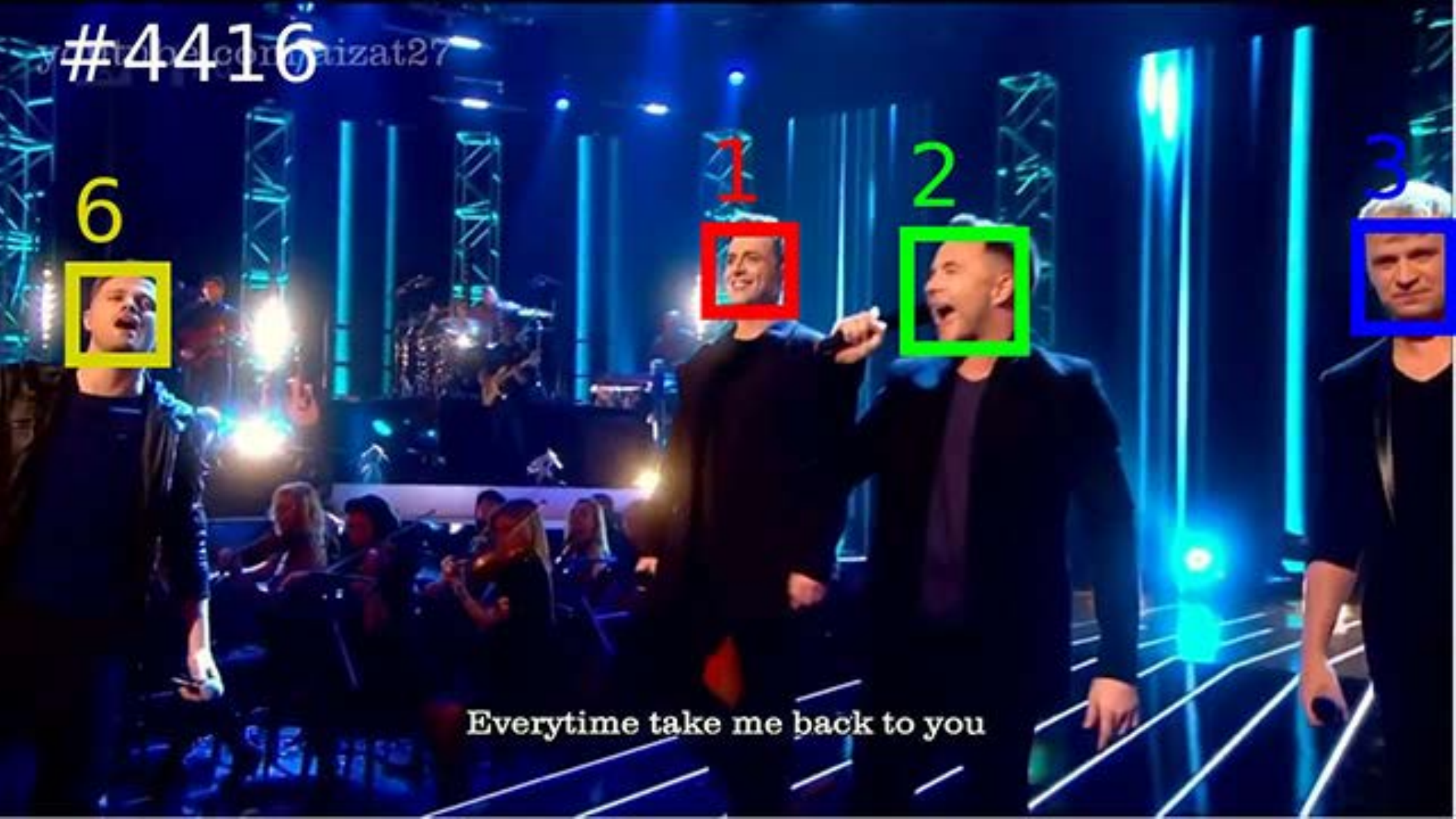} \\

\hspace{-1.0mm}\includegraphics[width=3.4cm]{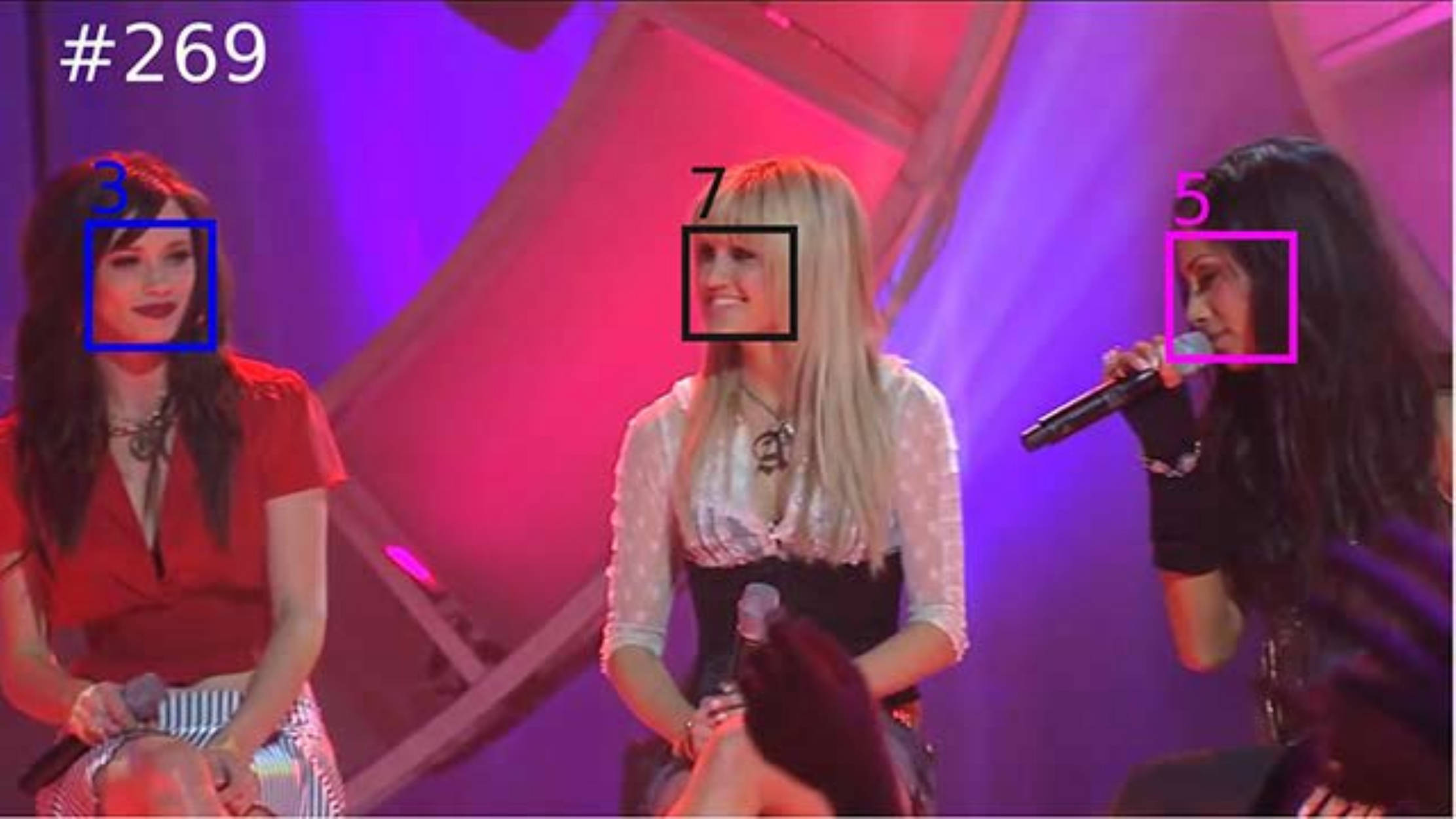} &
\hspace{-2.5mm}\includegraphics[width=3.4cm]{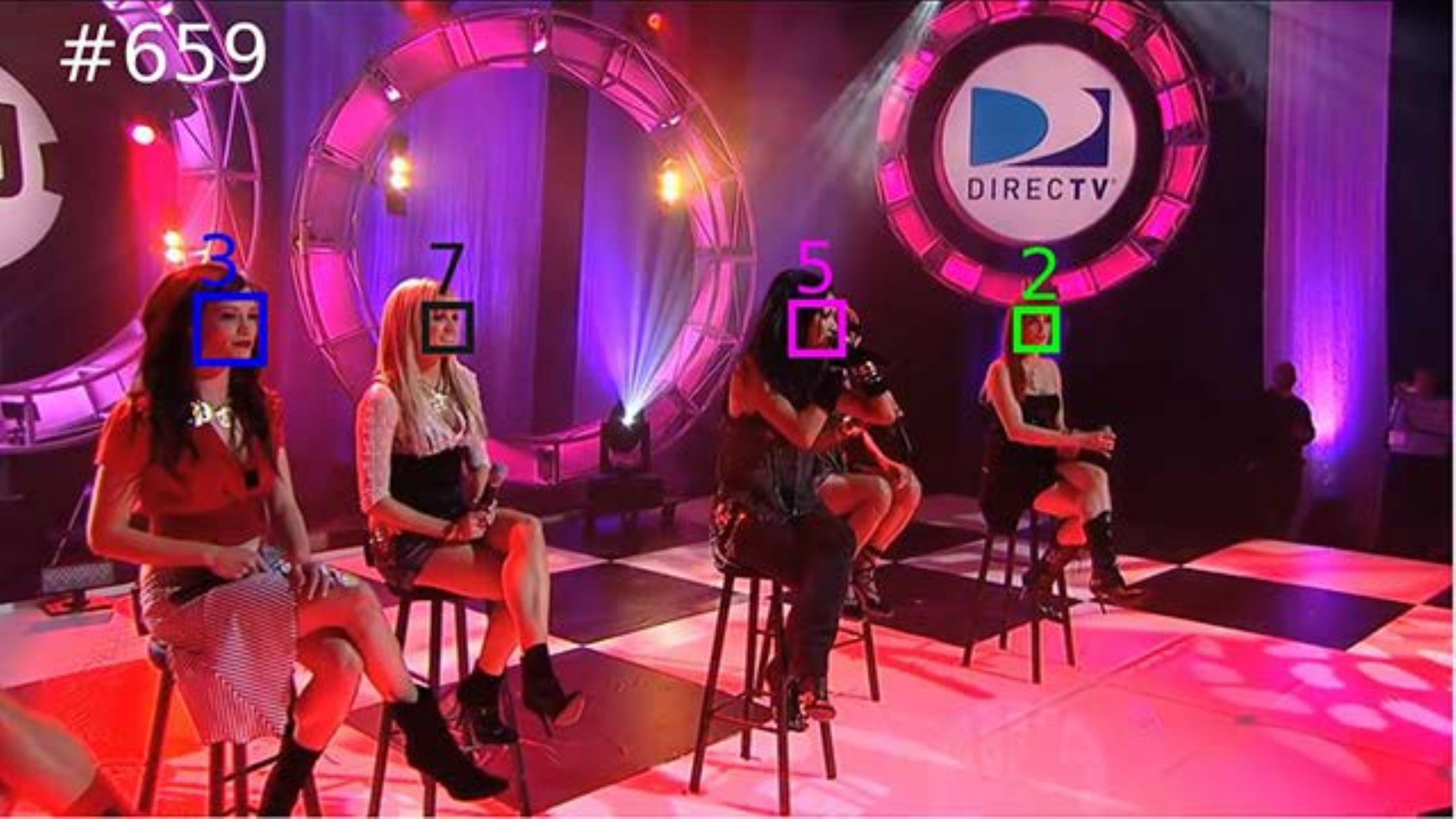} &
\hspace{-2.5mm}\includegraphics[width=3.4cm]{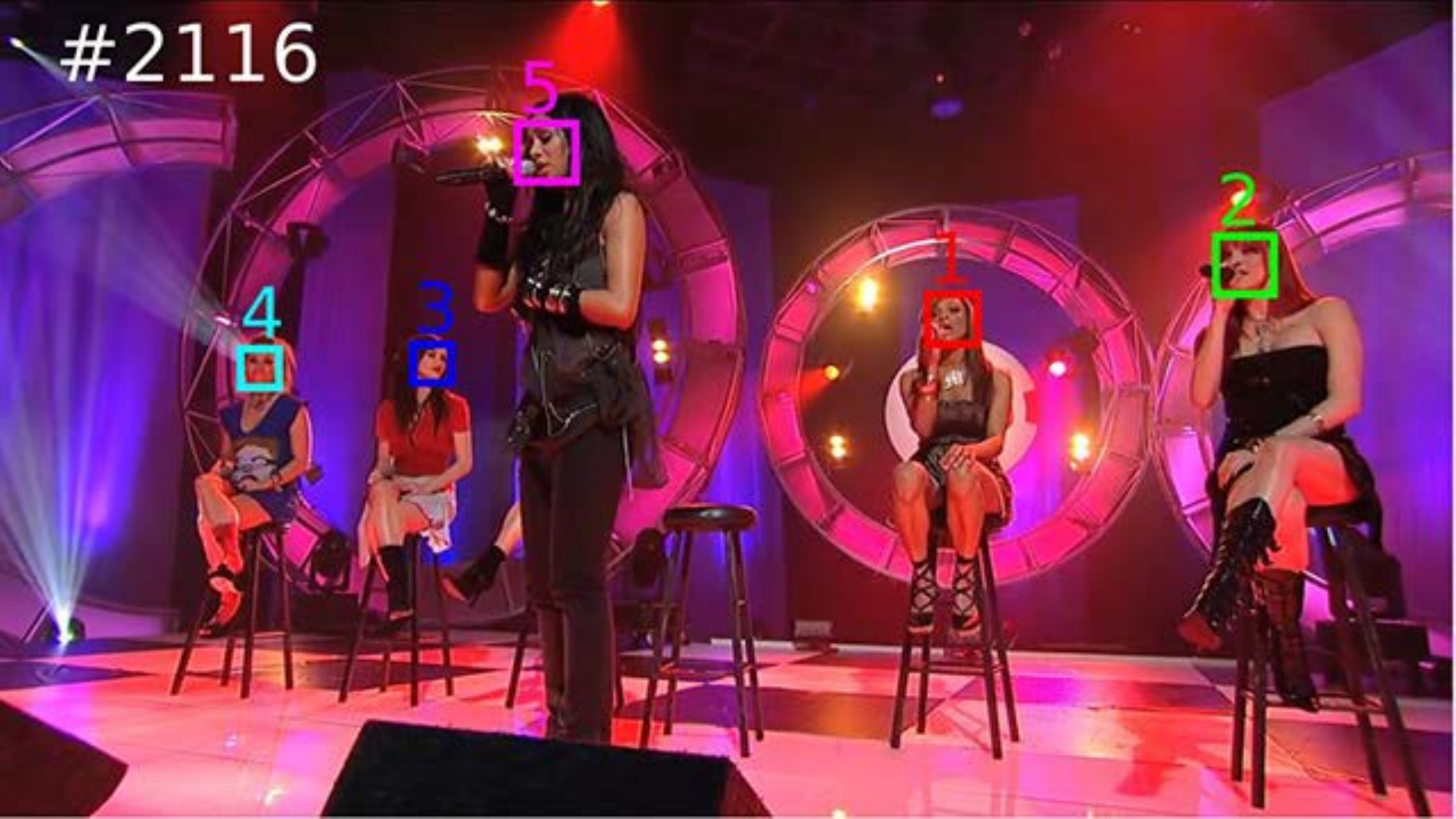} &
\hspace{-2.5mm}\includegraphics[width=3.4cm]{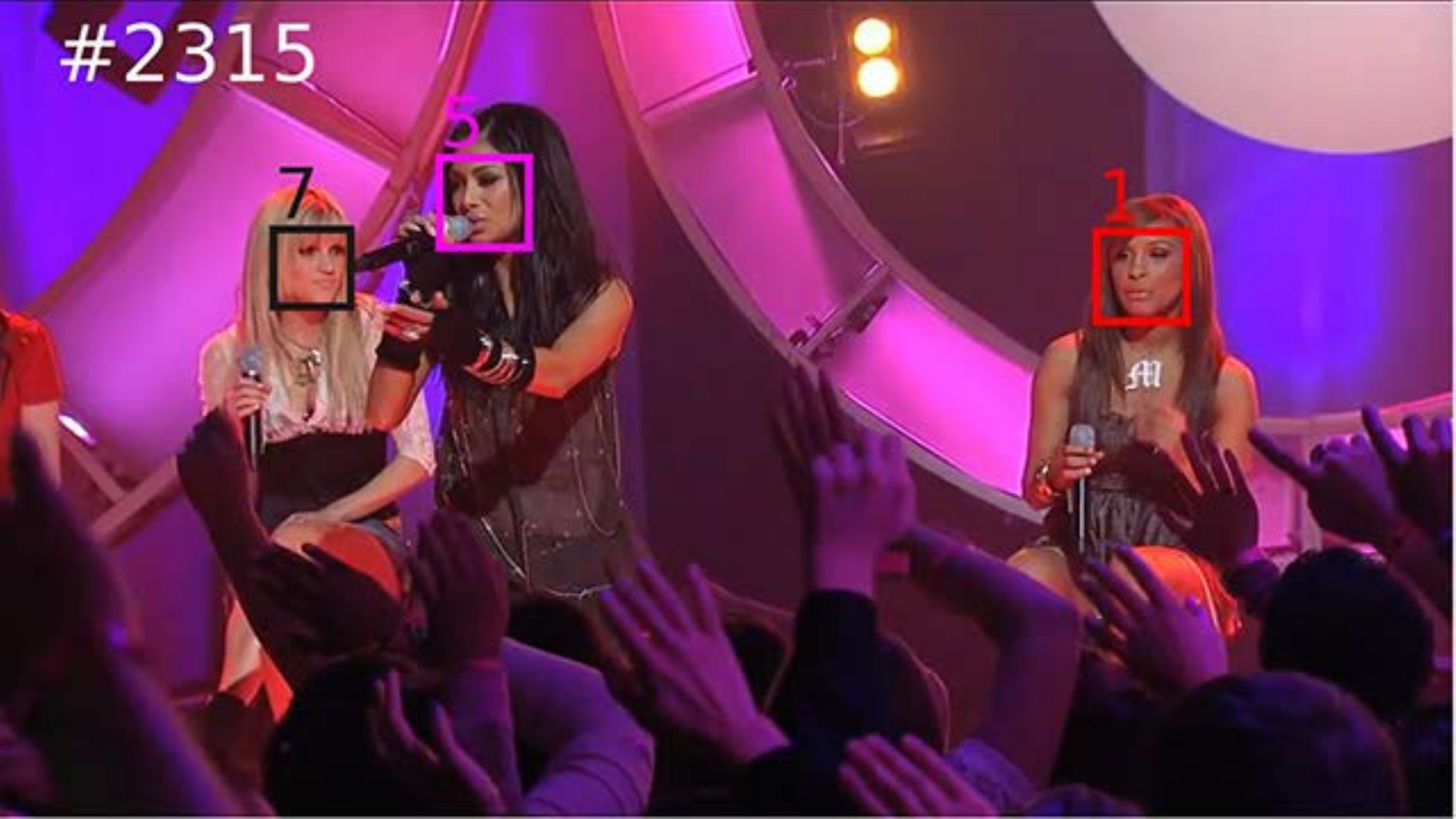} &
\hspace{-2.5mm}\includegraphics[width=3.4cm]{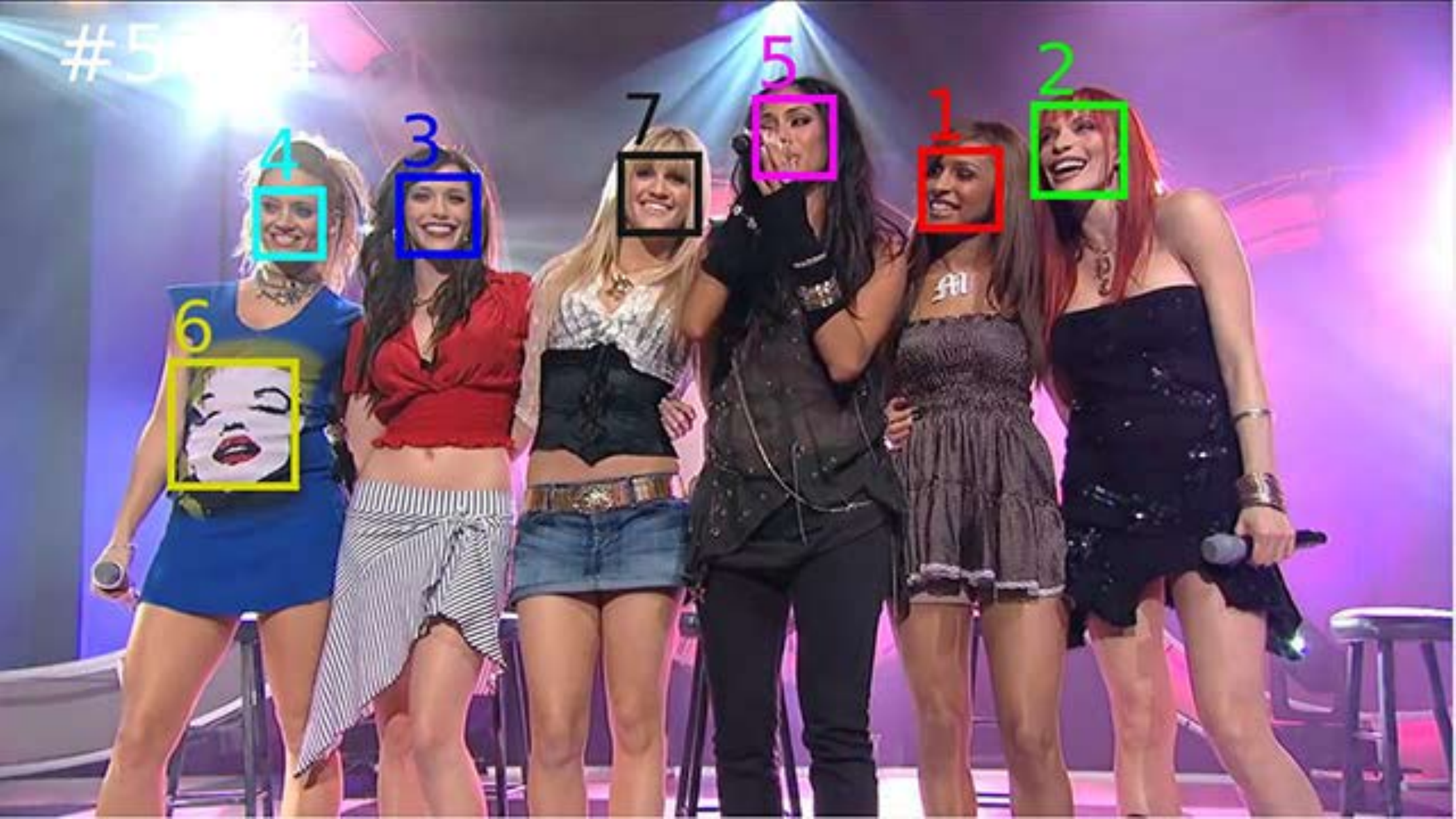} \\
\end{tabular}
\vspace{-1mm}
\caption{\textbf{Tracking results on YouTube Music video dataset}. Shown from the top to bottom are \textsc{Hello Bubble}, \textsc{Apink}, \textsc{Darling}, \textsc{T-ara}, \textsc{Bruno Mars}, \textsc{Girls Aloud}, \textsc{Westlife} and \textsc{Pussycat Dolls}. The faces of the different people are color coded.
}
\vspace{-3mm}
\label{fig:samples}
\end{figure*}

\setlength{\figwidth}{0.19\textwidth}

\begin{figure*}[t]
% \scriptsize
\centering

% BUFFY02
\rotatebox[origin=c]{90}{\textsc{Buffy02}} \hfill
\begin{minipage}{\figwidth} \centering
\includegraphics[width=\linewidth]{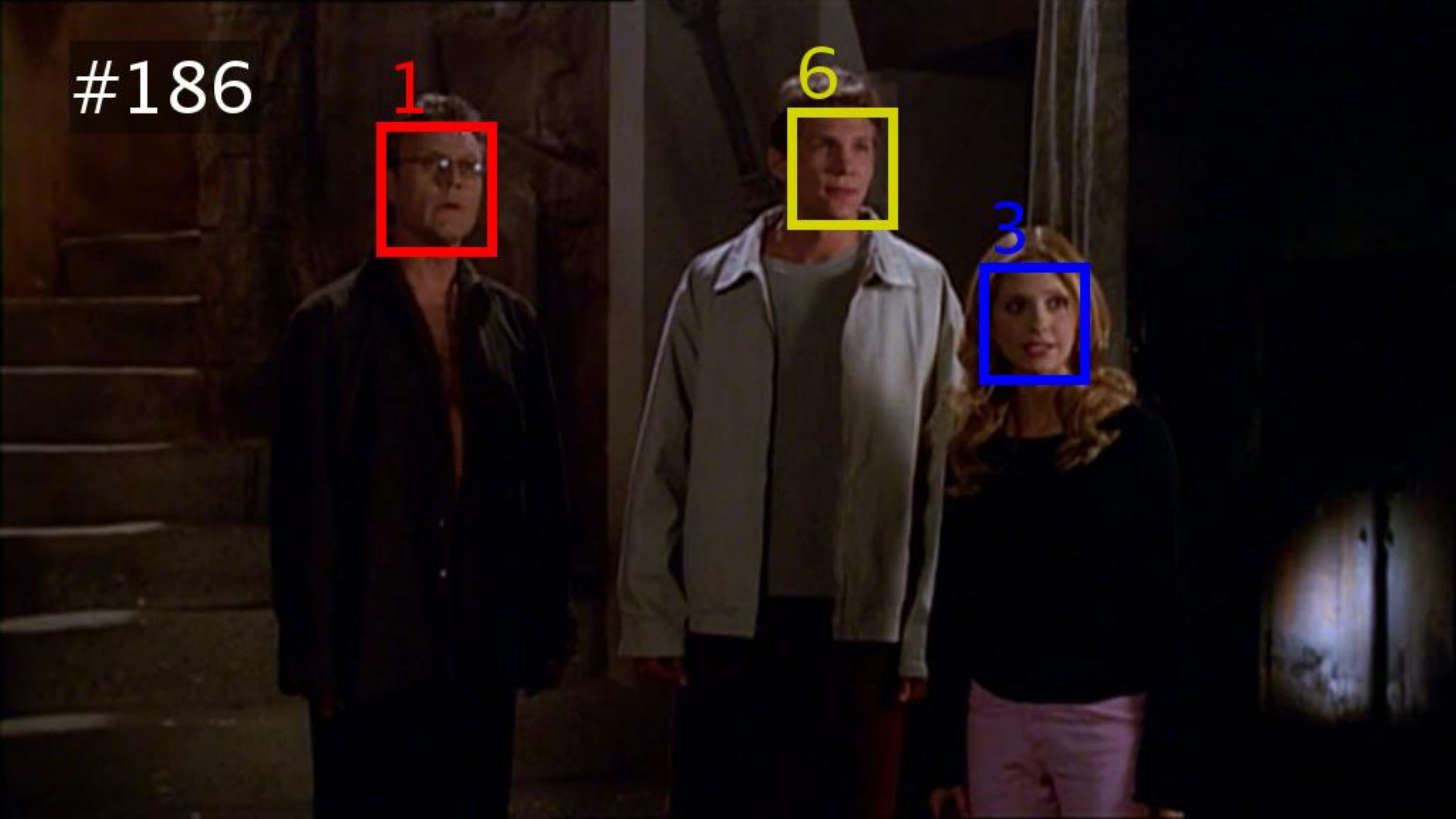}
\end{minipage}\hfill
\begin{minipage}{\figwidth} \centering
\includegraphics[width=\linewidth]{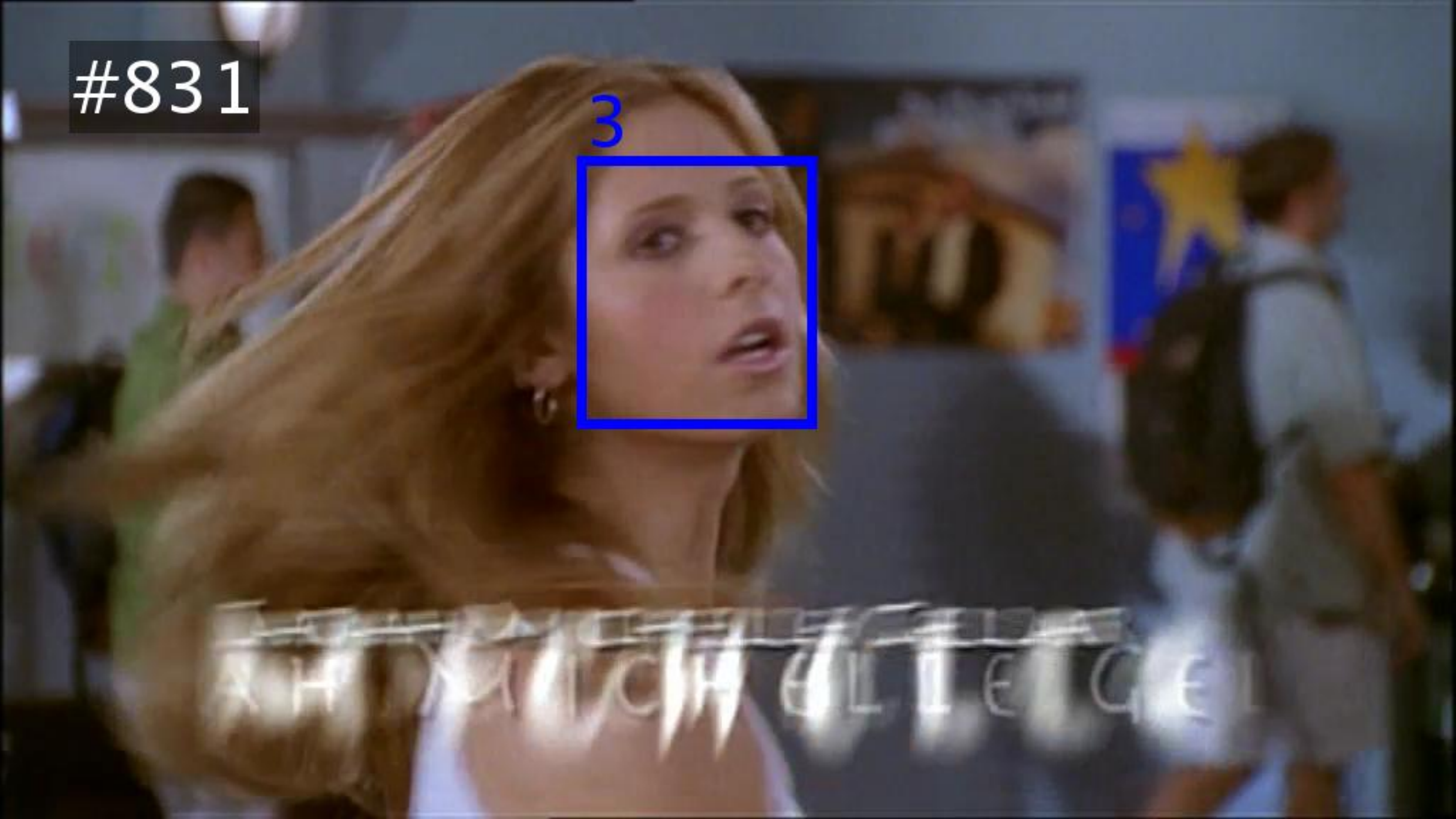}
\end{minipage}\hfill
\begin{minipage}{\figwidth} \centering
\includegraphics[width=\linewidth]{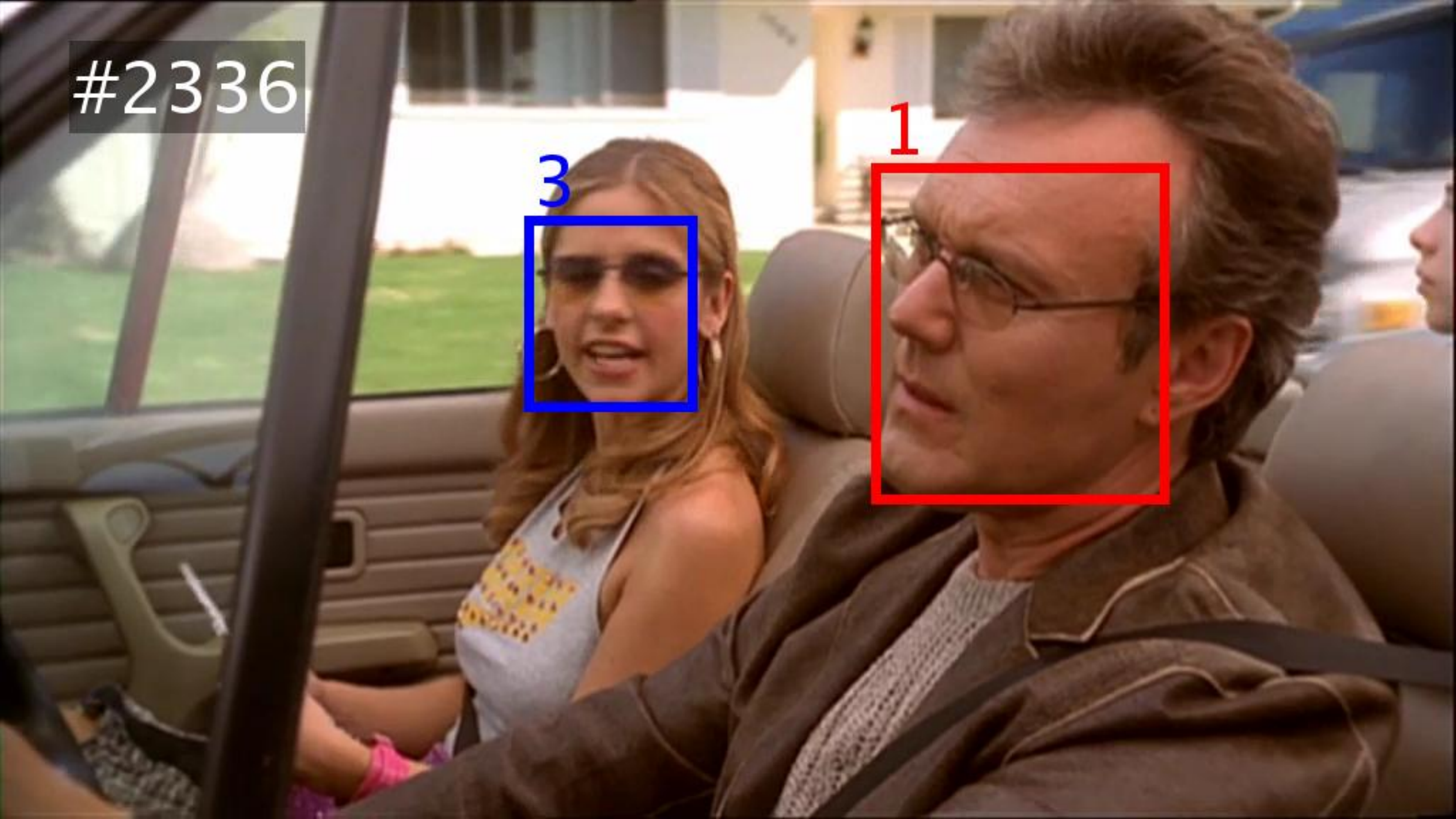}
\end{minipage}\hfill
\begin{minipage}{\figwidth} \centering
\includegraphics[width=\linewidth]{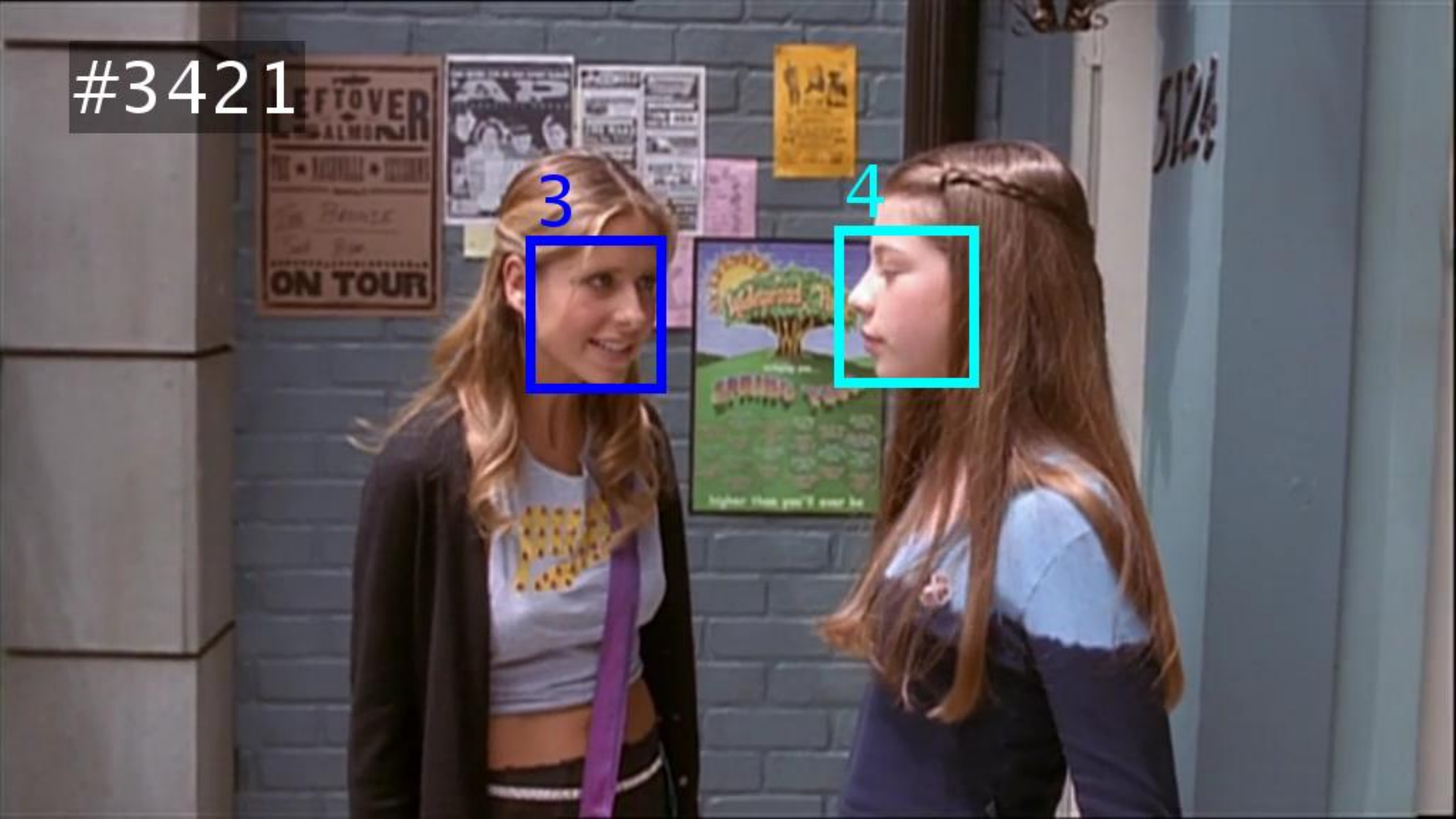}
\end{minipage}\hfill
\begin{minipage}{\figwidth} \centering
\includegraphics[width=\linewidth]{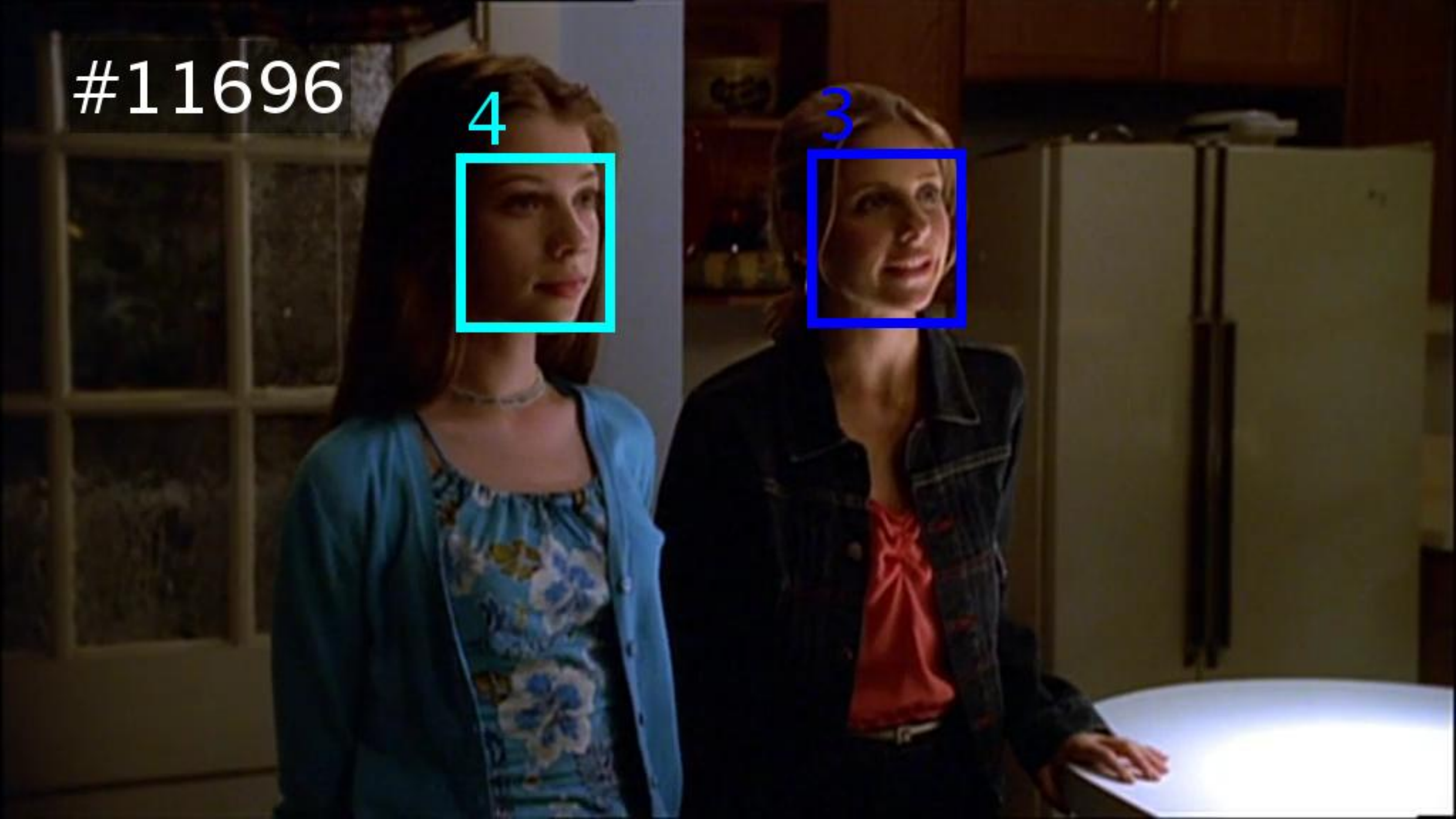}
\end{minipage}\hfill
\\
\vspace{1mm}

% BUFFY 05
\rotatebox[origin=c]{90}{\textsc{Buffy05}} \hfill
\begin{minipage}{\figwidth} \centering
\includegraphics[width=\linewidth]{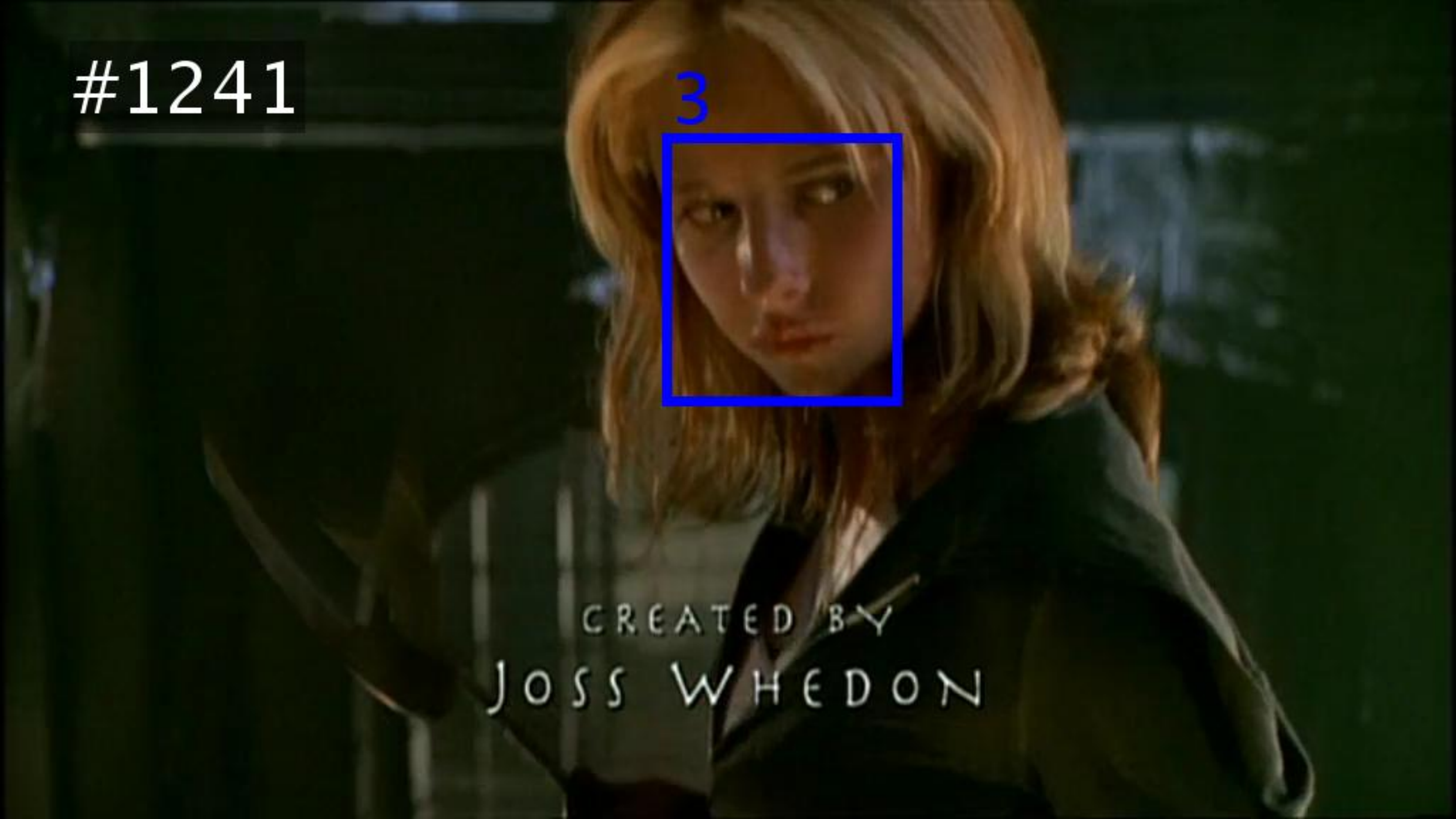}
\end{minipage}\hfill
\begin{minipage}{\figwidth} \centering
\includegraphics[width=\linewidth]{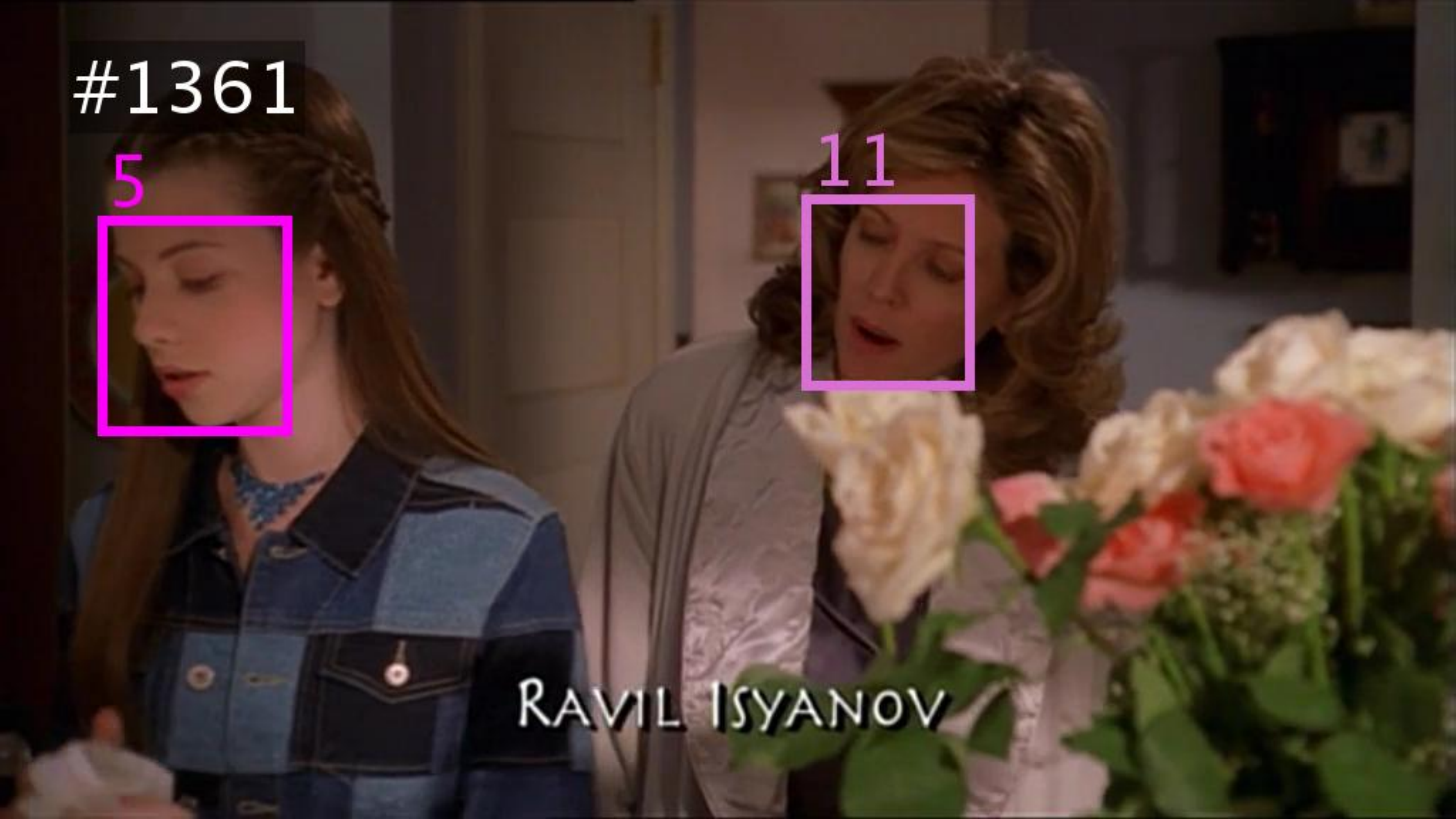}
\end{minipage}\hfill
\begin{minipage}{\figwidth} \centering
\includegraphics[width=\linewidth]{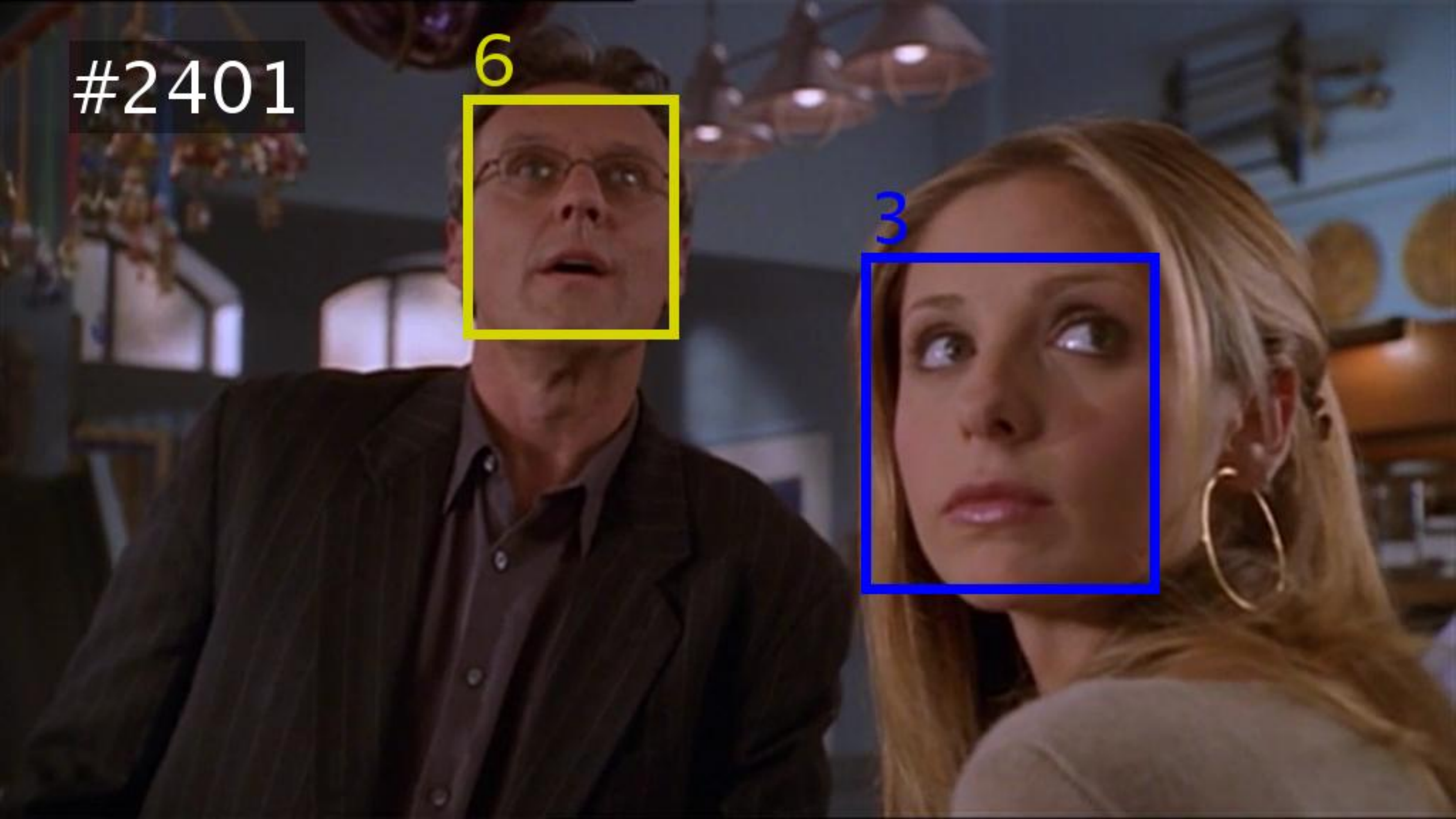}
\end{minipage}\hfill
\begin{minipage}{\figwidth} \centering
\includegraphics[width=\linewidth]{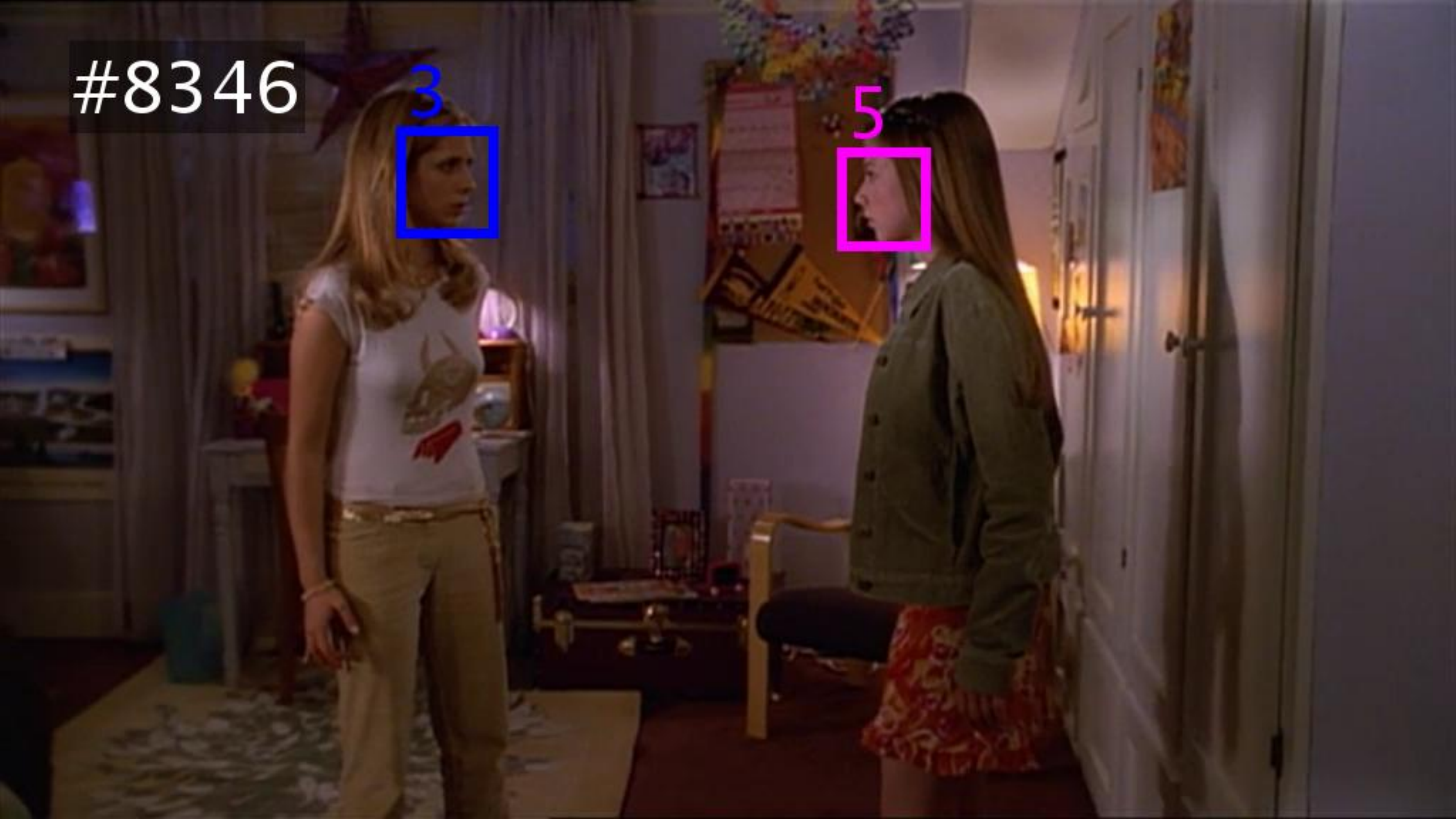}
\end{minipage}\hfill
\begin{minipage}{\figwidth} \centering
\includegraphics[width=\linewidth]{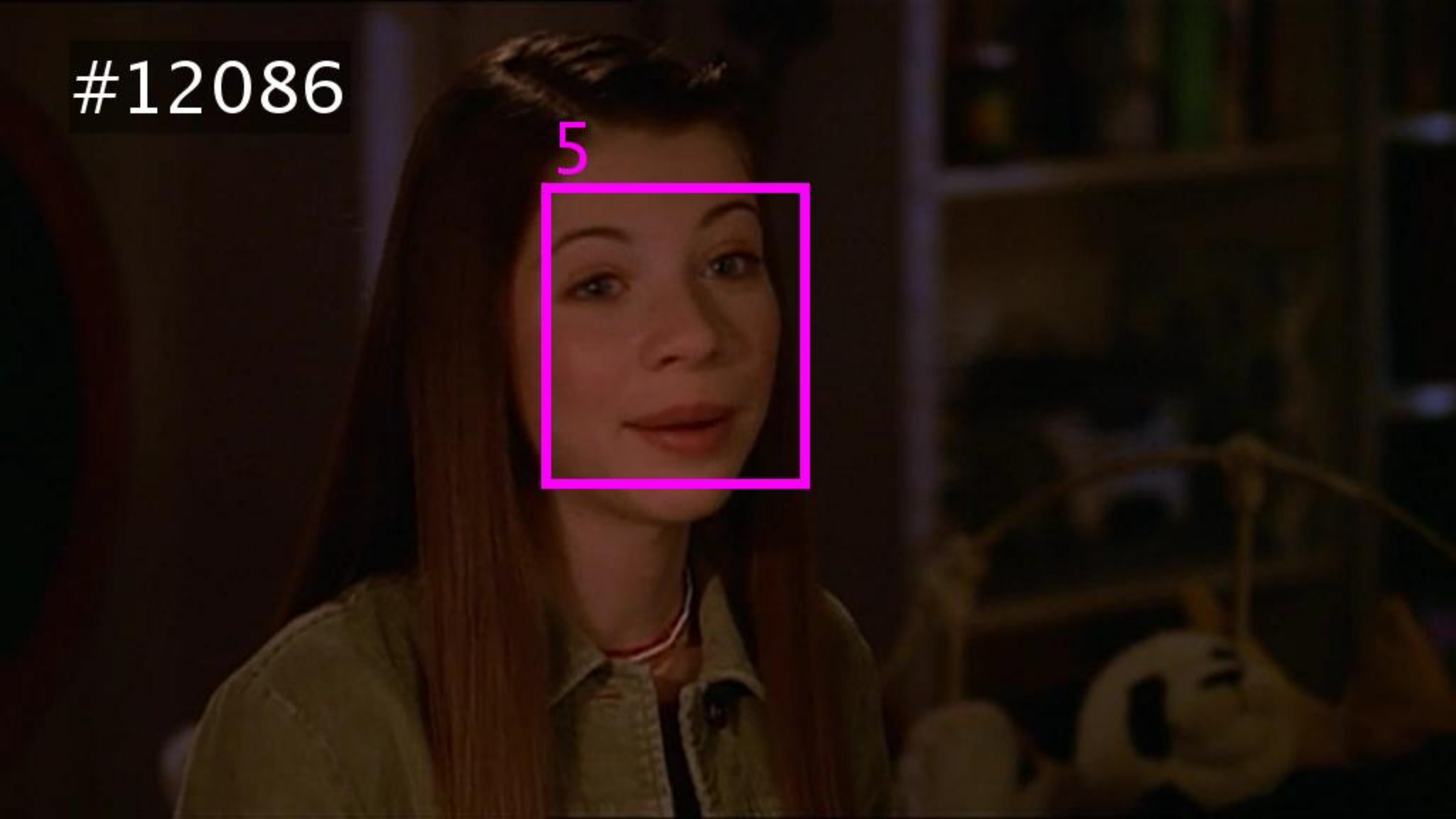}
\end{minipage}\hfill
\\
\vspace{1mm}

% BUFFY 05
\rotatebox[origin=c]{90}{\textsc{Buffy06}} \hfill
\begin{minipage}{\figwidth} \centering
\includegraphics[width=\linewidth]{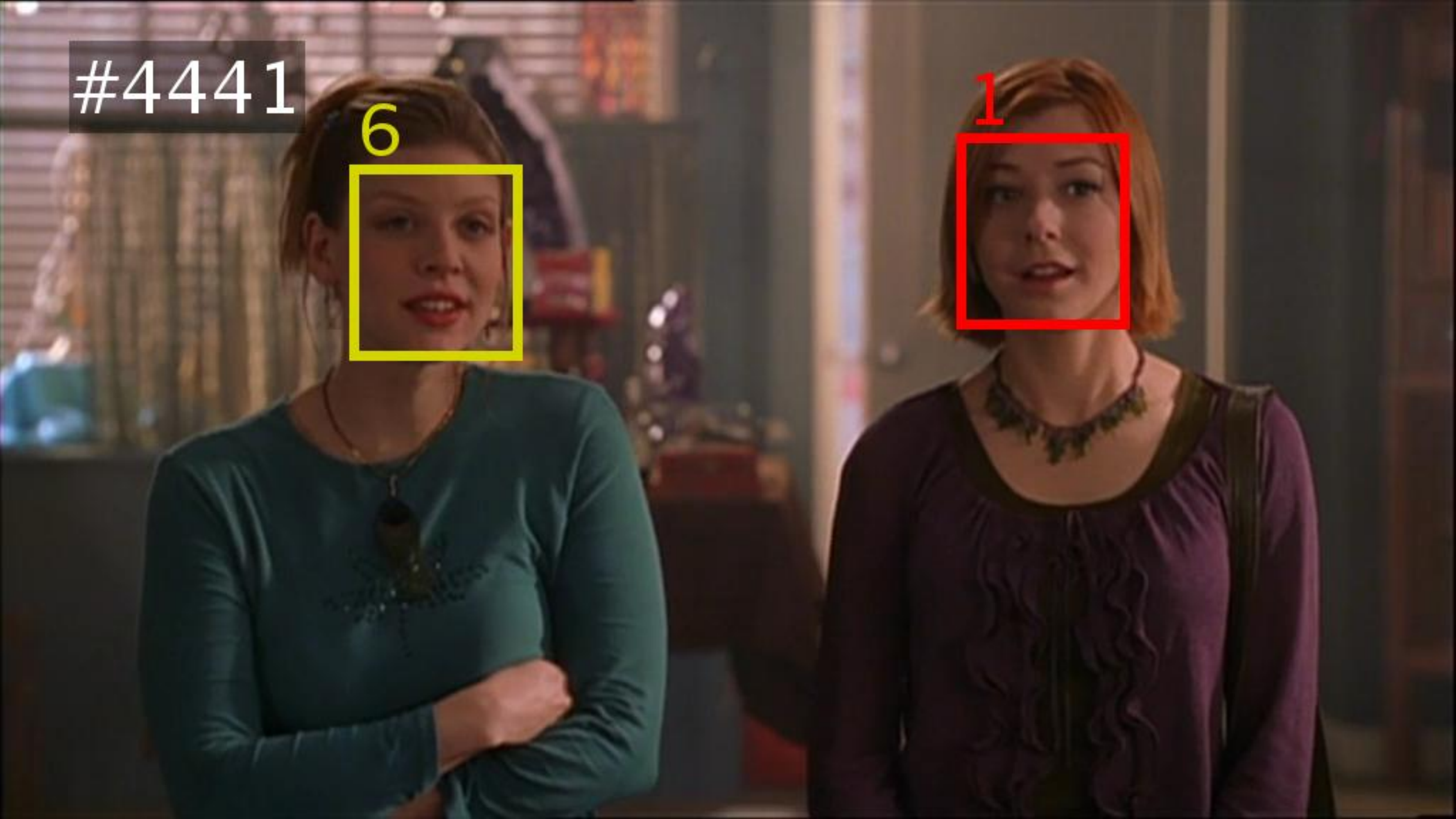}
\end{minipage}\hfill
\begin{minipage}{\figwidth} \centering
\includegraphics[width=\linewidth]{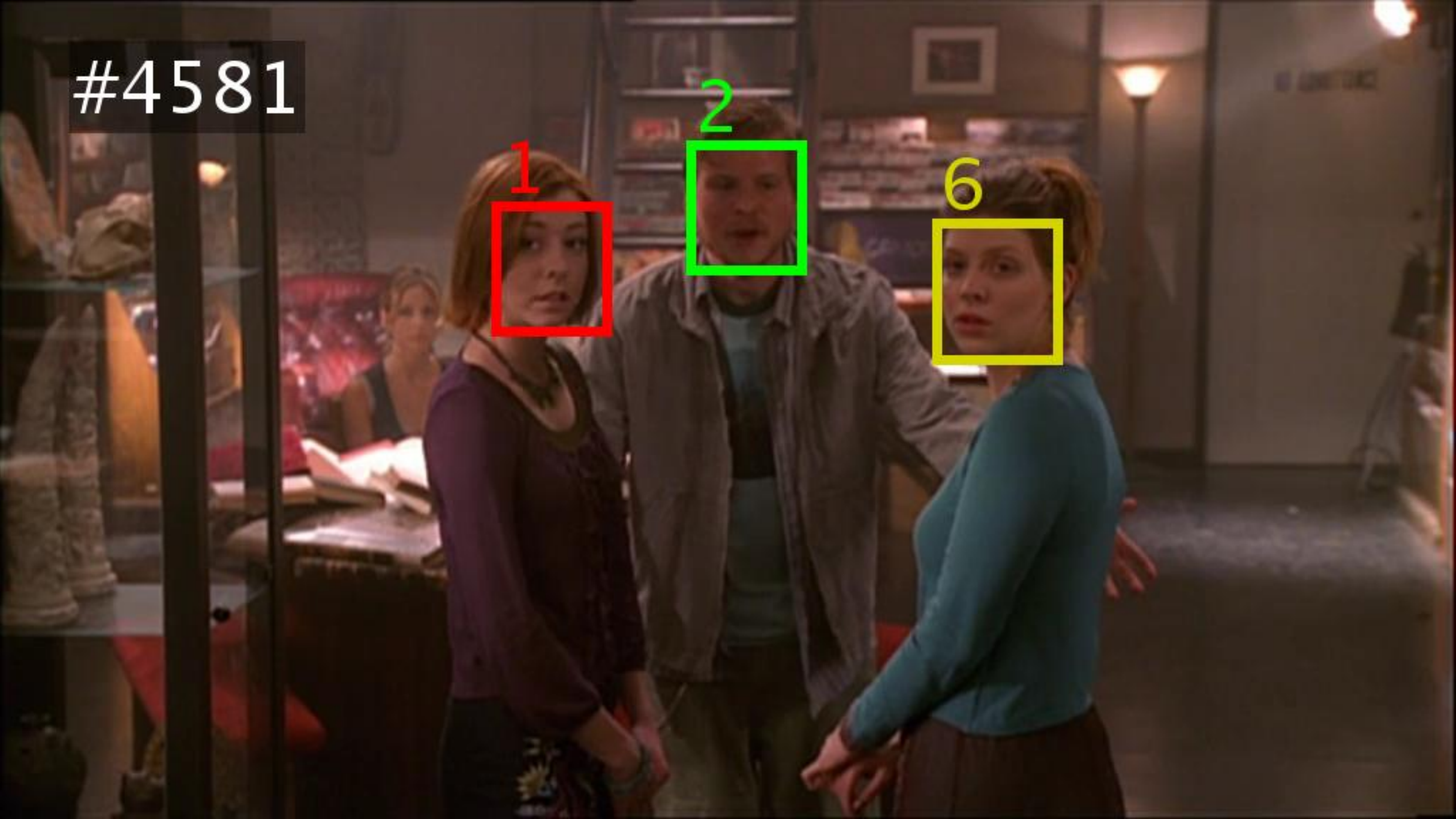}
\end{minipage}\hfill
\begin{minipage}{\figwidth} \centering
\includegraphics[width=\linewidth]{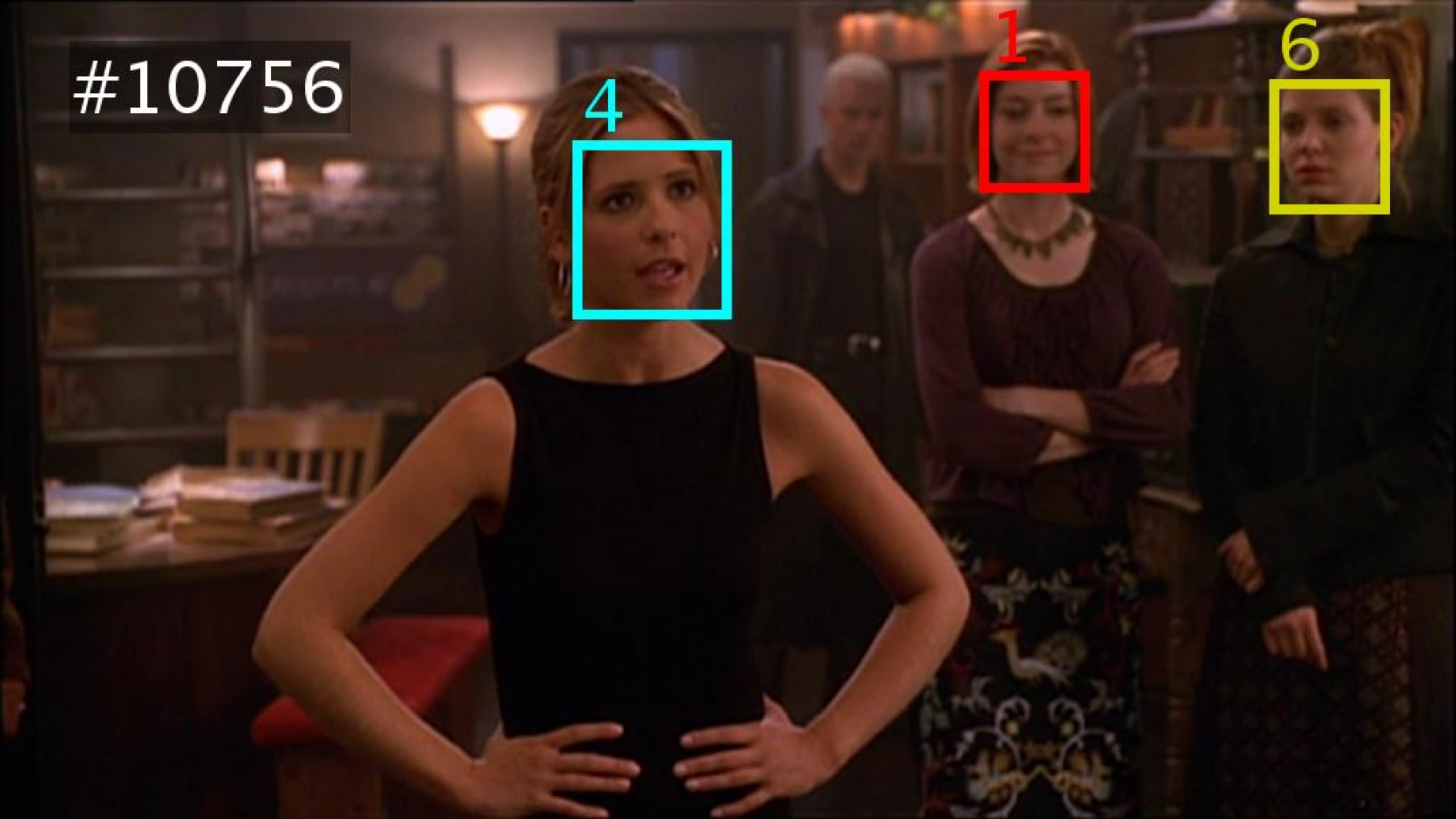}
\end{minipage}\hfill
\begin{minipage}{\figwidth} \centering
\includegraphics[width=\linewidth]{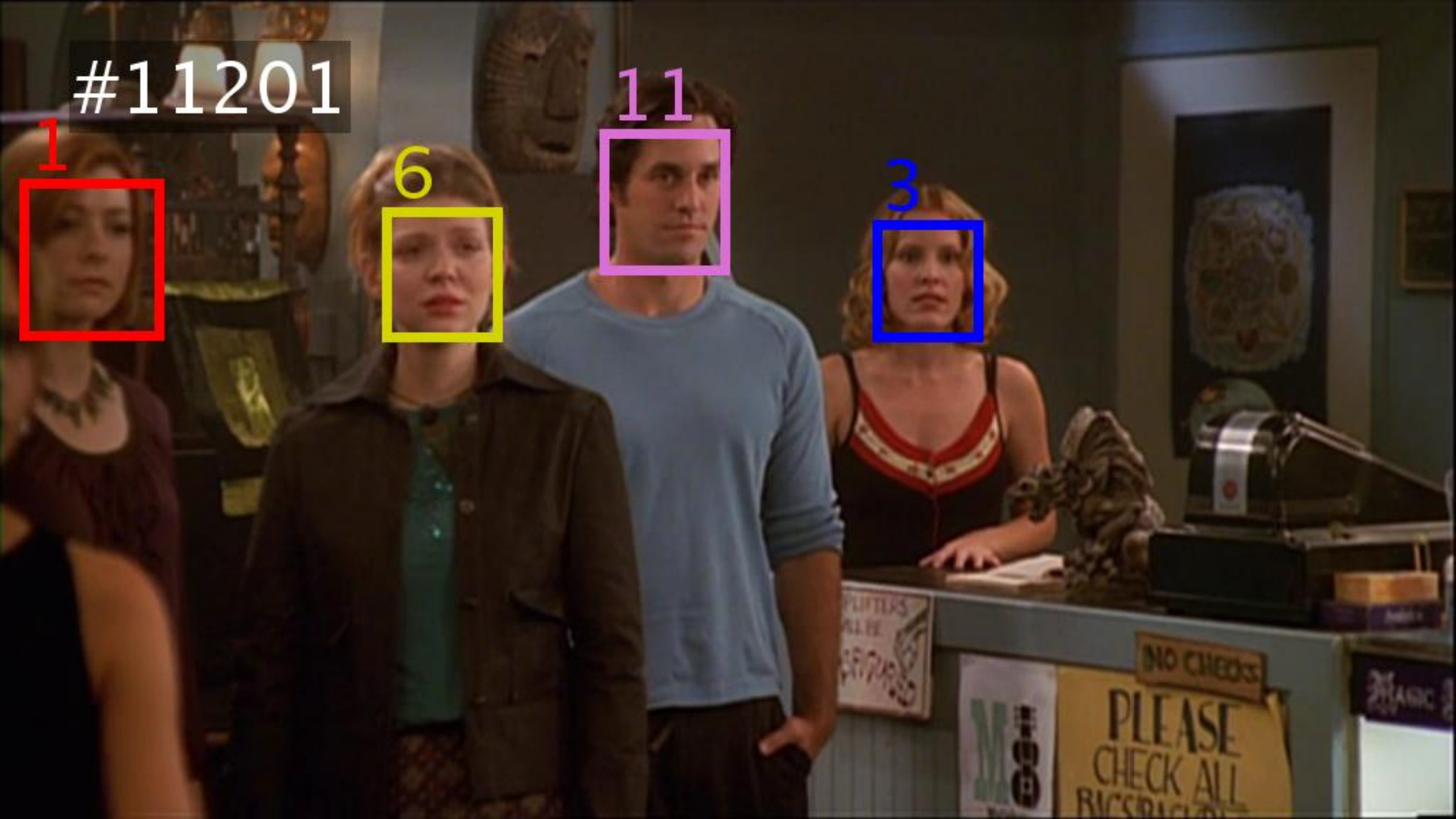}
\end{minipage}\hfill
\begin{minipage}{\figwidth} \centering
\includegraphics[width=\linewidth]{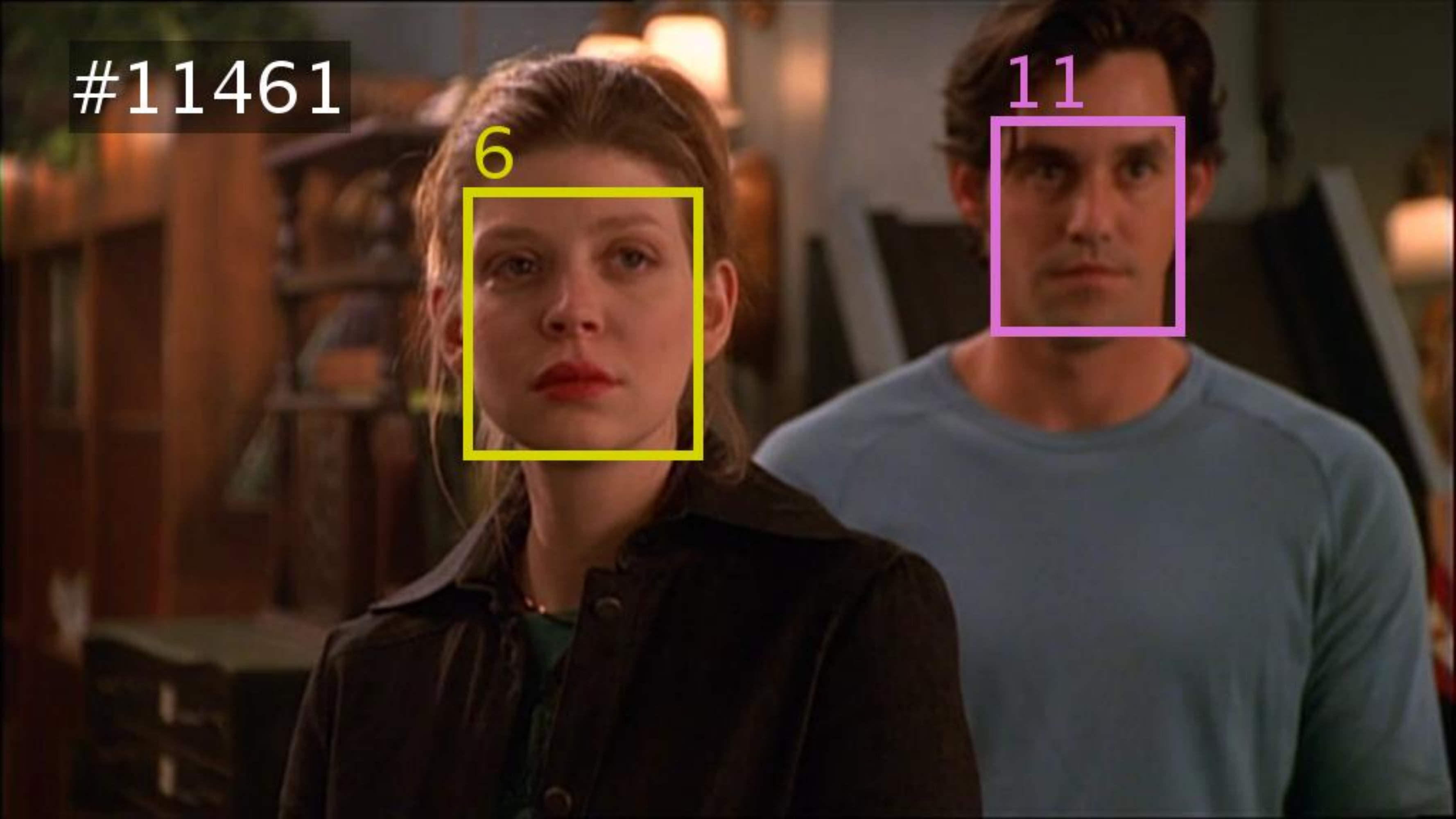}
\end{minipage}\hfill
\\
\vspace{1mm}

% BBT 01
\rotatebox[origin=c]{90}{\textsc{BBT01}} \hfill
\begin{minipage}{\figwidth} \centering
\includegraphics[width=\linewidth]{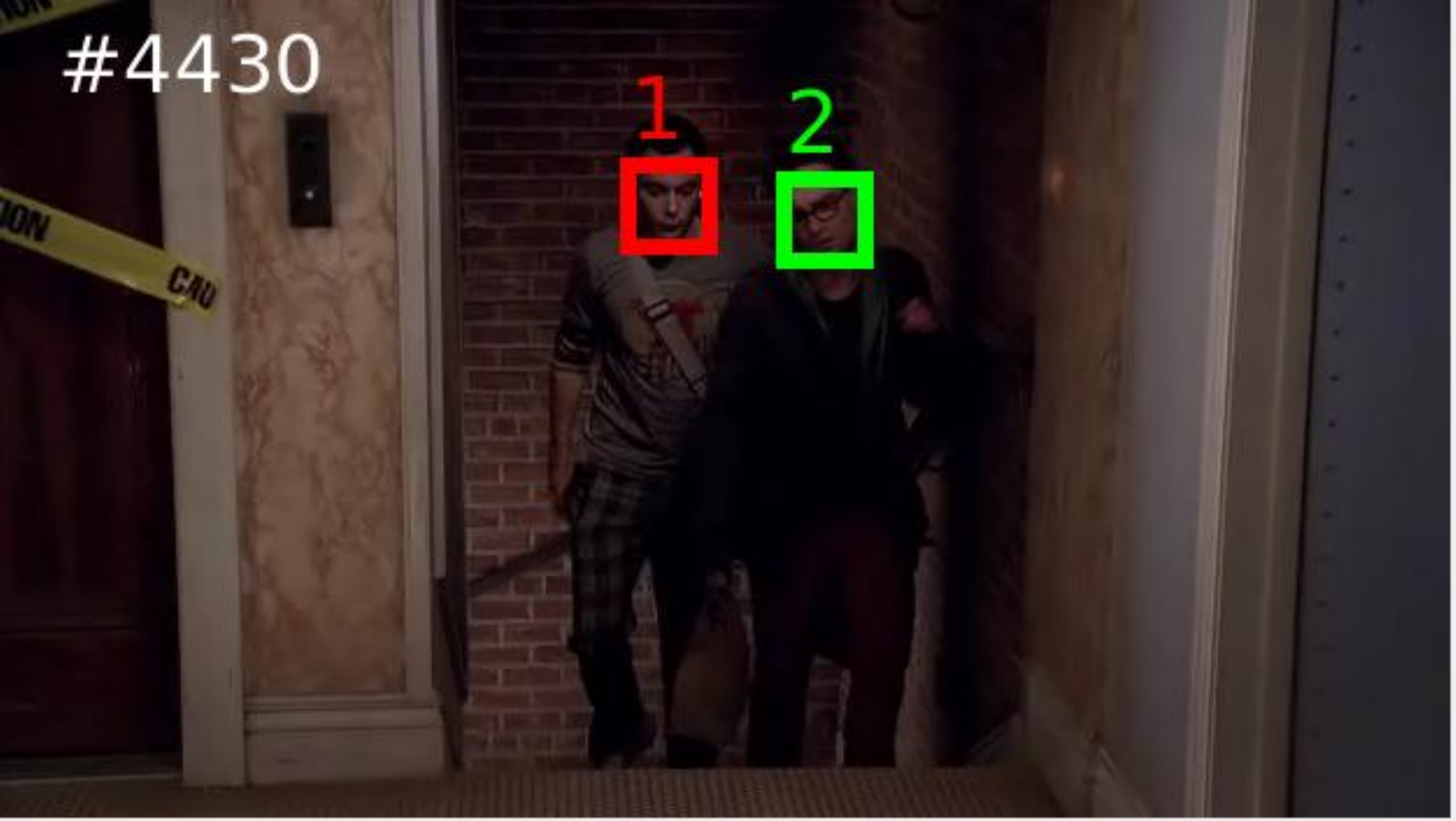}
\end{minipage}\hfill
\begin{minipage}{\figwidth} \centering
\includegraphics[width=\linewidth]{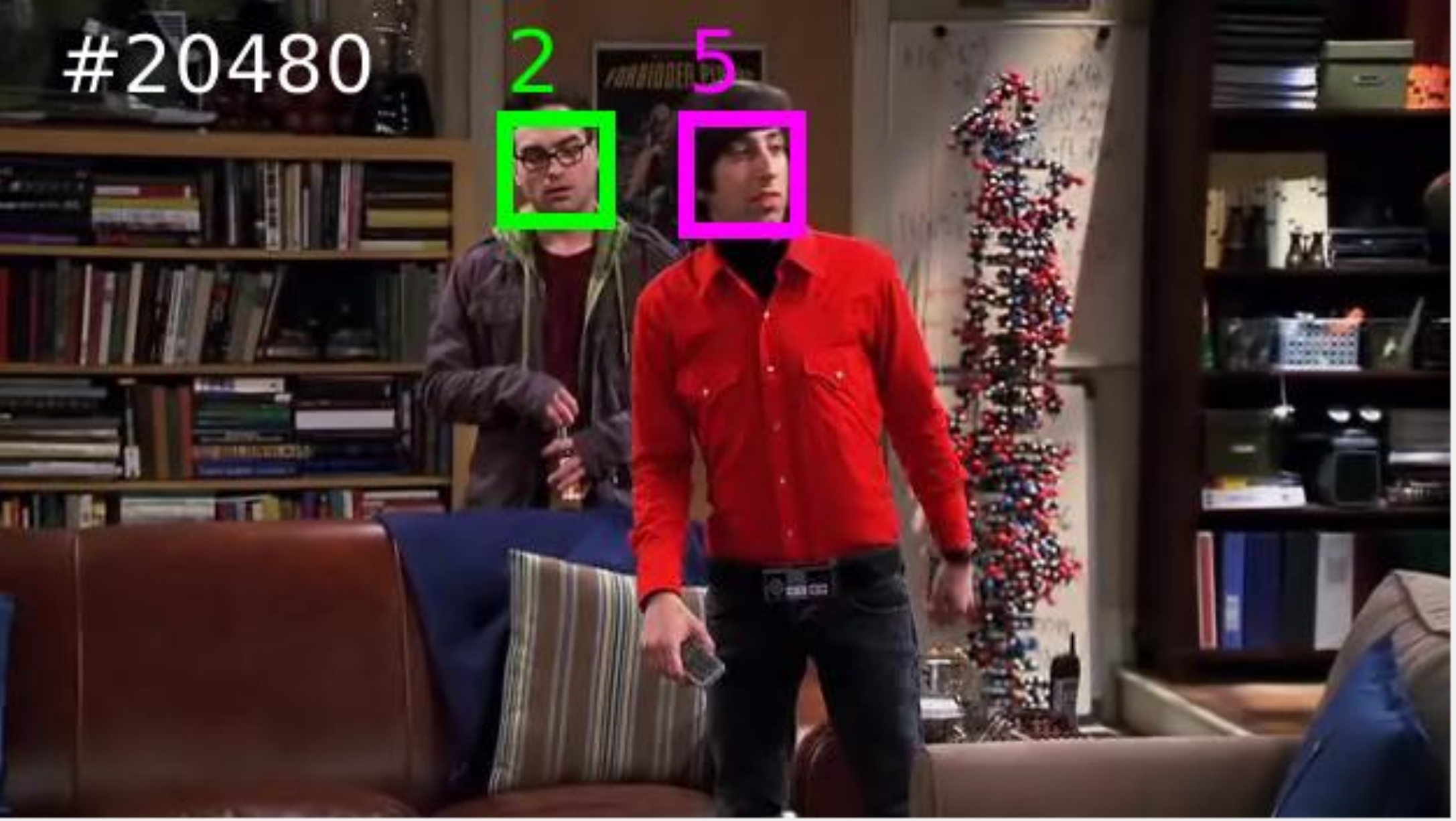}
\end{minipage}\hfill
\begin{minipage}{\figwidth} \centering
\includegraphics[width=\linewidth]{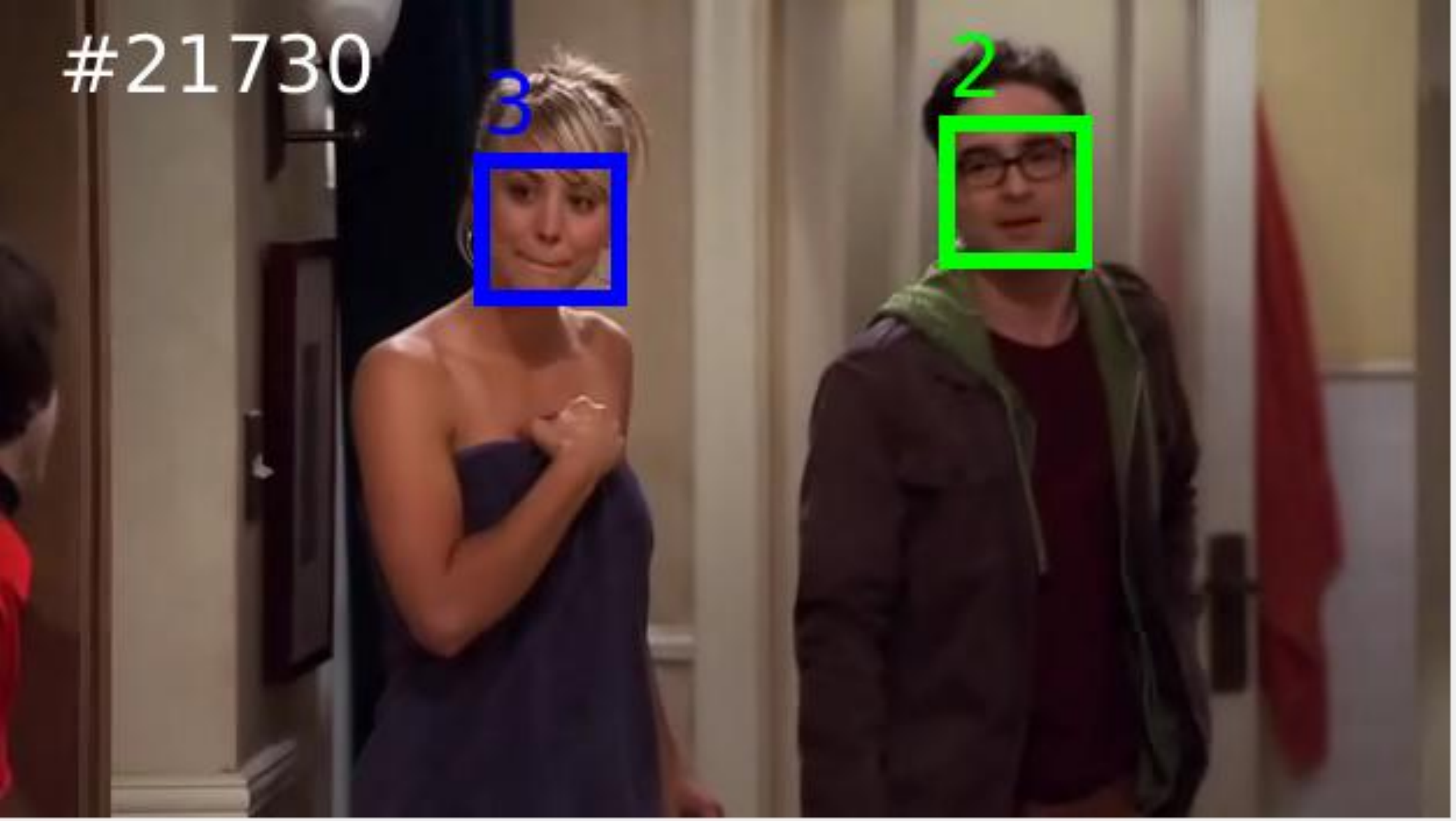}
\end{minipage}\hfill
\begin{minipage}{\figwidth} \centering
\includegraphics[width=\linewidth]{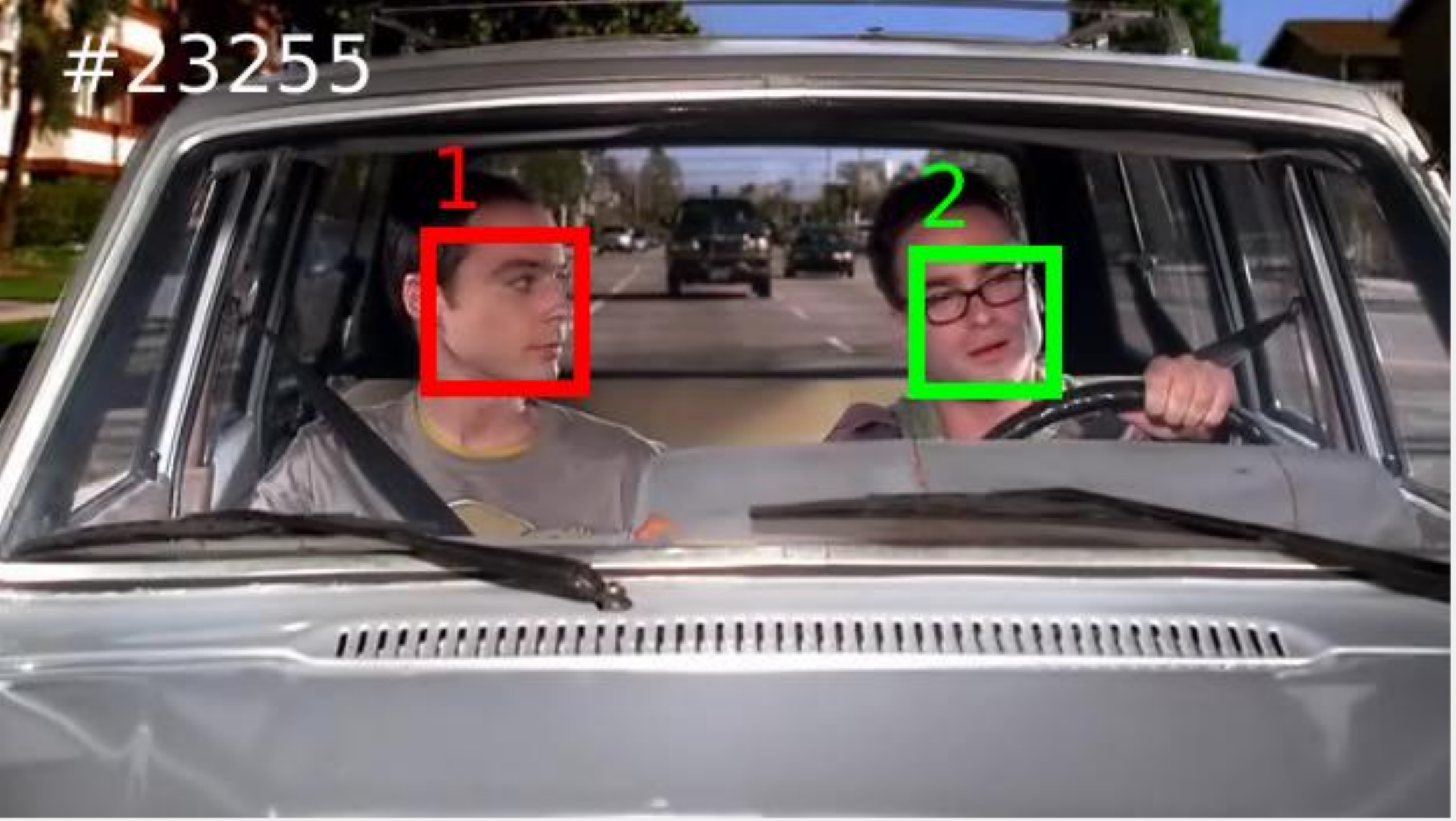}
\end{minipage}\hfill
\begin{minipage}{\figwidth} \centering
\includegraphics[width=\linewidth]{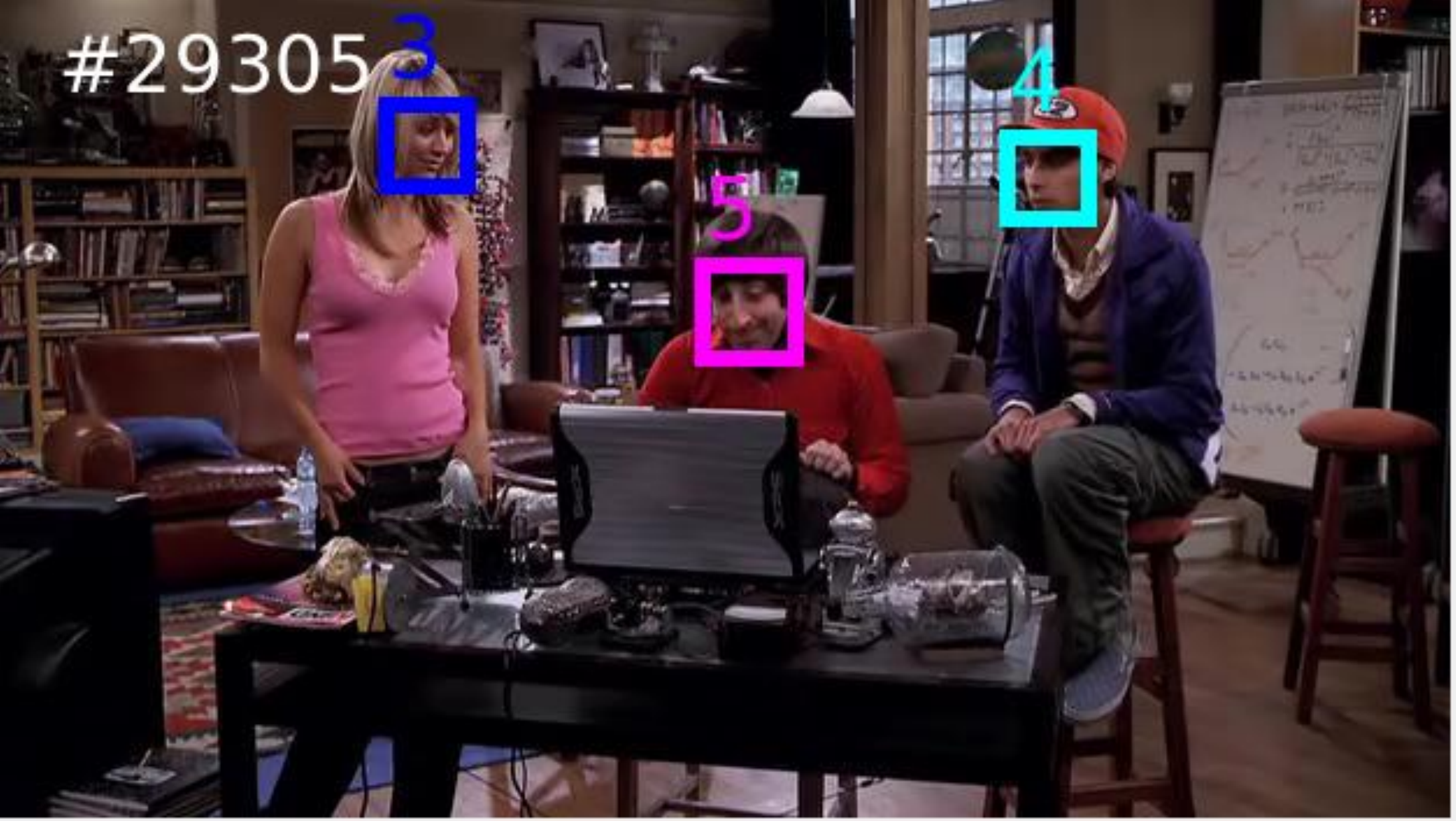}
\end{minipage}\hfill
\\
\vspace{1mm}

% BBT 02
\rotatebox[origin=c]{90}{\textsc{BBT02}} \hfill
\begin{minipage}{\figwidth} \centering
\includegraphics[width=\linewidth]{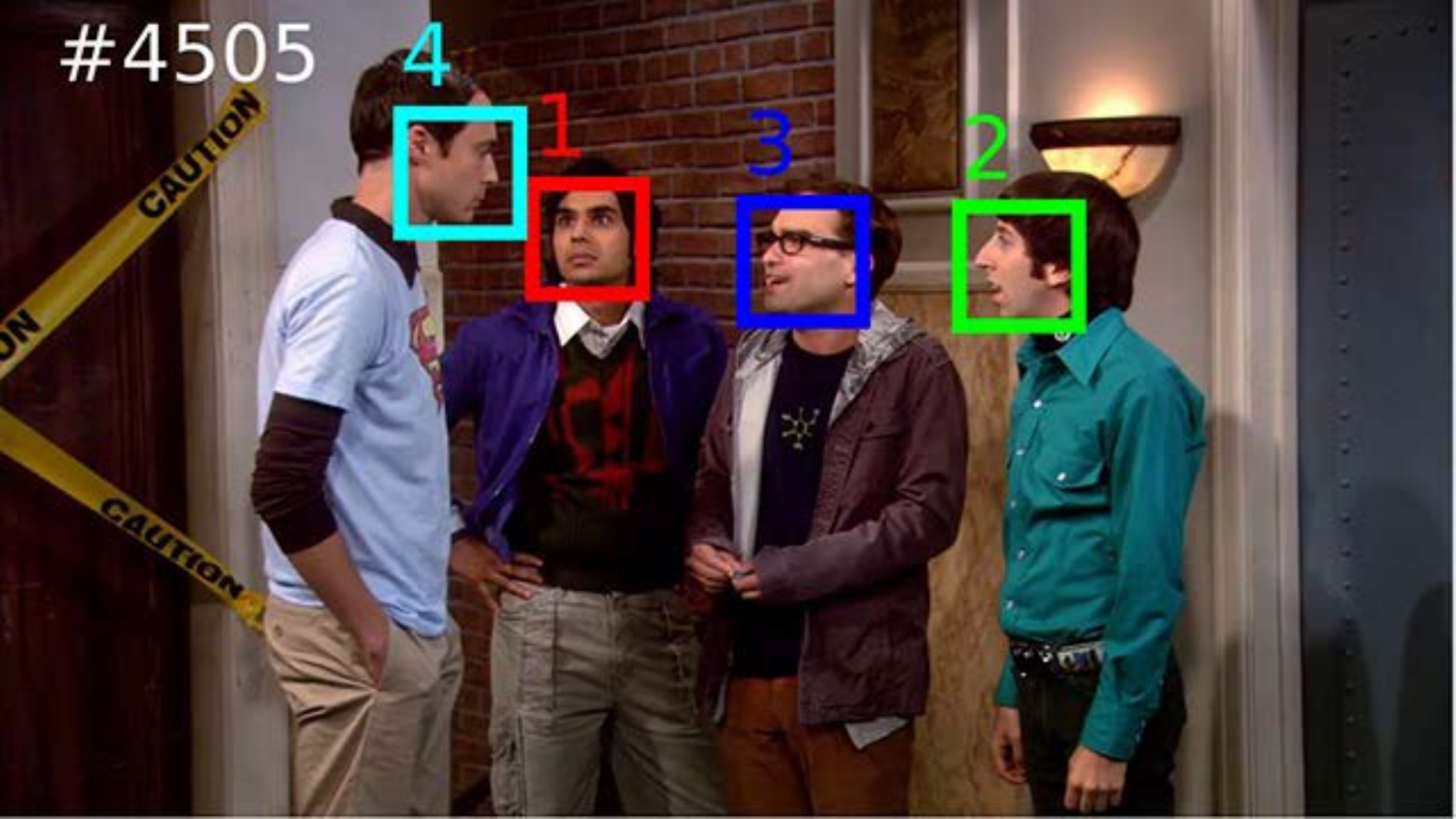}
\end{minipage}\hfill
\begin{minipage}{\figwidth} \centering
\includegraphics[width=\linewidth]{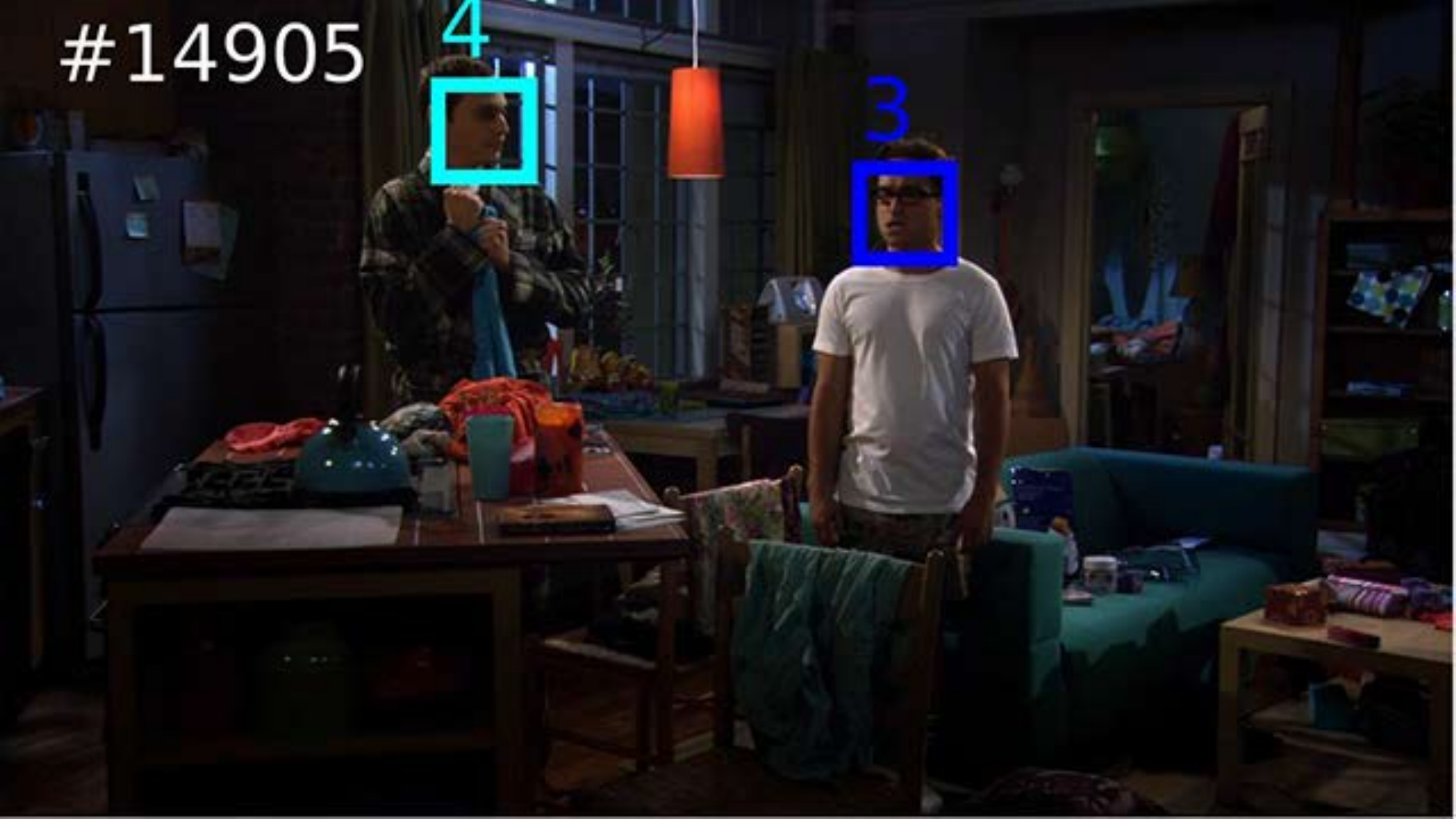}
\end{minipage}\hfill
\begin{minipage}{\figwidth} \centering
\includegraphics[width=\linewidth]{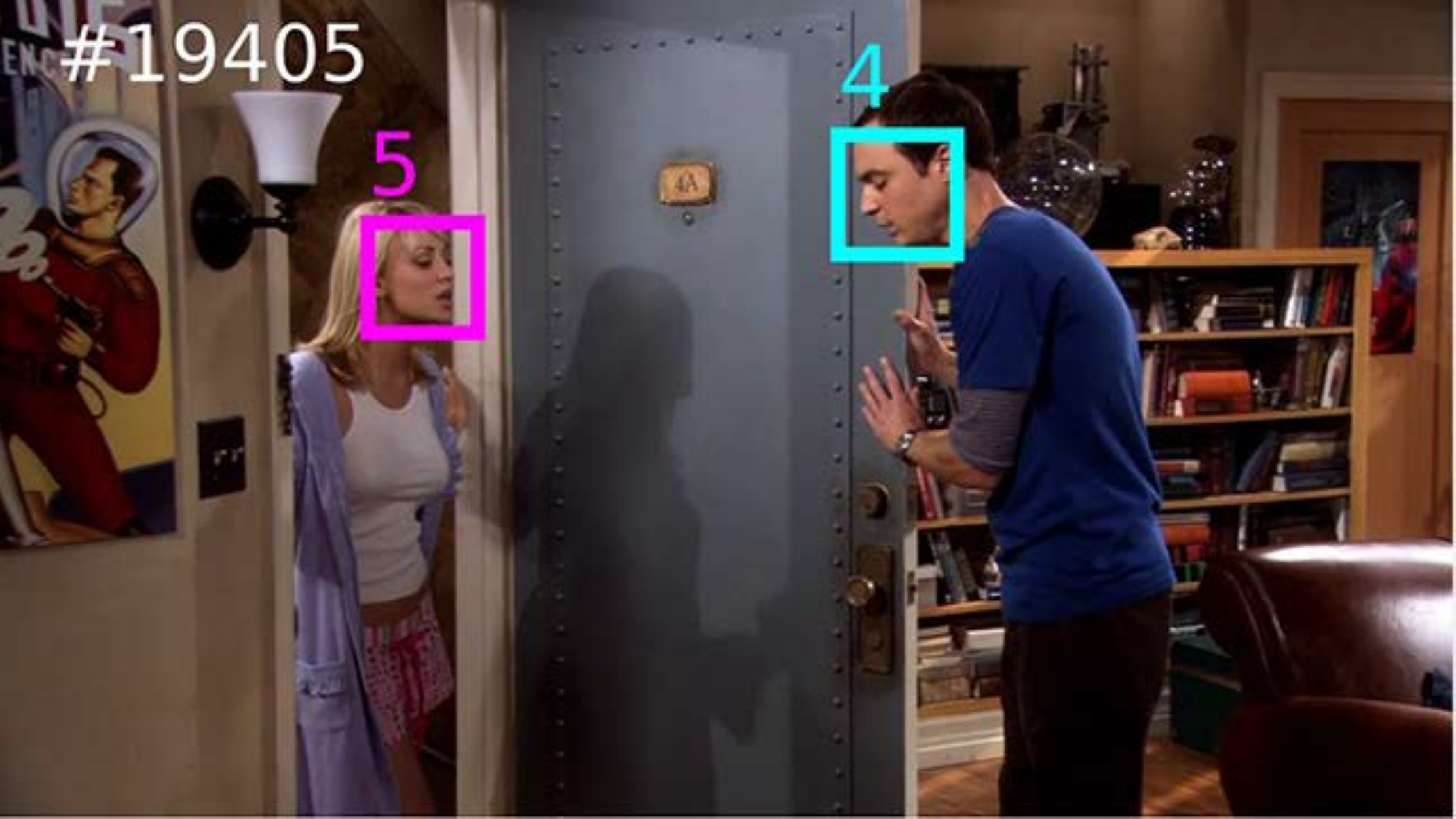}
\end{minipage}\hfill
\begin{minipage}{\figwidth} \centering
\includegraphics[width=\linewidth]{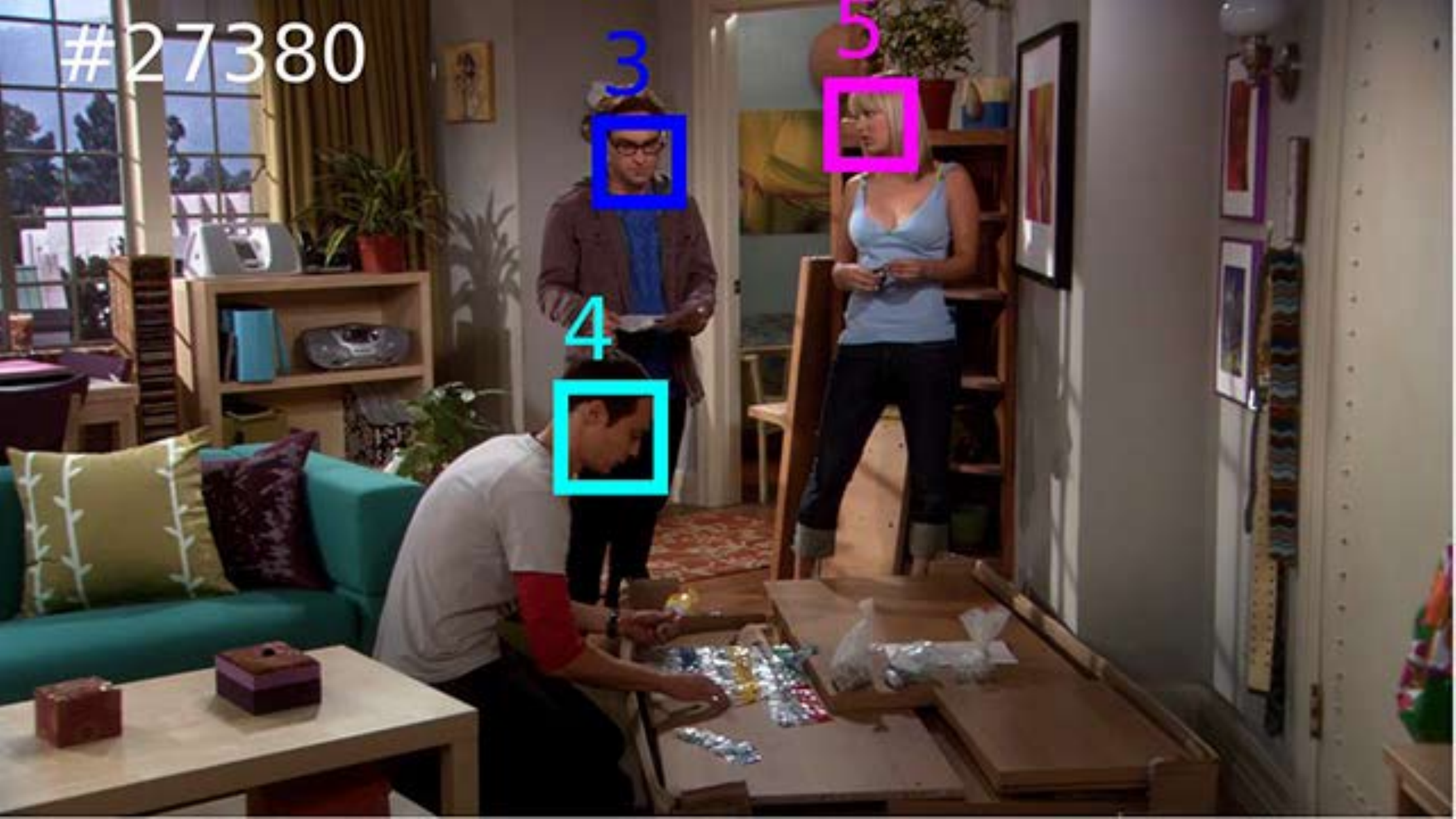}
\end{minipage}\hfill
\begin{minipage}{\figwidth} \centering
\includegraphics[width=\linewidth]{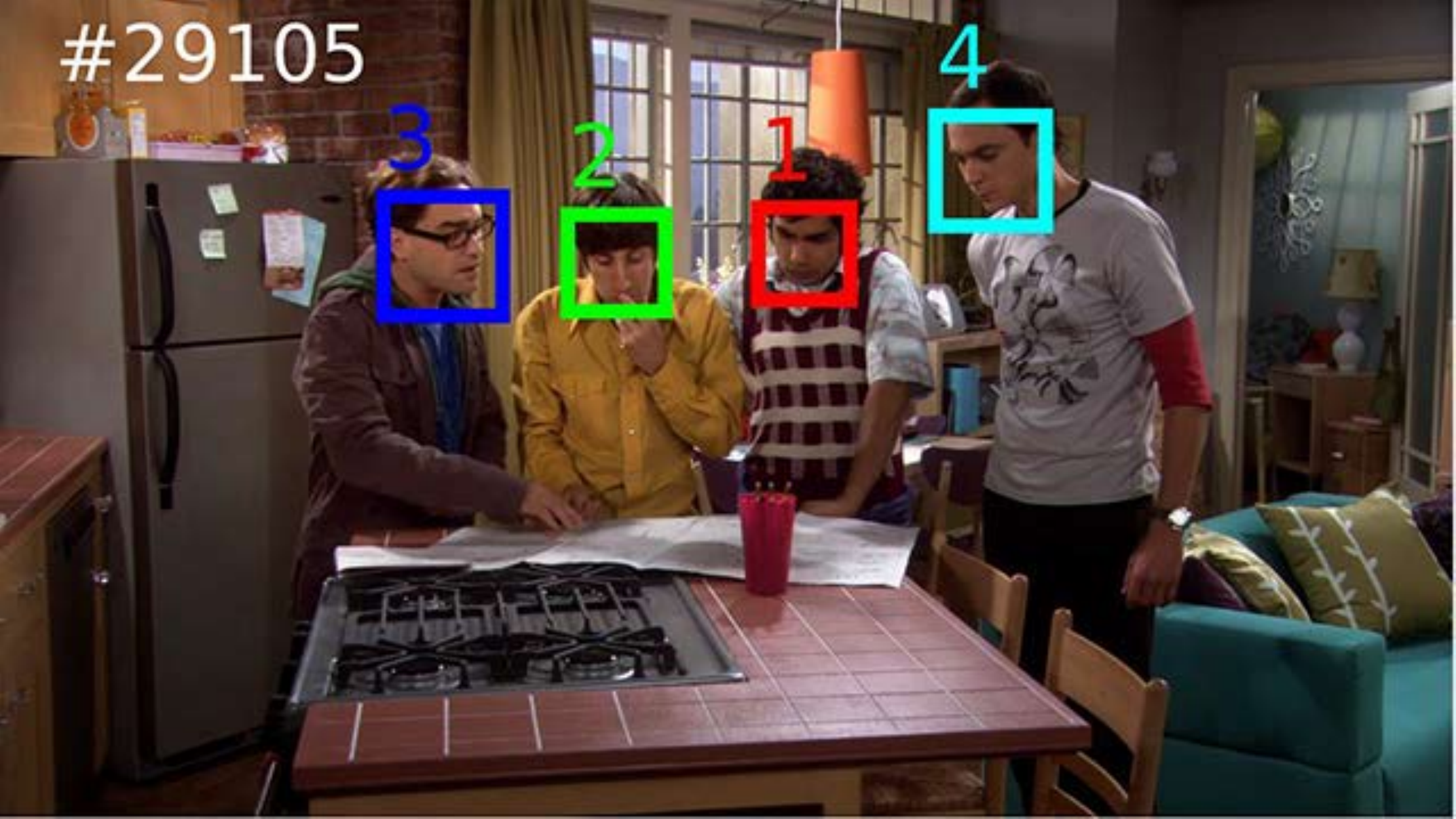}
\end{minipage}\hfill
\\
\vspace{1mm}

% BBT 02
\rotatebox[origin=c]{90}{\textsc{BBT05}} \hfill
\begin{minipage}{\figwidth} \centering
\includegraphics[width=\linewidth]{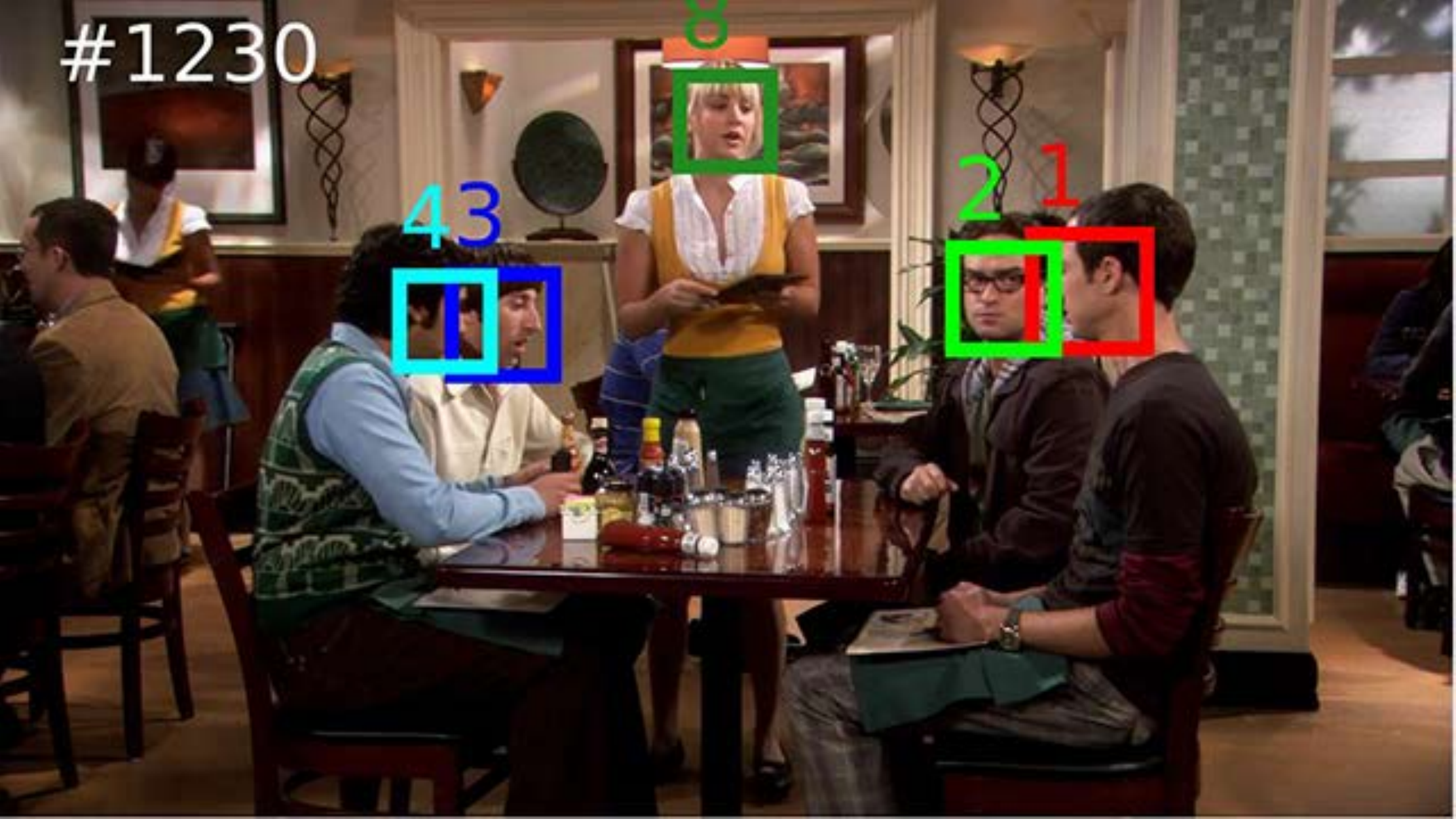}
\end{minipage}\hfill
\begin{minipage}{\figwidth} \centering
\includegraphics[width=\linewidth]{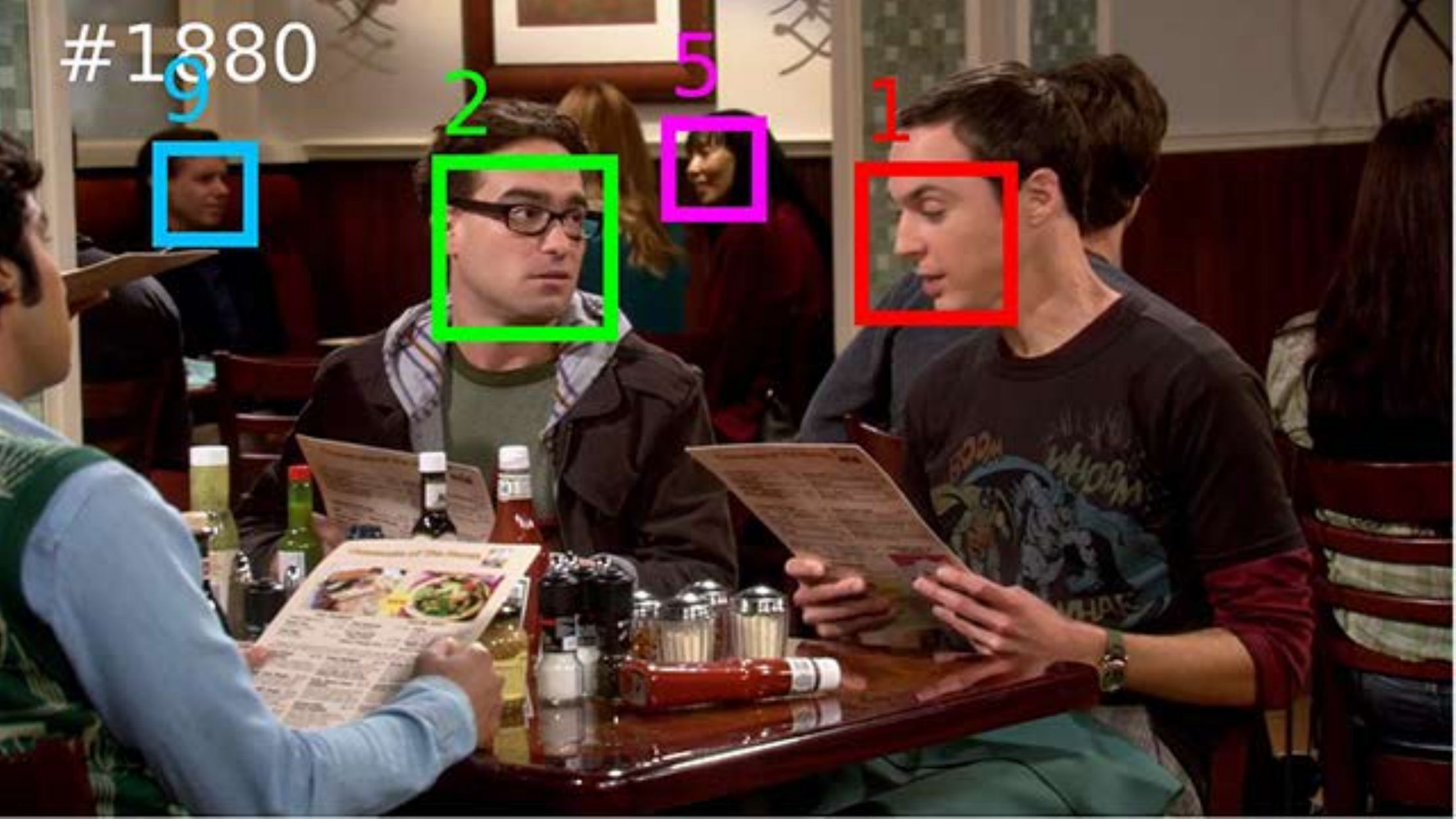}
\end{minipage}\hfill
\begin{minipage}{\figwidth} \centering
\includegraphics[width=\linewidth]{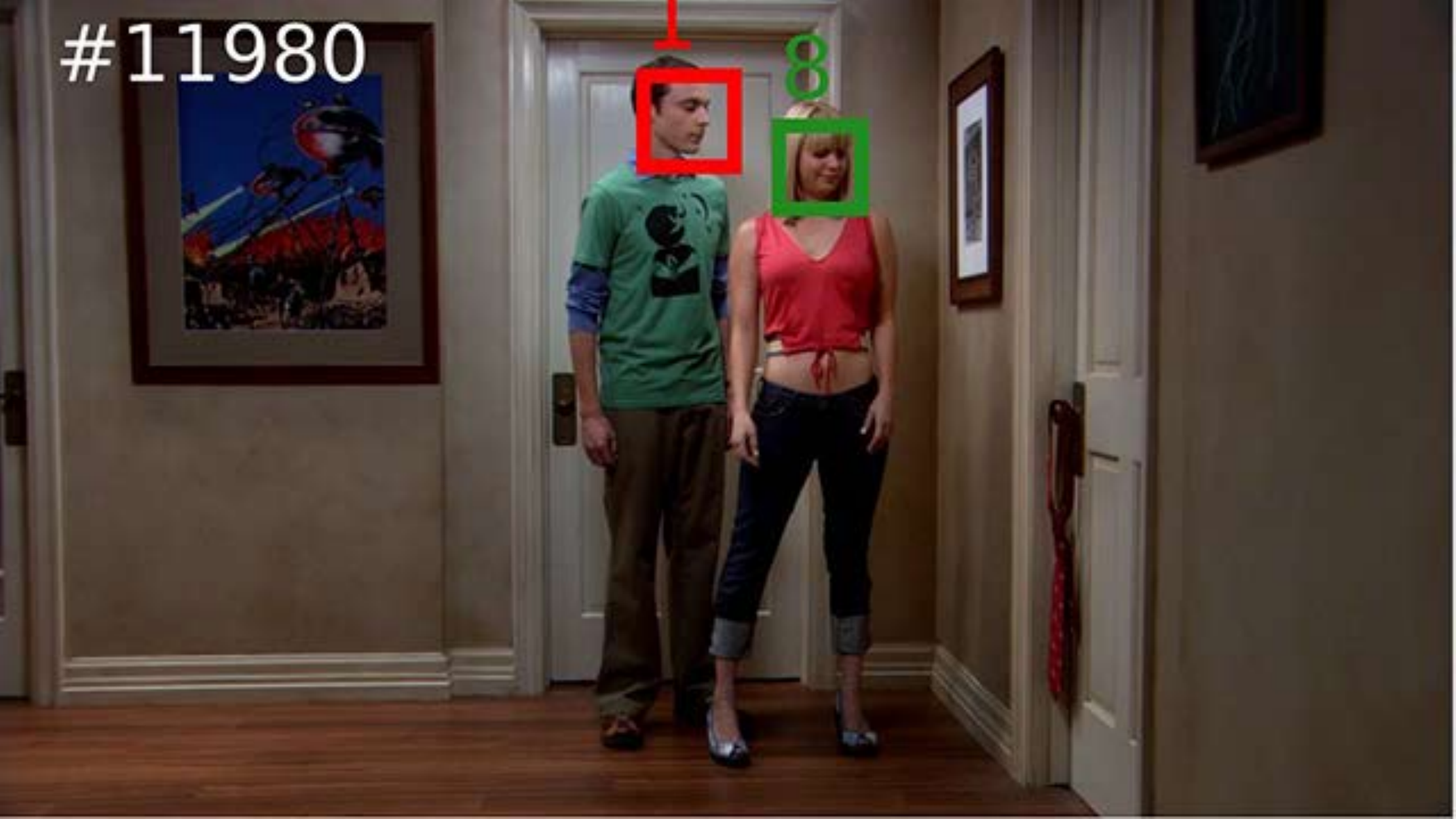}
\end{minipage}\hfill
\begin{minipage}{\figwidth} \centering
\includegraphics[width=\linewidth]{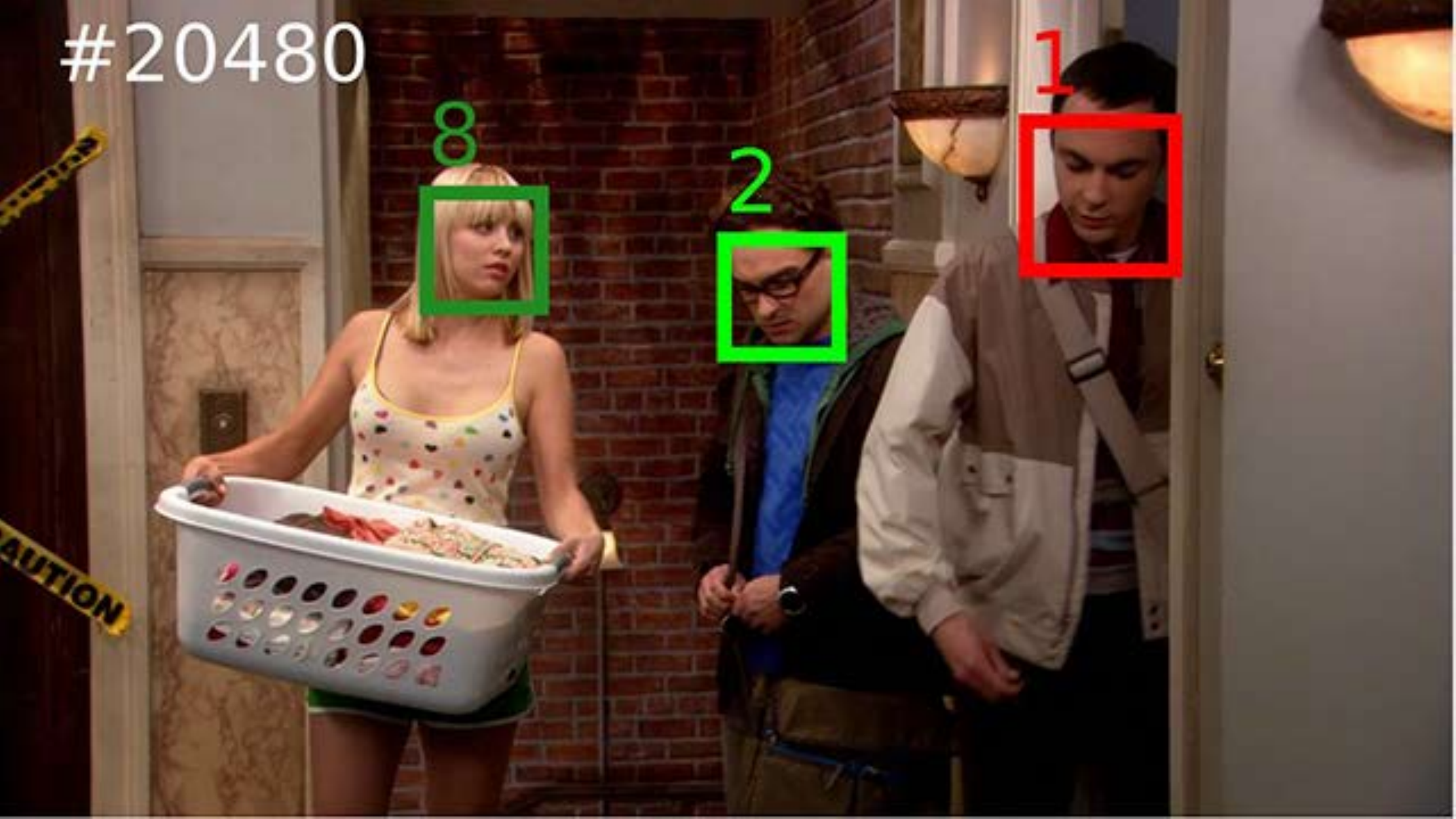}
\end{minipage}\hfill
\begin{minipage}{\figwidth} \centering
\includegraphics[width=\linewidth]{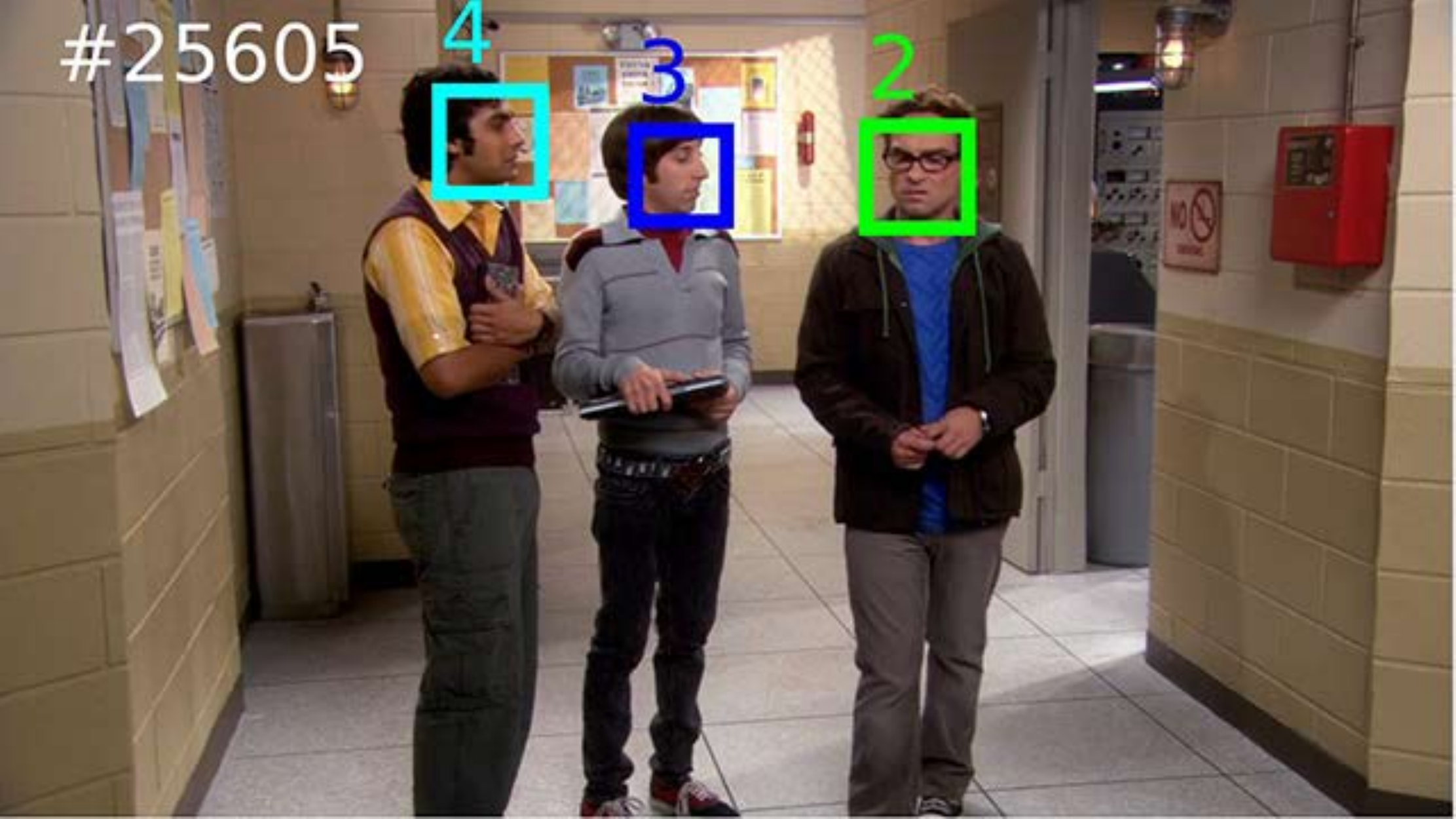}
\end{minipage}\hfill

\vspace{-1mm}
\caption{\textbf{Tracking results on Buffy and BBT dataset}. The faces of the different people are color coded.
}
\label{fig:samples1}
\vspace{-3mm}
\end{figure*}

\vspace{1mm}
\noindent \textbf{Qualitative results.}
Figure~\ref{fig:samples} shows sample tracking results of our algorithm with Ours-SymTriplet-Contx features on all eight music videos.
Figure~\ref{fig:samples1} shows the results on three BUFFY videos and three selected BBT sequences.
The numbers and the colors indicate the inferred identities of the targets.
The proposed algorithm is able to track multiple faces well despite large appearance variations in unconstrained videos.
In Figure~\ref{fig:samples}, for example, there are significant changes in scale and appearance (due to makeup and hairstyle) in the \textsc{Hello Bubble} sequence (first row).
In the fourth row, the six singers have similar looks and thus make multi-face tracking particularly challenging within and across shots.
Nonetheless, our approach can distinguish the faces and track them reliably with few id switches.
The results in other rows illustrate that our method is able to generate correct identities and trajectories when the same person appears in different shots or different scenes.

\vspace{-3mm}
\subsection{Pedestrian Tracking Across Cameras}
\vspace{-1mm}
\label{sec:MPT}

In this section, we show that the proposed method for learning adaptive discriminative features from tracklets is also applicable to other objects, e.g., pedestrians or cars in surveillance videos.
We validate our approach on the task of pedestrian tracking from multiple non-overlapping cameras.

The problem of multiple target tracking across cameras is challenging as we need to re-identify people from different images acquired at different viewing angles and imaging conditions.
In unconstrained scenes, the appearances of people also exhibit significant differences across cameras.
The motion cues of people are unreliable due to the non-overlapping views without knowing camera configurations apriori.
The re-identification problem becomes even more challenging when a large number of people needs to be tracked across views.

Similar to pre-training a CNN using face recognition dataset for learning identity-preserving features, we first train a CNN for people re-identification using the Market1501 dataset~\cite{zheng2015scalable} containing 32,668 images of 1,501 identities.
We evaluate our method on the DukeMTMC~\cite{ristani2016performance} dataset which contains surveillance footage from 8 cameras with approximately 85 minutes of videos for each one.
%
% All the videos have a high resolution of $1920 \times 1080$ pixels and captured at 60 frames per second.

%In the dataset, camera 2 and 5 are disjoint and have the most number of people (934 in total with 311 individuals appearing in both cameras).
%
We conduct the experiment using images from camera 2 and 5 because they are disjoint and have the most number of people.
We first use the two-threshold strategy to generate tracklets on the videos from both cameras.
Next, we collect training samples based on the tracklets using spatio-temporal and contextual constraints.
Similar to the experiments on multi-face tracking, we fine-tune the pre-trained CNN with the discovered training samples using the SymTriplet loss function.

\begin{figure}[t]
\scriptsize
\centering
\begin{tabular}{{c}{c}}
%\hspace{-1.0mm}\includegraphics[width=3.4cm]{figs_exp7/cam2_124114.pdf} &
\hspace{0mm}\includegraphics[width=3.5cm]{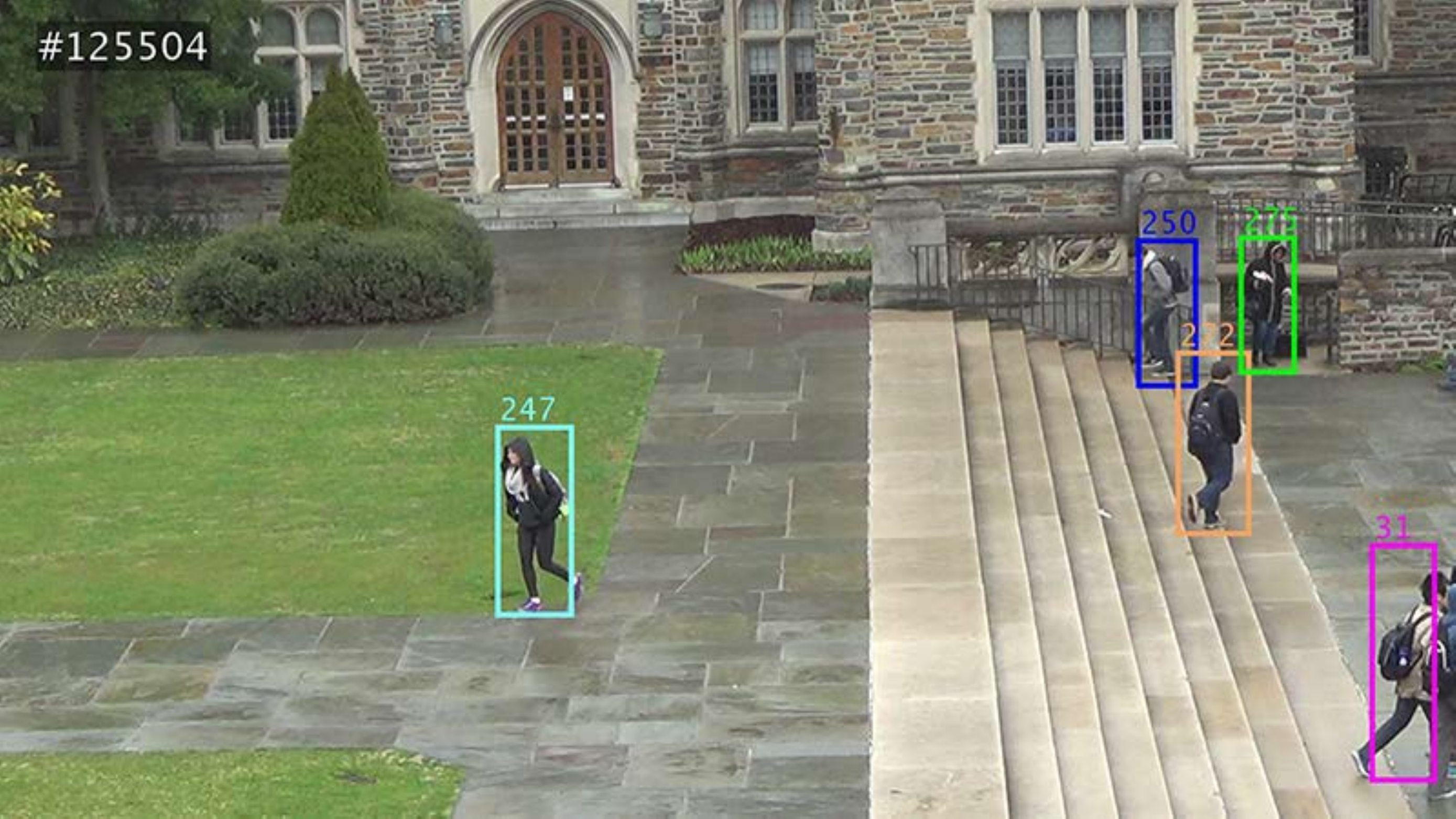} &
%\hspace{-2.5mm}\includegraphics[width=3.4cm]{figs_exp7/cam2_125517.pdf} &
%\hspace{-2.5mm}\includegraphics[width=3.4cm]{figs_exp7/cam2_130445.pdf} &
\hspace{-2.5mm}\includegraphics[width=3.5cm]{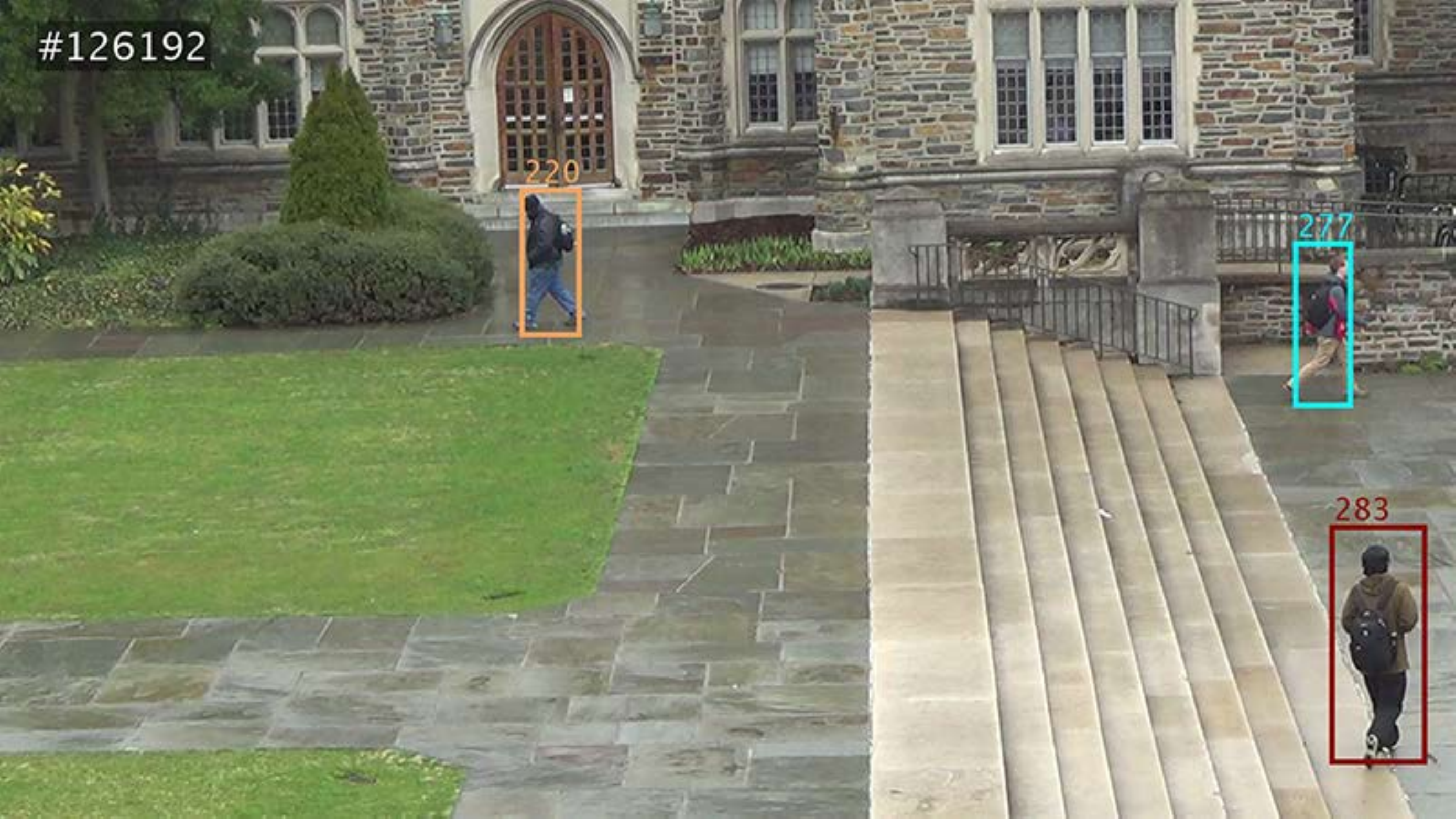} \\

%\hspace{-1.0mm}\includegraphics[width=3.4cm]{figs_exp7/cam5_128781.pdf} &
\hspace{0mm}\includegraphics[width=3.5cm]{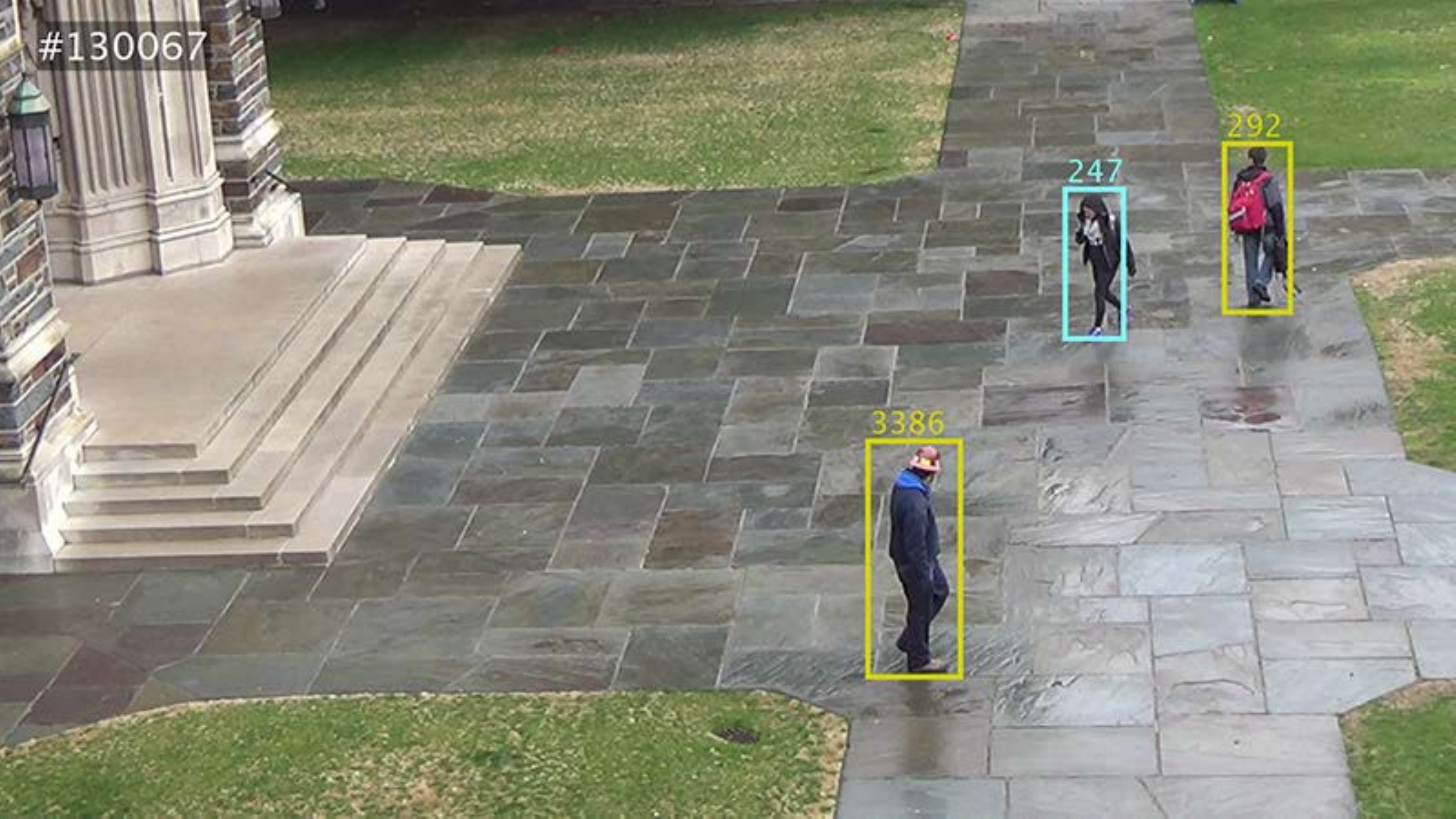} &
%\hspace{-2.5mm}\includegraphics[width=3.4cm]{figs_exp7/cam5_127720.pdf} &
%\hspace{-2.5mm}\includegraphics[width=3.4cm]{figs_exp7/cam5_136078.pdf} &
\hspace{-2.5mm}\includegraphics[width=3.5cm]{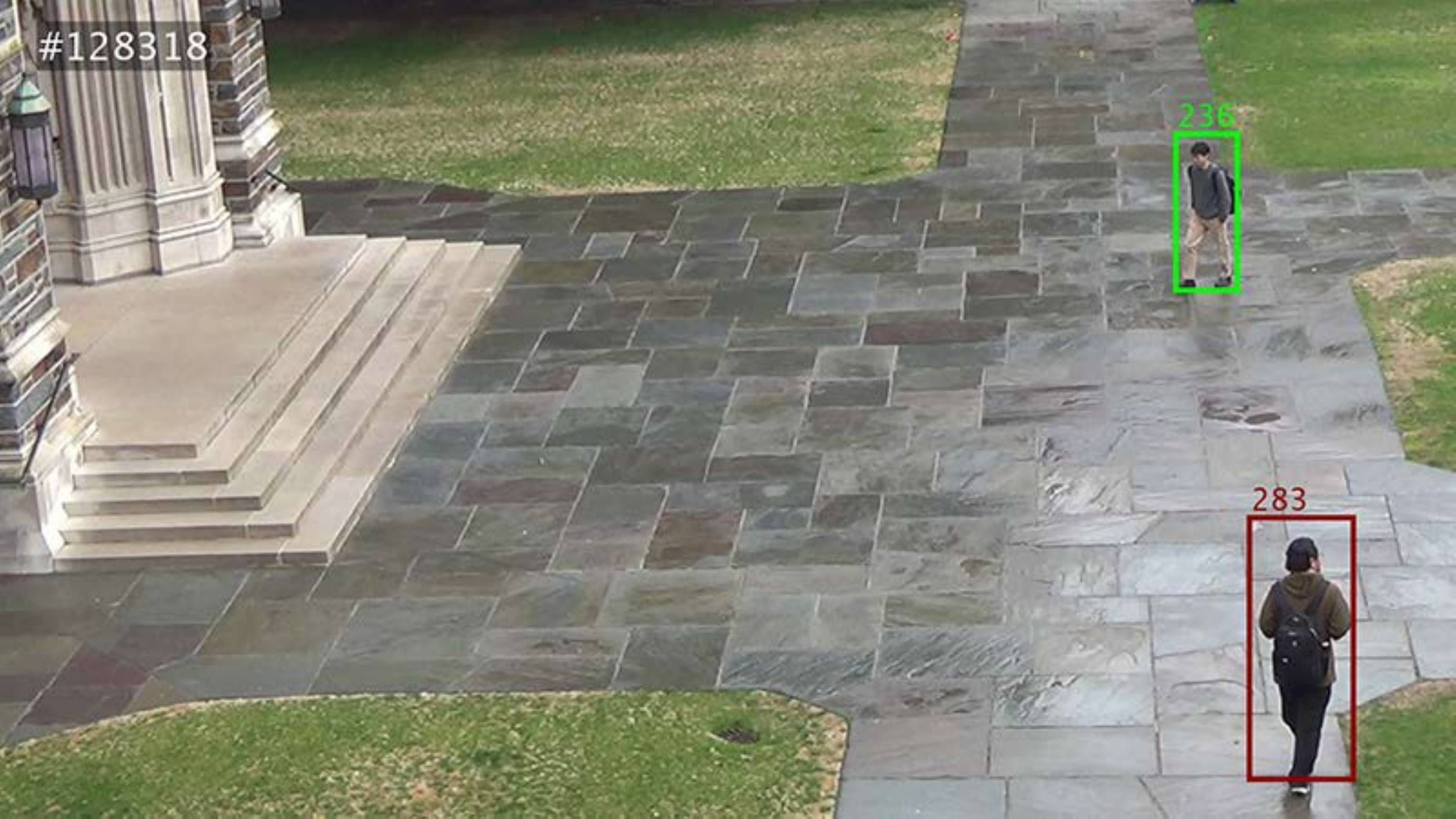}\\
\end{tabular}
\vspace{-1mm}
\caption{\textbf{Sample pedestrian tracking results}.
Shown from the top to bottom are Camera 2 and Camera 5 of the DukeMTMC dataset. The different people are color coded.
}
\vspace{-3mm}
\label{fig:pedsamples}
\end{figure}

After extracting the learned features for each detection, we first link the tracklets within one camera into camera-level trajectories.
We then group these camera-level trajectories into tracking results across the two cameras.
Following~\cite{ristani2016performance}, we measure the tracking performance using identification precision (IDP), identification recall (IDR), and the corresponding F1 score IDF1, as well as other metrics.
The identification precision (recall) is the fraction of computed (ground truth) detections that are correctly identified.
The IDF1 metric is the ratio of correctly identified detections over the average number of ground-truth and computed detections.
Both ID precision and ID recall indicate tracking trade-offs, while the IDF1 score allows ranking all methods on a single scale that balances identification precision and recall through the harmonic mean.
Table~\ref{tab:MPT} shows the tracking results on both cameras in the DukeMTMC dataset.
Overall, the proposed method performs favorably against the other methods in~\cite{ristani2016performance} in term of IDS, MOTA, IDP and IDF1.
We show sample visual results of the DukeMTMC datset in Figure \ref{fig:pedsamples}.
% JB: There are NO person 237. It's impossible to read digits in the figure. Highlight them directly in the figure. Please use the SAME color and ID to match the persons.
Person 237 and person 283 both appear in Camera 2 and Camera 5, and are correctly matched across cameras with our method.
%}

\begin{table}[t]
\scriptsize
\centering
\caption{Tracking results on the DukeMTMC datset.}
\vspace{-1mm}
\resizebox{\linewidth}{!}{
\begin{tabular}{@{}lccccccc@{}}
\toprule
\multicolumn{8}{c}{Camera 2} \\
\midrule
\textbf{Method} &\textbf{IDS}$\downarrow$ &\textbf{Frag}$\downarrow$ &\textbf{MOTA}$\uparrow$ &\textbf{MOTP}$\uparrow$ &\textbf{IDP}$\uparrow$ &\textbf{IDR}$\uparrow$ &\textbf{IDF1}$\uparrow$ \\ \midrule
Ergys Ristani et al.~\cite{ristani2016performance}
&866&1929&49.2\%&61.7\%
&69.1\%&{63.8\%}&66.3\%\\
Ours-SymTriplet-Contx
&802&{2018}& {51.4}& 60.9 &69.6\% &64.8\% &{66.7\%}
\\
\bottomrule
\toprule
\multicolumn{8}{c}{Camera 5} \\
\midrule
\textbf{Method} &\textbf{IDS}$\downarrow$ &\textbf{Frag}$\downarrow$ &\textbf{MOTA}$\uparrow$ &\textbf{MOTP}$\uparrow$ &\textbf{IDP}$\uparrow$ &\textbf{IDR}$\uparrow$ &\textbf{IDF1}$\uparrow$ \\ \midrule
Ergys Ristani et al.~\cite{ristani2016performance}
&162&292&73.1\%&70.5\%
&84.9\%&{68.0\%}&75.5\%\\
Ours-SymTriplet-Contx
&147&{316}& {76.2\%}& 68.5\% &86.2\% &69.5\% &{77.0\%}
\\
\bottomrule
\end{tabular}
}
\label{tab:MPT}
\vspace{-3mm}
\end{table}

\vspace{-3mm}
\subsection{Discussions}
\vspace{-1mm}
\label{sec:discussions}

\begin{figure*}[t]
\setlength{\belowcaptionskip}{0.cm}
\scriptsize
\centering
\begin{tabular}{{c}{c}{c}{c}{c}}
\includegraphics[width=3.0cm]{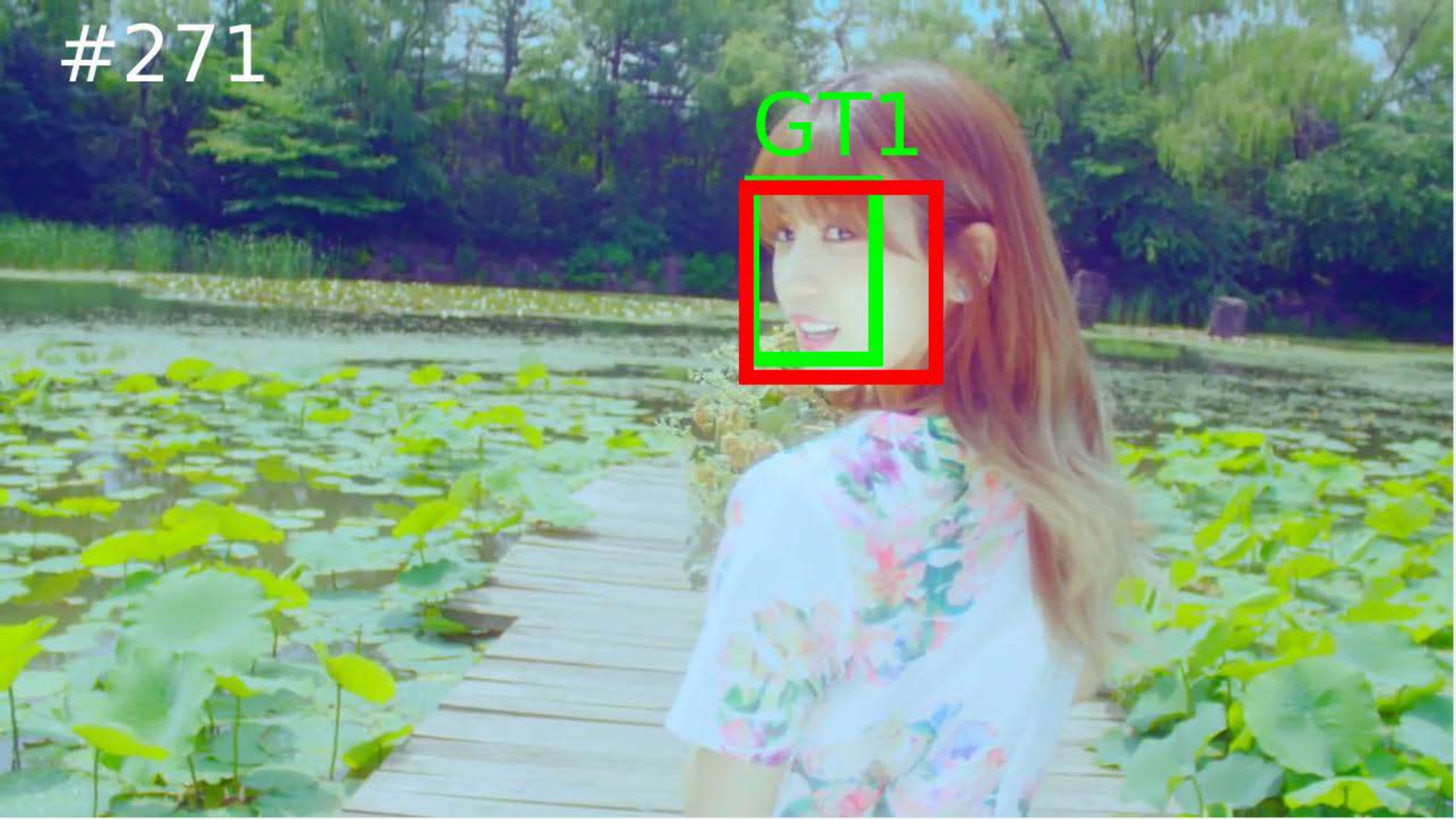} &
\hspace{-1.0mm}\includegraphics[width=3.0cm]{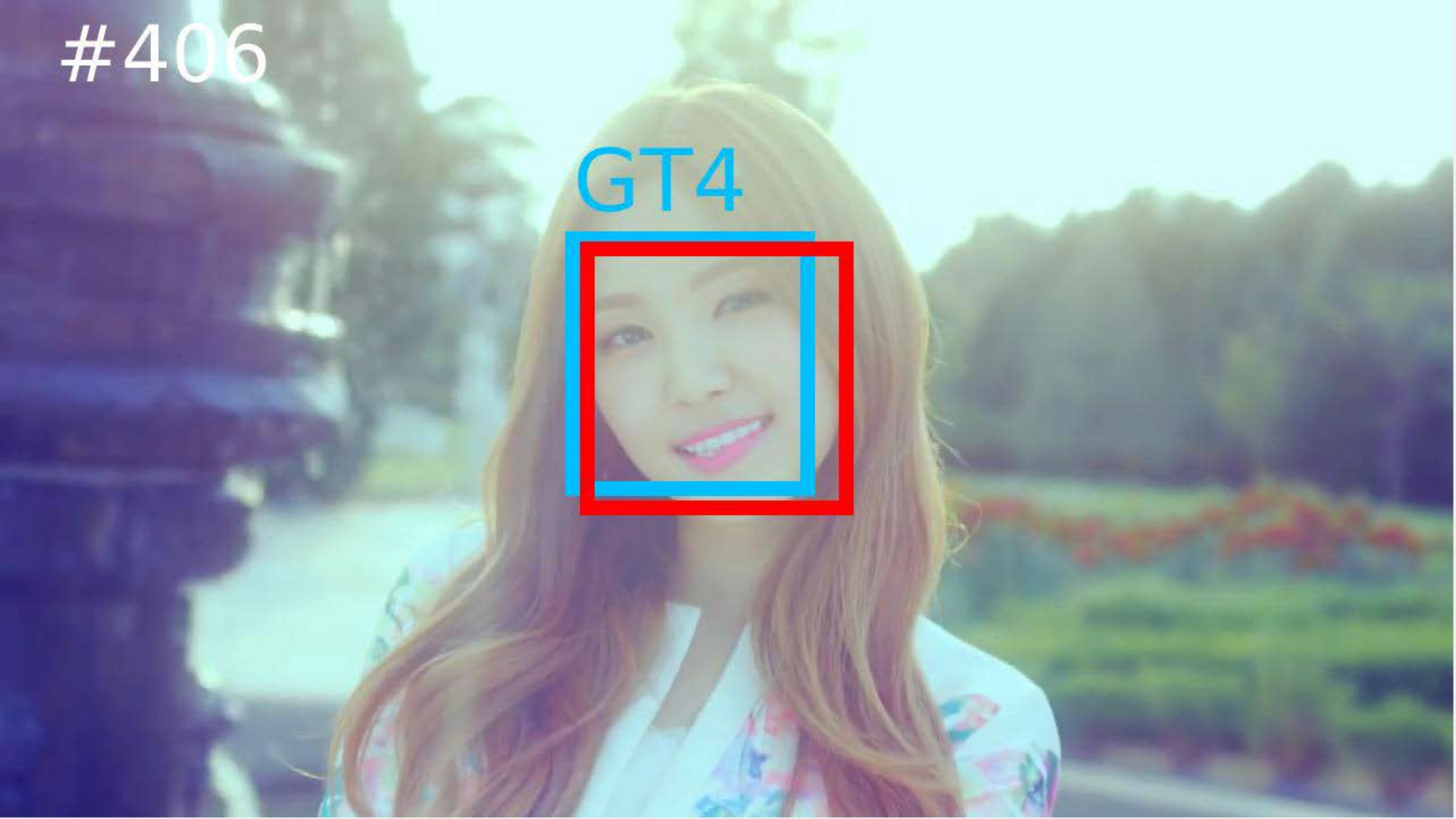} &
\hspace{-1.0mm}\includegraphics[width=3.0cm]{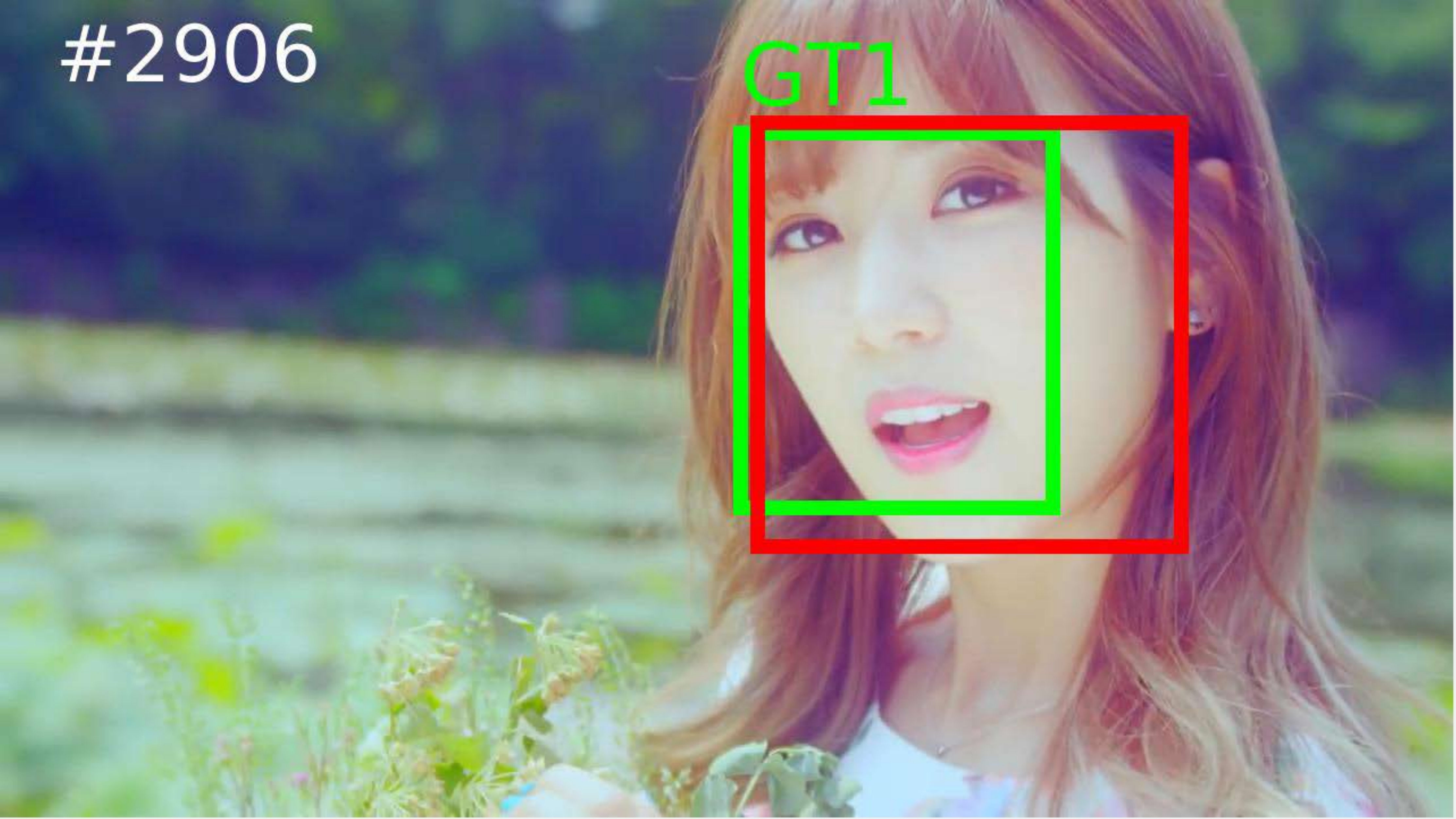} &
\hspace{-1.0mm}\includegraphics[width=3.0cm]{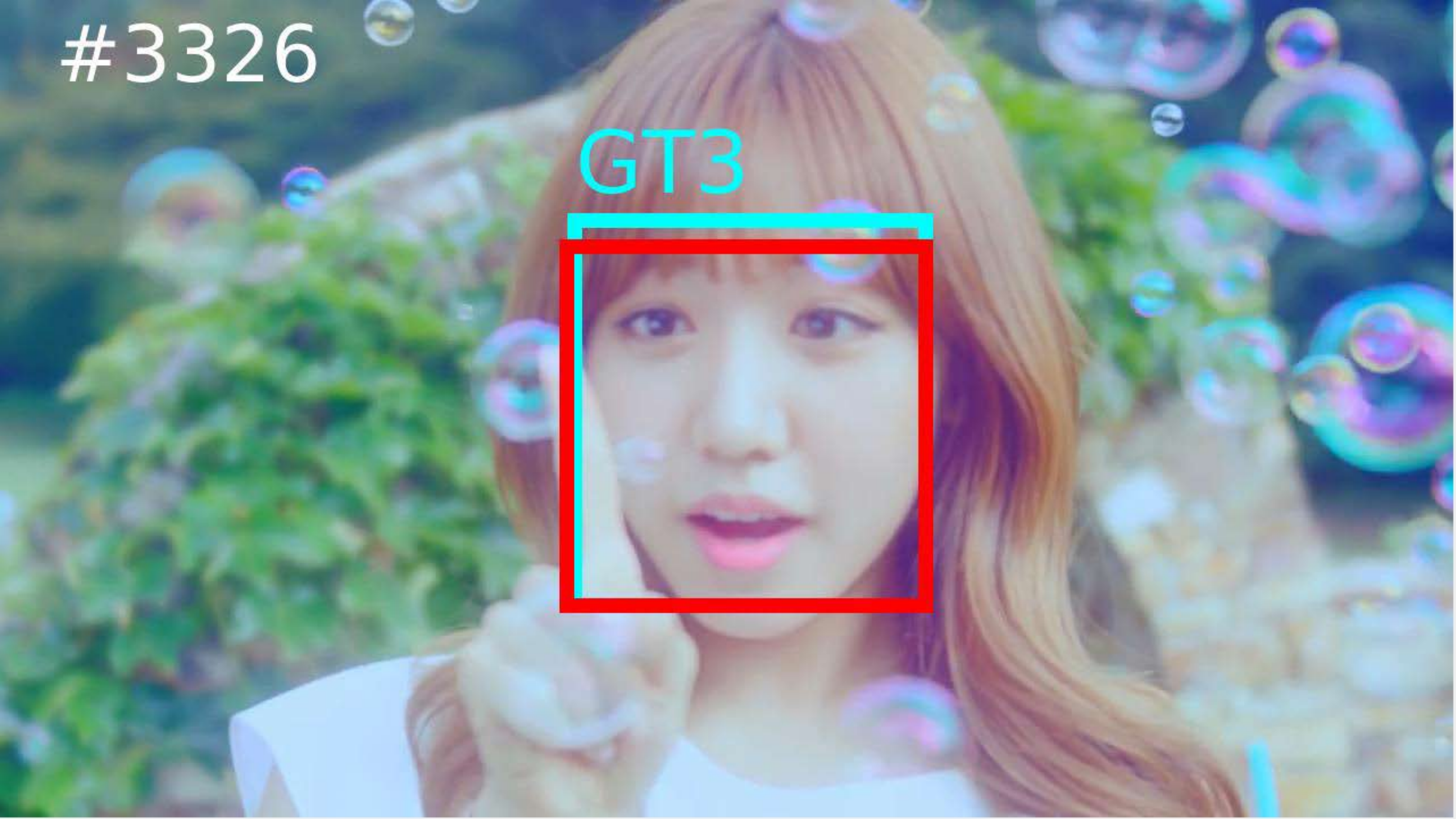} &
\hspace{-1.0mm}\includegraphics[width=3.0cm]{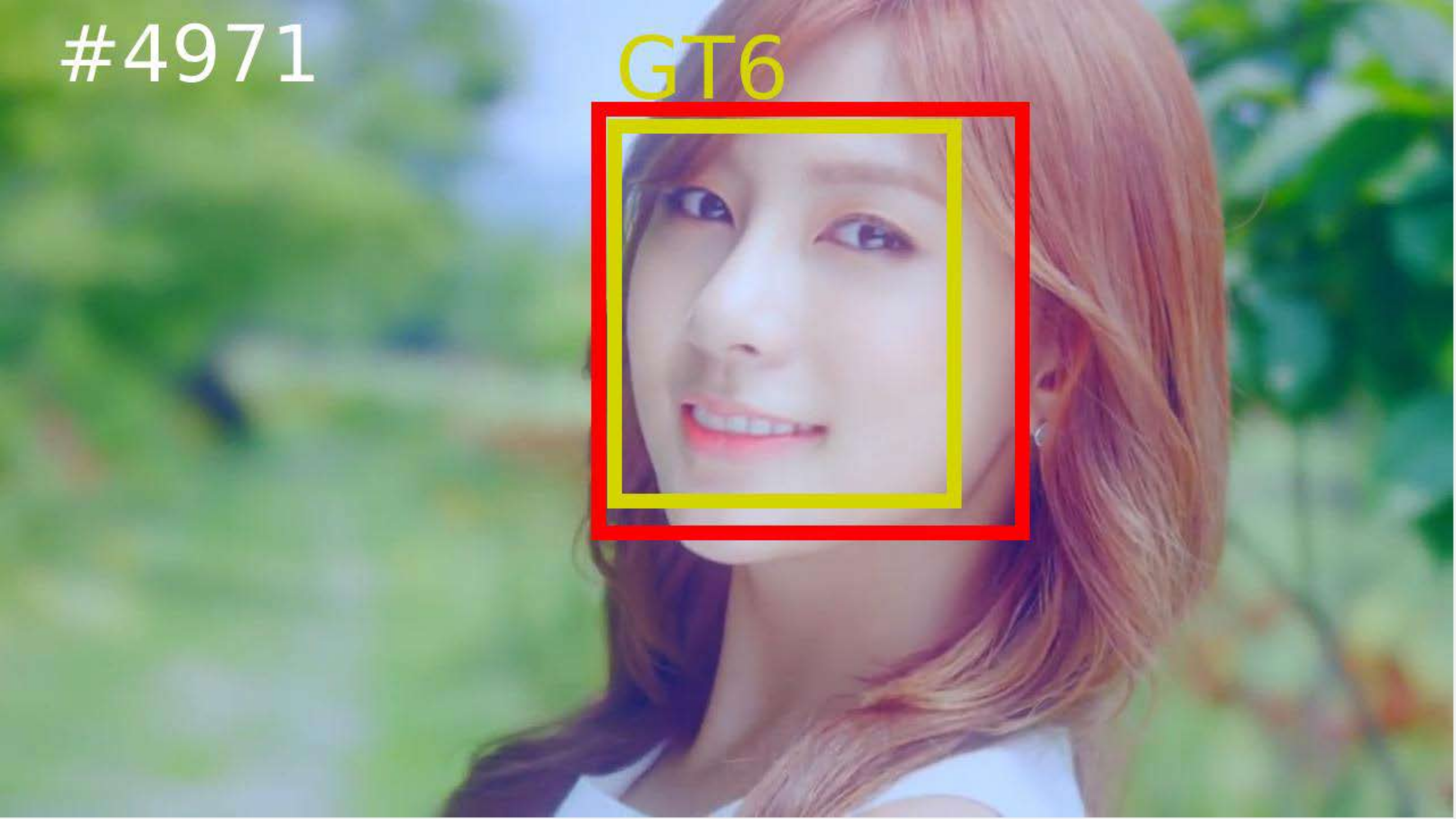} \\

\includegraphics[width=3.0cm]{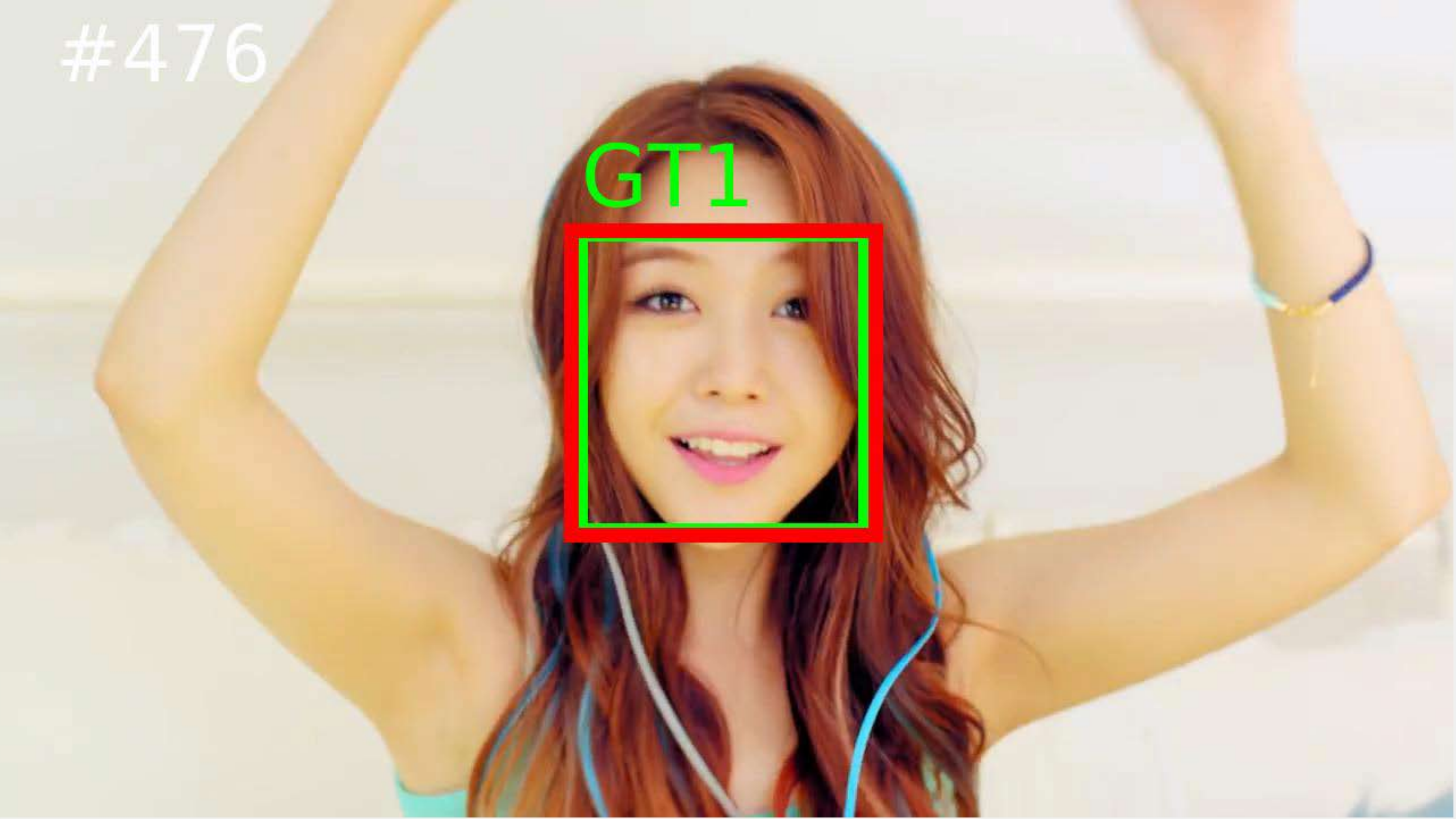} &
\hspace{-1.0mm}\includegraphics[width=3.0cm]{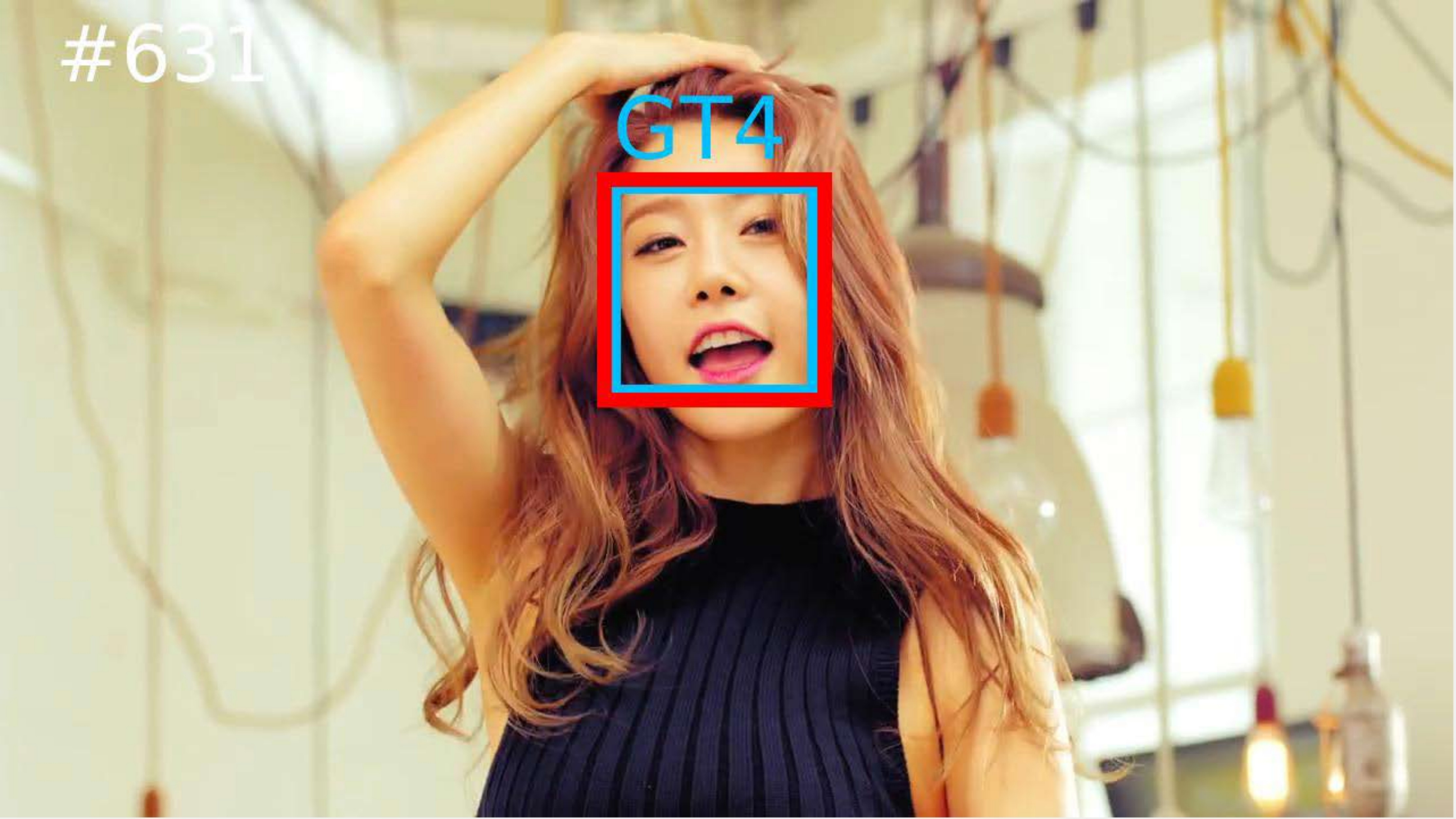} &
\hspace{-1.0mm}\includegraphics[width=3.0cm]{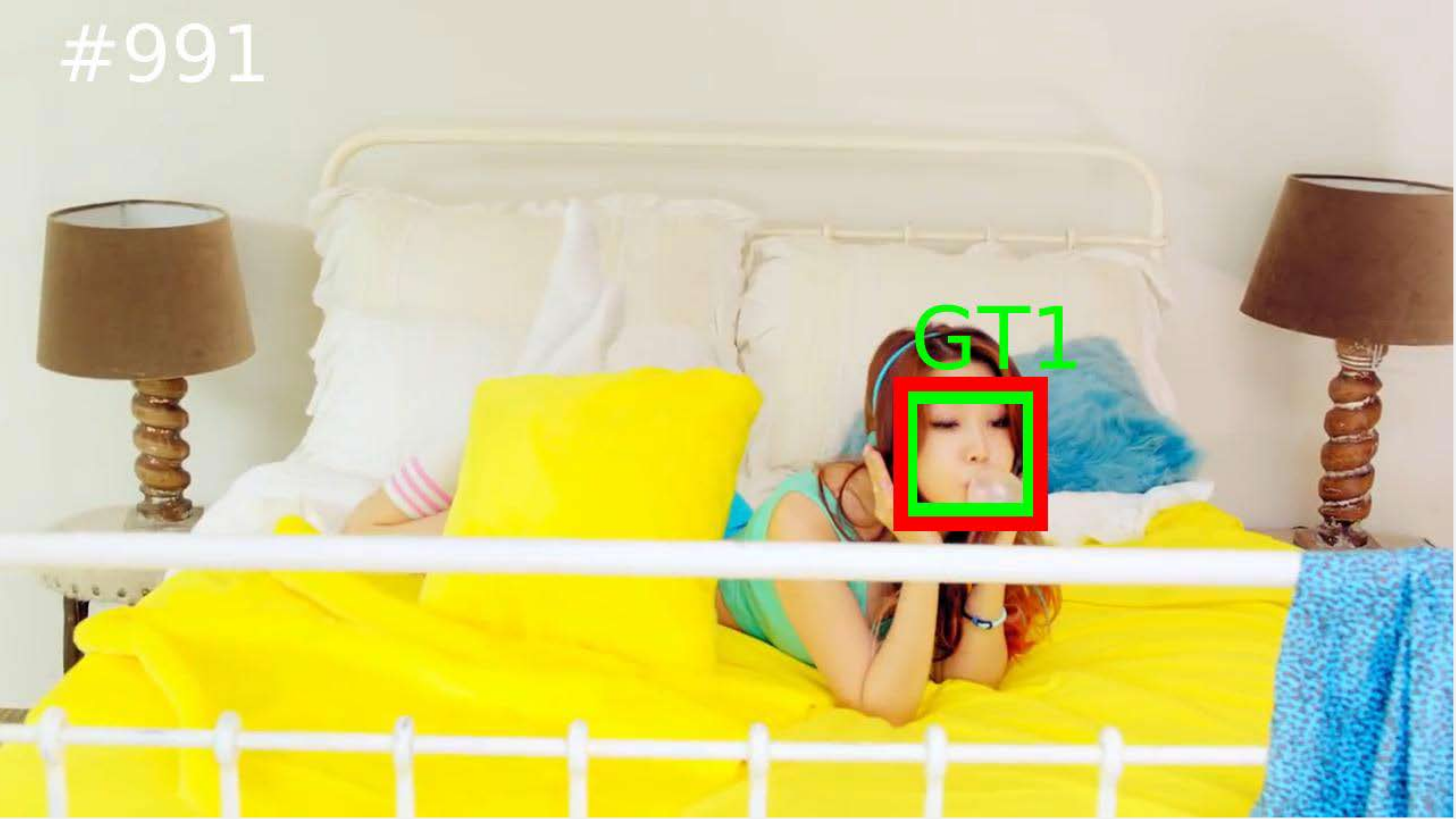} &
\hspace{-1.0mm}\includegraphics[width=3.0cm]{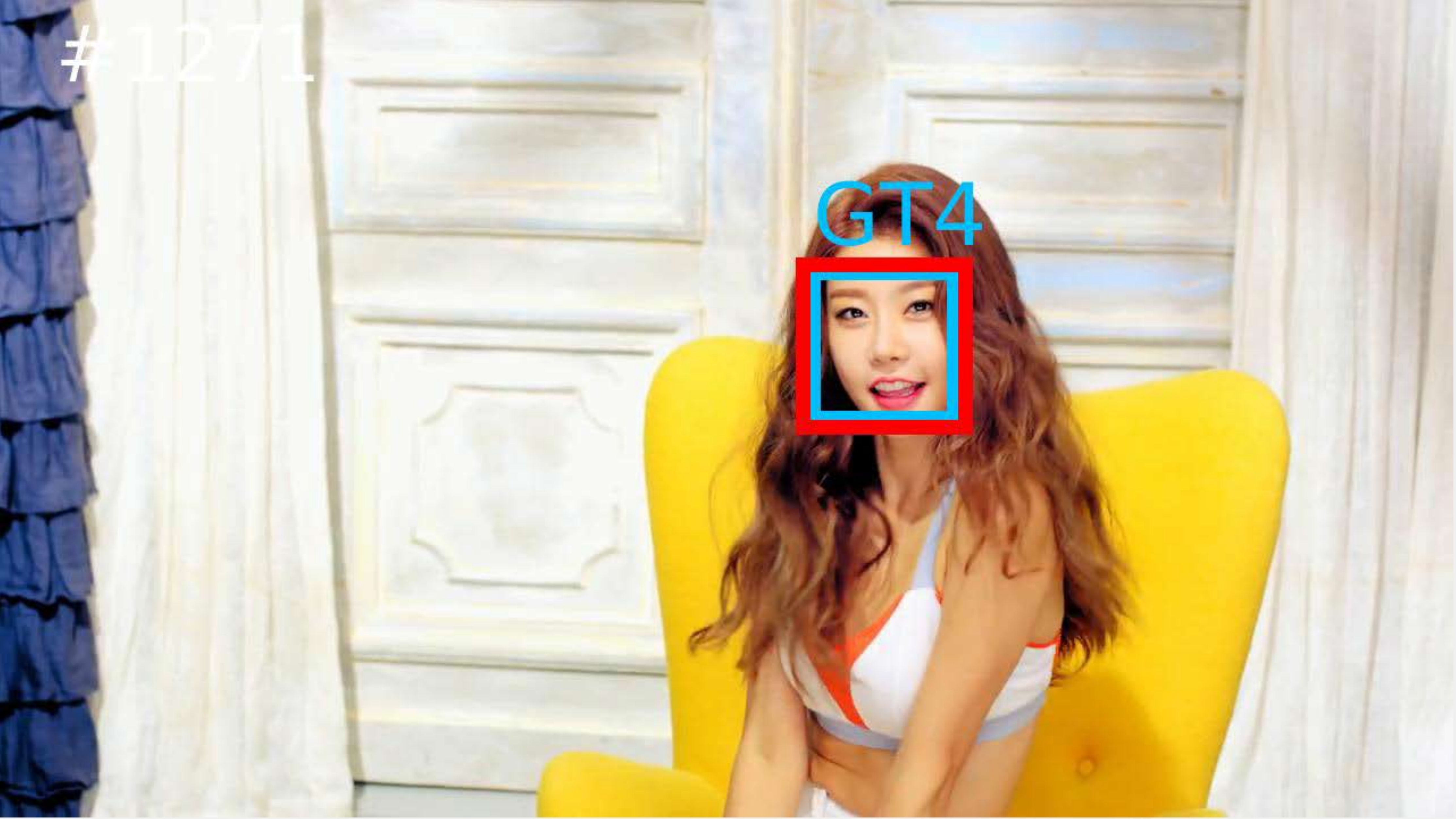} &
\hspace{-1.0mm}\includegraphics[width=3.0cm]{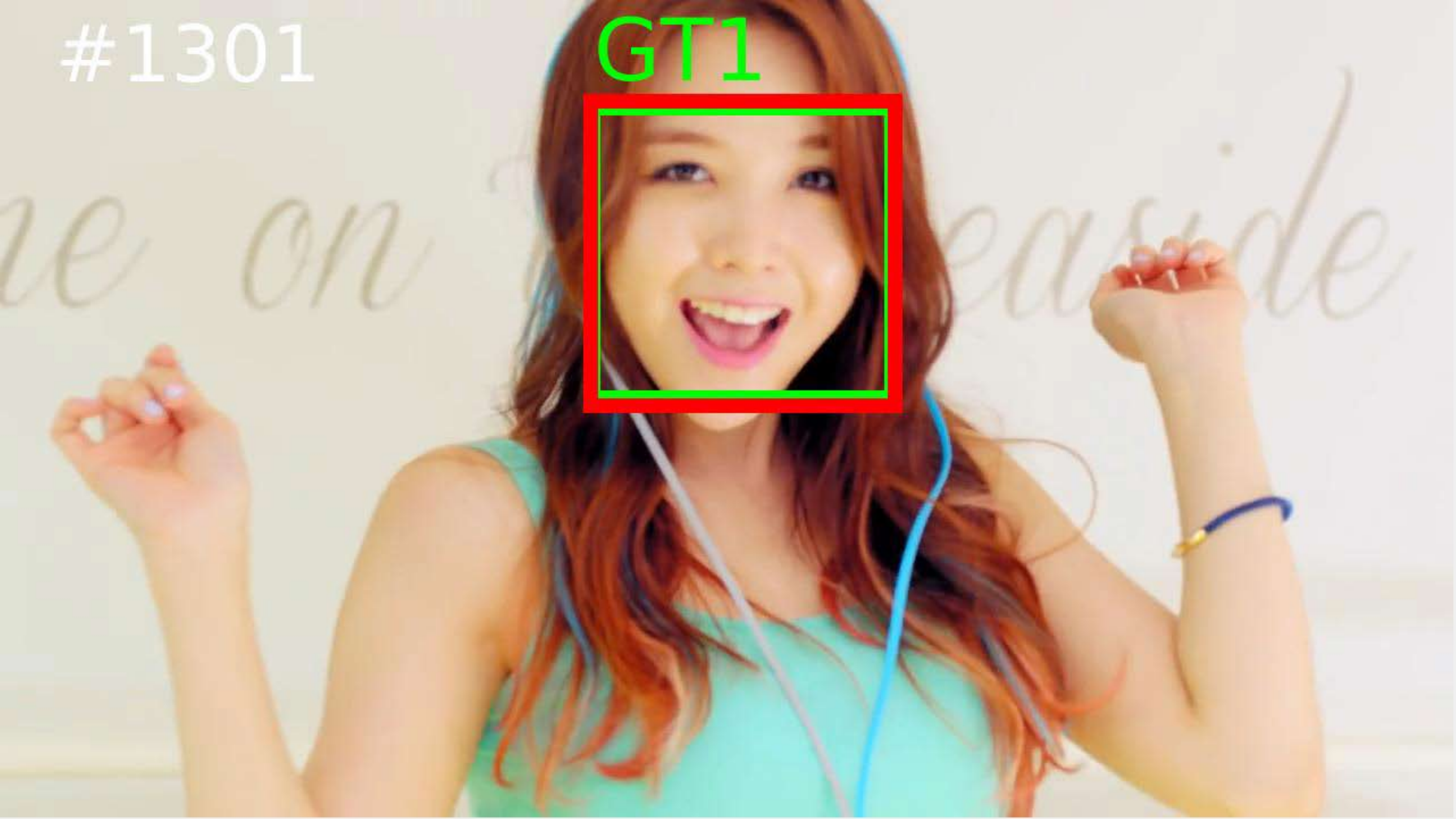} \\
\end{tabular}
\vspace{-1mm}
\caption{\textbf{Failure cases}. Our method incorrectly identifies different persons as the same one across shots on the \textsc{Apink} and \textsc{Darling} sequences. Numbers and colors of rectangles indicate the \emph{ground truth} identities of persons. The red rectangles show the predicted locations and are tracked as one person by our method. On the \textsc{Apink} sequence on the top row, Persons 1, 3, 4 and 6 are incorrectly assigned with the same identity. On the \textsc{Darling} sequence on the bottom row, our method incorrectly identifies Persons 1 and 4 as the same one across shots.}
\label{fig:fail}
\vspace{-3mm}
\end{figure*}

While the proposed algorithm performs favorably against the state-of-the-art face tracking and clustering methods in handling challenging video sequences, there are three main limitations.
First, as our algorithm takes face detections as inputs, the tracking performance depends on whether faces can be reliably detected.
For example, in the fourth row of Figure~\ref{fig:samples}, the leftmost person was not detected in frame 419 and next few images due to occlusion.
In addition, falsely detected faces could be incorrectly linked as a trajectory, e.g., the Marilyn Monroe image on the T-shirt in frame 5,704 in the eighth row of Figure~\ref{fig:samples}.

Second, the proposed algorithm may not perform well on sequences where many shots contain only one single person.
We show in Figure~\ref{fig:fail} two failure cases in the \textsc{Darling} and \textsc{Apink} sequences.
In such cases, the proposed method does not generate negative face pairs for training the Siamese/triplet network for distinguishing similar faces.
As such, different persons are incorrectly identified as the same one.
One remedy is to exploit other weak supervision signals (e.g., scripts, voice, contextual information) to generate visual constraints for different scenarios.

Third, the CNN fine-tuning process is time-consuming.
It takes around 1 hour on a NVIDIA GT980Ti GPU for 10,000 back-propagation iterations.
There are two approaches that may alleviate this issue.
First, we may use faster training algorithms~\cite{lin2015neural}.
Second, for TV Sitcom episodes we can use one or a few videos for feature adaptation and apply the learned features to all other episodes.
Note that we only need to adapt features once as the main characters are the same.
In Table~\ref{tab:feats_videos}, we train Ours-SymTriplet features on \textsc{BBT02}
(referred to as Ours-SymTriplet-BBT02) and evaluate on other episodes.
Although the weight purity of Ours-SymTriplet-BBT02 is slightly inferior to that of Ours-SymTriplet, it still outperforms the pre-trained and VGG-Face features.

%We note several methods exploit body descriptors for improving character tracking in TV Sitcom videos~\cite{tapaswi2012knock}.
%%
%However, these body descriptors may not be helpful for music videos as a person may be in completely different outfits in different scenes, e.g., the first row in Figure~\ref{fig:samples}.

%========================================================
\vspace{-3mm}
\section{Conclusions}
\vspace{-1mm}
In this paper, we tackle the multi-face tracking problem in unconstrained videos by learning video-specific features.
We first pre-train a CNN on a large-scale face recognition dataset to learn identity-preserving face representation.
We then adapt the pre-trained CNN using training samples extracted through the spatio-temporal and contextual constraints.
To learn discriminative features for handling large appearance variations of faces presented in a specific video, we propose the SymTriplet loss function.
Using the learned features for modeling face tracklets, we use a hierarchical clustering algorithm link face tracklets across multiple shots.
In addition to multi-face tracking, we demonstrate that the proposed algorithm can also be applied to other domains such as pedestrian tracking across multiple cameras.
Experimental results show that the proposed algorithm outperforms the state-of-the-art methods in terms of clustering accuracy and tracking performance.
As the performance of our approach depends on the automatically discovered visual constraints in the video, we believe that exploiting multi-modal information (e.g., sound/script alignment) is a promising direction for further improvement.

\ifCLASSOPTIONcompsoc
\vspace{-4mm}
\section*{Acknowledgments}
\vspace{-1mm}
\else
\section*{Acknowledgment}
\fi

The work is supported by National Basic Research Program of China (973 Program, 2015CB351705), NSFC (61332018, 61703344), Office of Naval Research (N0014-16-1-2314), R\&D programs by NRF (2014R1A1A2058501) and MSIP/IITP (IITP-2016-H8601-16-1005) of Korea, NSF CAREER (1149783) and gifts from Adobe, Panasonic, NEC, and NVIDIA.

\ifCLASSOPTIONcaptionsoff
%\newpage
\fi

%{
%\scriptsize
\vspace{-1mm}
\bibliographystyle{IEEEtran}
\bibliography{IEEEabrv,FaceTracking}
%}

\begin{IEEEbiography}
[{\includegraphics[width=1in,height=1.25in,clip,keepaspectratio]{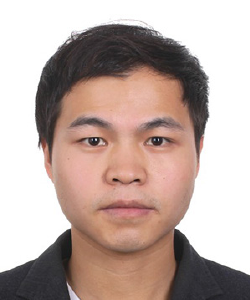}}]{Shun Zhang}
received the Ph.D. degree in Institute of Artificial Intelligence and Robotics from Xi'an Jiaotong University. He is currently an assistant professor in School of Electronics and Information at Northwestern Polytechnical University.
His research interests include Machine Learning, Computer Vision and Human-Computer Interaction, with a focus on visual tracking, object detection, image classification, feature extraction and sparse representation.
\end{IEEEbiography}
\vspace{-15mm}
% if you will not have a photo at all:
\begin{IEEEbiography}
[{\includegraphics[width=1in,height=1.25in,clip,keepaspectratio]{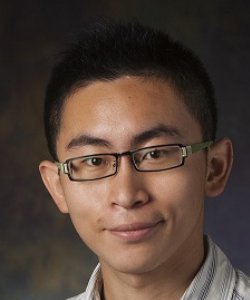}}]{Jia-Bin Huang}
is an assistant professor in the Electrical and Computer Engineering department at Virginia Tech. He received the PhD degree in the Department of Electrical and Computer Engineering at the University of Illinois, Urbana-Champaign. His main research interests are in the area of computer vision, computer graphics, and machine learning.
\end{IEEEbiography}
\vspace{-15mm}
\begin{IEEEbiography}
[{\includegraphics[width=1in,height=1.25in,clip,keepaspectratio]{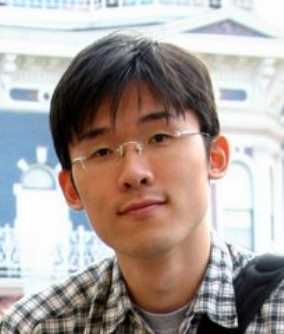}}]{Jongwoo Lim}
graduated from Seoul National University, Seoul, Korea, in 1997, and received his MS degree in 2003 and PhD degree in 2005 from University of Illinois at Urbana-Champaign, USA. He worked at Honda Research Institute USA Inc., Mountain View, CA, USA, as a senior scientist from 2005 to 2011, and at Google Inc., Mountain View, CA, USA, as a software engineer from 2011 to 2012. Currently, he is an associate professor in Division of Computer Science and Engineering at Hanyang University.
%His research interests include computer vision, robotics, and machine learning.
\end{IEEEbiography}
\vspace{-15mm}
\begin{IEEEbiography}[{\includegraphics[width=1in,height=1.25in,clip,keepaspectratio]{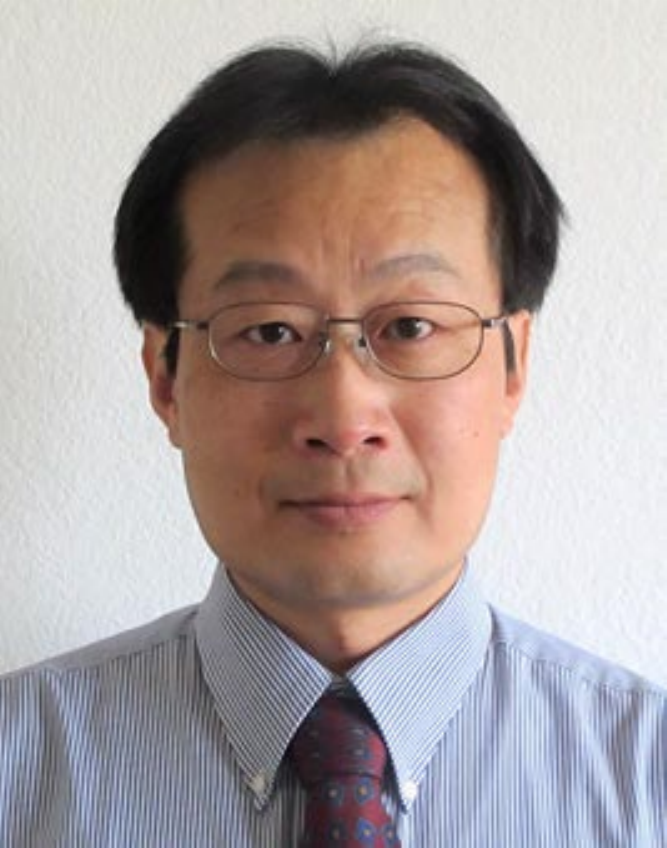}}]{Yihong Gong}
received his B.S., M.S., and Ph.D. degrees in electrical engineering from the University of Tokyo, Japan in 1987, 1989, and 1992, respectively.
In 1992, he joined Nanyang Technological University, Singapore as an assistant professor with the School of Electrical and Electronic Engineering.
From 1996 to 1998, he was a Project Scientist with the Robotics Institute, Carnegie Mellon University, USA.
Since 1999 he worked for the Silicon Valley branch, NEC Labs America as a group leader, department head, and branch manager.
He joined Xian Jiaotong University, China as a distinguished professor in 2012.
%His research interests include image and video analysis, multimedia database systems, and machine learning.
\end{IEEEbiography}
\vspace{-15mm}
\begin{IEEEbiography}[{\includegraphics[width=1in,height=1.25in,clip,keepaspectratio]{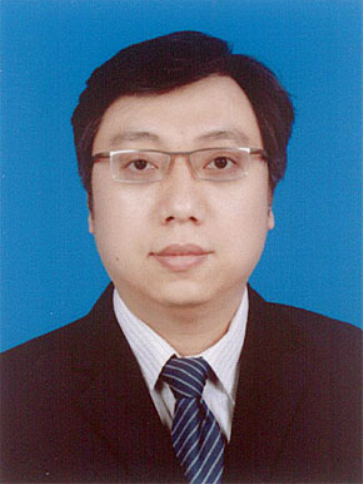}}]{Jinjun Wang}
received the B.E. and M.E. degree from Huazhong University of Science and Technology, China, in 2000 and 2003 respectively, and received the Ph.D. degree from Nanyang Technological University, Singapore, in 2006.
Currently, he is with Xi'an Jiaotong University, China, as a professor.
Prior to join XJTU, he worked in NEC Laboratories America and Epson Research USA form 2006 to 2013 as a research scientist and senior research scientist.
His research interests include computer vision, pattern classification, image/video enhancement and editing, content-based multimedia retrieval, etc.
\end{IEEEbiography}
\vspace{-15mm}
\begin{IEEEbiography}
[{\includegraphics[width=1in,height=1.25in,clip,keepaspectratio]{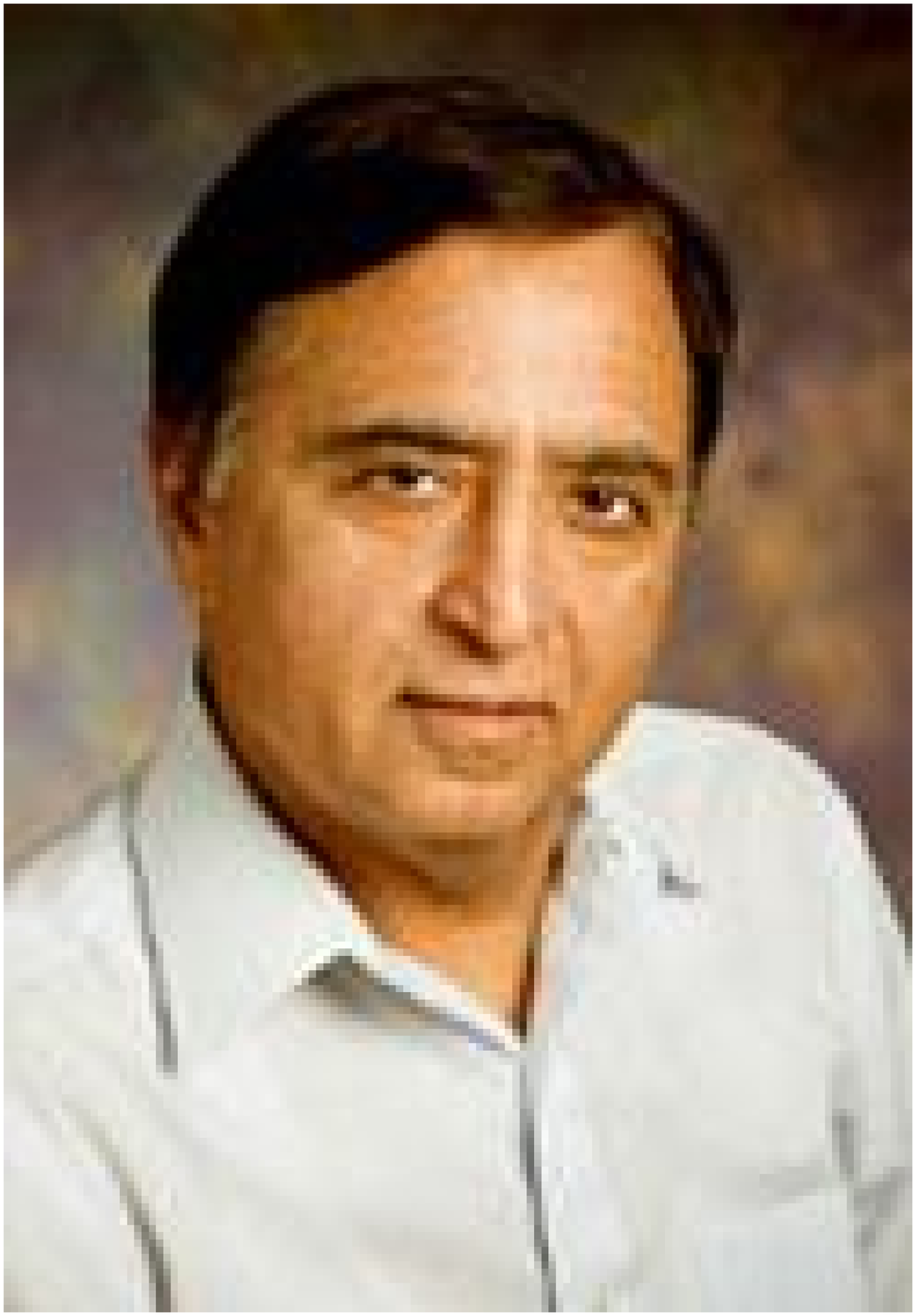}}]{Narendra Ahuja}
is the Donald Biggar Willet Professor in the Department of Electrical and Computer Engineering at the University of Illinois at Urbana-Champaign. He is a Fellow of the IEEE, American Association for Artificial Intelligence, International Association for Pattern Recognition, Association for Computing Machinery, American Association for the Advancement of Science, and International Society for Optical Engineering. Narendra is on the editorial boards of several journals. He was the Founding Director of the International Institute of Information Technology, Hyderabad where he continues to serve as Director International.
%He is a fellow of the IEEE, the AAAI, the IAPR, the ACM, the AAAS, and the SPIE.
\end{IEEEbiography}
\vspace{-15mm}
\begin{IEEEbiography}
[{\includegraphics[width=1in,height=1.25in,clip,keepaspectratio]{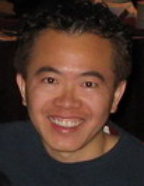}}]{Ming-Hsuan Yang}
is a professor in Electrical Engineering and Computer Science at University of California, Merced.
He received the PhD degree in Computer Science from the University of Illinois at Urbana-Champaign in 2000.
Yang has served as an associate editor of the IEEE Transactions on Pattern Analysis and Machine Intelligence,
International Journal of Computer Vision, Image and Vision Computing, Computer Vision and Image Understanding, and Journal of Artificial Intelligence Research. He received the NSF CAREER award in 2012 and the Google Faculty Award in 2009.
%He is a senior member of the IEEE and the ACM.
\end{IEEEbiography}

\end{document}